\documentclass[10pt,twocolumn,letterpaper]{article} 

\usepackage{cvpr}              % To produce the CAMERA-READY version
\usepackage[accsupp]{axessibility}
\usepackage{array}
\usepackage{multirow}
\usepackage{tabularx}

\usepackage{algorithm}
\usepackage{algorithmic}
\usepackage{amsfonts}
\usepackage{amsmath} 
\usepackage{float}
\usepackage{xcolor}
\usepackage{rotating}

\definecolor{idcolor}{RGB}{0, 70, 130}     
\definecolor{action1color}{RGB}{214, 39, 40}  
\definecolor{action2color}{RGB}{31, 119, 180} 
\definecolor{action3color}{RGB}{44, 160, 44} 
\definecolor{action4color}{RGB}{255, 127, 14} 
\definecolor{action5color}{RGB}{148, 103, 189} % A nice violet
\definecolor{action6color}{RGB}{140, 86, 75}   % A warm brown
\definecolor{action7color}{RGB}{227, 119, 194} % A magenta/pink
\definecolor{action8color}{RGB}{23, 190, 207}   % A bright cyan
\newcommand{\promptpart}[2]{{\color{#1}#2}}

\usepackage{newfloat}
\usepackage{listings}

\definecolor{cvprblue}{rgb}{0.21,0.49,0.74}
\usepackage[pagebackref,breaklinks,colorlinks,allcolors=cvprblue]{hyperref}

\def\paperID{} % *** Enter the Paper ID here
\def\confName{CVPR}
\def\confYear{2026}

\title{RealDiffusion: Physics-informed Attention for Multi-character Storybook Generation}

\author{
  Qi Zhao\textsuperscript{1,3}
  \quad
  Jun Chen\textsuperscript{2,3,$\dagger$}
  \quad
  Ivor Tsang\textsuperscript{4}
  \quad
  Guang Dai\textsuperscript{3}
  \\[6pt]
  \textsuperscript{1}Xi'an Jiaotong University, China
  \quad  \textsuperscript{2}Zhejiang Normal University, China
  \\
  \textsuperscript{3}SGIT AI Lab, State Grid Corporation of China
  \\
  \textsuperscript{4}Centre for Frontier AI Research (CFAR), A*STAR, Singapore
  \\[4pt]
  {\tt\small shmilyqi7@gmail.com \quad junc.change@gmail.com \quad Ivor\_tsang@a-star.edu.sg \quad guang.gdai@gmail.com}
}

\begin{document}
\twocolumn[{%
\renewcommand\twocolumn[1][]{#1}%
\maketitle
\begin{center}
\centering
\vspace{-2.5em}
\begin{tabular}{@{\hspace{-10mm}}c@{\hspace{0mm}}c}

\multicolumn{2}{c}{\parbox{\linewidth}{\centering\small\textbf{Prompt 1:} \promptpart{idcolor}{Art style of a classic comic book, two heroes,} \promptpart{action1color}{sharing pizza,} \promptpart{action2color}{racing to tower,} \promptpart{action3color}{watching city,} \promptpart{action4color}{celebrating.}}} \\

% \begin{tabular}{c}
%     \includegraphics[width=0.13\linewidth]{figure_cvpr/fig1_1/row_2_col_1.pdf}
%     \includegraphics[width=0.13\linewidth]{figure_cvpr/fig1_1/row_2_col_2.pdf}
%     \includegraphics[width=0.13\linewidth]{figure_cvpr/fig1_1/row_2_col_3.pdf}
%     \includegraphics[width=0.13\linewidth]{figure_cvpr/fig1_1/row_2_col_4.pdf}
% \end{tabular}
% &
% \begin{tabular}{c}
%     \includegraphics[width=0.13\linewidth]{figure_cvpr/fig1_2/row_2_col_1.pdf}
%     \includegraphics[width=0.13\linewidth]{figure_cvpr/fig1_2/row_2_col_2.pdf}
%     \includegraphics[width=0.13\linewidth]{figure_cvpr/fig1_2/row_2_col_3.pdf}
%     \includegraphics[width=0.13\linewidth]{figure_cvpr/fig1_2/row_2_col_4.pdf}
% \end{tabular}

\begin{tabular}{c}
    \includegraphics[width=0.13\linewidth]{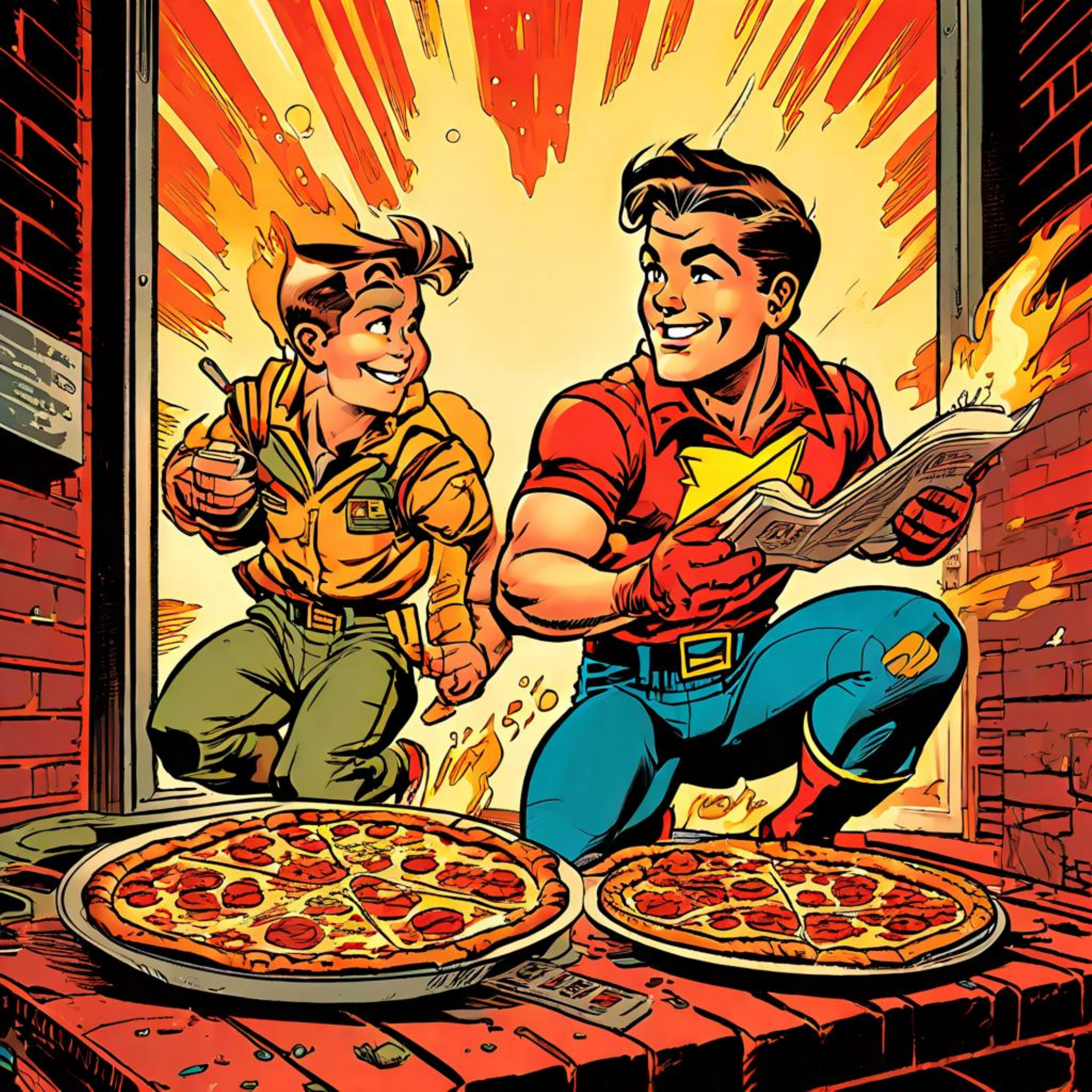}
    \includegraphics[width=0.13\linewidth]{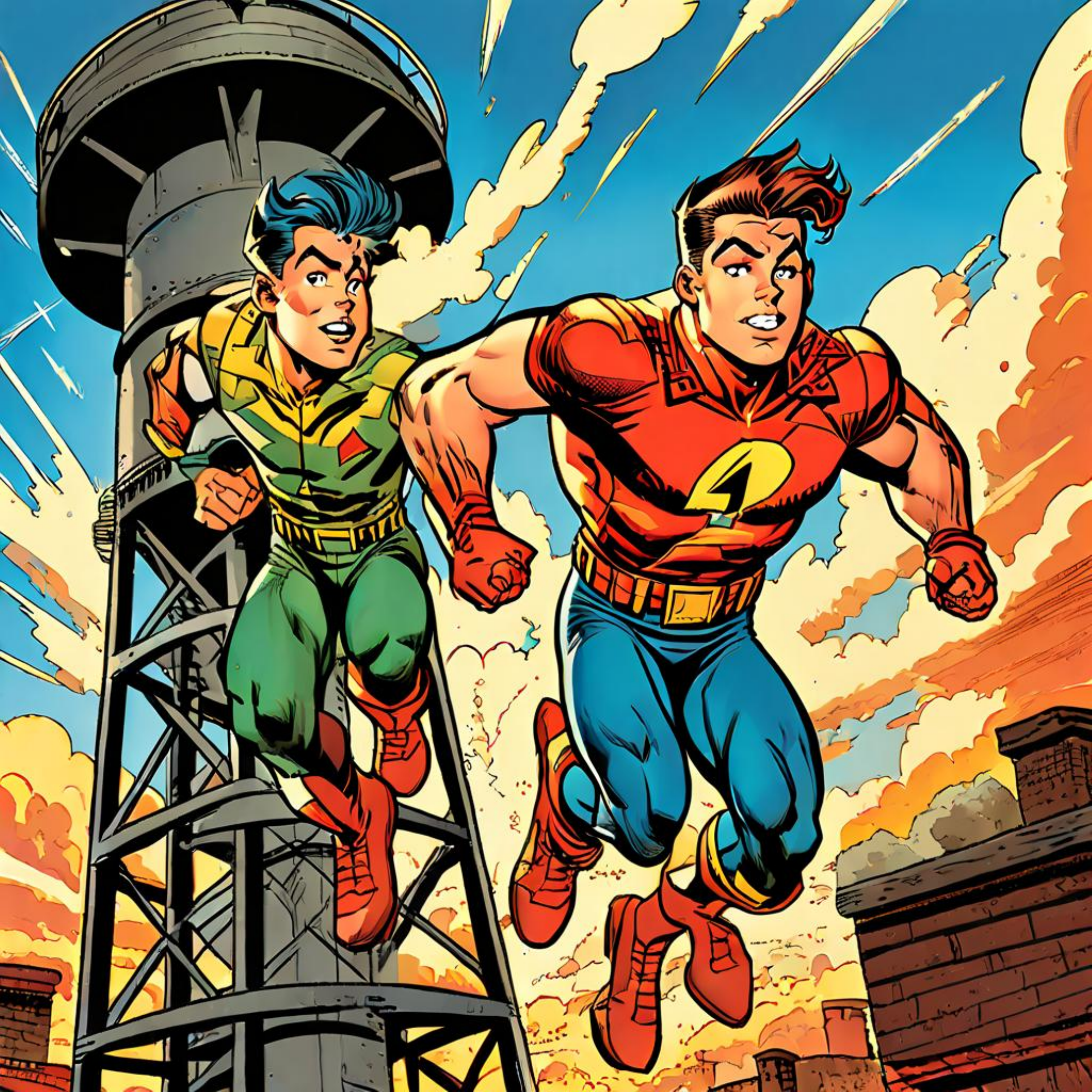}
    \includegraphics[width=0.13\linewidth]{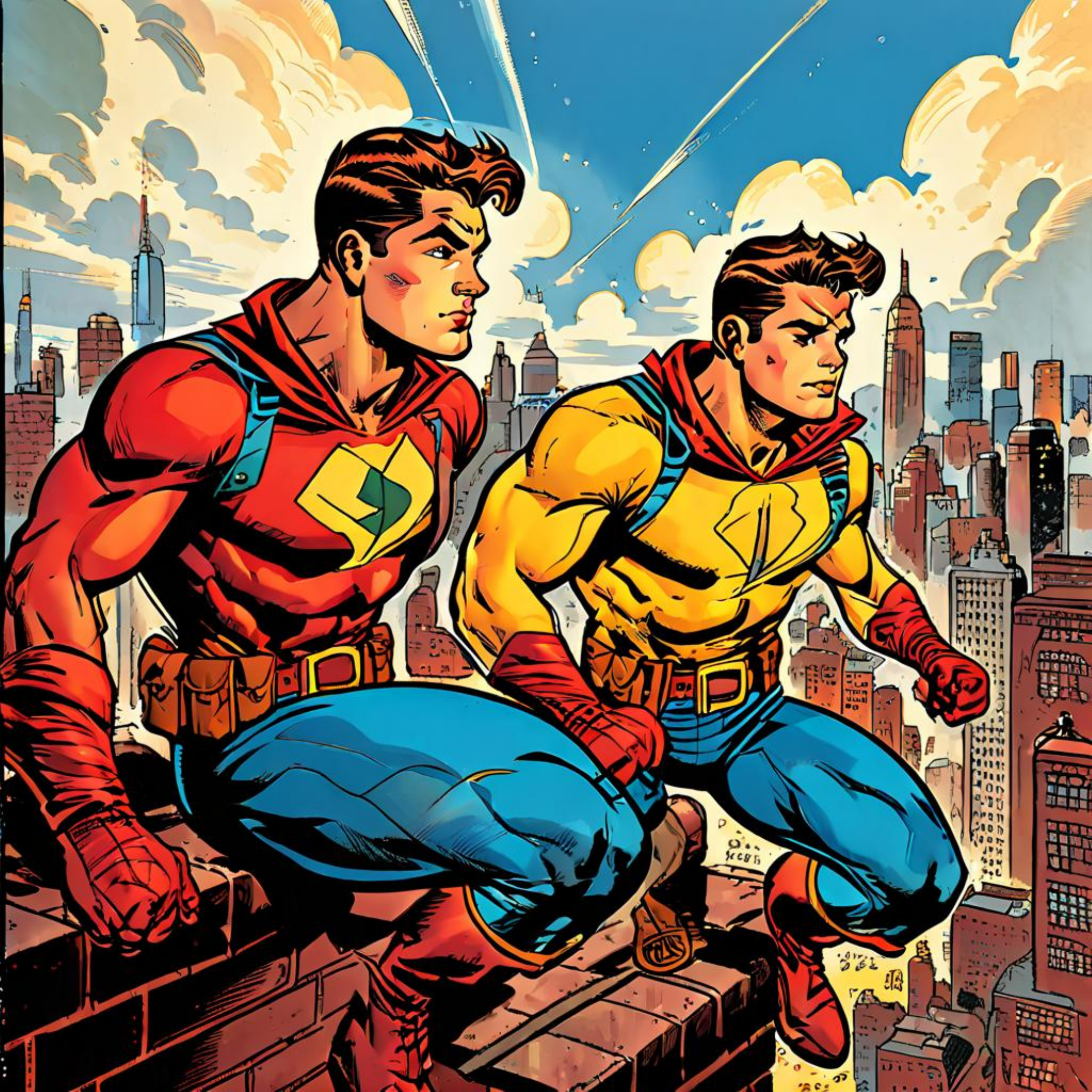}
    \includegraphics[width=0.13\linewidth]{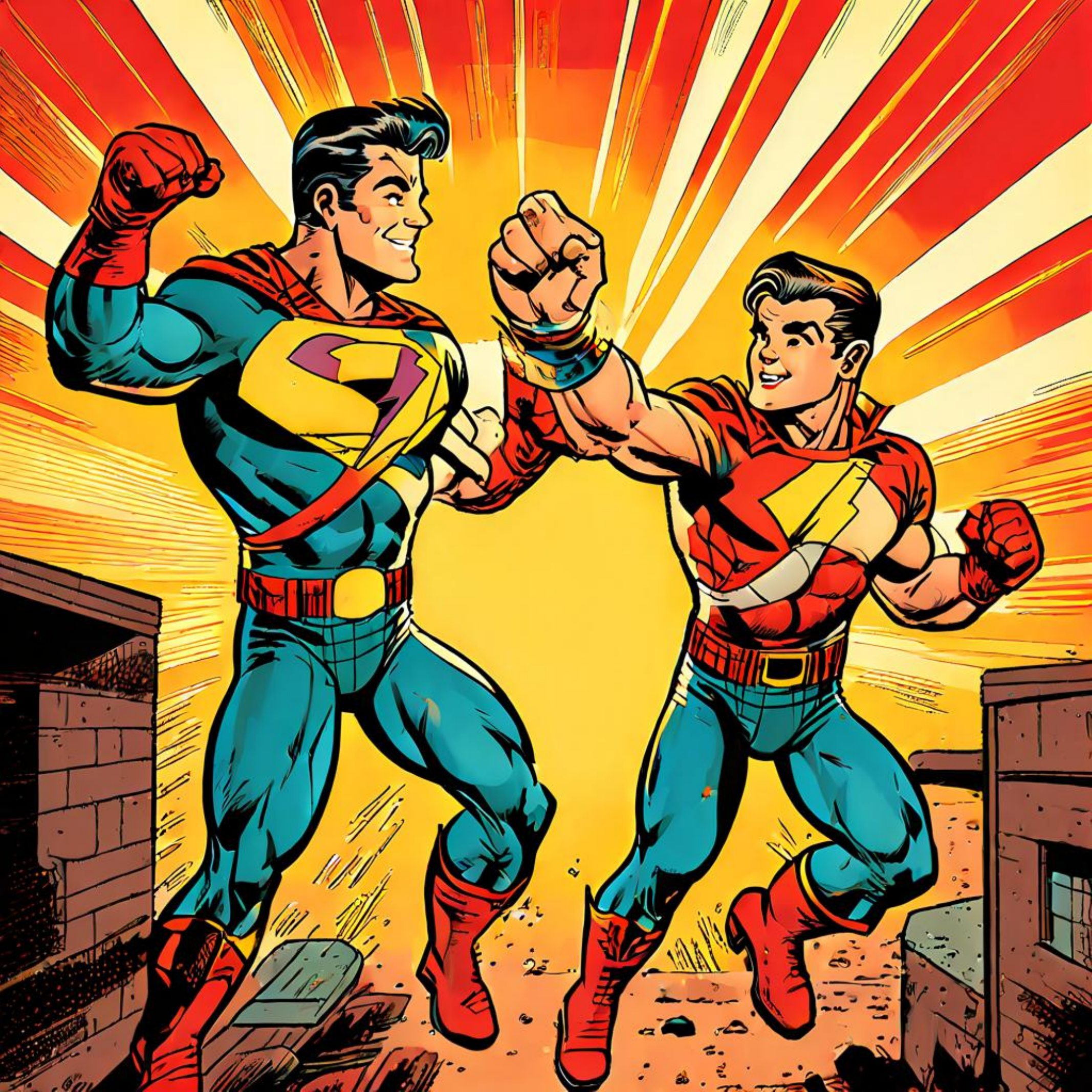}
\end{tabular}
&
\begin{tabular}{c}
    \includegraphics[width=0.13\linewidth]{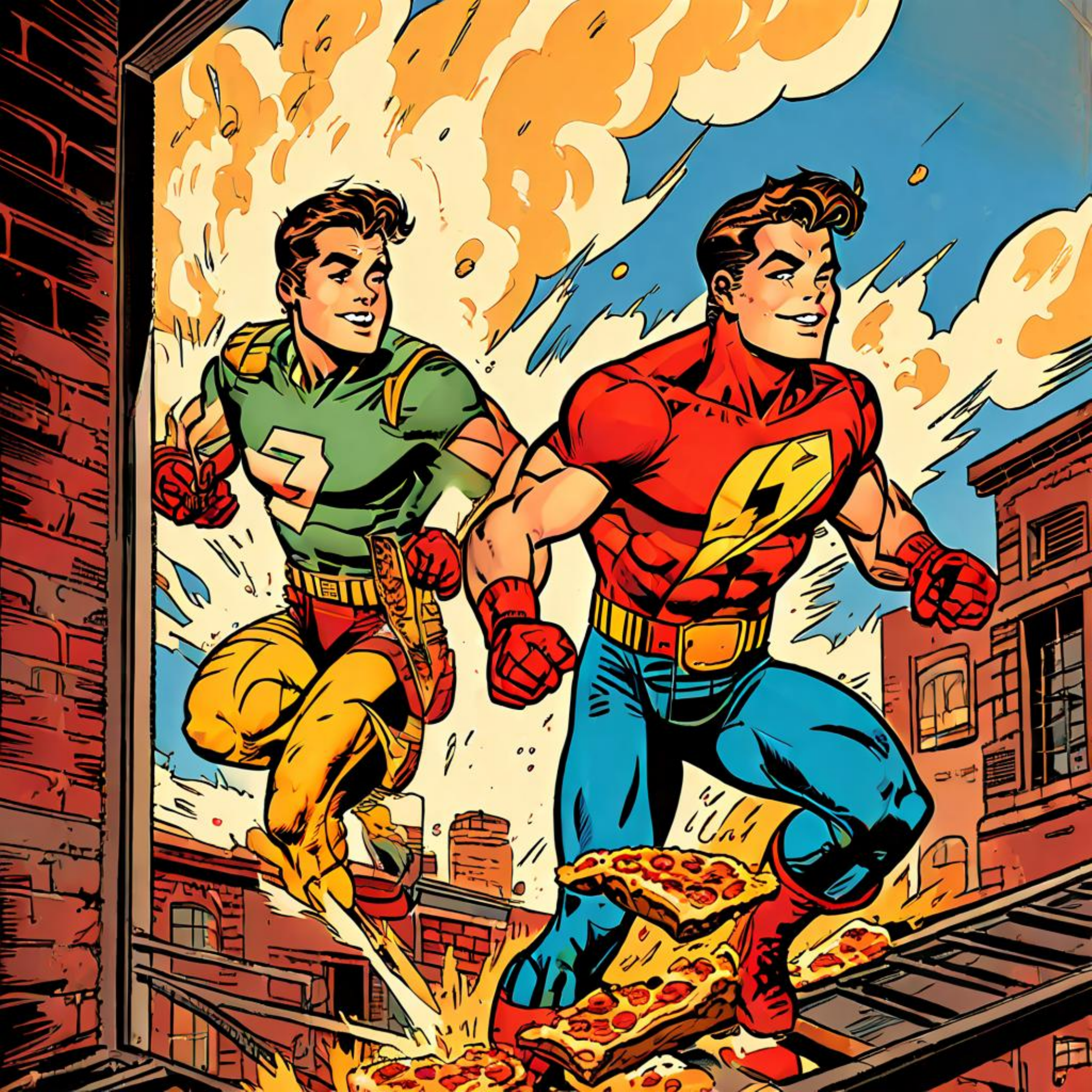}
    \includegraphics[width=0.13\linewidth]{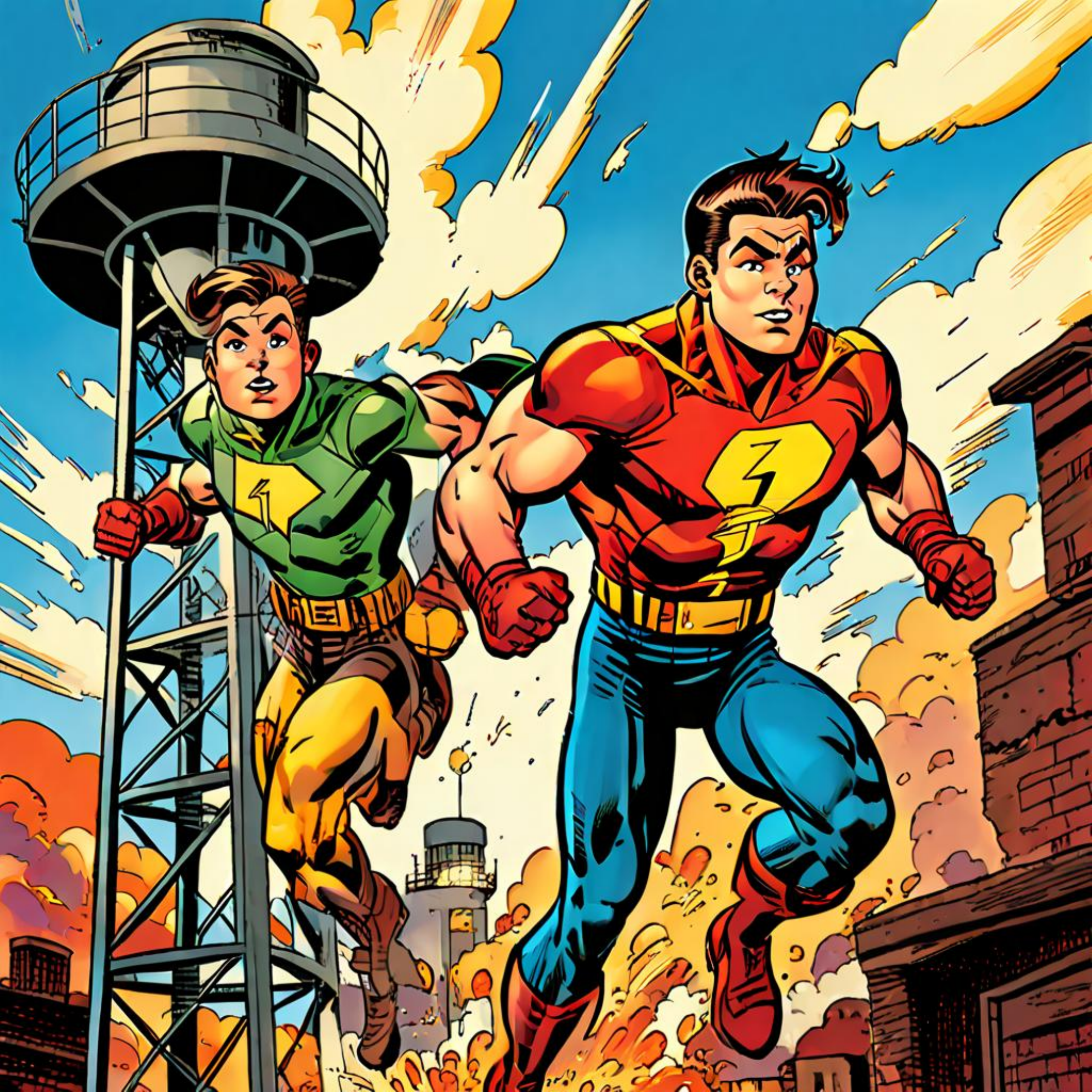}
    \includegraphics[width=0.13\linewidth]{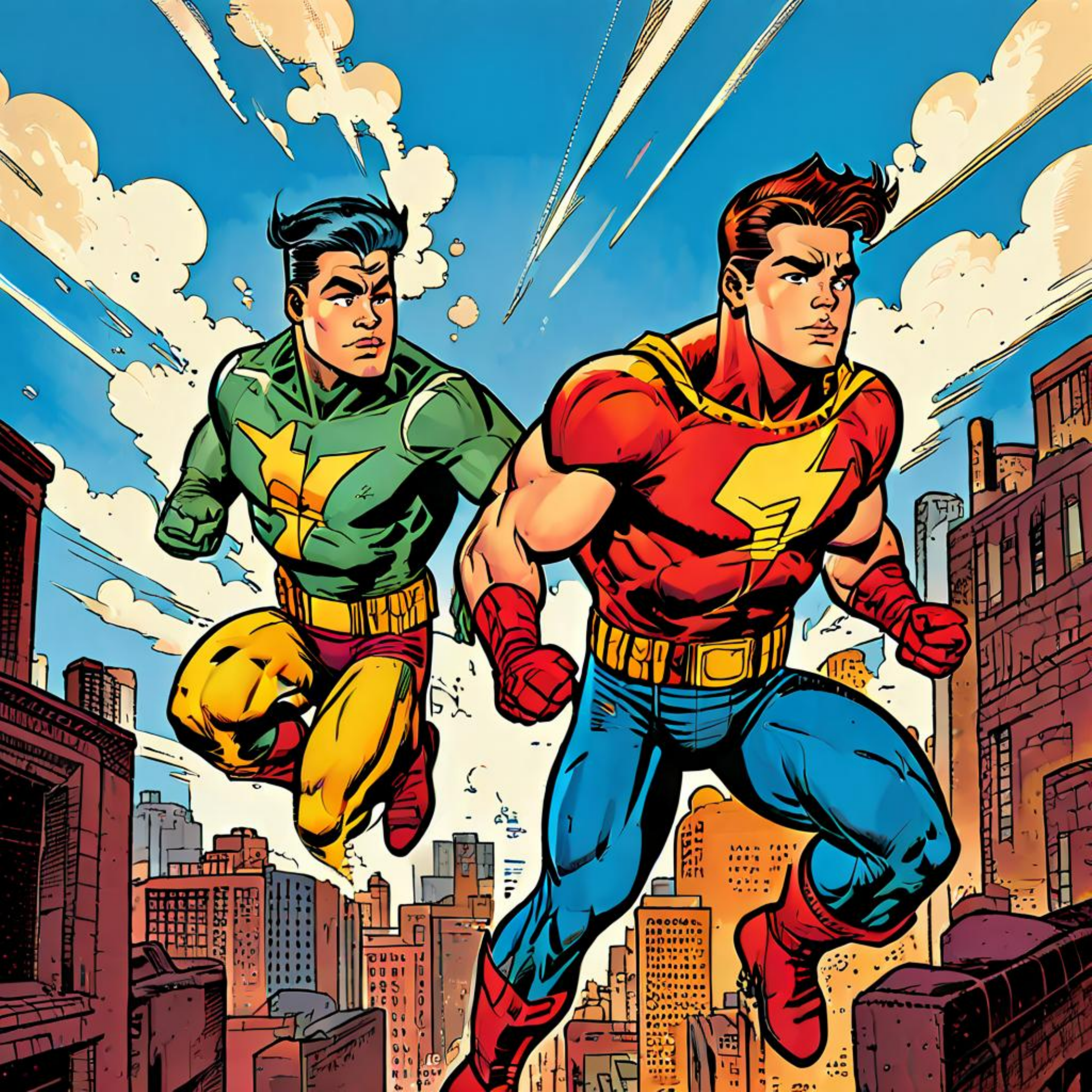}
    \includegraphics[width=0.13\linewidth]{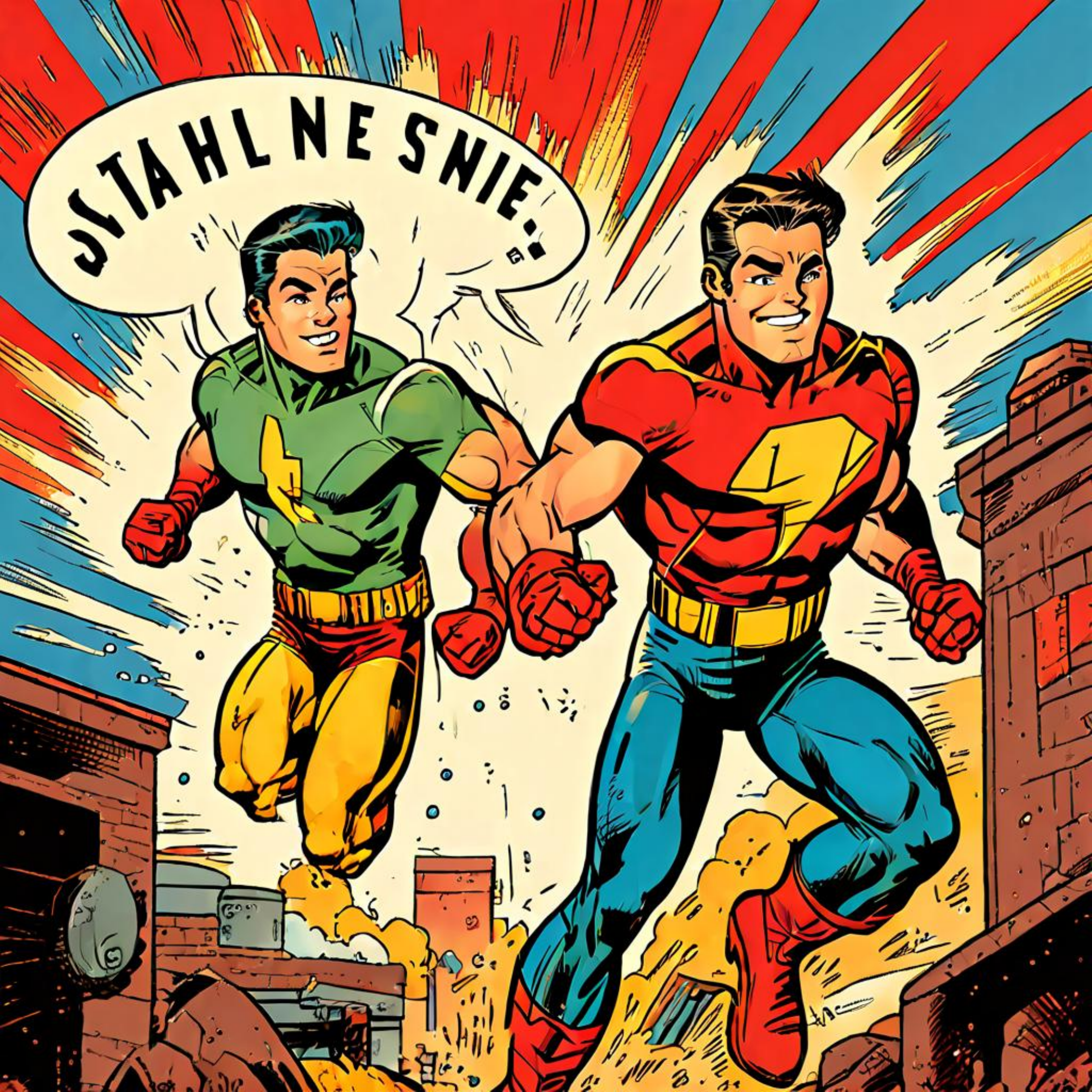}
\end{tabular}
\\

\multicolumn{2}{c}{\parbox{\linewidth}{\centering\small\textbf{Prompt 2:} \promptpart{idcolor}{Ghibli style anime, brother and sister,} \promptpart{action1color}{running in grass,} \promptpart{action2color}{sharing watermelon,} \promptpart{action3color}{wading in stream,} \promptpart{action4color}{looking from train.}}} \\

% \begin{tabular}{c}
%     \includegraphics[width=0.13\linewidth]{figure_cvpr/fig1_3/row_2_col_1.pdf}
%     \includegraphics[width=0.13\linewidth]{figure_cvpr/fig1_3/row_2_col_2.pdf}
%     \includegraphics[width=0.13\linewidth]{figure_cvpr/fig1_3/row_2_col_3.pdf}
%     \includegraphics[width=0.13\linewidth]{figure_cvpr/fig1_3/row_2_col_5.pdf}
% \end{tabular}
% &
% \begin{tabular}{c}
%     \includegraphics[width=0.13\linewidth]{figure_cvpr/fig1_4/row_2_col_1.pdf}
%     \includegraphics[width=0.13\linewidth]{figure_cvpr/fig1_4/row_2_col_2.pdf}
%     \includegraphics[width=0.13\linewidth]{figure_cvpr/fig1_4/row_2_col_3.pdf}
%     \includegraphics[width=0.13\linewidth]{figure_cvpr/fig1_4/row_2_col_5.pdf}
% \end{tabular}
\begin{tabular}{c}
    \includegraphics[width=0.13\linewidth]{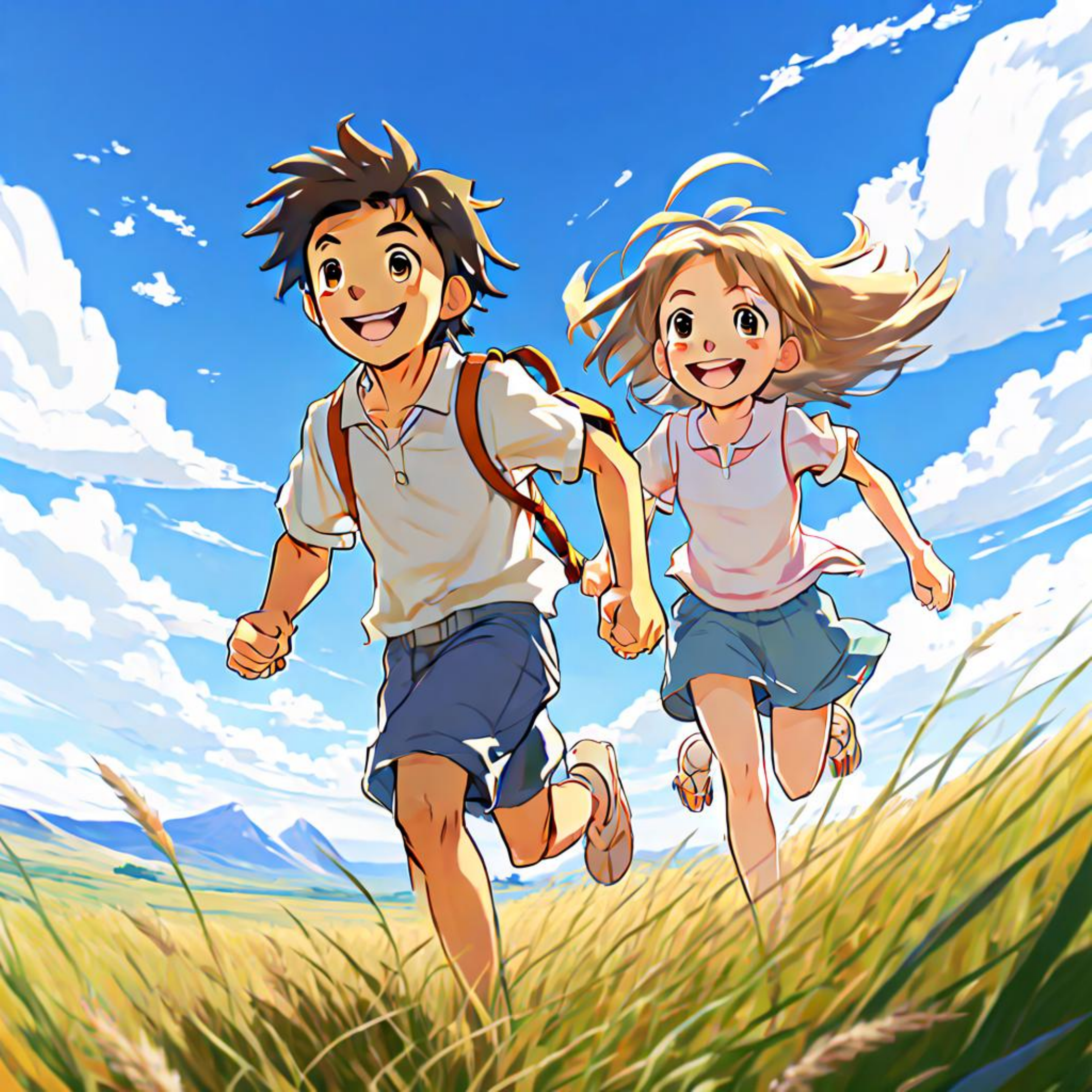}
    \includegraphics[width=0.13\linewidth]{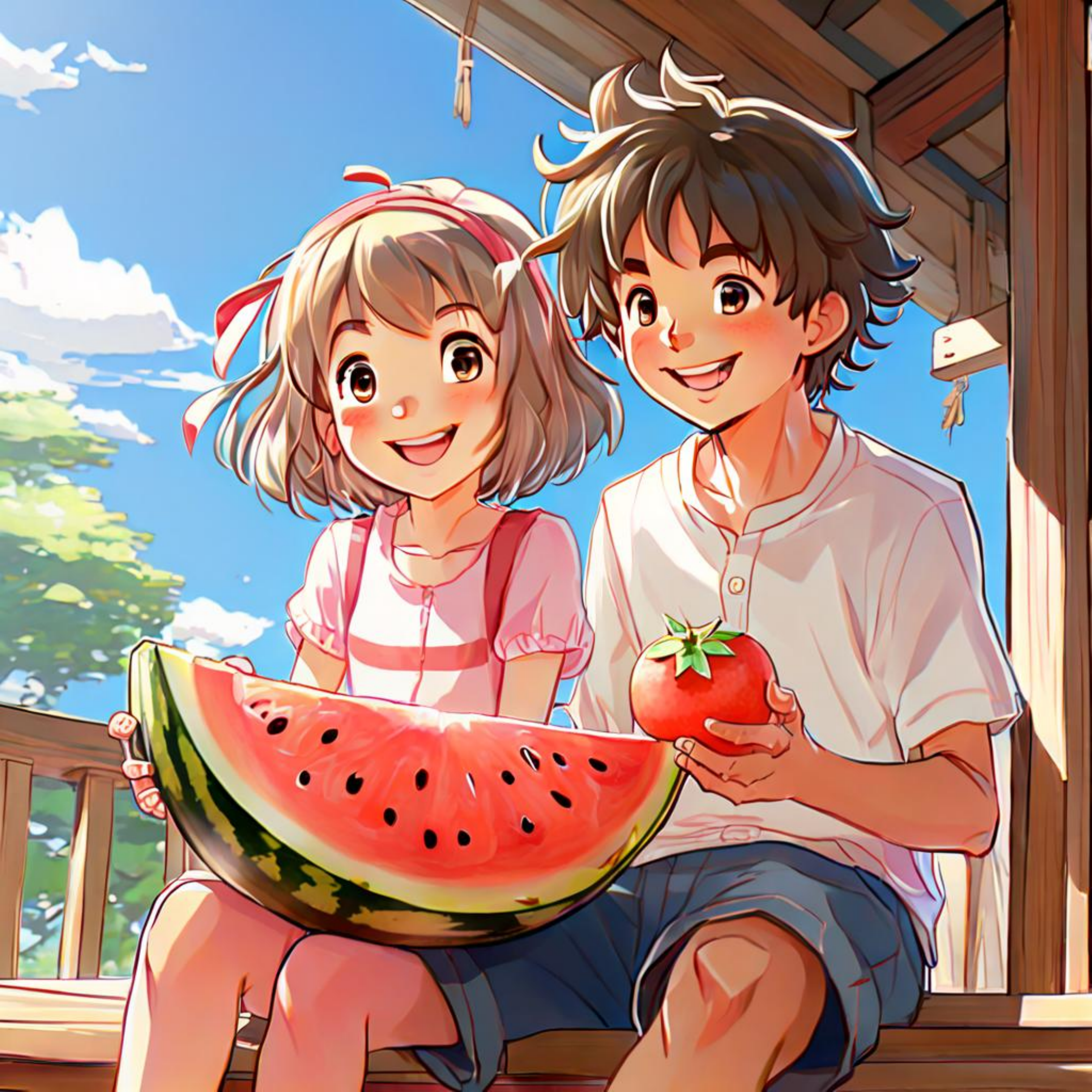}
    \includegraphics[width=0.13\linewidth]{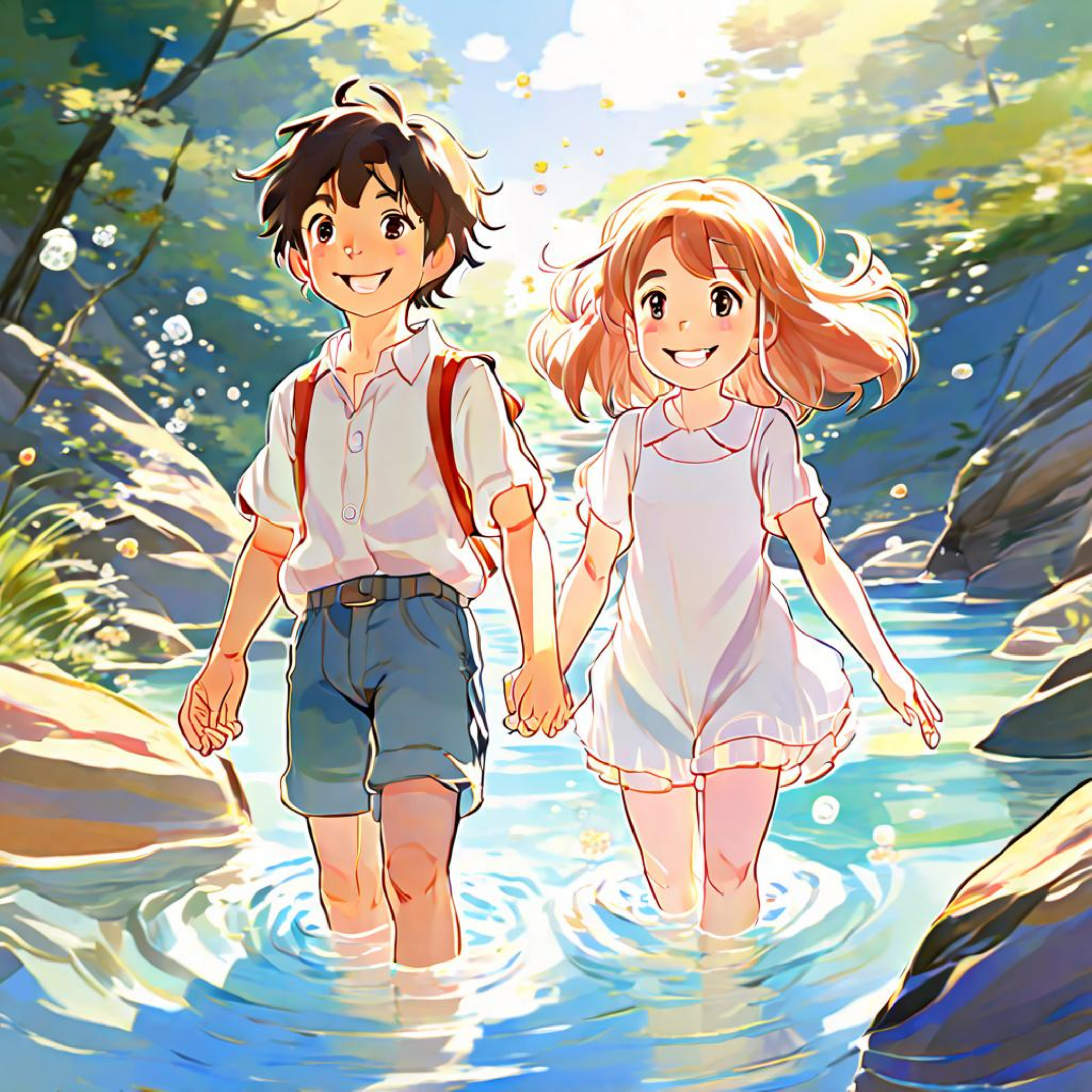}
    \includegraphics[width=0.13\linewidth]{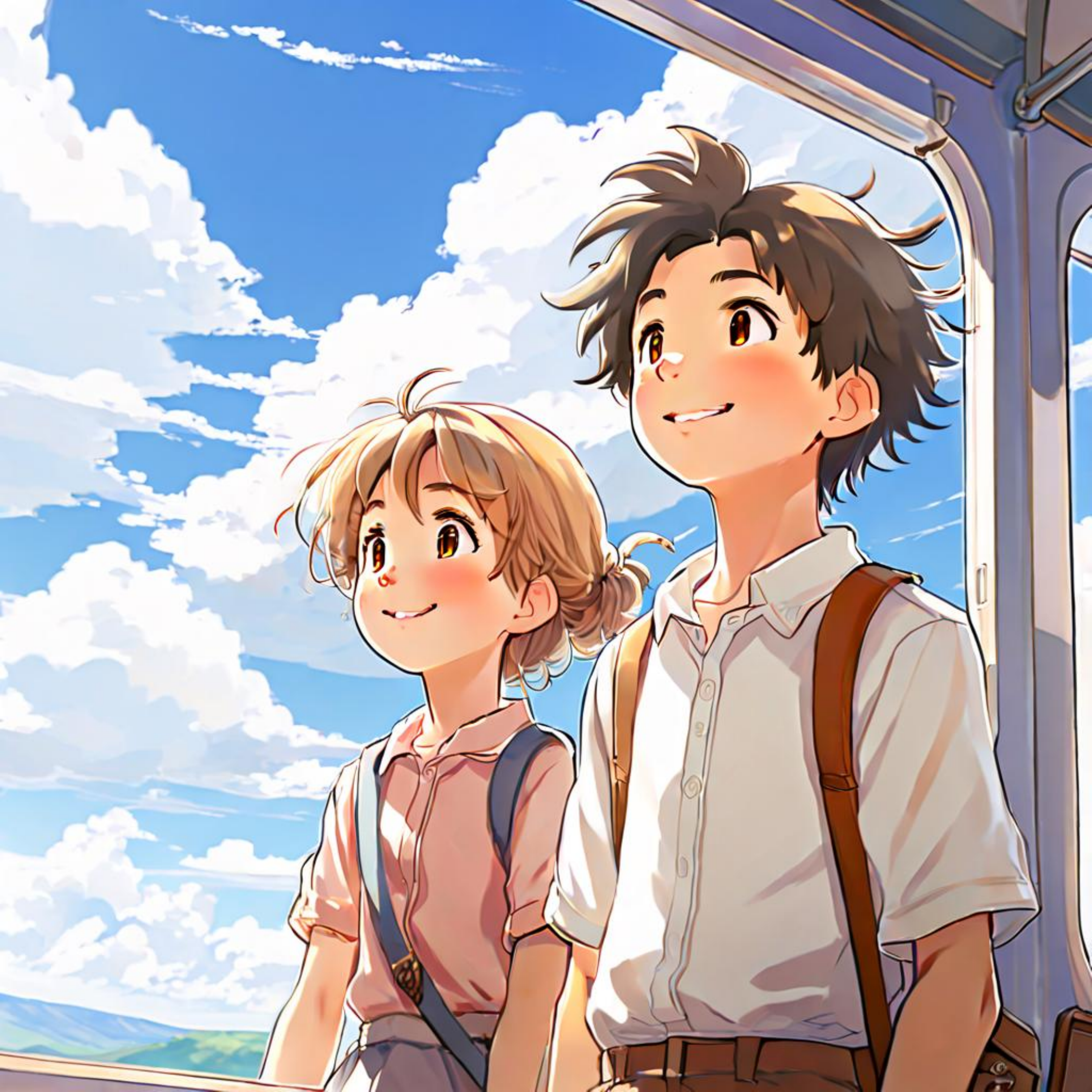}
\end{tabular}
&
\begin{tabular}{c}
    \includegraphics[width=0.13\linewidth]{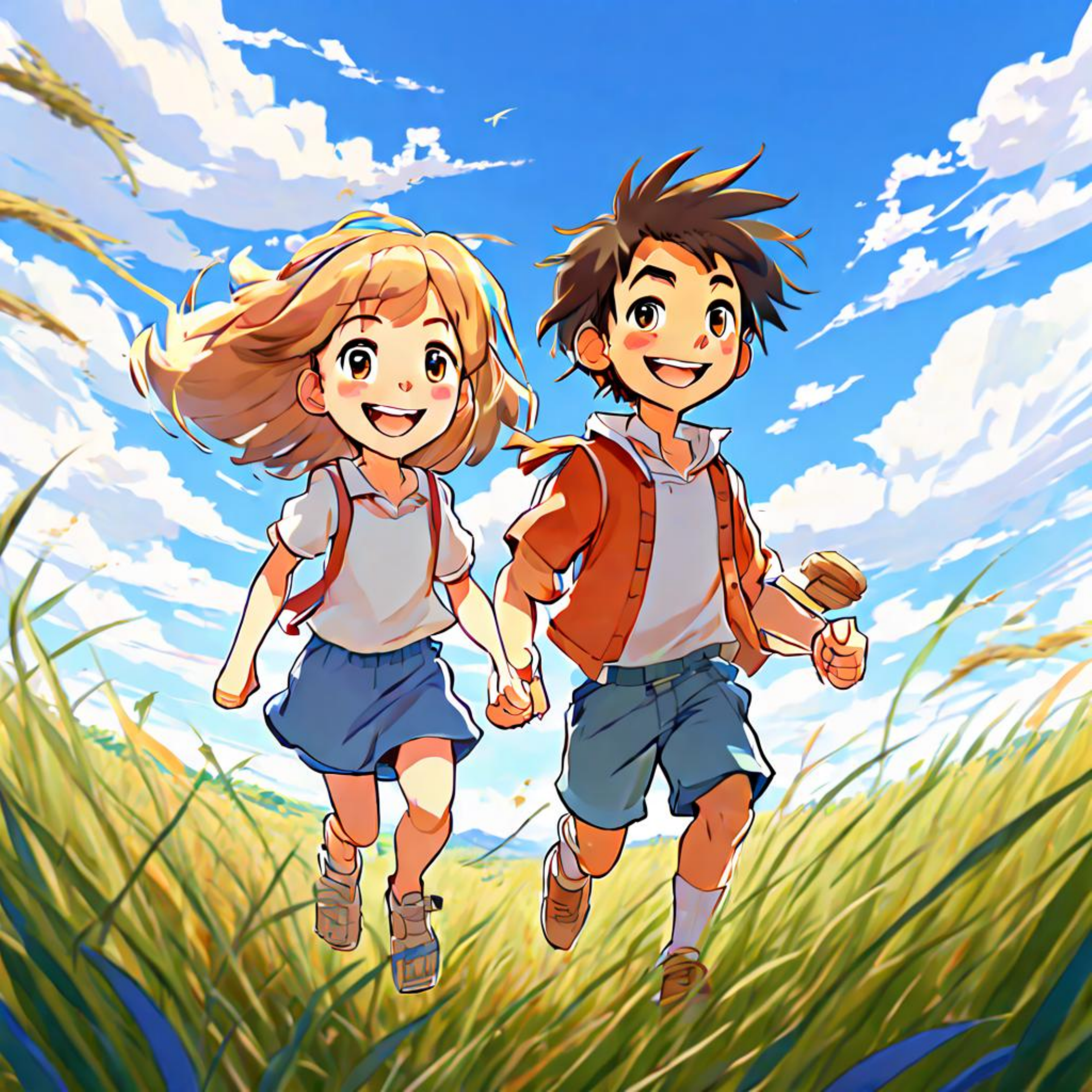}
    \includegraphics[width=0.13\linewidth]{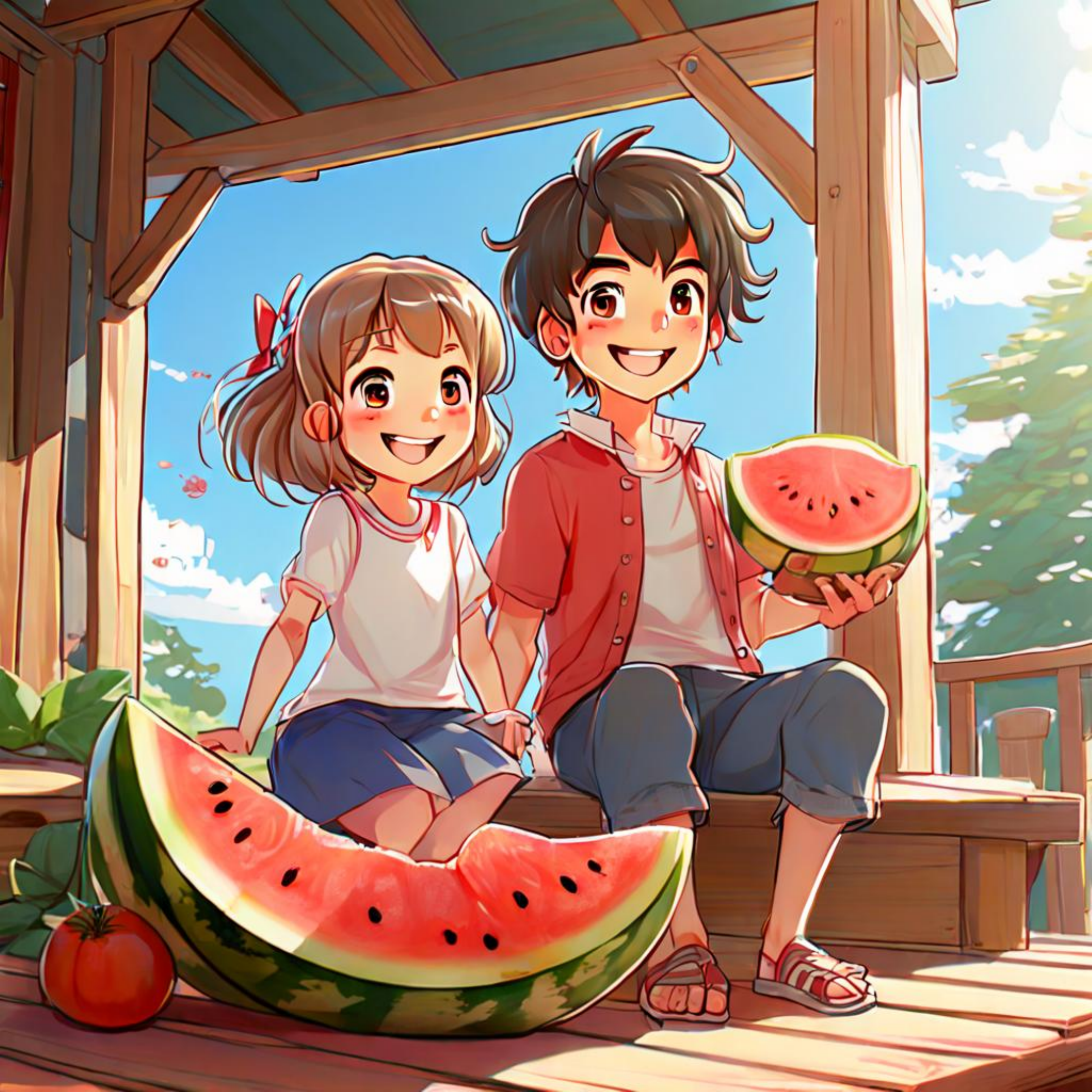}
    \includegraphics[width=0.13\linewidth]{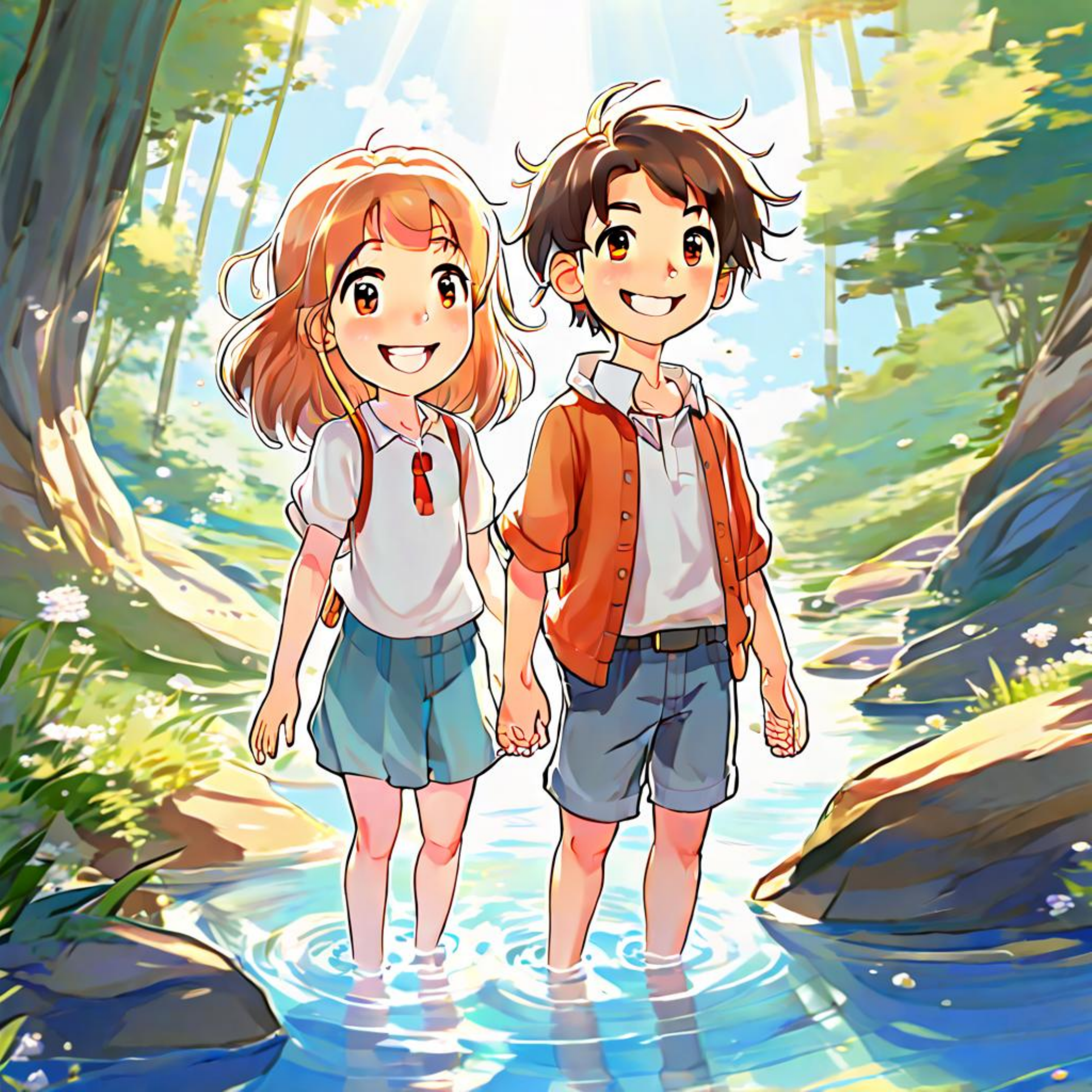}
    \includegraphics[width=0.13\linewidth]{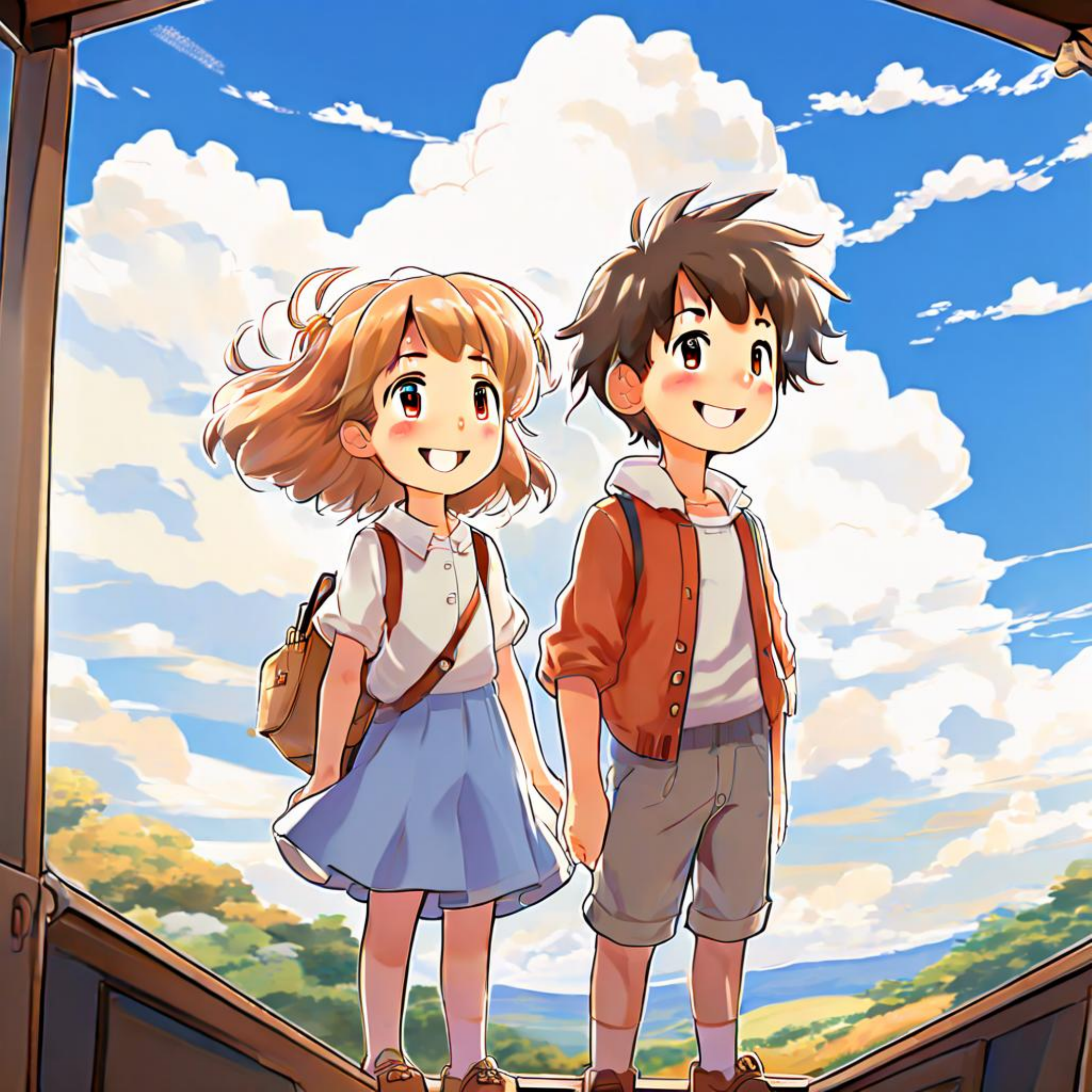}
\end{tabular}
\\
\multicolumn{2}{c}{\parbox{\linewidth}{\centering\small\textbf{Prompt 3:} \promptpart{idcolor}{Modern Chinese animation style,} \promptpart{action1color}{sitting,} \promptpart{action2color}{playing a melody,} \promptpart{action3color}{releasing floating lanterns,} \promptpart{action4color}{reading bamboo.}}} \\
% \begin{tabular}{c}
%     \includegraphics[width=0.13\linewidth]{figure_cvpr/fig1_5/row_2_col_1.pdf}
%     \includegraphics[width=0.13\linewidth]{figure_cvpr/fig1_5/row_2_col_2.pdf}
%     \includegraphics[width=0.13\linewidth]{figure_cvpr/fig1_5/row_2_col_3.pdf}
%     \includegraphics[width=0.13\linewidth]{figure_cvpr/fig1_5/row_2_col_5.pdf}
% \end{tabular}
% &
% \begin{tabular}{c}
%     \includegraphics[width=0.13\linewidth]{figure_cvpr/fig1_6/row_2_col_1.pdf}
%     \includegraphics[width=0.13\linewidth]{figure_cvpr/fig1_6/row_2_col_2.pdf}
%     \includegraphics[width=0.13\linewidth]{figure_cvpr/fig1_6/row_2_col_3.pdf}
%     \includegraphics[width=0.13\linewidth]{figure_cvpr/fig1_6/row_2_col_4.pdf}
% \end{tabular}
\begin{tabular}{c}
    \includegraphics[width=0.13\linewidth]{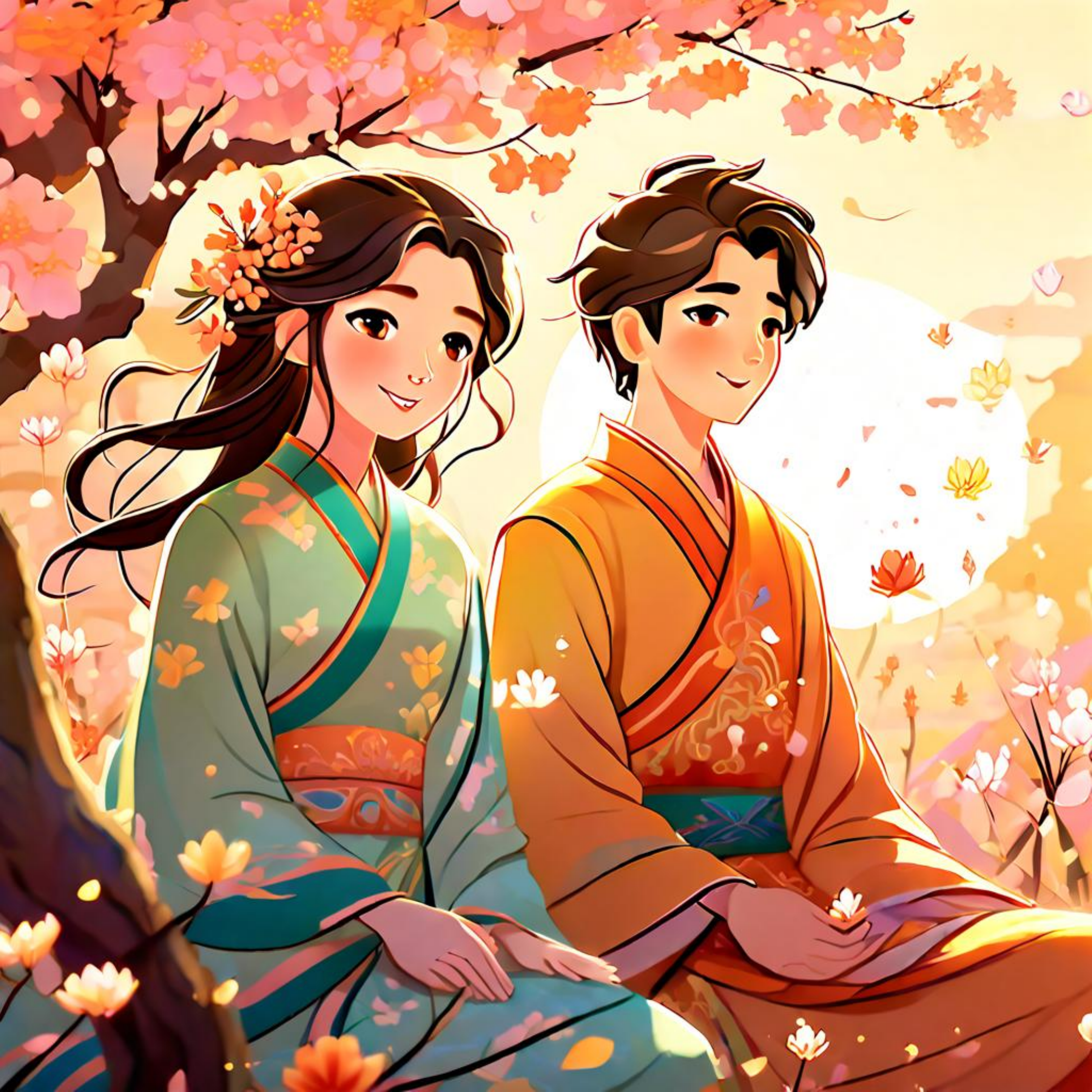}
    \includegraphics[width=0.13\linewidth]{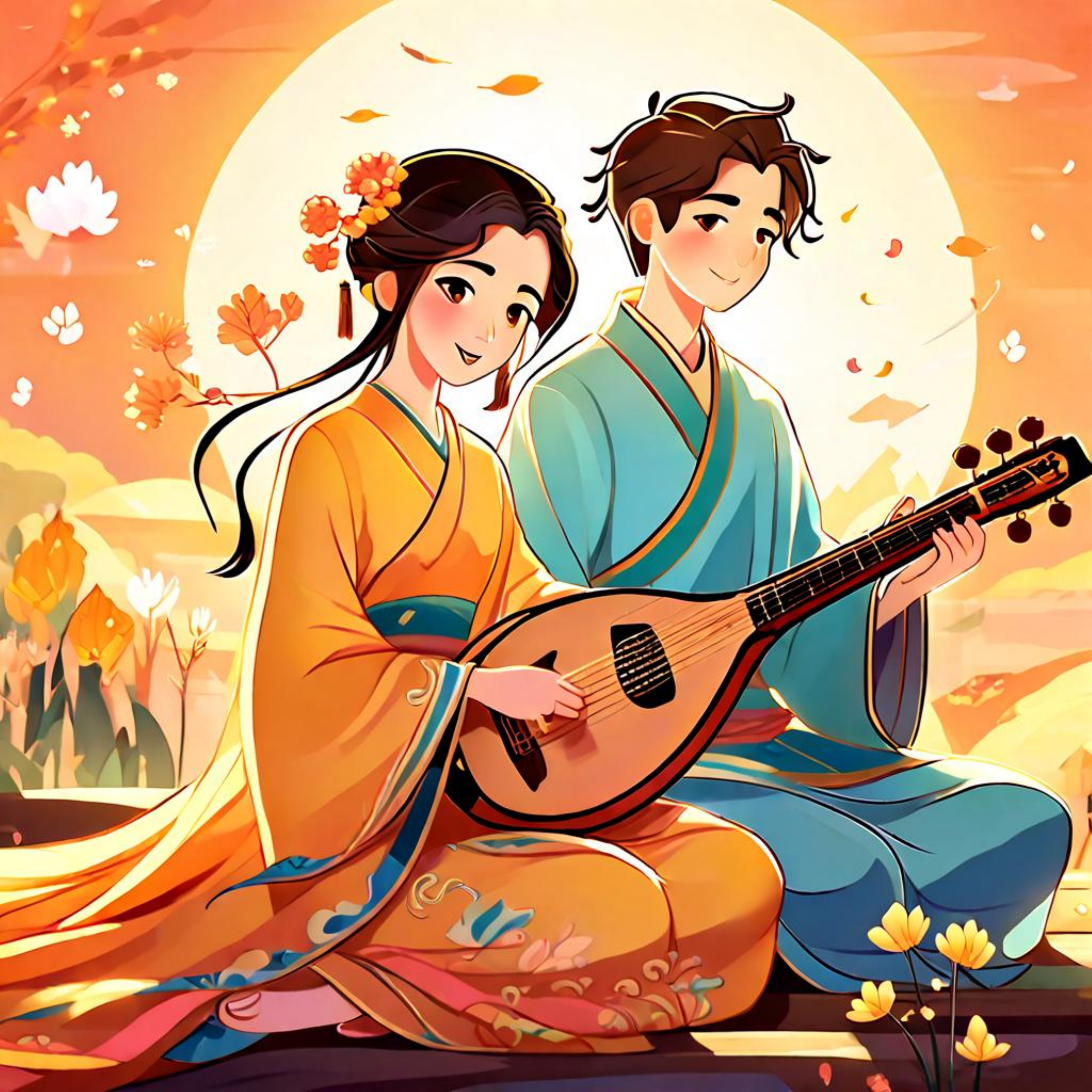}
    \includegraphics[width=0.13\linewidth]{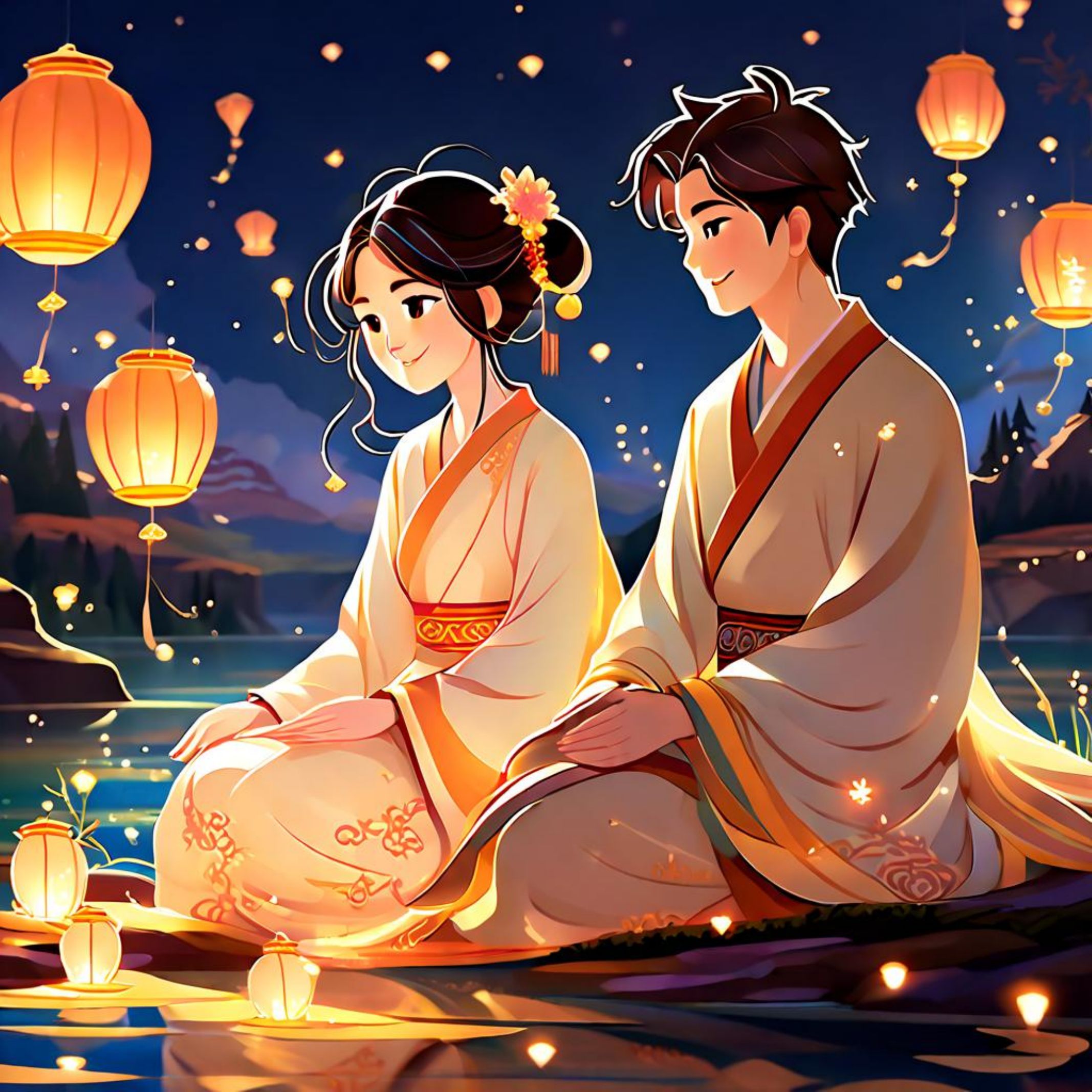}
    \includegraphics[width=0.13\linewidth]{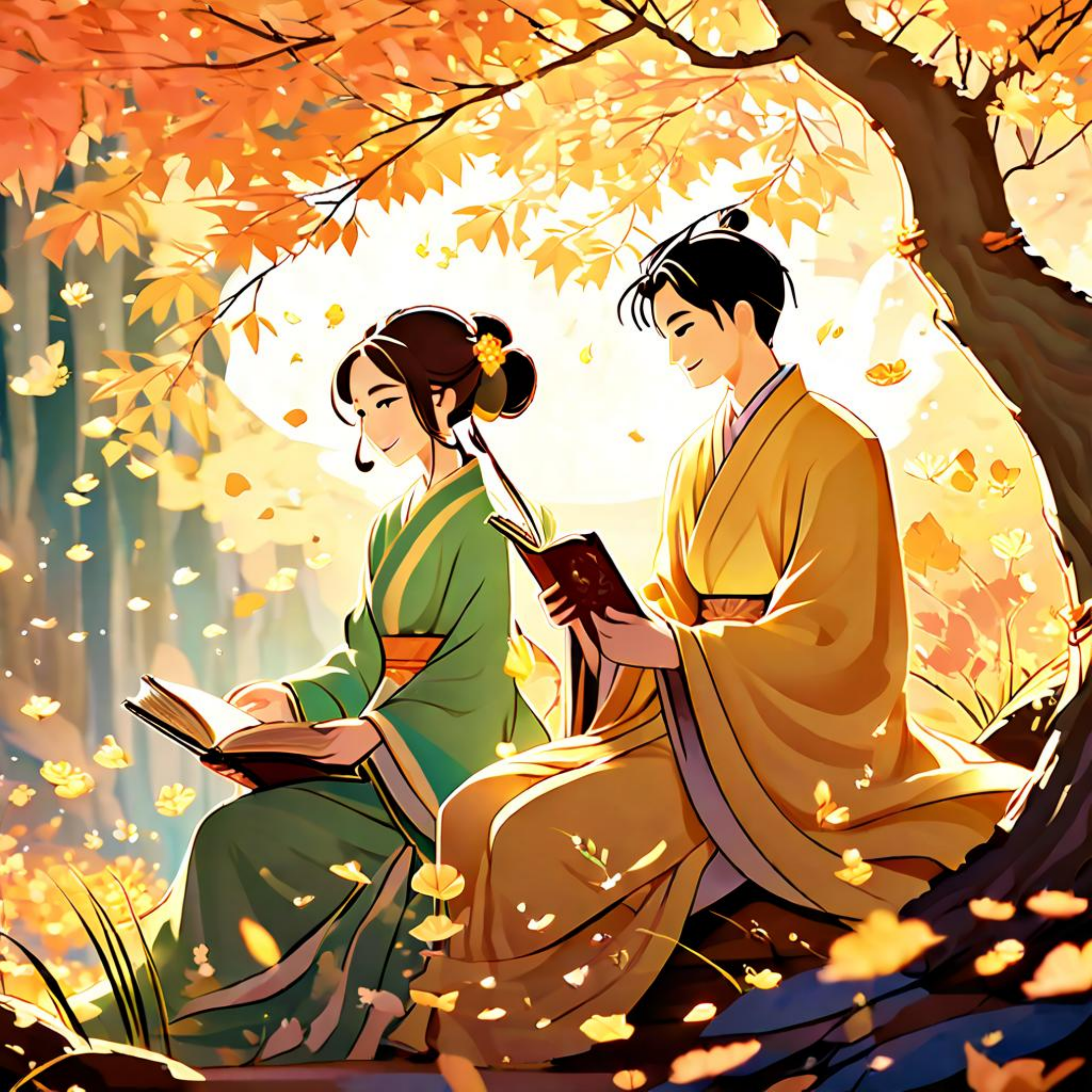}
\end{tabular}
&
\begin{tabular}{c}
    \includegraphics[width=0.13\linewidth]{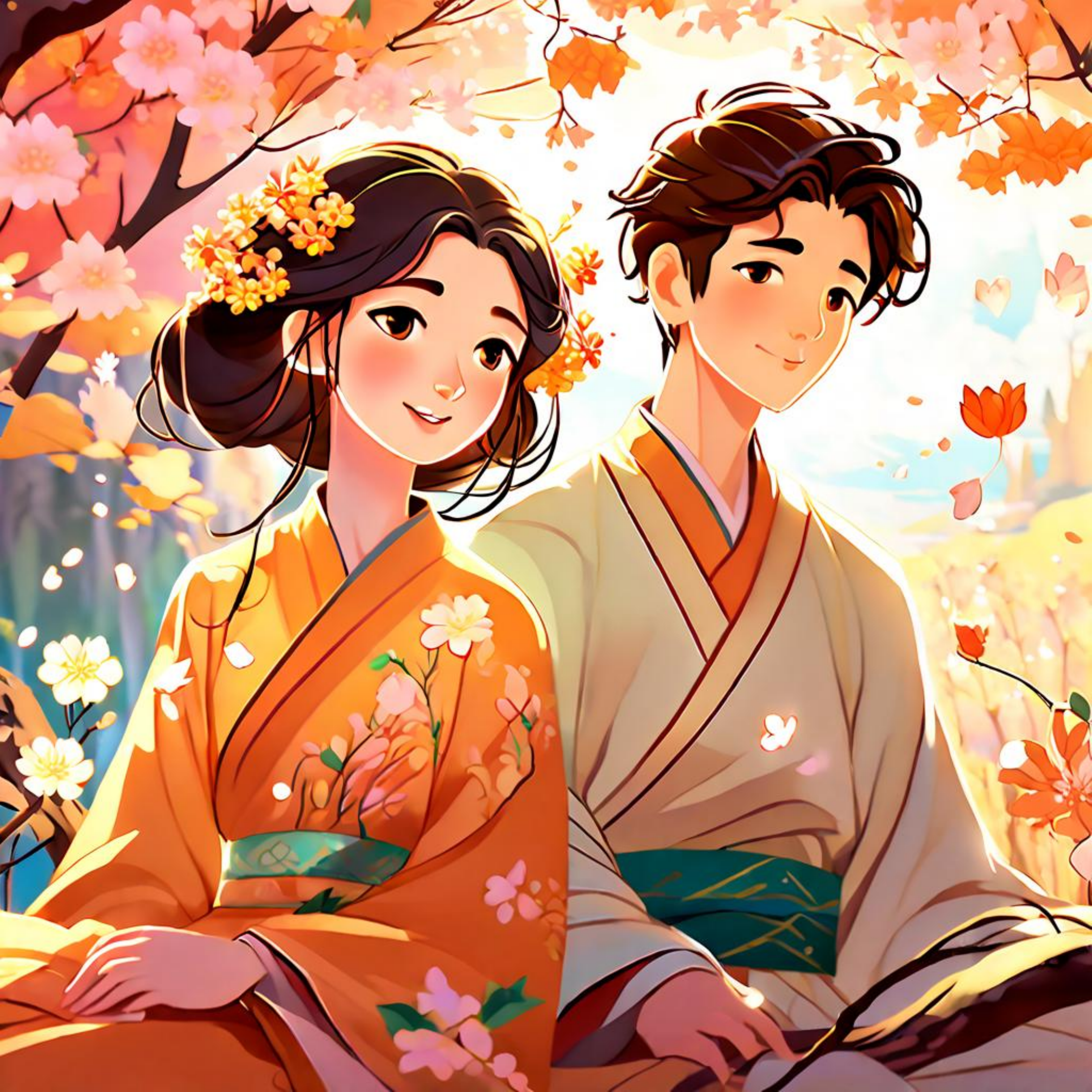}
    \includegraphics[width=0.13\linewidth]{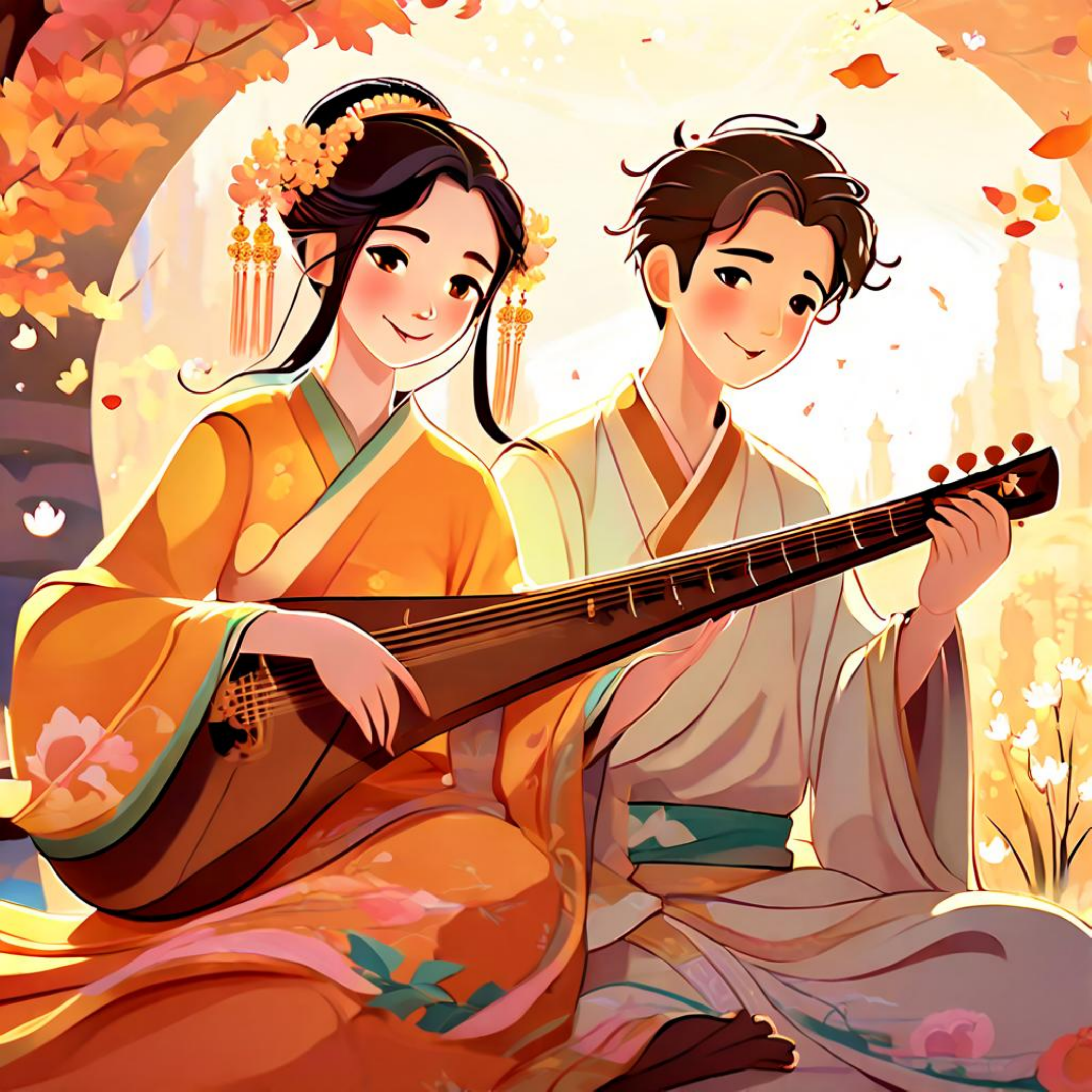}
    \includegraphics[width=0.13\linewidth]{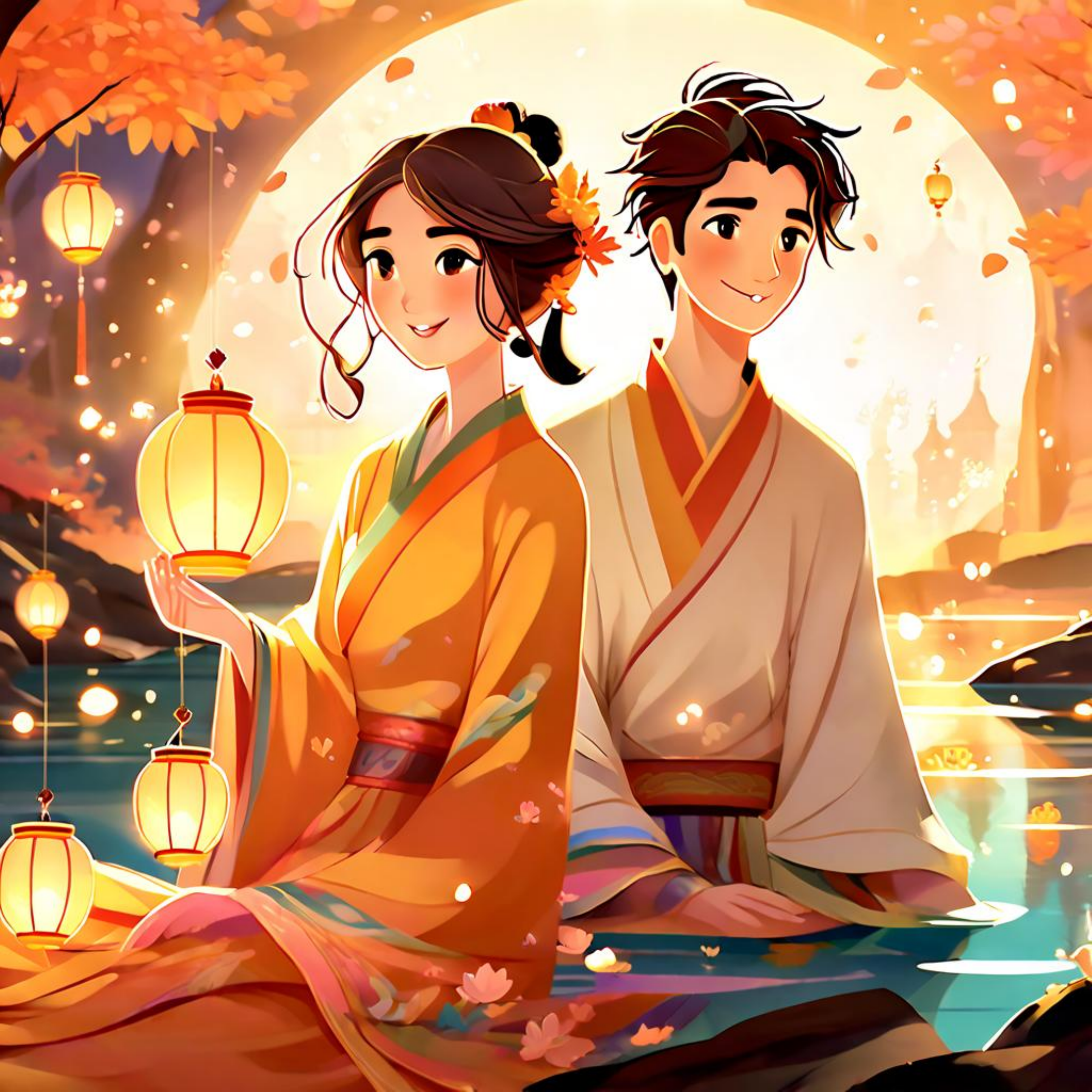}
    \includegraphics[width=0.13\linewidth]{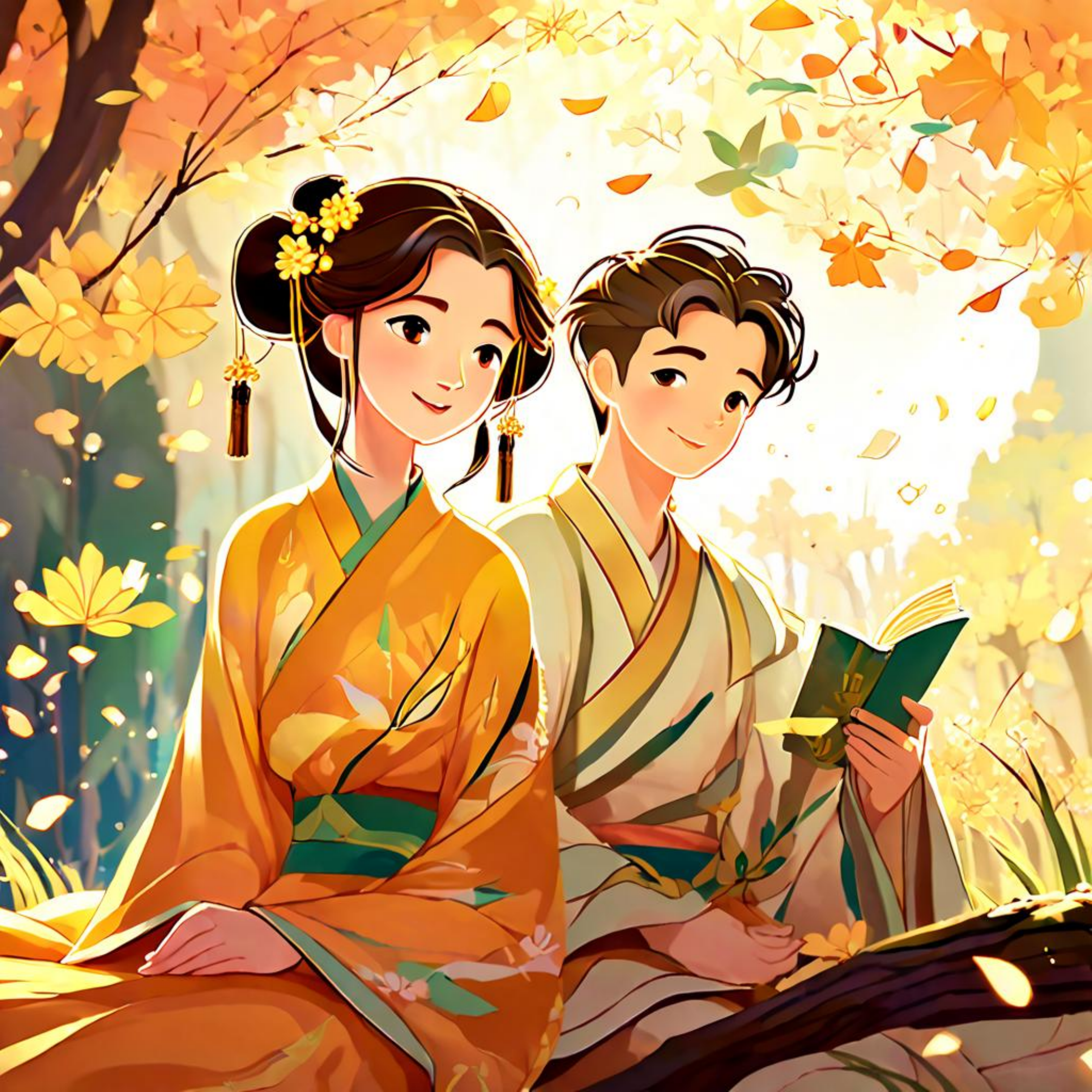}
\end{tabular}
\\
% $\alpha=0.1$ & $\alpha=0.9$\\
\end{tabular}
\captionof{figure}{Our framework tunes the balance between storytelling dynamism (left column) and robust character coherence (right column).}

\label{fig:fig1}
\end{center}
}]

\renewcommand{\thefootnote}{\fnsymbol{footnote}}
% \footnotetext[1]{This work was completed during the internship at SGIT AI Lab.}
\footnotetext[2]{Corresponding author.}

\begin{abstract}
While modern diffusion models excel at generating diverse single images, extending this to sequential generation reveals a fundamental challenge: balancing narrative dynamism with multi-character coherence. Existing methods often falter at this trade-off, leading to artifacts where characters lose their identity or the story stagnates. To resolve this critical tension, we introduce RealDiffusion, a unified framework designed to reconcile robust coherence with narrative dynamism. Heat diffusion serves as a dissipative prior that averages neighboring features along the sequence and removes high-frequency noise within the subject region. This suppresses attribute drift and stabilizes identity across frames. A region-aware stochastic process then introduces small perturbations that explore nearby modes and prevent collapse so the story maintains pose change and scene evolution. We thus introduce a lightweight, training-free Physics-informed Attention mechanism that injects controllable physical priors into the self-attention layers during inference. By modeling feature evolution as a configurable physical system, our method regularizes spatio-temporal relationships without suppressing intentional, prompt-driven changes. Extensive experiments demonstrate that RealDiffusion achieves substantial gains in character coherence while preserving narrative dynamism, outperforming state-of-the-art approaches. Code is available at \url{https://github.com/ShmilyQi-CN/RealDiffusion}.

\end{abstract}    
\section{Introduction}
\begin{figure*}[t]
    \centering
    \includegraphics[width=\textwidth]{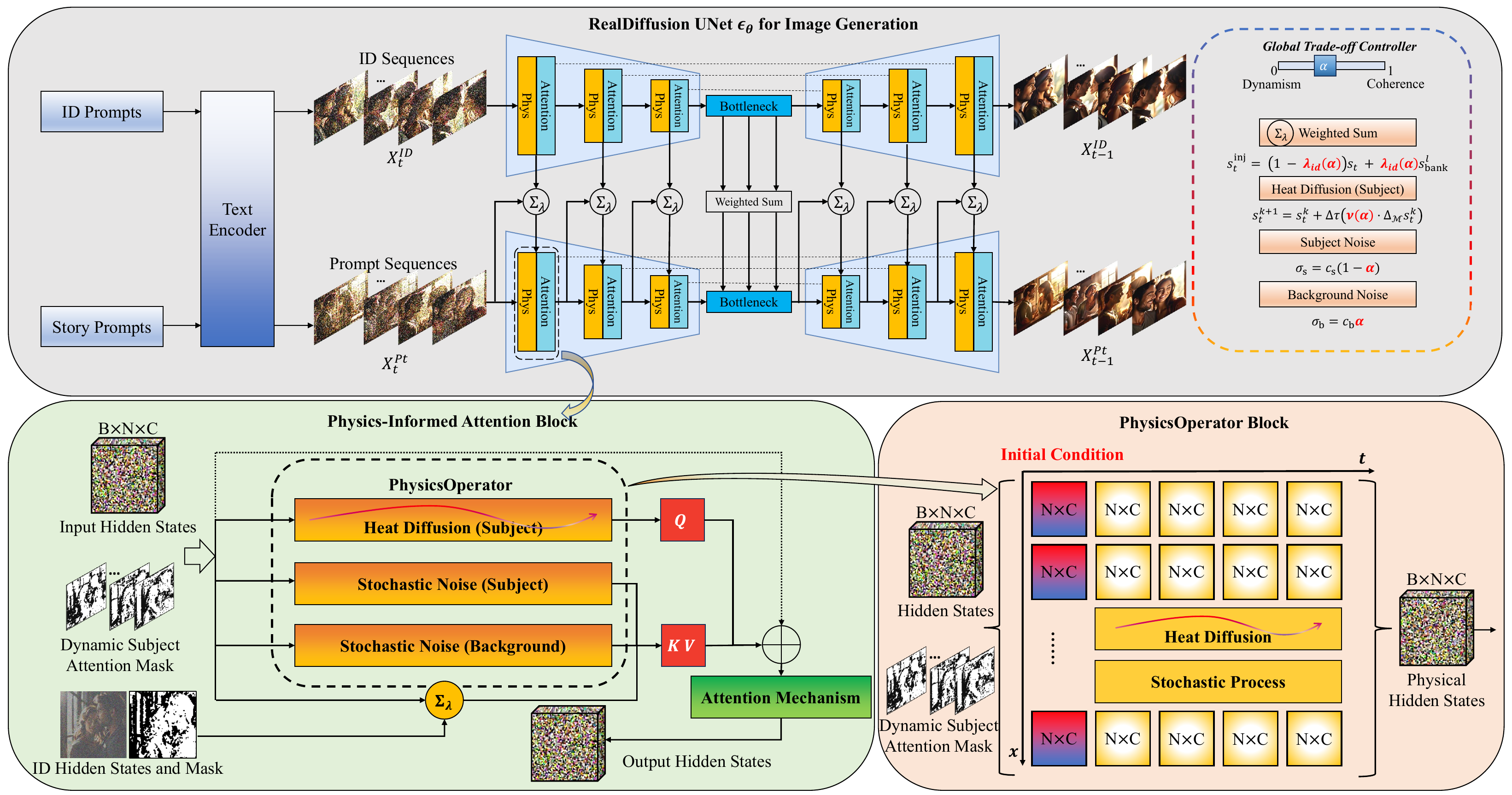}
    \caption{Overview of RealDiffusion. Our framework incorporates Physics-informed Attention into a U-Net, governed by $\alpha$. The module combines insulated heat diffusion for coherence with region-aware stochasticity for dynamism, guided by dynamic subject masks.}
    \label{fig:pipeline}
\end{figure*}

In recent years, diffusion models have rapidly evolved, demonstrating remarkable potential in content generation tasks including image synthesis~\cite{rombach2022high, saharia2022photorealistic, esser2024scaling}, 3D object generation~\cite{poole2022dreamfusion, lin2023magic3d}, and video production~\cite{ho2022imagen, singer2022make, blattmann2023align}. Leveraging large-scale pre-training on massive datasets, these models now outperform earlier approaches in producing high-fidelity and coherent generative content.

% However, maintaining narrative coherence for multiple characters remains a critical challenge.
Current approaches range from slow fine-tuning methods like DreamBooth~\cite{ruiz2023dreambooth} and Textual Inversion~\cite{gal2022image}, to zero-shot ID injection with encoders like IP-Adapter~\cite{ye2023ip}, PhotoMaker~\cite{li2024photomaker}, and InstantID~\cite{wang2024instantid}, but these sacrifice full-body coherence for facial fidelity. The most relevant line of work involves training-free methods that enforce coherence by manipulating internal features. This is achieved through shared self-attention~\cite{zhou2024storydiffusion, tewel2024training}, addressing attention biases~\cite{wang2025characonsist}, prompt engineering~\cite{liu2025one}, or novel sampling techniques~\cite{liconsistent}. Despite these advances, such methods frequently fail in intricate multi-character scenarios, resulting in artifacts like attribute swapping and illogical spatial layouts~\cite{gong2023talecrafter}. Our work distinguishes itself by specifically targeting these challenges in multi-character story generation. %

Diverse image generation is largely a solved problem. The critical frontier is maintaining coherence, especially in multi-character scenarios with their inherent trade-off between narrative dynamism and character coherence. We draw inspiration from physical systems to resolve this. Heat diffusion offers a natural smoothing kernel to enforce coherence, while a stochastic process provides a complementary kernel to preserve dynamism. Unlike established physics-informed methods like PINNs that require costly training~\cite{raissi2019physics}, our framework uses these principles as direct, training-free computational tools at inference.

We introduce RealDiffusion, a novel framework for mastering this trade-off, illustrated in Figure~\ref{fig:pipeline}. It features a training-free Physics-informed Attention mechanism guided by dynamic masks. Its core is our PhysicsOperator, which employs an insulated heat diffusion kernel and a stochastic kernel in tandem. A \textbf{Global Trade-off Controller} $\alpha$ governs the system by strengthening heat diffusion and foundational ID injection, empowering users to select their desired balance between coherence and dynamism.

Our contributions are summarized as follows:
\begin{itemize}
    \item \textbf{RealDiffusion:} a novel, training-free controllable framework that explicitly balances coherence–dynamism and excels on multi-character stories.
    \item \textbf{Physics-informed Attention:} a training-free, mask-guided self-attention mechanism; \emph{insulated heat diffusion} smooths features and suppresses attribute drift, while \emph{region-aware stochasticity} preserves evolution.
    \item \textbf{Results \& metrics:} state-of-the-art storybook quality and new sequence-level metrics for coherence and dynamism.
\end{itemize}

\begin{figure*}[t]
\centering
% Set the space between columns to a very small value to fit 8 images.
\setlength{\tabcolsep}{1pt} 
% Optional: You can slightly reduce the default font size within the figure for captions if needed.
\small 

\begin{tabular}{cc} % Outer table for vertical label + image row

%---------------------------------------------------
% ROW 1: The Generated Story Sequence
%---------------------------------------------------
\begin{tabular}{c}
    \vspace{1mm}\rotatebox[origin=c]{90}{\textbf{Story Sequence}}
\end{tabular}
&
% A single-row table containing 8 images.
% The width is set to 0.118\linewidth. 8 * 0.118 = 0.944, leaving space for margins.
\begin{tabular}{cccccccc}
    \includegraphics[width=0.118\linewidth]{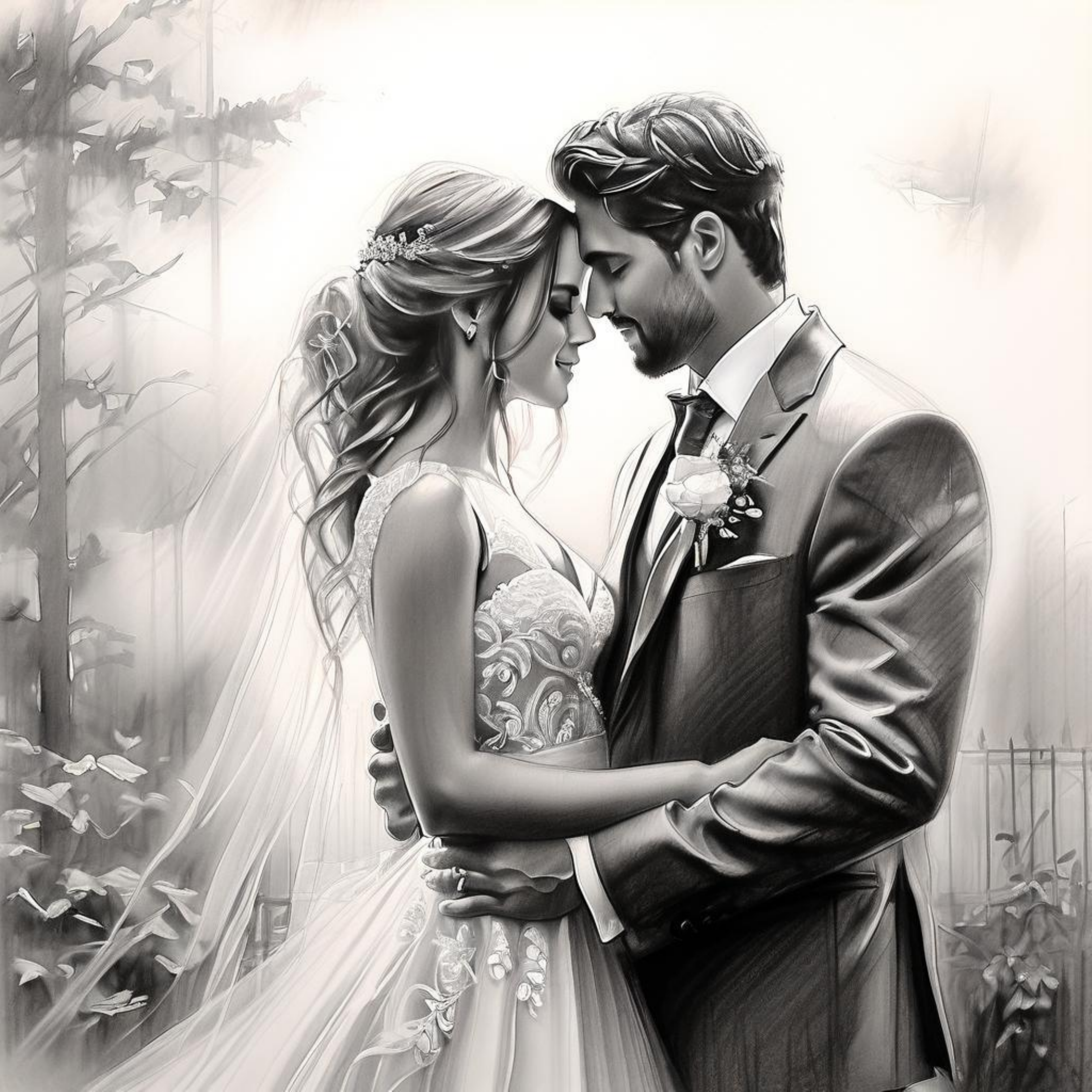} &
    \includegraphics[width=0.118\linewidth]{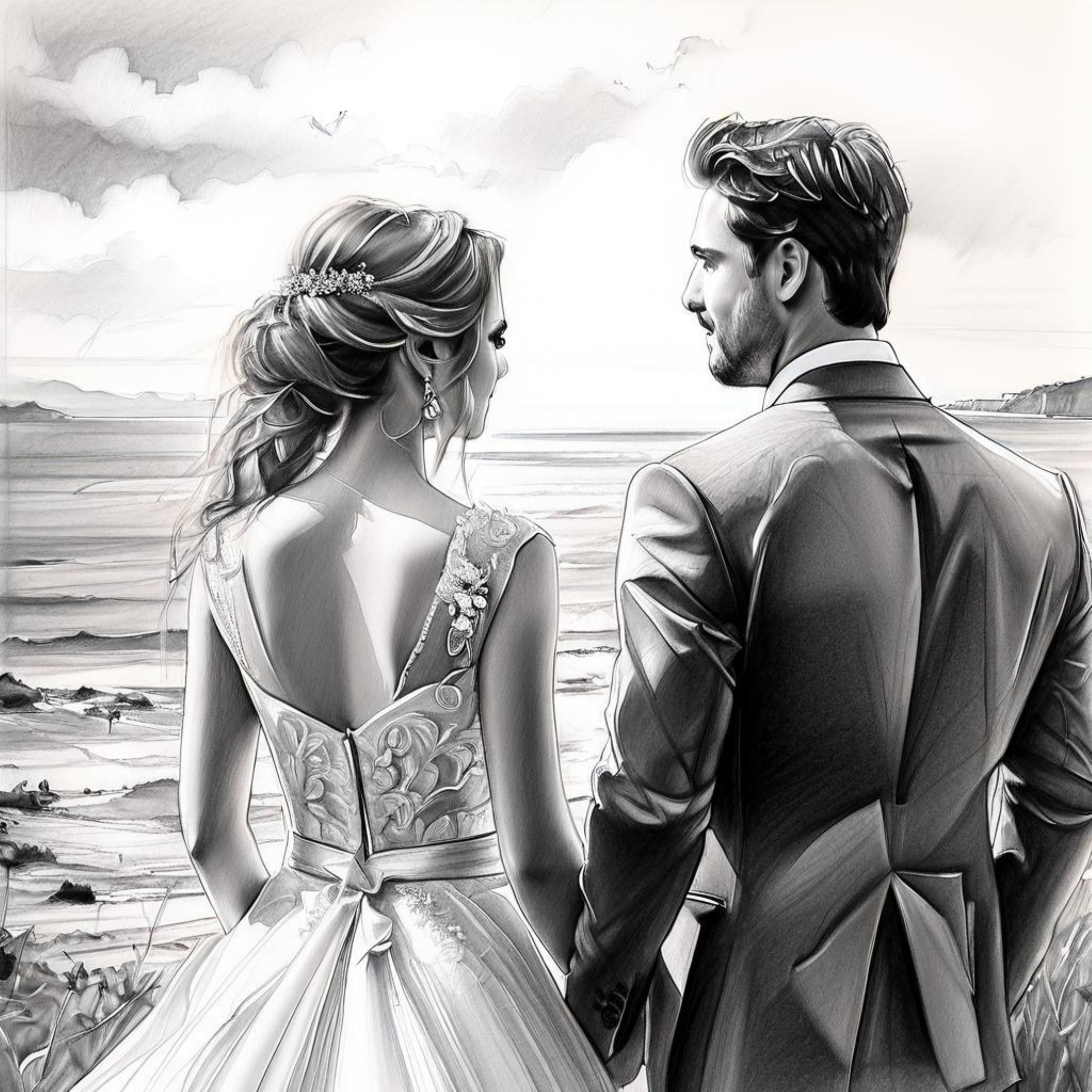} &
    \includegraphics[width=0.118\linewidth]{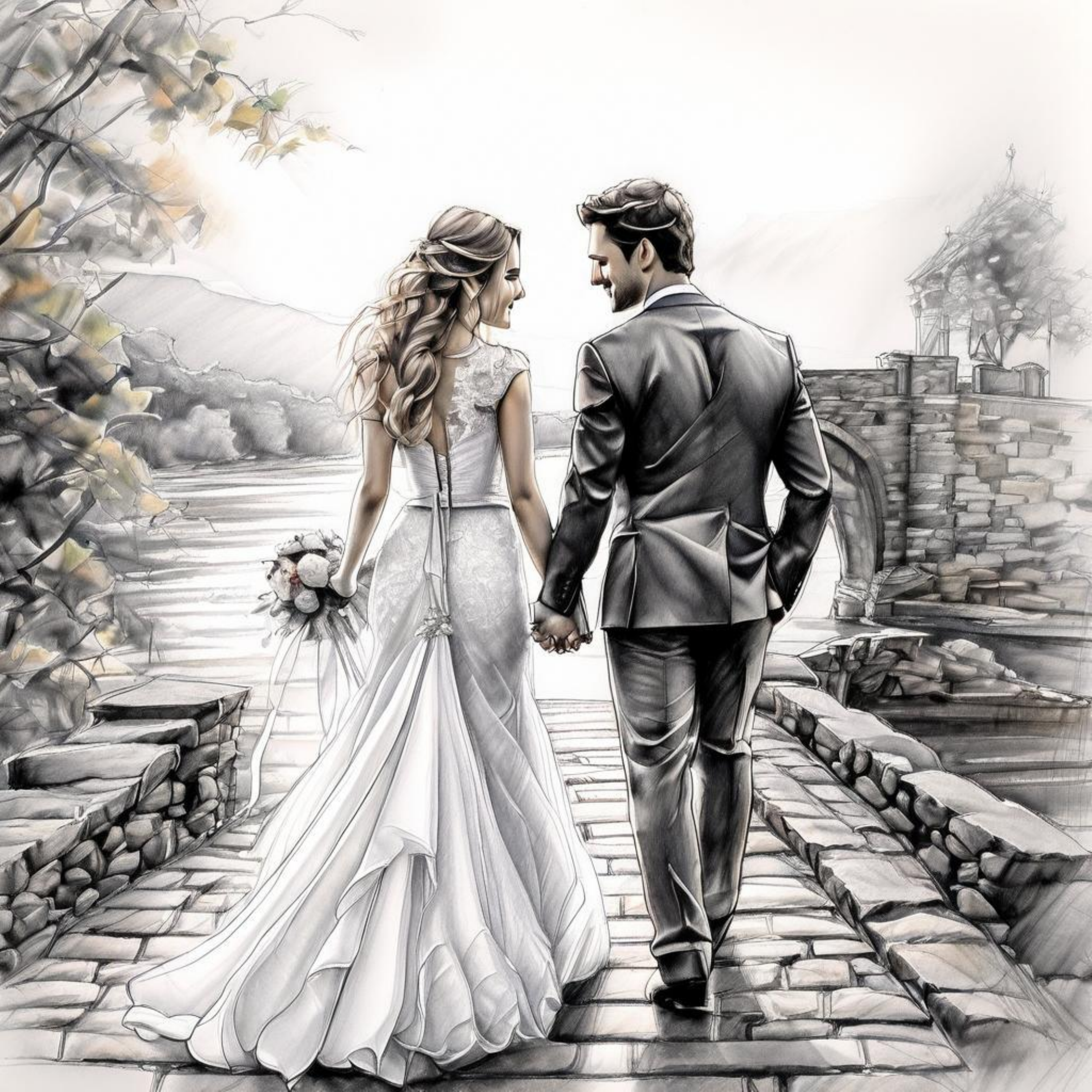} &
    \includegraphics[width=0.118\linewidth]{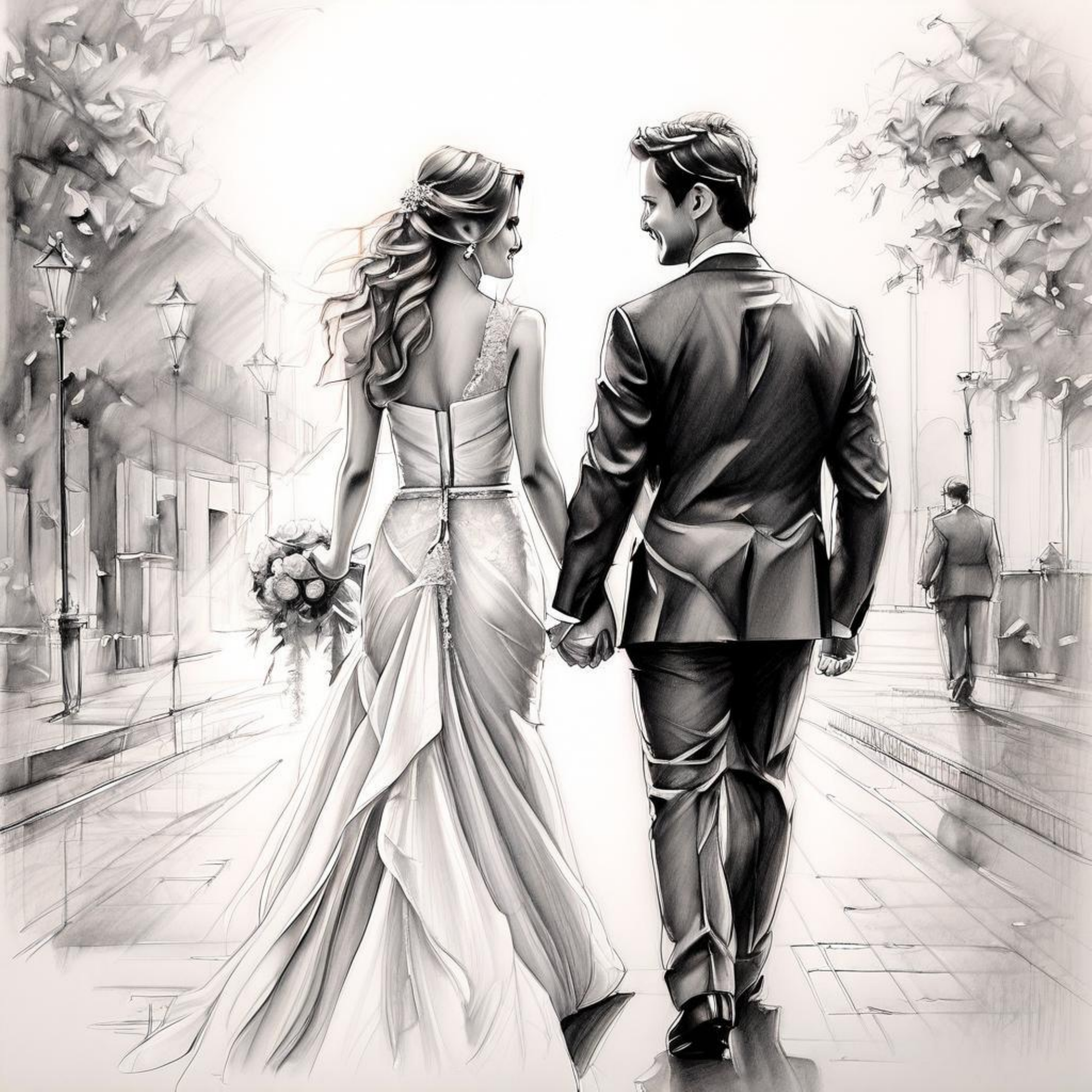} &
    \includegraphics[width=0.118\linewidth]{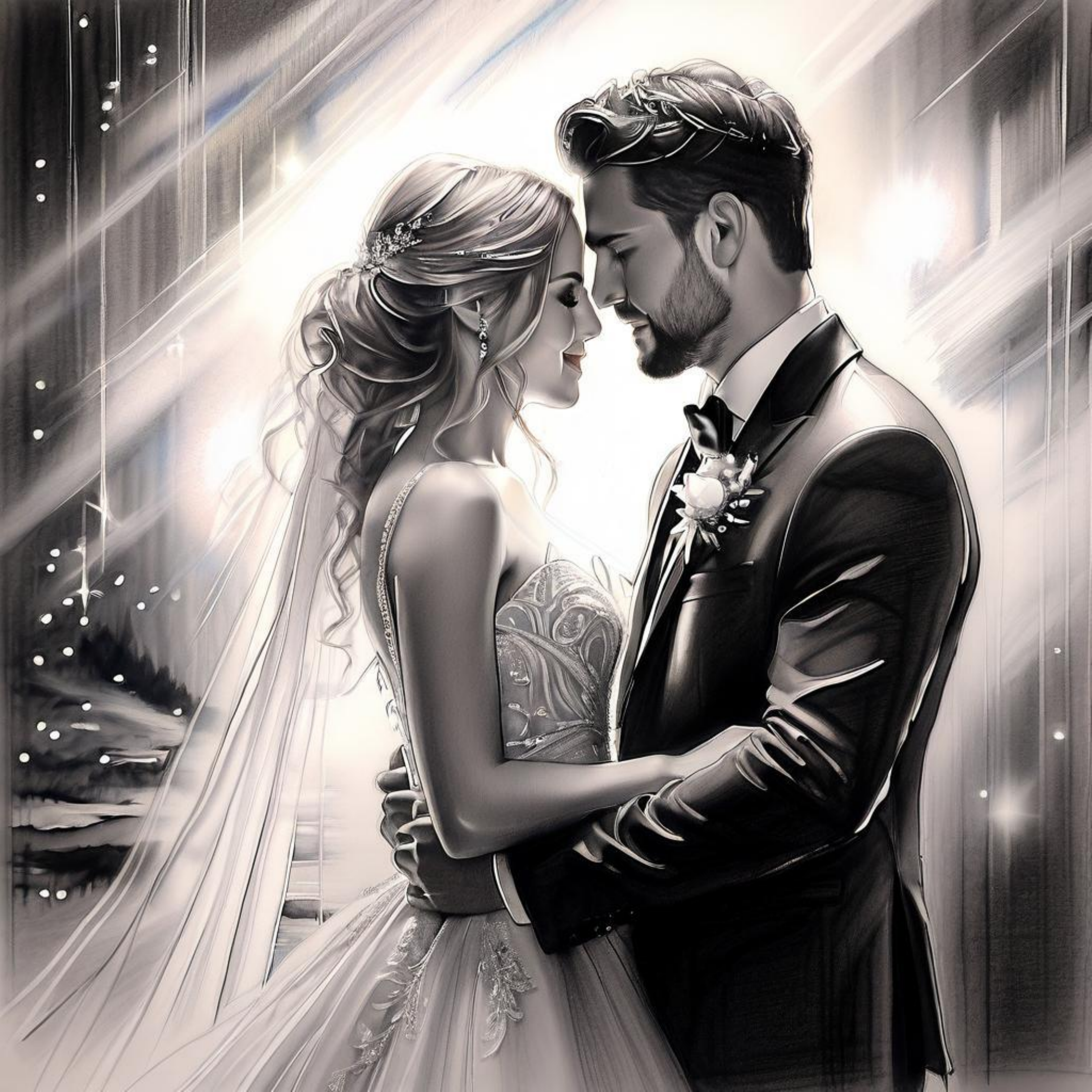} &
    \includegraphics[width=0.118\linewidth]{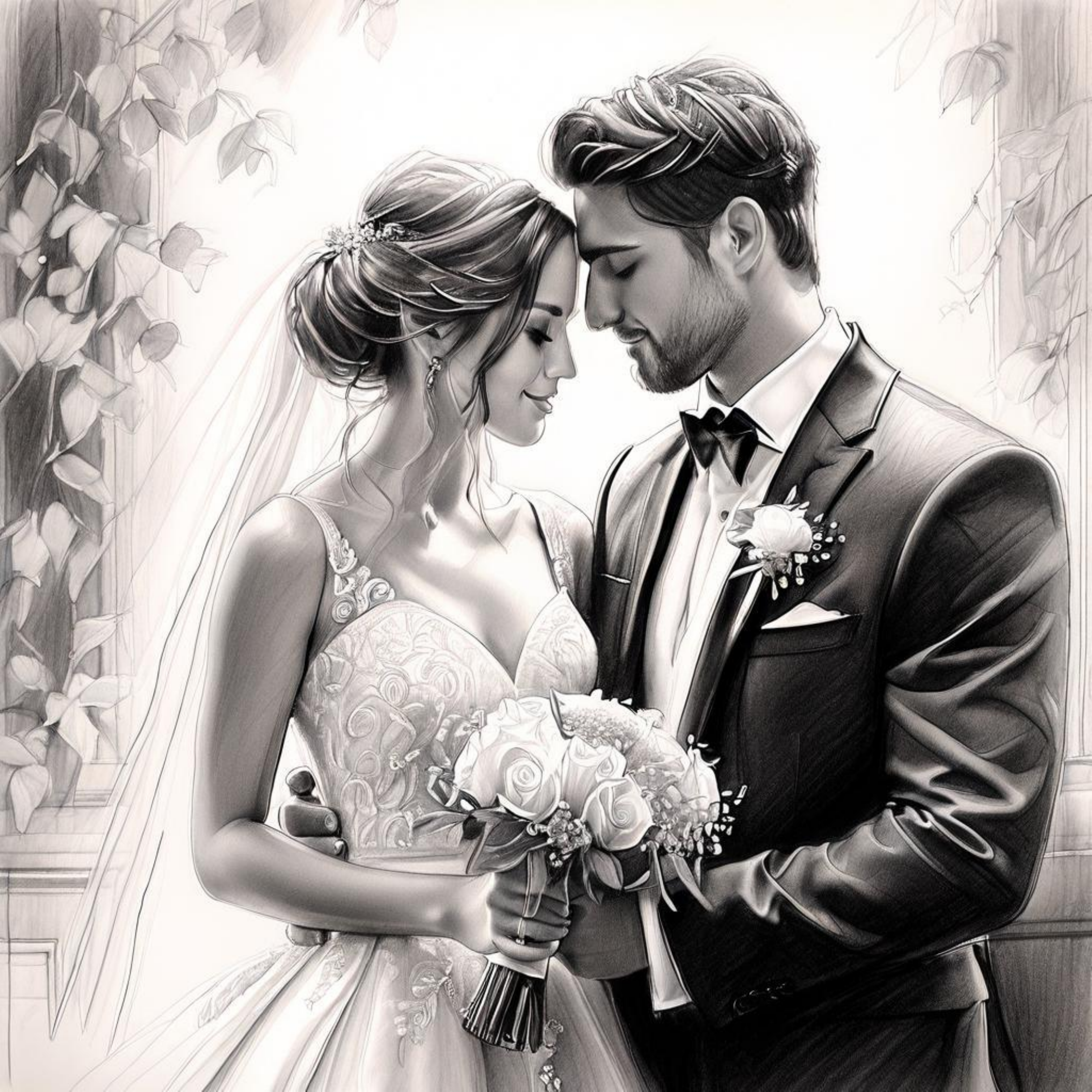} &
    \includegraphics[width=0.118\linewidth]{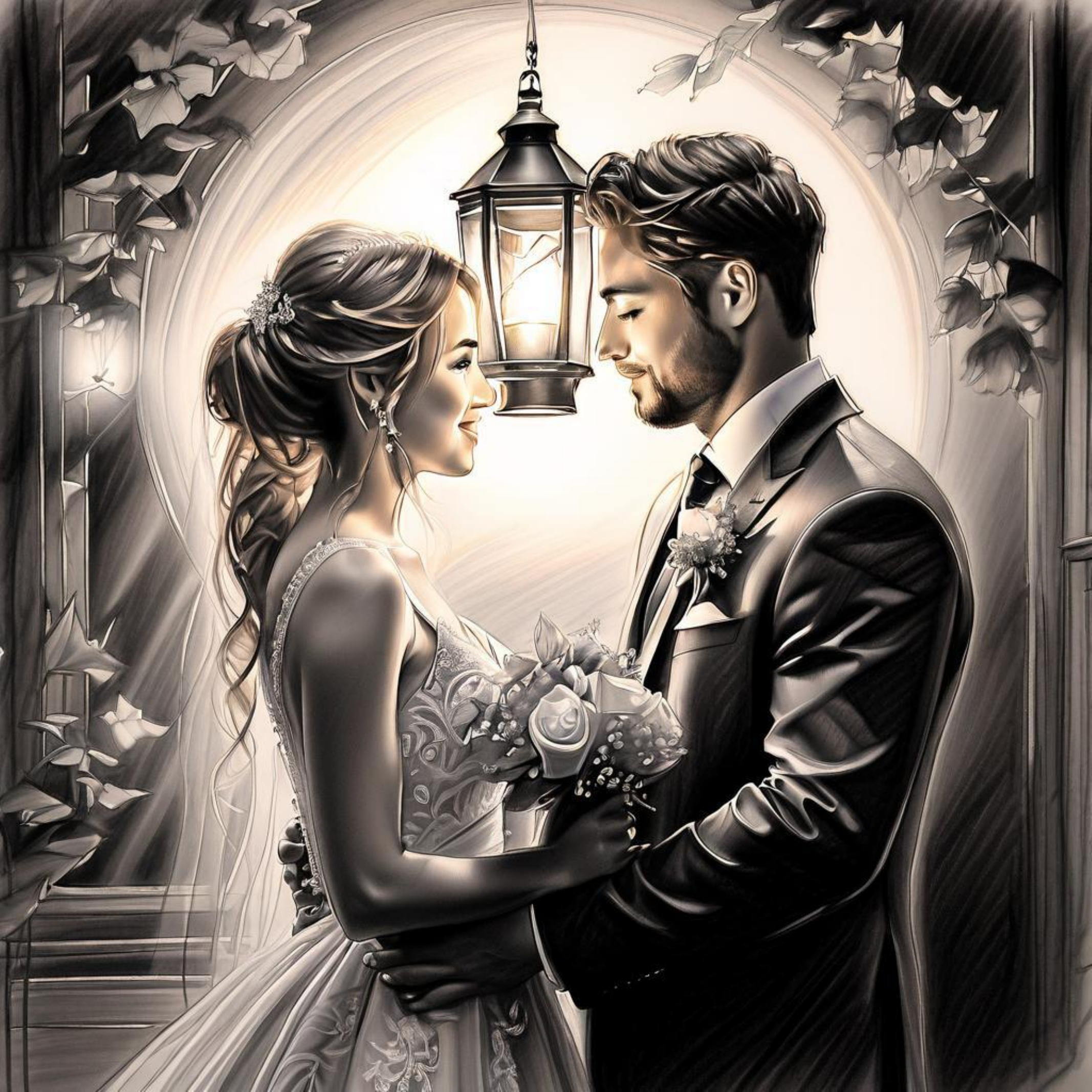} &
    \includegraphics[width=0.118\linewidth]{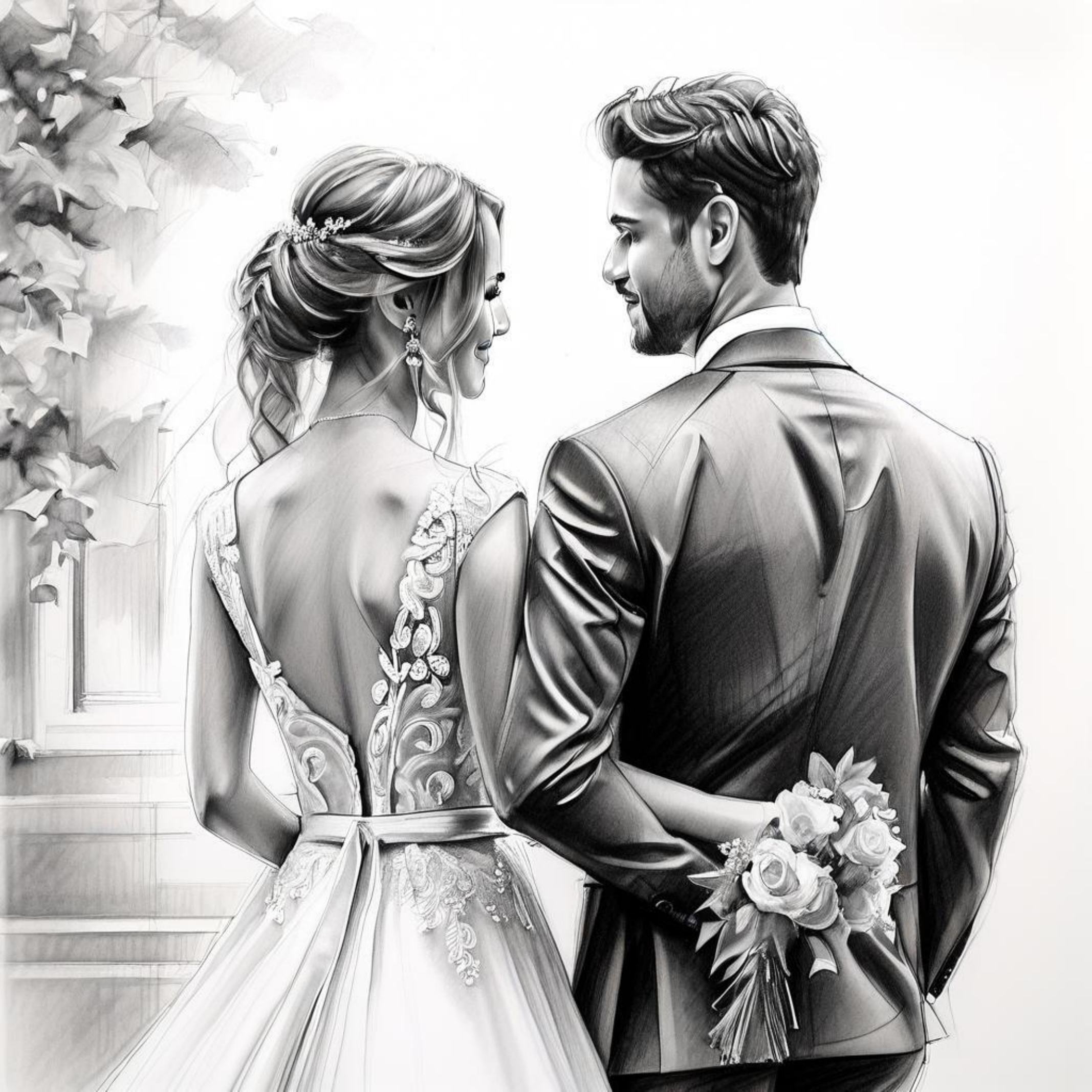}
\end{tabular}
\\ % End of the first main row

% Add a larger vertical space between the story and the masks.
\addlinespace[0mm] 

%---------------------------------------------------
% ROW 2: The Corresponding Dynamic Masks
%---------------------------------------------------
\begin{tabular}{c}
    \vspace{1mm}\rotatebox[origin=c]{90}{\textbf{Dynamic Masks}}
\end{tabular}
&
\begin{tabular}{cccccccc}
    \includegraphics[width=0.118\linewidth]{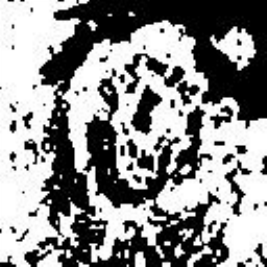} &
    \includegraphics[width=0.118\linewidth]{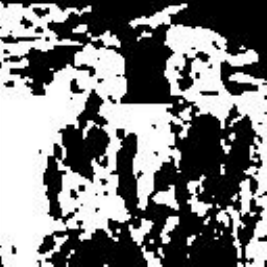} &
    \includegraphics[width=0.118\linewidth]{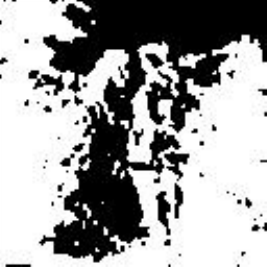} &
    \includegraphics[width=0.118\linewidth]{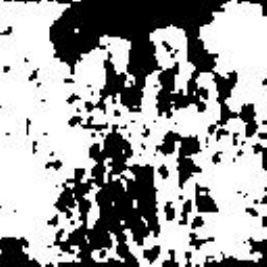} &
    \includegraphics[width=0.118\linewidth]{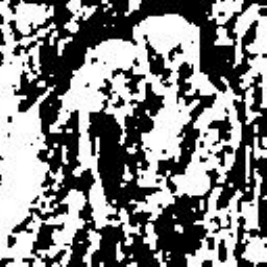} &
    \includegraphics[width=0.118\linewidth]{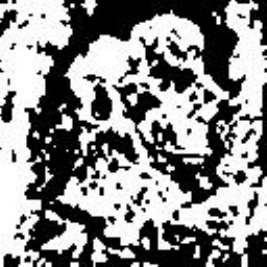} &
    \includegraphics[width=0.118\linewidth]{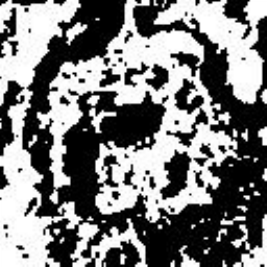} &
    \includegraphics[width=0.118\linewidth]{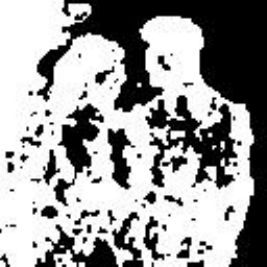}
\end{tabular}
\\ % End of the second main row

\end{tabular}
\caption{
    An 8-frame story sequence with its dynamic masks.
    Prompt:
    \promptpart{idcolor}{A pencil sketch of, bride and groom;}
    \promptpart{action1color}{in morning fog,}
    \promptpart{action2color}{looking at horizon,}
    \promptpart{action3color}{on a stone bridge,}
    \promptpart{action4color}{walking side-by-side,}
    \promptpart{action5color}{under aurora,}
    \promptpart{action6color}{sharing a quiet moment,}
    \promptpart{action7color}{under lantern light,}
    \promptpart{action8color}{standing back-to-back.}
}
\label{fig:masks}
\end{figure*}
\section{Related work}

\subsection{Diffusion Models}
Diffusion models~\cite{sohl2015deep} have become the dominant paradigm in generative modeling. Their formulation as Denoising Diffusion Probabilistic Models (DDPMs)~\cite{ho2020denoising} and their unification with score-based models via stochastic differential equations (SDEs)~\cite{song2020score} provided a solid theoretical foundation. The introduction of Latent Diffusion Models (LDMs)~\cite{rombach2022high} was a pivotal efficiency breakthrough, enabling high-resolution synthesis by operating in a compressed latent space. Architecturally, while early models relied on U-Net~\cite{ronneberger2015u}, recent state-of-the-art models increasingly adopt Transformer-based backbones~\cite{peebles2023scalable}, showcasing superior scaling properties. Controllability is a key strength, achieved through techniques like Classifier-Free Guidance~\cite{ho2022classifier} for better text alignment and external frameworks like ControlNet~\cite{zhang2023adding} for explicit spatial guidance.

\subsection{Storybook Generation}
Extending single-image diffusion to multi-frame storybooks foregrounds the challenge of temporal coherence, especially in multi-character settings. Prior work spans three main directions. (i) \emph{Personalization-based} methods embed subject identity via test-time or few-shot adaptation—DreamBooth~\cite{ruiz2023dreambooth}, Textual Inversion~\cite{gal2022image}, and more efficient LoRA variants~\cite{hu2022lora, ryu2023low} or multi-concept composition~\cite{kumari2023multi}—but require per-subject training and risk drift/overfitting. (ii) \emph{Encoder-based ID injection} conditions generation on image-encoder features, with IP-Adapter~\cite{ye2023ip} and successors such as PhotoMaker~\cite{li2024photomaker} and InstantID~\cite{wang2024instantid} improving identity fidelity yet sometimes sacrificing prompt faithfulness or full-body coherence. (iii) \emph{Training-free} approaches intervene at inference: attention manipulation (StoryDiffusion~\cite{zhou2024storydiffusion}, ConsiStory~\cite{tewel2024training}, CharaConsist~\cite{wang2025characonsist}) and multi-subject composition (FastComposer~\cite{xiao2025fastcomposer}); prompt engineering that concatenates all frame descriptions with token reweighting (OnePromptOneStory~\cite{liu2025one}); and novel samplers such as Zigzag Sampling~\cite{liconsistent}. Despite progress, complex scenes still exhibit artifacts (e.g., attribute swapping or illogical layouts~\cite{gong2023talecrafter}) and most training-free methods lack an explicit, user-controllable knob to balance narrative dynamism and character coherence—a gap our RealDiffusion addresses.

\subsection{Physics-informed Neural Networks (PINNs)}
Physics-informed Neural Networks (PINNs) represent a class of deep learning models designed to solve partial differential equations (PDEs)~\cite{raissi2019physics}. They achieve this by embedding the physical laws described by the PDEs directly into the neural network's loss function. This ensures the learned solution adheres to the specified physical dynamics, boundary conditions, and initial states. The standard PINN paradigm, however, requires retraining for any change in PDE parameters, a limitation discussed in recent surveys~\cite{cuomo2022scientific, torres2025adaptive}. To overcome this, operator learning and meta-learning approaches have been proposed to learn a mapping from PDE parameters to the solution function, allowing for near-instantaneous inference after an initial, costly training phase~\cite{desai2021one, psaros2022meta}. Our work draws conceptual inspiration from this field, not by training a network to solve a PDE, but by using the mathematical principles of physical processes (specifically, heat diffusion) as a computational tool to regularize the generative process at inference time.

\begin{table*}[htbp]
\centering
\caption{Physical priors for temporal regularization. The frame index $t$ is treated as a spatial dimension, regularized over a virtual time $\tau$.}

\label{tab:pde_ablation_rules}
\begin{tabular}{l|l|l}
\hline
\textbf{Physical Law} & \textbf{Conceptual Basis / PDE Form} & \textbf{Discrete Update Rule (Variational Form)} \\ \hline
\rule{0pt}{3.5ex}
Ori (Baseline) & 
N/A (No physical prior) & 
$s_t^{k+1} = s_t^k$ \\
\rule{0pt}{3.5ex}
Burgers' Equation & 
$\frac{\partial s}{\partial \tau} + s \frac{\partial s}{\partial t} = 0$ & 
$s_t^{k+1} = s_t^k - \Delta\tau \cdot s_t^k \frac{s_{t+1}^k - s_{t-1}^k}{2}$ \\
\rule{0pt}{3.5ex}
Wave Equation &
$\frac{\partial^2 s}{\partial \tau^2} = c^2 \frac{\partial^2 s}{\partial t^2}$ &
$s_t^{k+1} = 2s_t^k - s_t^{k-1} + C (s_{t+1}^k - 2s_t^k + s_{t-1}^k)$ \\
\rule{0pt}{3.5ex}
Conservation Law & 
$\frac{\partial s}{\partial \tau} + \nabla_t \cdot \mathbf{F}(s) = 0$ & 
$s_t^{k+1} = s_t^k - \Delta\tau \frac{F(s_{t+1}^k) - F(s_{t-1}^k)}{2}$ \\
\rule{0pt}{3.5ex}
Elasticity & 
$\rho \frac{\partial^2 s}{\partial \tau^2} = \nabla_t \cdot \sigma$ & 
$s_t^{k+1} = 2s_t^k - s_t^{k-1} + C_e (s_{t+1}^k - 2s_t^k + s_{t-1}^k)$ \\ % <-- The key change is here
\rule{0pt}{3.5ex}
\textbf{Ours (Heat Diff.)} & 
$\frac{\partial s}{\partial \tau} = \nu(\alpha) \nabla_t^2 s$ & 
$s_t^{k+1} = s_t^k + \Delta\tau \cdot \nu(\alpha) (s_{t+1}^k - 2s_t^k + s_{t-1}^k)$ \\ \hline
\end{tabular}
\end{table*}

% --- Controller mappings (symbols unified with text) ---
% \begin{equation}
% \label{eq:alpha_panel}
% \boxed{
% \begin{aligned}
% \nu(\alpha) &= \nu_{\text{base}}\,\alpha,\\
% \lambda_{\mathrm{id}}(\alpha) &= \lambda_{\max}\,\alpha,\\
% \sigma_{s}(\alpha) &= \sigma_{s,\max}\,(1-\alpha),\\
% \sigma_{b}(\alpha) &= \sigma_{b,\text{base}}+(\sigma_{b,\max}-\sigma_{b,\text{base}})\,\alpha,\\
% \sigma(x,\alpha) &= \sigma_{s}(\alpha)\,\mathbf{1}_{x\in \mathcal M_t}+\sigma_{b}(\alpha)\,\mathbf{1}_{x\notin \mathcal M_t}.
% \end{aligned}}
% \end{equation}

% % --- Alpha-controlled control law inside PIA ---
% \begin{equation}
% \label{eq:pia_control_law}
% \boxed{
% \begin{aligned}
% \text{(phys)}\quad
% s_t^{(k+1)} &= s_t^{(k)}
% + \underbrace{\Delta\tau\,\nu(\alpha)\,\Delta_{\mathcal M}\,s_t^{(k)}}_{\text{coherence}}
% + \underbrace{\sqrt{2\Delta\tau}\,\sigma(x,\alpha)\,\mathcal N(0,I)}_{\text{dynamism}},\\[2pt]
% s_t^{\text{phys}} &\coloneqq s_t^{(k+1)},\\[2pt]
% \text{(ID inject)}\quad
% s_t^{\text{id}} &= \mathcal M_t \odot \Big((1-\lambda_{\mathrm{id}}(\alpha))\,s_t+\lambda_{\mathrm{id}}(\alpha)\,s_{\text{ref}}\Big)
% + (1-\mathcal M_t)\odot s_t,\\[2pt]
% \text{(attn inputs)}\quad
% Q &= s_t^{\text{phys}} W_Q,\qquad
% K = s_t^{\text{id}} W_K,\qquad
% V = s_t^{\text{id}} W_V.
% \end{aligned}}
% \end{equation}

\section{RealDiffusion}
\label{sec:realdiffusion}

Rather than learning coherence, we control it at inference time. We view multi-frame generation as a controllable dynamical system and inject lightweight physics-based priors directly into attention. The overall architecture is shown in Figure~\ref{fig:pipeline}. We introduce the essential notation here. The system operates on a sequence of per-frame hidden features $S=\{s_t\}_{t=1}^T$, where the subscript $t$ denotes the frame index. These features are guided by a dynamic binary subject mask $\mathcal{M}_t$. The entire process is governed by $\alpha\in[0,1]$. This knob simultaneously modulates four key parameters, each scaled by a base constant $c$. The full sampling procedure is detailed in Algorithm~\ref{alg:full_sampling}.

\subsection{Dynamic Mask Generation}
\label{sec:mask_generation}
To localize interventions to subjects, we produce a binary mask $\mathcal{M}_t$ at each sampling step without training. We capture cross-attention maps that correspond to the subject tokens, average them over a recent window to stabilize noise, and binarize the result using Otsu's method~\cite{Otsu}. The mask sequence gates all subsequent operations, ensuring that regularization acts on the subject and avoids background leakage. Example masks are illustrated in Figure~\ref{fig:masks}.

\subsection{PhysicsOperator}
\label{sec:physics_operator}
The PhysicsOperator regularizes the temporal sequence $\{s_t\}$ via a dissipative coherence kernel and a region-aware dynamism kernel, tuned by $\alpha$. The choice of the coherence kernel is critical. As detailed in Table~\ref{tab:pde_ablation_rules}, we select the Heat Diffusion equation because it is a dissipative system that guarantees a stable smoothing process towards equilibrium. This contrasts with conservative systems (e.g., Wave, Elasticity) that cause feature oscillations, and non-linear systems (e.g., Burgers') that risk unpredictable artifacts. This theoretical stability is reflected empirically, as heat diffusion offers a significantly wider and more robust parameter tuning range than the highly sensitive alternatives.

\paragraph{Coherence via insulated heat diffusion.}
To suppress high-frequency temporal incoherence, we use the heat equation as a dissipative prior, whose continuous form is:
\begin{equation}
    \frac{\partial s(x,\tau)}{\partial \tau} = \nabla \cdot \big(\nu(\alpha)\,\nabla s(x,\tau)\big),
    \label{eq:heat_pde}
\end{equation}
where $x$ is the spatial position and $\tau$ is a virtual time. For the discrete feature sequence, we employ periodic boundary conditions, allowing the first and last frames to influence each other for temporal coherence. The core of our method is the \emph{insulated} Laplacian $\Delta_{\mathcal{M}}$, which confines the smoothing process to the subject regions:
\begin{equation}
    \Delta_{\mathcal{M}}\, s_t = s'_{t+1} - 2s_t + s'_{t-1}.
    \label{eq:insulated_laplacian}
\end{equation}
The insulation is achieved by defining the effective neighbors $s'_{t\pm1}$ conditionally based on the dynamic masks:
\begin{equation}
    s'_{t\pm1} = \mathcal{M}_{t\pm1}\odot s_{t\pm1} + (1-\mathcal{M}_{t\pm1})\odot s_t.
    \label{eq:effective_neighbors}
\end{equation}
As defined in Eq.~\ref{eq:effective_neighbors}, if a neighboring feature at frame $t\pm1$ is outside the subject mask (i.e., $\mathcal{M}_{t\pm1}$ is zero), its value is replaced by the current feature $s_t$. This effectively creates a zero-flux boundary at the mask's edge, ensuring the diffusion process only occurs within the subject region and preventing information leakage from the background.

\paragraph{Dynamism via region-aware stochasticity.}
To avoid over-smoothing and preserve narrative dynamism, we inject controlled noise with region-aware intensities. The intensities differ inside ($\sigma_{\mathrm{s}}$) and outside ($\sigma_{\mathrm{b}}$) the mask $\mathcal{M}_t$ and are both governed by $\alpha$. The subject noise intensity $\sigma_{\mathrm{s}}$ is inversely tied to $\alpha$ to stabilize the subject, while the background noise intensity $\sigma_{\mathrm{b}}$ is modulated by $\alpha$ to preserve scene dynamism. We combine these into a single, spatially-varying noise intensity tensor $\sigma_t$:
\begin{equation}
    \sigma_t = \mathcal{M}_t \cdot \sigma_{\mathrm{s}} + (1 - \mathcal{M}_t) \cdot \sigma_{\mathrm{b}}.
    \label{eq:spatial_noise}
\end{equation}
This composite intensity $\sigma_t$ is then used in the stochastic part of the full update rule (Eq.~\ref{eq:final_update}), providing a compact representation of our region-aware dynamism kernel.

\paragraph{Discrete update.}
Combining both kernels with a forward Euler step in $\tau$, the per-iteration update is
\begin{equation}
\begin{aligned}
    s_t^{k+1} = s_t^{k}
    \;&+\; \underbrace{\Delta\tau \left( \nu(\alpha) \cdot \Delta_{\mathcal{M}} s_t^{k} \right)}_{\text{coherence}}
    \;\\
    &+\; \underbrace{\sqrt{2 \Delta\tau} \cdot \sigma_t\, \mathcal{N}(0, I)}_{\text{dynamism}} \, ,
\end{aligned}
\label{eq:final_update}
\end{equation}
where $\Delta\tau$ is the virtual time step and $\mathcal{N}(0,I)$ is standard normal noise. The coefficients introduced in Eq.~\ref{eq:final_update} are simple, intuitive functions of the control parameter $\alpha$. Their precise formulations are detailed in Algorithm~\ref{alg:full_sampling}.

\paragraph{Physics-informed attention.}
We integrate this process into attention via a disentangled intervention: $Q$ are built from the PhysicsOperator-smoothed states to ensure temporal coherence, while $(K,V)$ are built from identity-enhanced states that use an ID bank injection with a strength proportional to $\alpha$. This allows attention to retrieve identity-faithful information using temporally smooth queries. The precise assignments are detailed in Algorithm~\ref{alg:full_sampling}.

% \subsection{Algorithm: Full Diffusion Sampling with RealDiffusion}
% \label{alg:full}
% We now give the full sampling procedure (e.g., DDIM/DPMSolver-style) augmented with dynamic masking, insulated diffusion, and identity-aware attention. All mappings of strengths to $\alpha$ are implemented as scalar assignments here to keep notation minimal.

% The single knob $\alpha$ simultaneously strengthens diffusion-driven coherence (via $\nu$), increases ID injection ($\lambda_{\text{id}}$), and suppresses subject noise while allowing modest background variability. At $\alpha\!\rightarrow\!1$ the system emphasizes stable identity and smooth temporal evolution; at $\alpha\!\rightarrow\!0$ it favors dynamism and pose dynamism. All evitality ffects are realized at inference without additional training and are confined to subject regions by $\mathcal{M}_t$.

\begin{algorithm}[t]
\caption{RealDiffusion Sampling}
\label{alg:full_sampling}
\small
\begin{algorithmic}[1]
\REQUIRE Prompt $p$; frames $T$; steps $L$; control $\alpha\!\in\![0,1]$; ID bank $\{s_{\text{bank}}^l\}_{l=1}^L$ ; PhysicsOperator iters $n$; virtual time step $\Delta\tau$. 
\STATE \parbox[t]{.9\linewidth}{
    \textbf{Notation:} \\
    \begin{tabular}{@{}l@{\ --\ }l}
        $Q, K, V$      & Queries, Keys, and Values in attention.\\
        $s_t^{\text{phys}}$ & Physically-smoothed features. \\
        $s_t^{\text{id}}$   & Identity-enhanced features. \\
        % $s_{\text{bank}}^\ell$ & Shared ID bank features for denoising step $\ell$. 
    \end{tabular}
}
\ENSURE Images $\{x_t^\ast\}_{t=1}^T$.

\STATE Encode $p$.
\STATE For $t\!=\!1..T$: sample $z_L^{(t)}\!\sim\!\mathcal N(0,I)$.
\STATE Set $\nu\!\leftarrow\!c_{\text{heat}}\alpha$;\; $\lambda_{\text{id}}\!\leftarrow\!c_{\text{id}}\alpha$;\;
$\sigma_{\mathrm{s}}\!\leftarrow\!c_{\mathrm{s}}(1-\alpha)$;\;
$\sigma_{\mathrm{b}}\!\leftarrow\!c_{\mathrm{b}}\alpha$.
\FOR{$\ell=L,\ldots,1$}
  \STATE Collect subject cross-attention maps.
  \STATE For each $t$: $\mathcal M_t\leftarrow\text{Otsu}(\text{AvgRecentAttn})$.
  \STATE Extract current hidden states $\{s_t\}_{t=1}^T$.
  \STATE Use $\sigma_{\mathrm{s}}$ inside $\mathcal M_t$, $\sigma_{\mathrm{b}}$ outside $\mathcal M_t$ to get $\sigma_t$ by Eq. ~\ref{eq:spatial_noise}.

  \STATE Apply Eq.~\ref{eq:final_update} $n$ iters with step $\Delta\tau$ to get $\{s_t^{\text{phys}}\}$.
  \STATE  $s_t^{\text{inj}}\!\leftarrow\!(1-\lambda_{\text{id}})s_t+\lambda_{\text{id}}s_{\text{bank}}^l$; 
  $s_t^{\text{id}}\!\leftarrow\!\mathcal M_t\!\odot\! s_t^{\text{inj}}+(1-\mathcal M_t)\!\odot\! s_t$.
  \STATE Build $Q$ from $s_t^{\text{phys}}$; build $(K,V)$ from $s_t^{\text{id}}$; U-Net to get $\hat\epsilon_\ell$.
  \STATE Scheduler update: $z_{\ell-1}^{(t)}$ from $z_\ell^{(t)}$ and $\hat\epsilon_\ell$.
\ENDFOR
\STATE Decode each $z_0^{(t)}$ with VAE to obtain $x_t^\ast$.
\RETURN $\{x_t^\ast\}_{t=1}^T$.
\end{algorithmic}
\end{algorithm}

\section{Experiments}
\label{sec:experiments}

This section experimentally validates RealDiffusion's effectiveness and controllability. We compare our method against state-of-the-art approaches, conduct ablation studies on the physical prior and the $\alpha$ controller, and introduce novel metrics to quantitatively measure the balance between coherence and dynamism.

\newsavebox{\imagebox}
\sbox{\imagebox}{\includegraphics[width=0.22\textwidth]{example-image-a}}

\begin{figure*}[!ht]
\centering
% We set column separation to a small value for a compact layout.
\setlength{\tabcolsep}{1.5pt} 
\small

\begin{tabular}{>{\centering\arraybackslash}m{0.2cm} c c}

% --- HEADER ROW ---
& 
% \multicolumn is used to make a title span across the 4 image columns.
\multicolumn{1}{c}{\parbox{0.45\linewidth}{\centering\textbf{Prompt A:} \promptpart{idcolor}{Impressionist oil painting, woman and child,} \promptpart{action1color}{watching sailboats,} \promptpart{action4color}{reading a book,} \promptpart{action3color}{chasing butterflies.}}} & 
\multicolumn{1}{c}{\parbox{0.45\linewidth}{\centering\textbf{Prompt B:} \promptpart{idcolor}{Ghibli anime, girl and fluffy creature,} \promptpart{action1color}{waiting at bus stop,} \promptpart{action4color}{running through grass,} \promptpart{action3color}{flying through the sky.}}} \\
% A thin line under the header for better structure.
\cmidrule(lr){2-2} \cmidrule(lr){3-3} \\

%---------------------------------------------------
% MODEL 1: IP-Adapter
%---------------------------------------------------
\vspace{-13.5mm}\rotatebox[origin=c]{90}{\small IP-Adapter} &
\begin{tabular}{cccc}
    \includegraphics[width=0.15\linewidth]{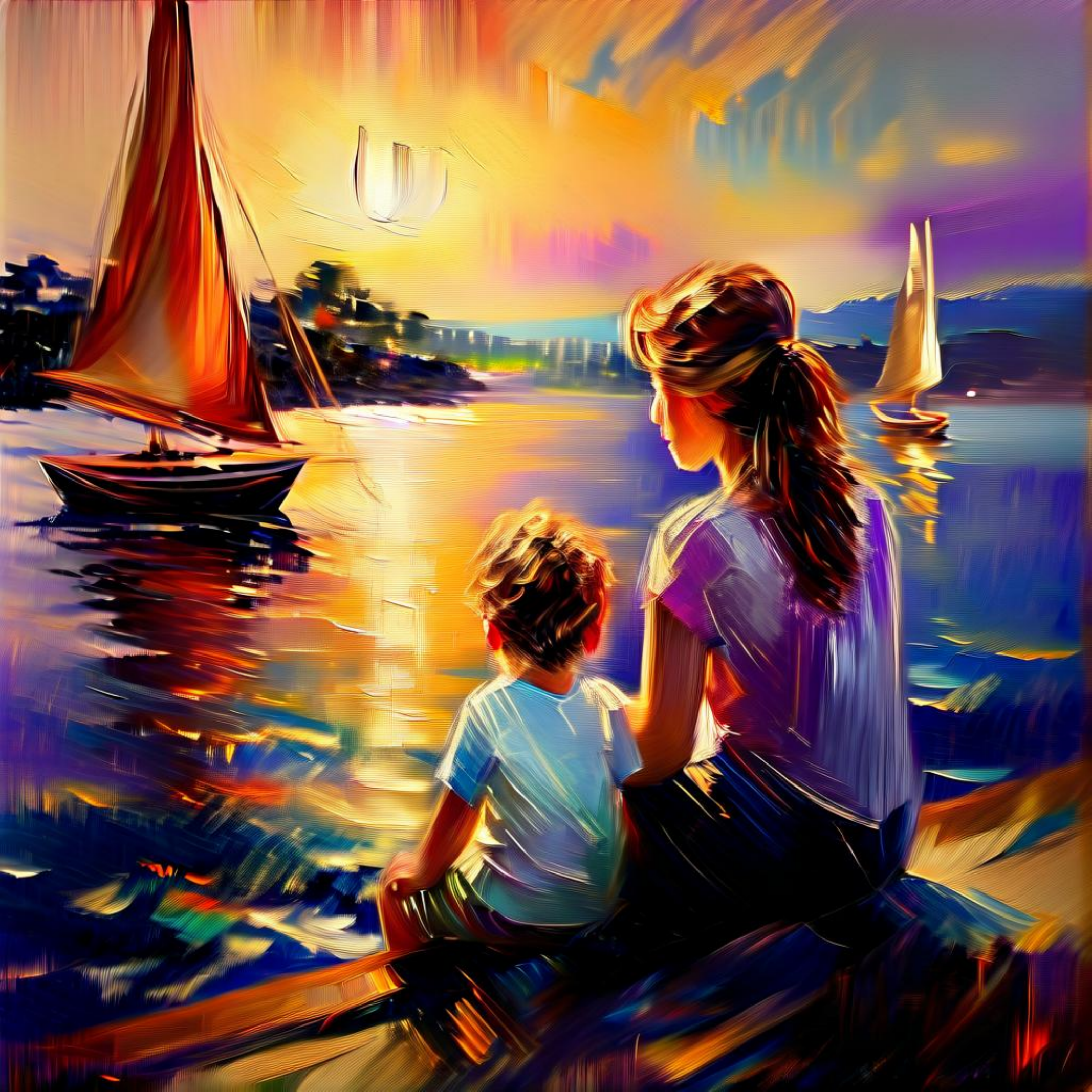} &
    \includegraphics[width=0.15\linewidth]{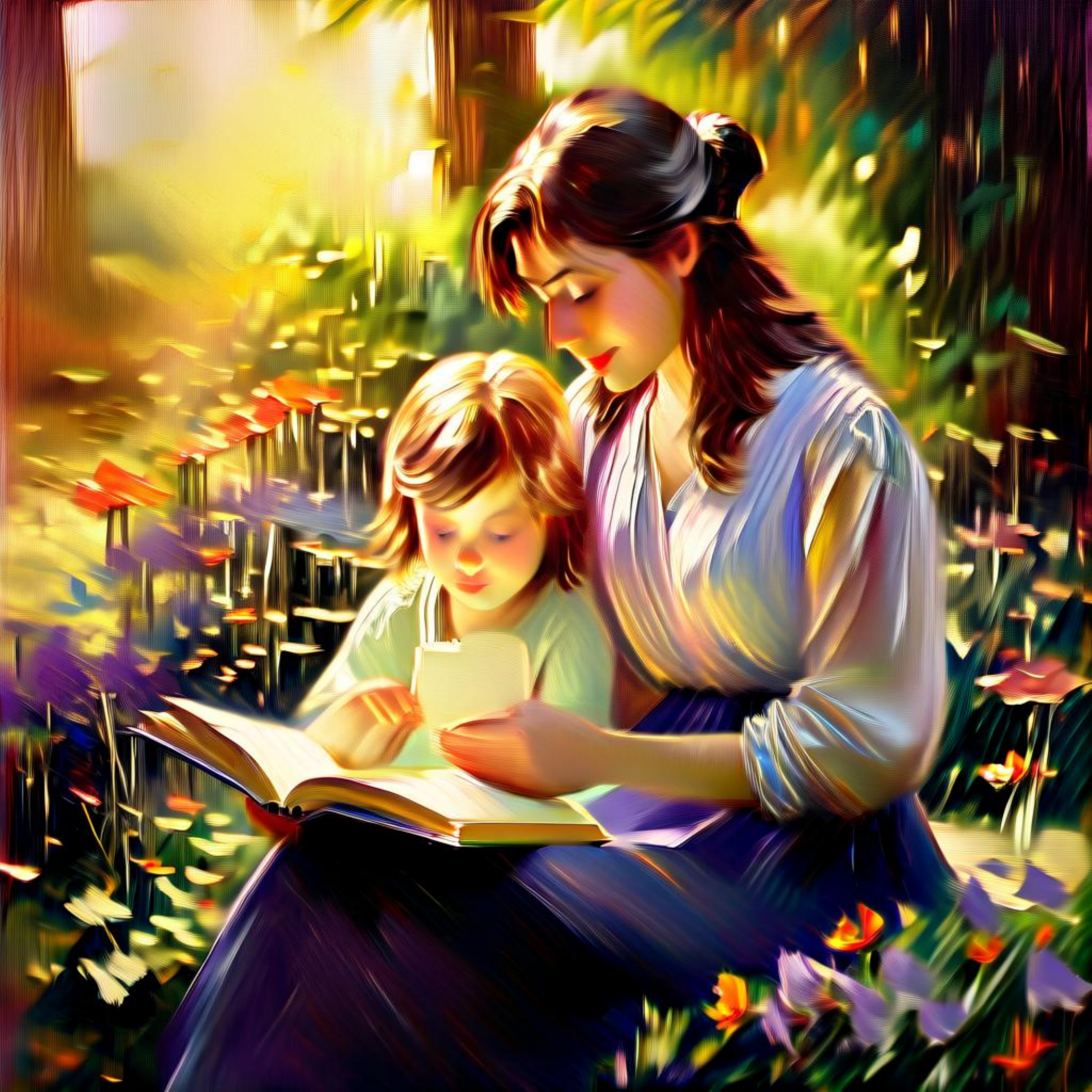} &
    \includegraphics[width=0.15\linewidth]{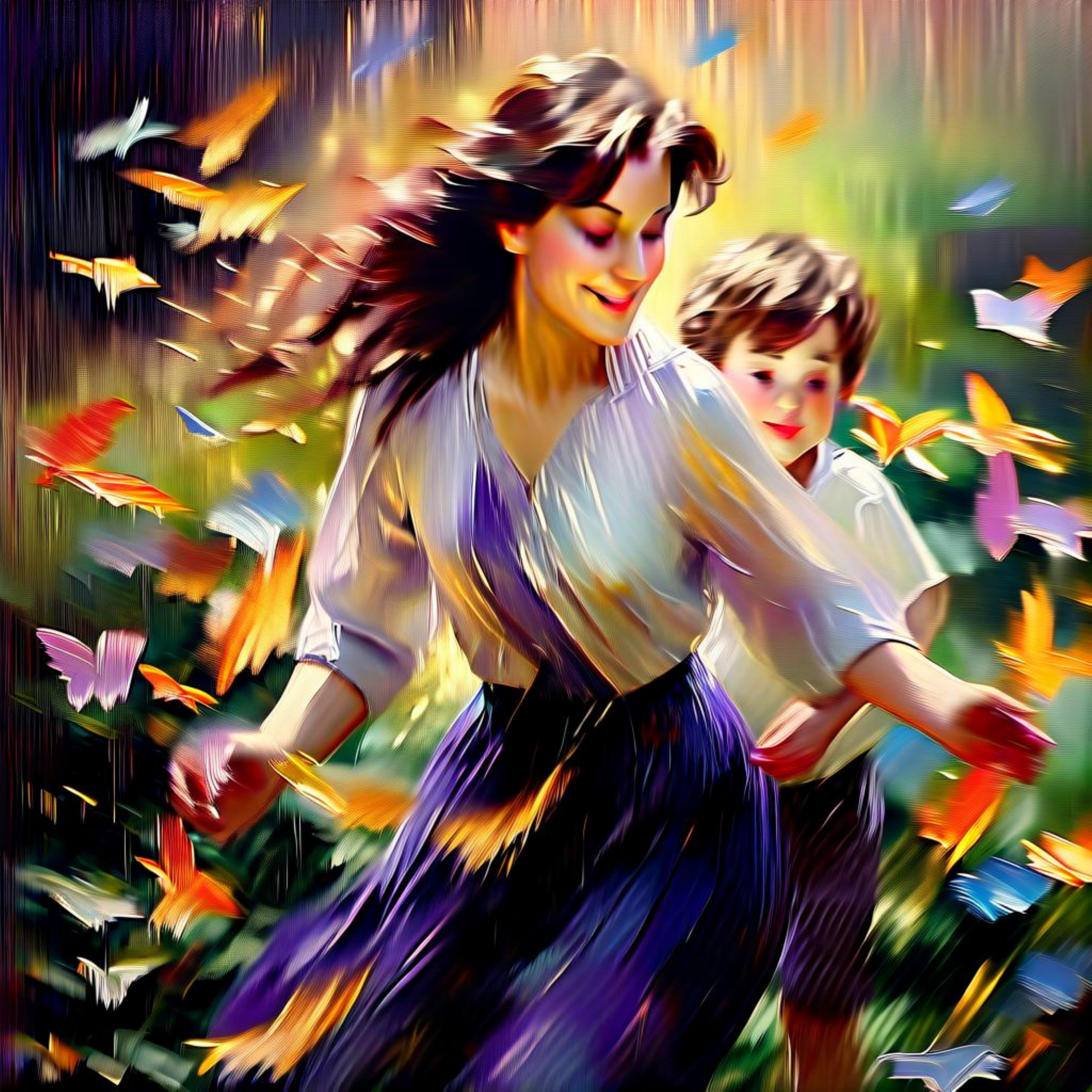} &
\end{tabular}
&
\begin{tabular}{cccc}
    \includegraphics[width=0.15\linewidth]{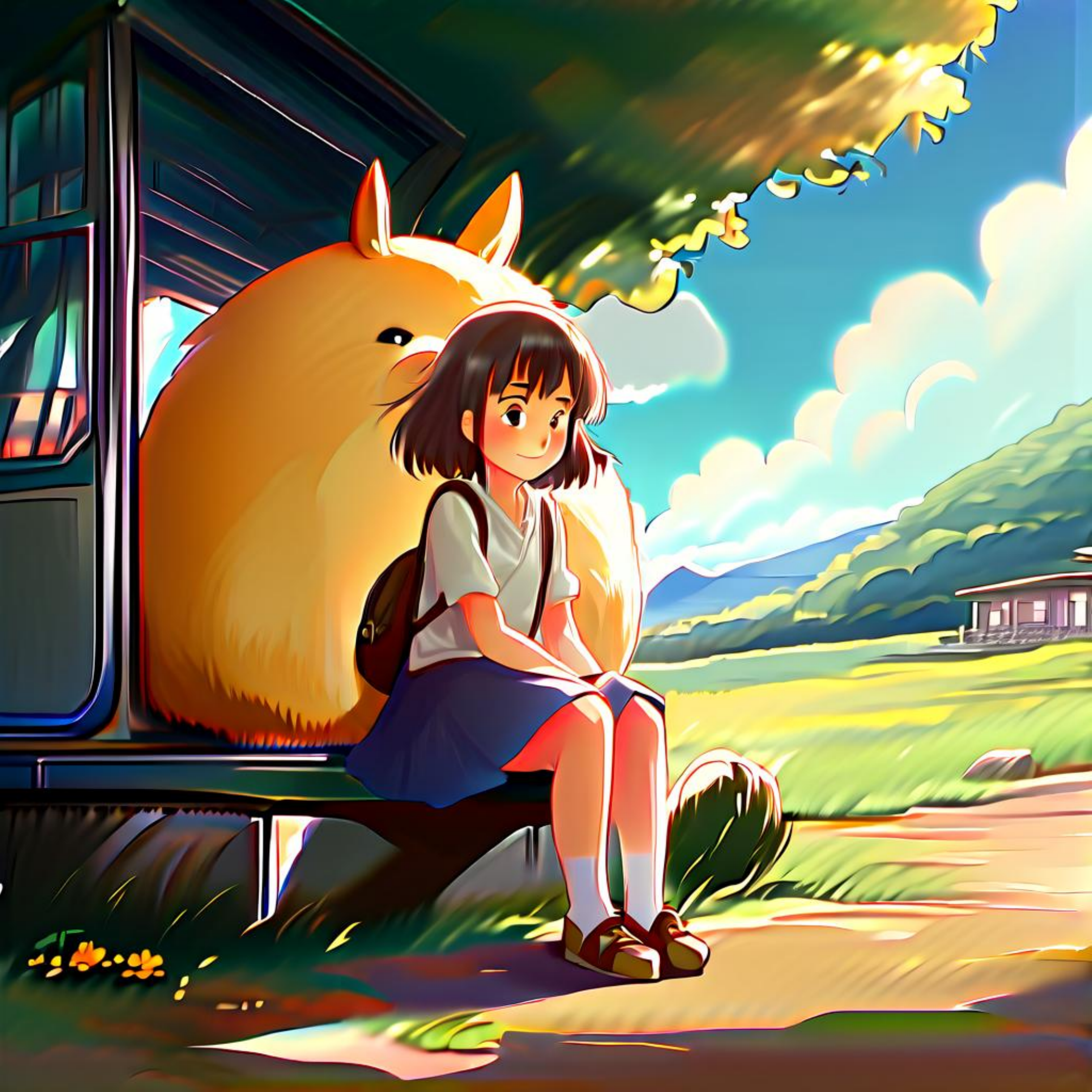} &
    \includegraphics[width=0.15\linewidth]{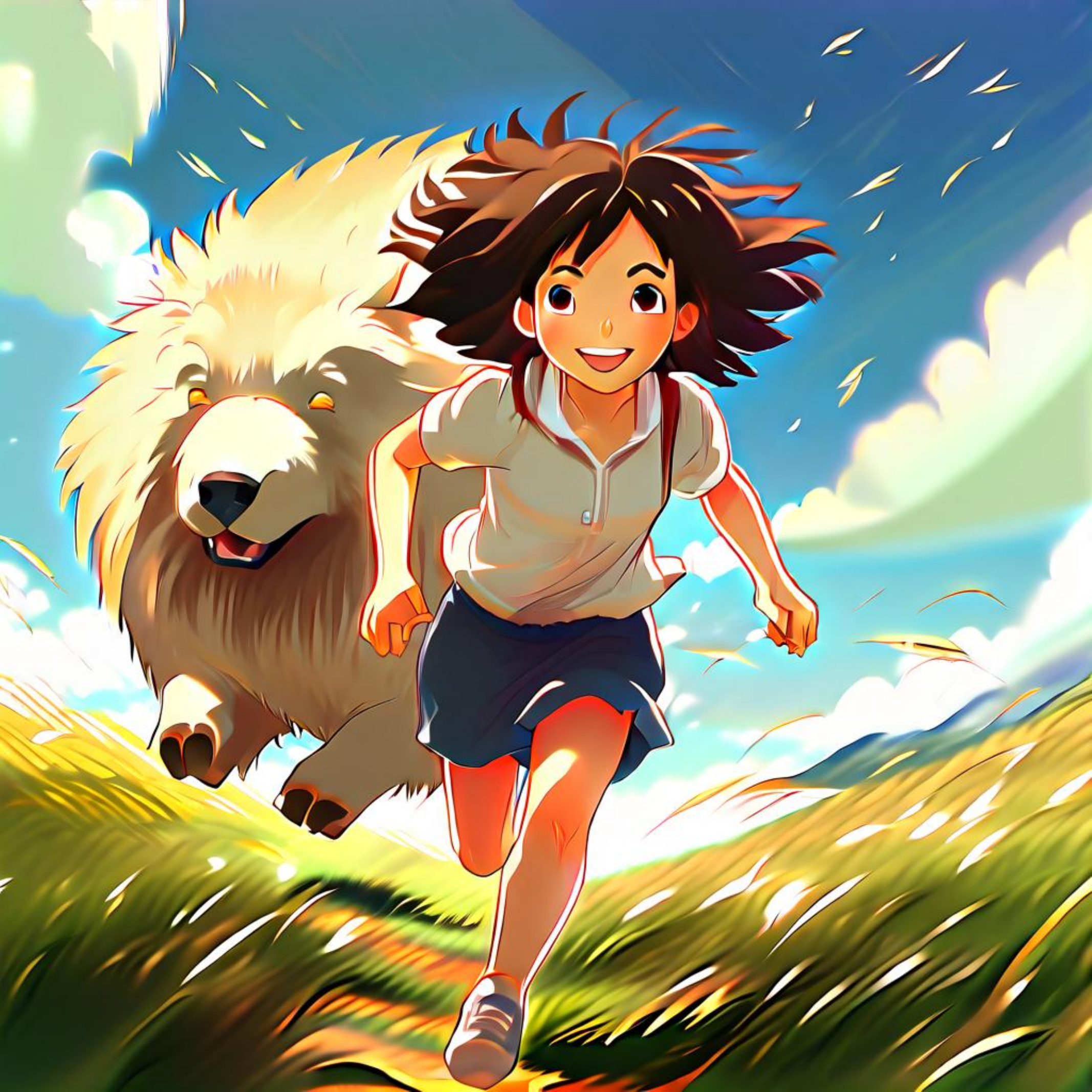} &
    \includegraphics[width=0.15\linewidth]{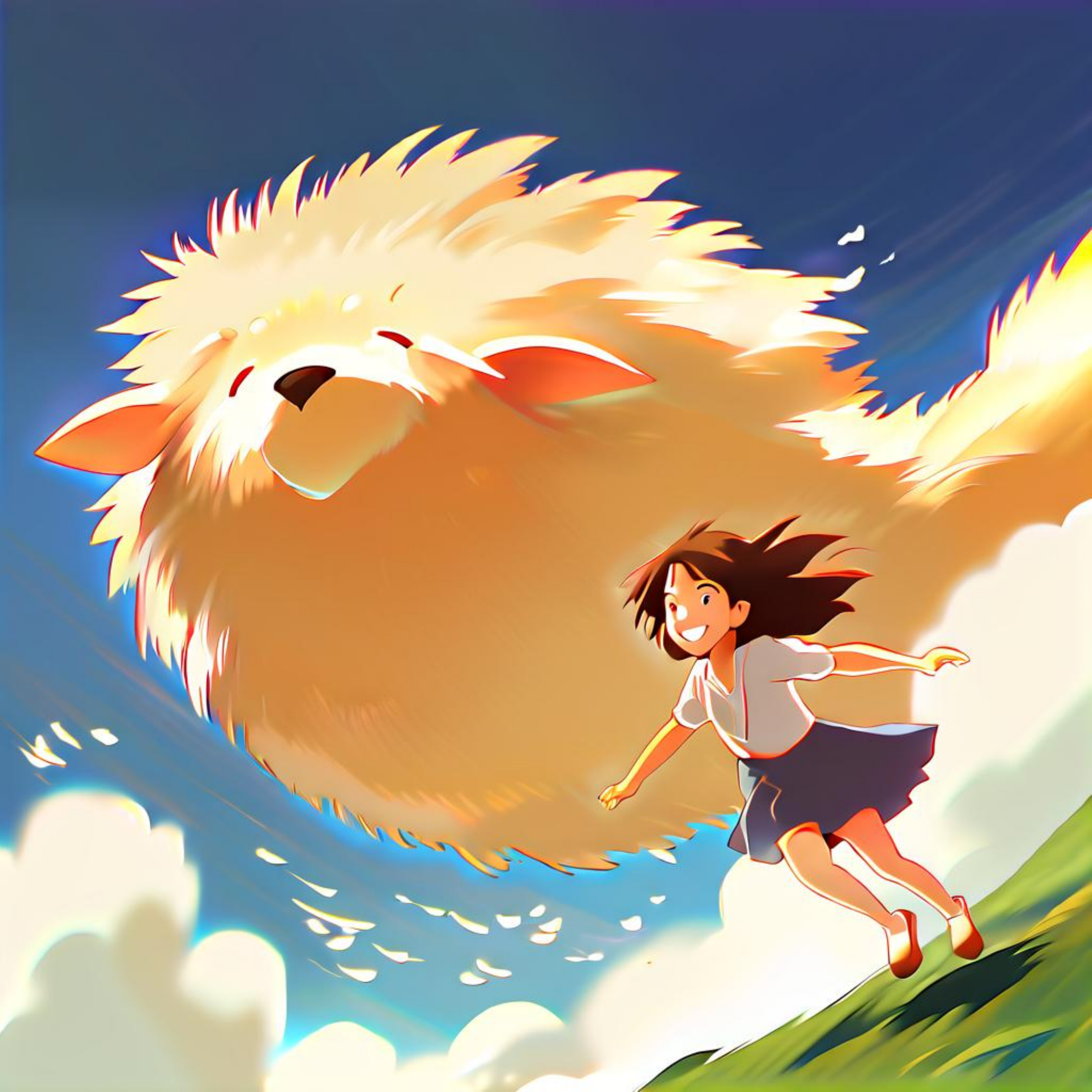}
\end{tabular}
\\ 
\vspace{-13.5mm}\rotatebox[origin=c]{90}{\small PhotoMaker} &
\begin{tabular}{cccc}
    \includegraphics[width=0.15\linewidth]{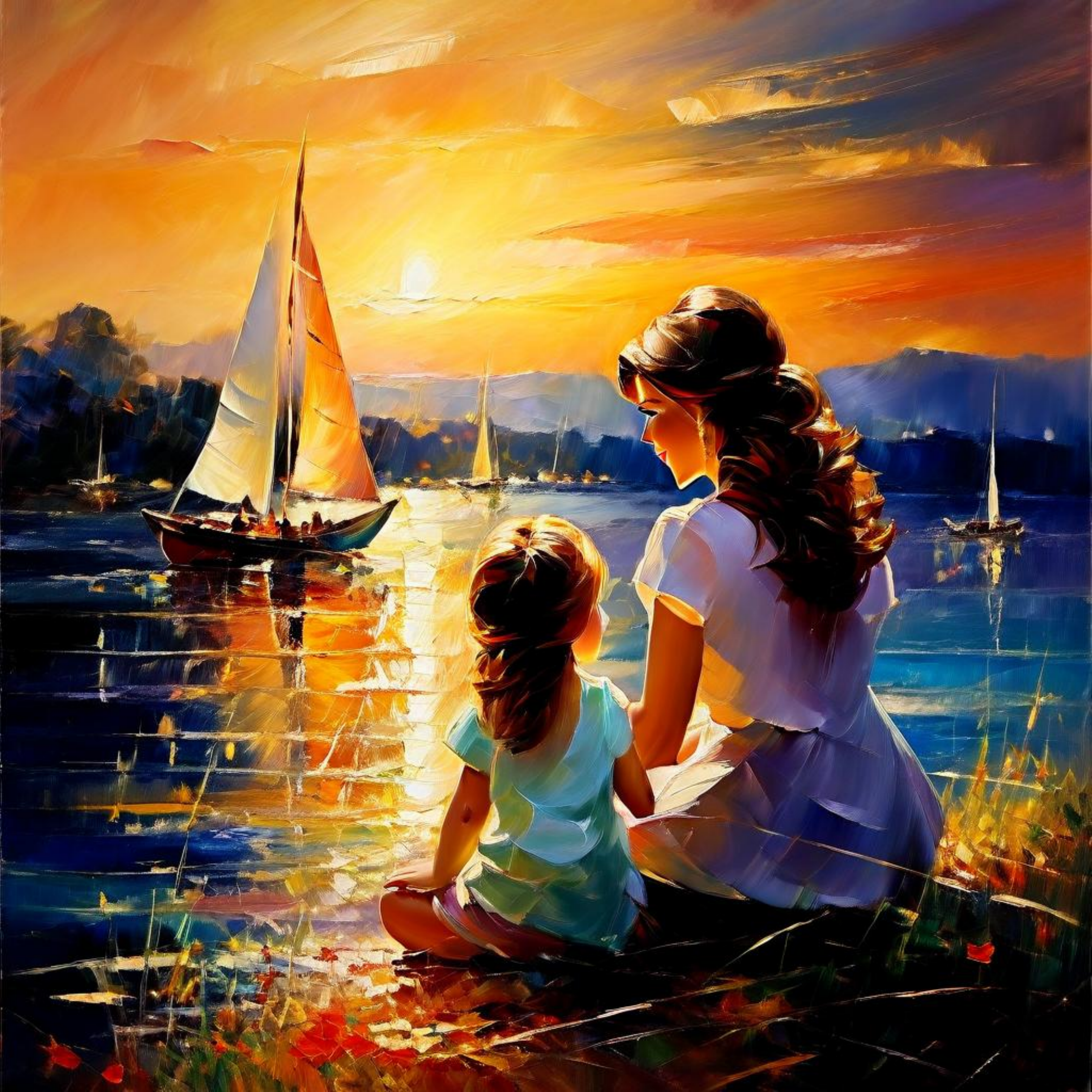} &
    \includegraphics[width=0.15\linewidth]{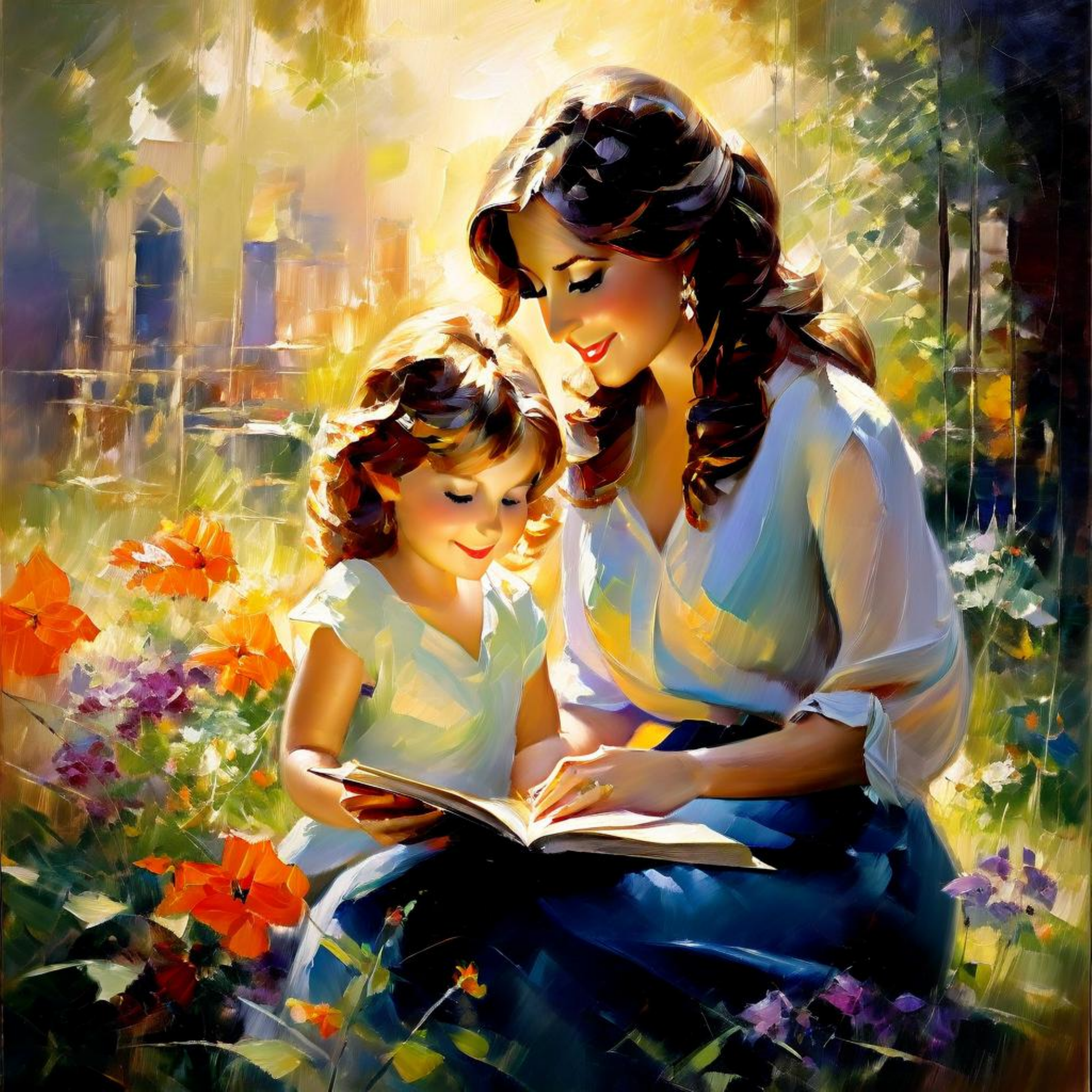} &
    \includegraphics[width=0.15\linewidth]{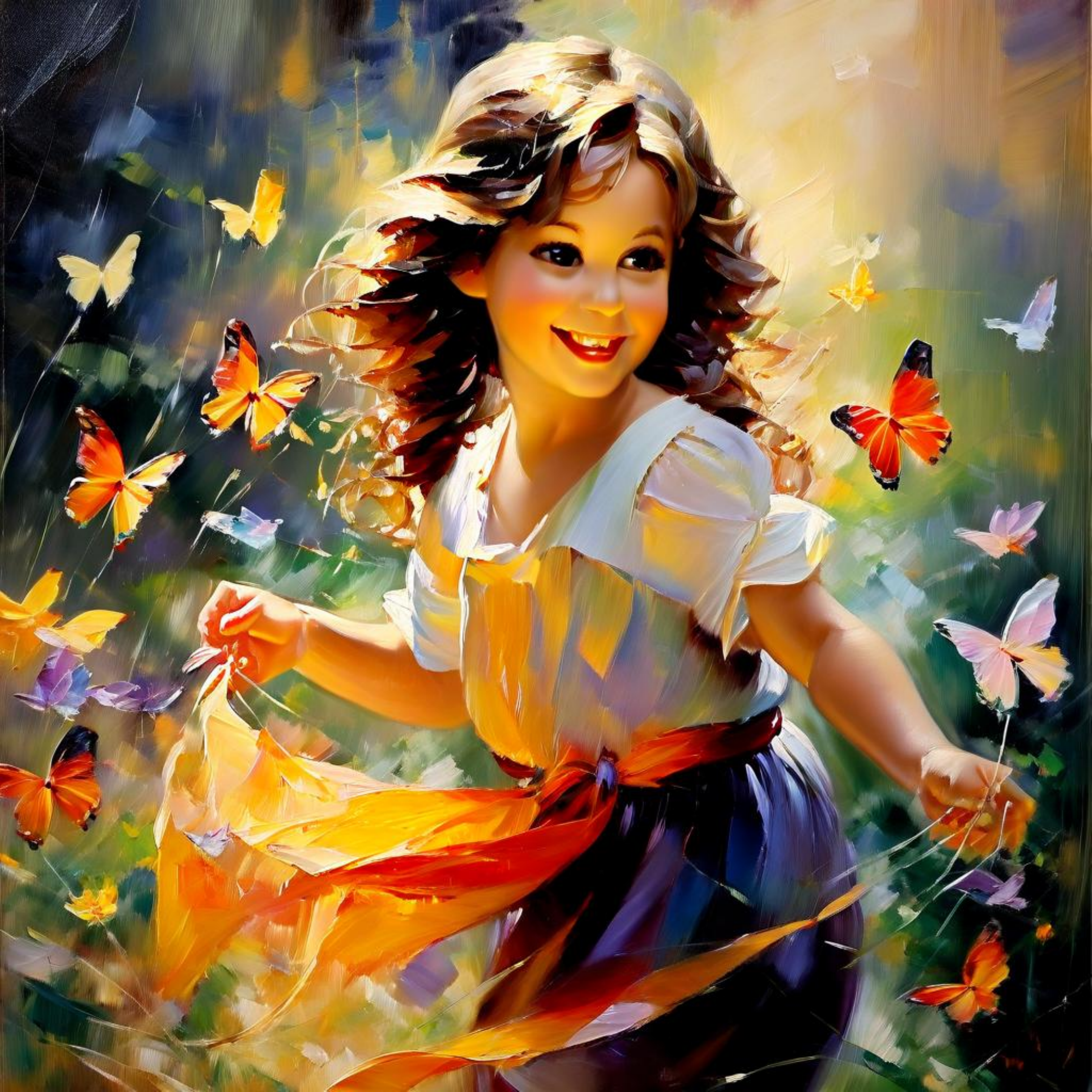} &
\end{tabular}
&
\begin{tabular}{cccc}
    \includegraphics[width=0.15\linewidth]{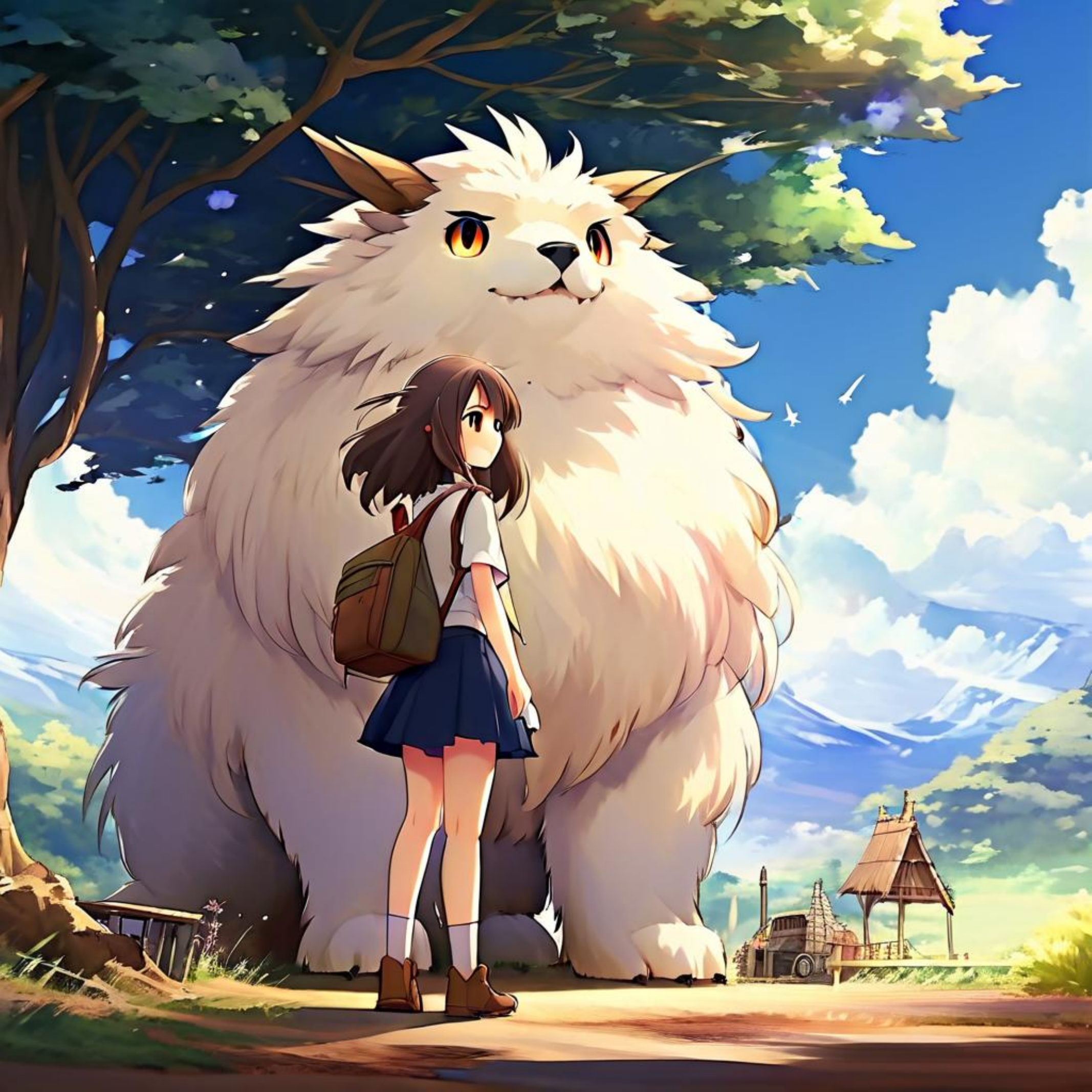} &
    \includegraphics[width=0.15\linewidth]{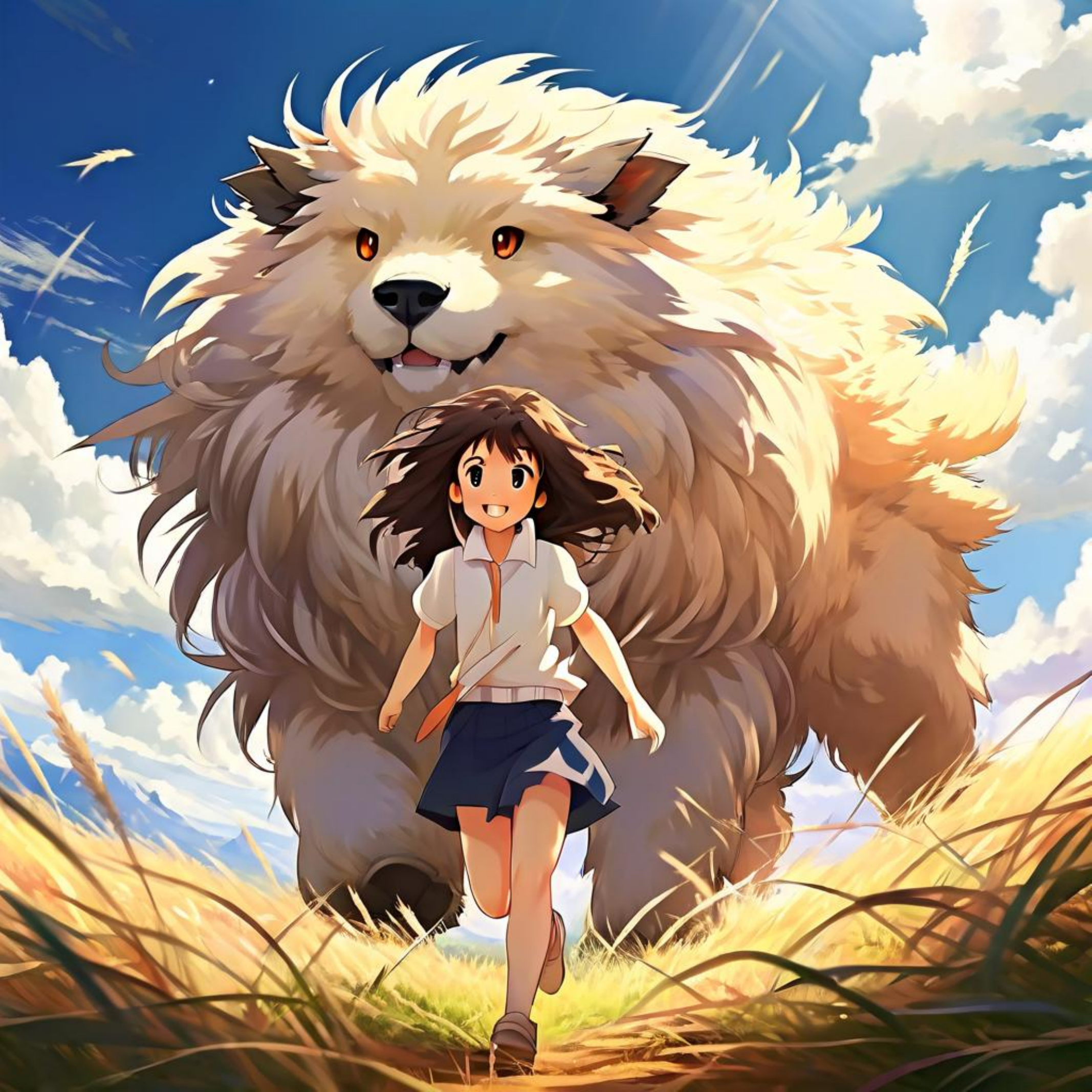} &
    \includegraphics[width=0.15\linewidth]{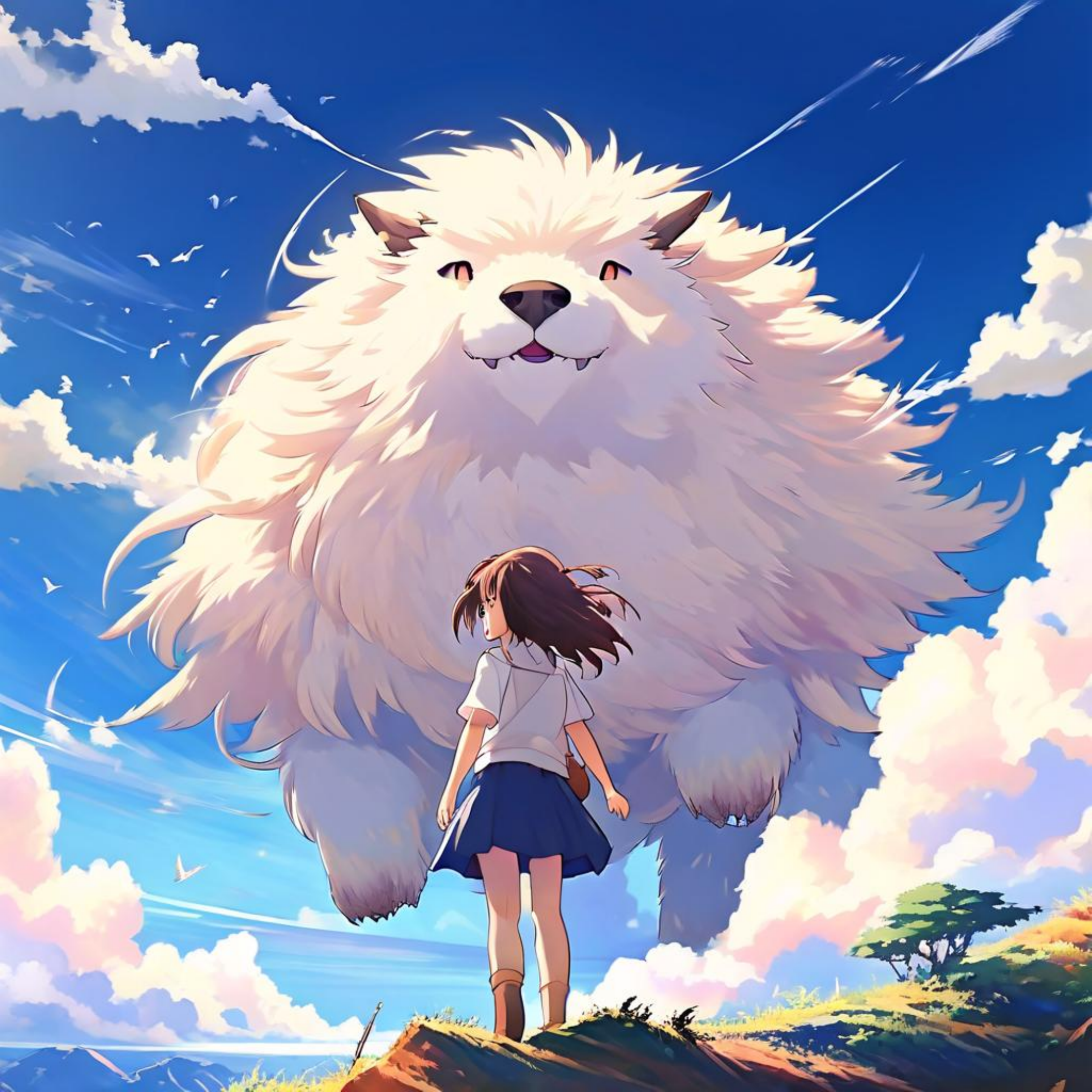}
\end{tabular}
\\ 
\vspace{-13.5mm}\rotatebox[origin=c]{90}{\small StoryDiffusion} &
\begin{tabular}{cccc}
    \includegraphics[width=0.15\linewidth]{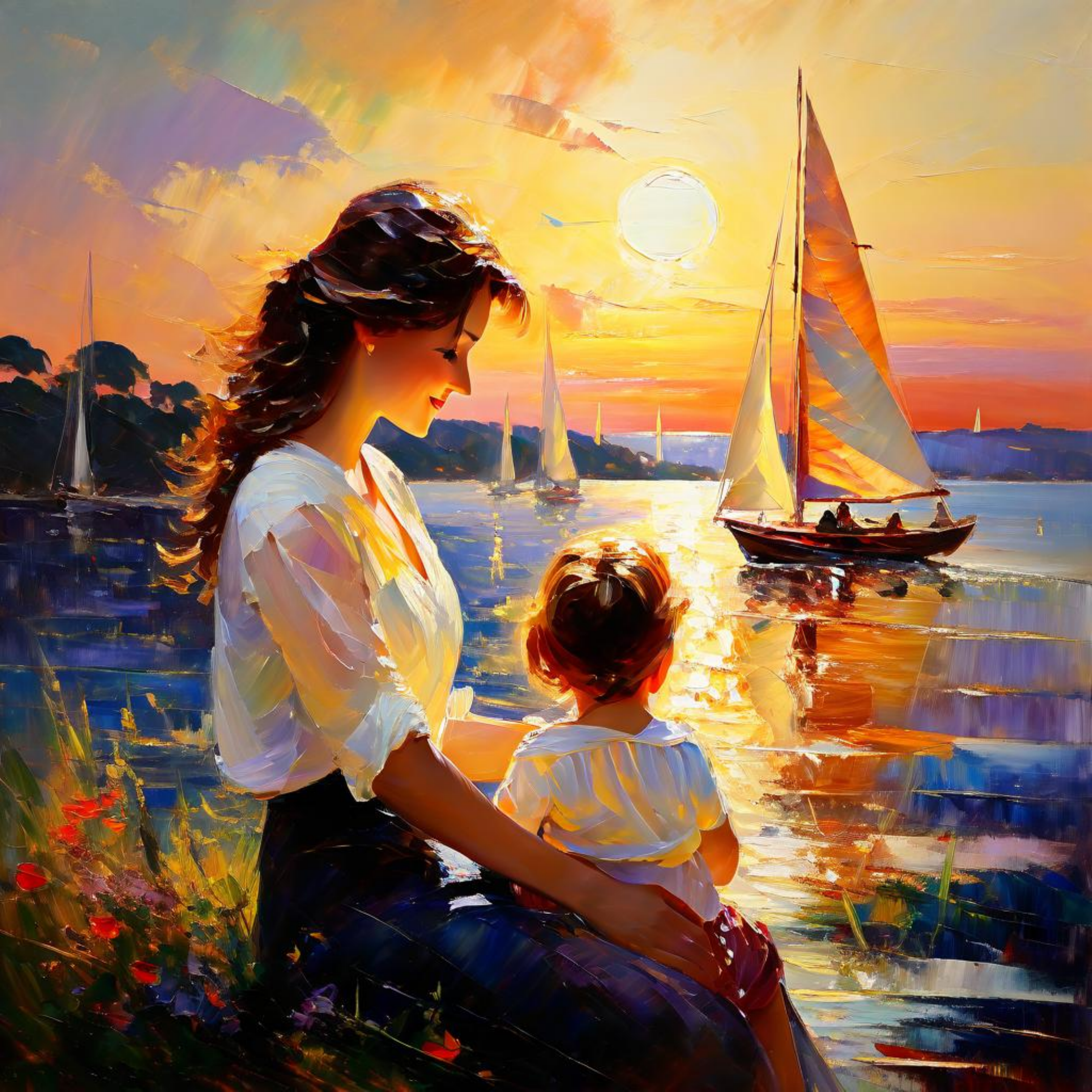} &
    \includegraphics[width=0.15\linewidth]{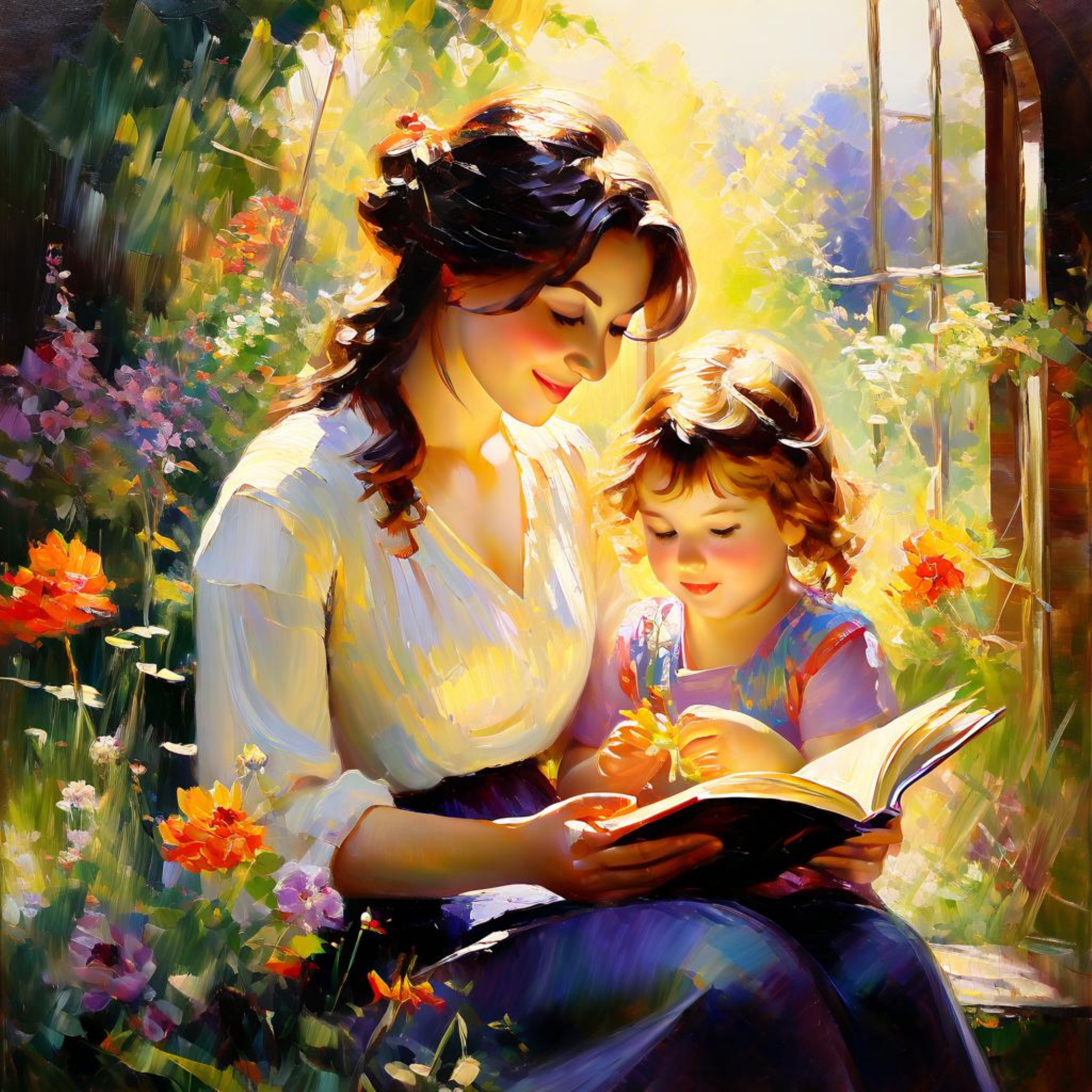} &
    \includegraphics[width=0.15\linewidth]{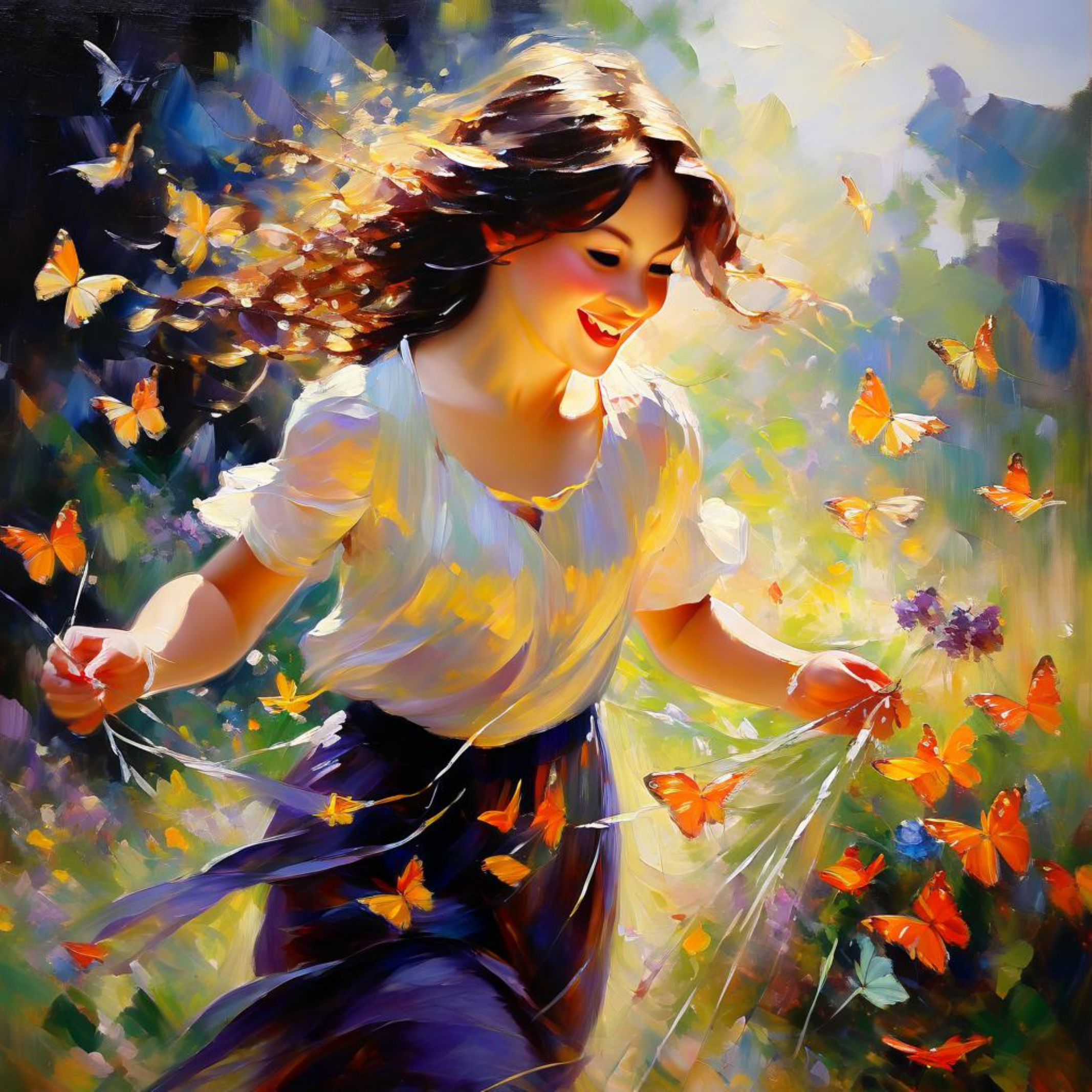} &
\end{tabular}
&
\begin{tabular}{cccc}
    \includegraphics[width=0.15\linewidth]{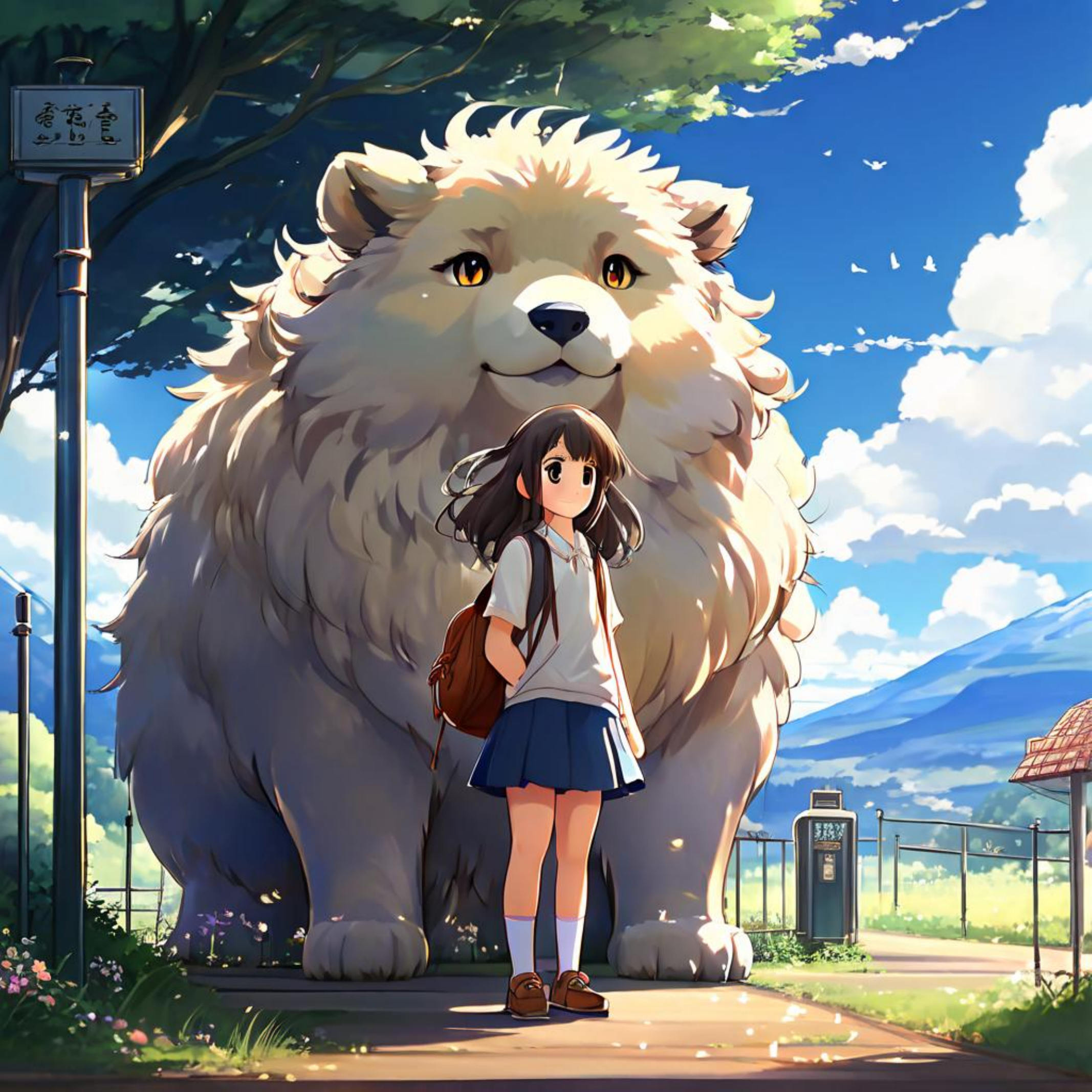} &
    \includegraphics[width=0.15\linewidth]{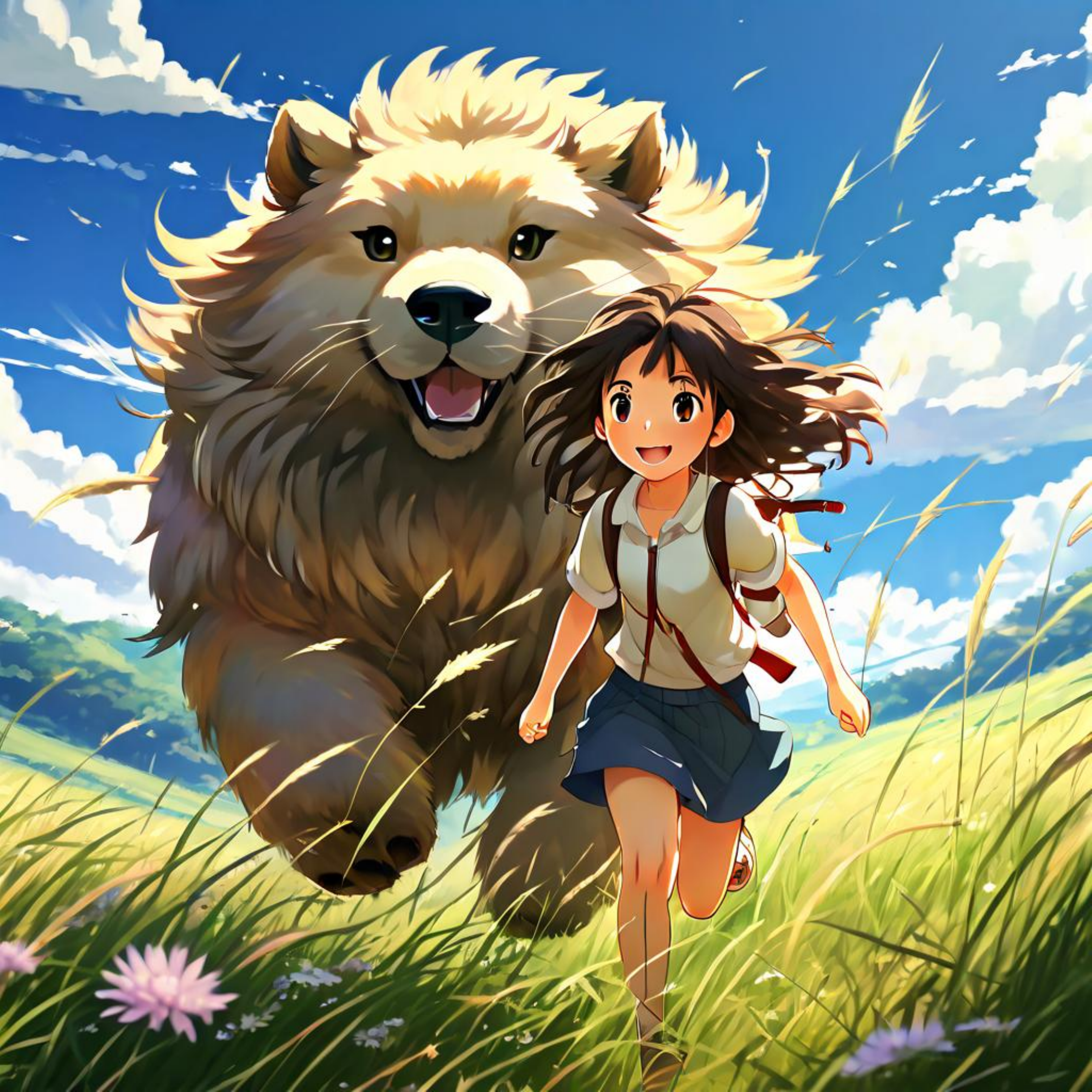} &
    \includegraphics[width=0.15\linewidth]{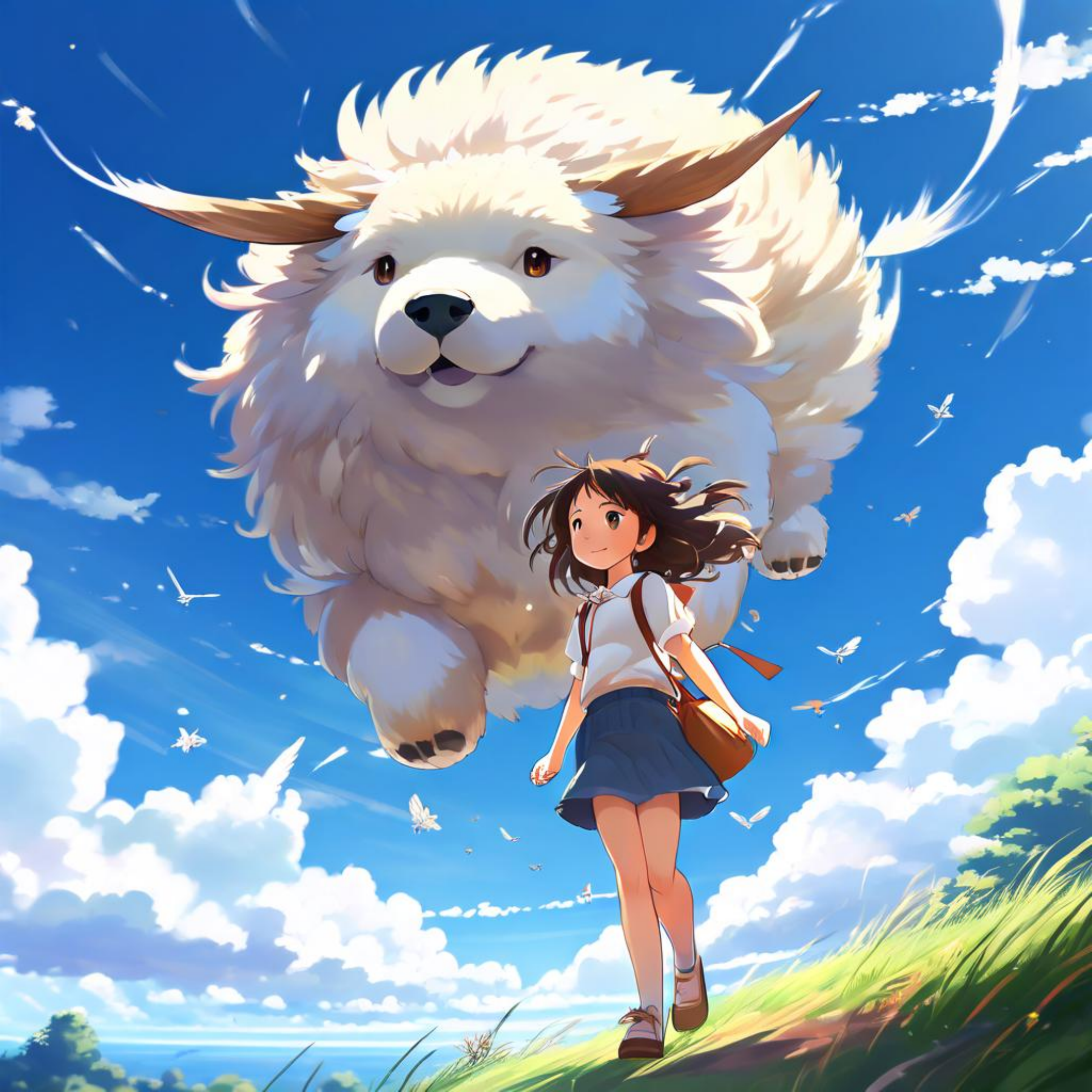}
\end{tabular}
\\
\vspace{-13.5mm}\rotatebox[origin=c]{90}{\small ConsiStory} &
\begin{tabular}{cccc}
    \includegraphics[width=0.15\linewidth]{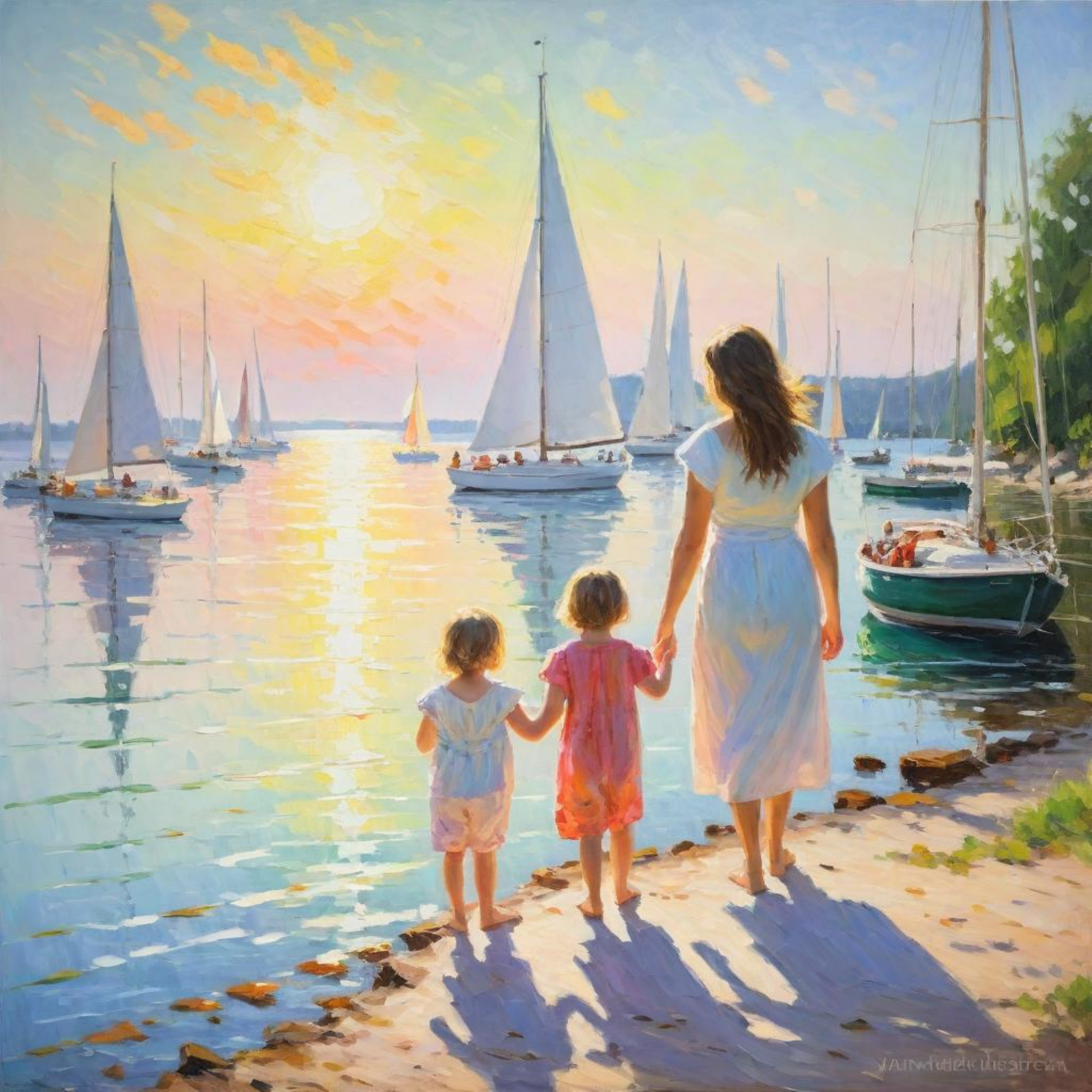} &
    \includegraphics[width=0.15\linewidth]{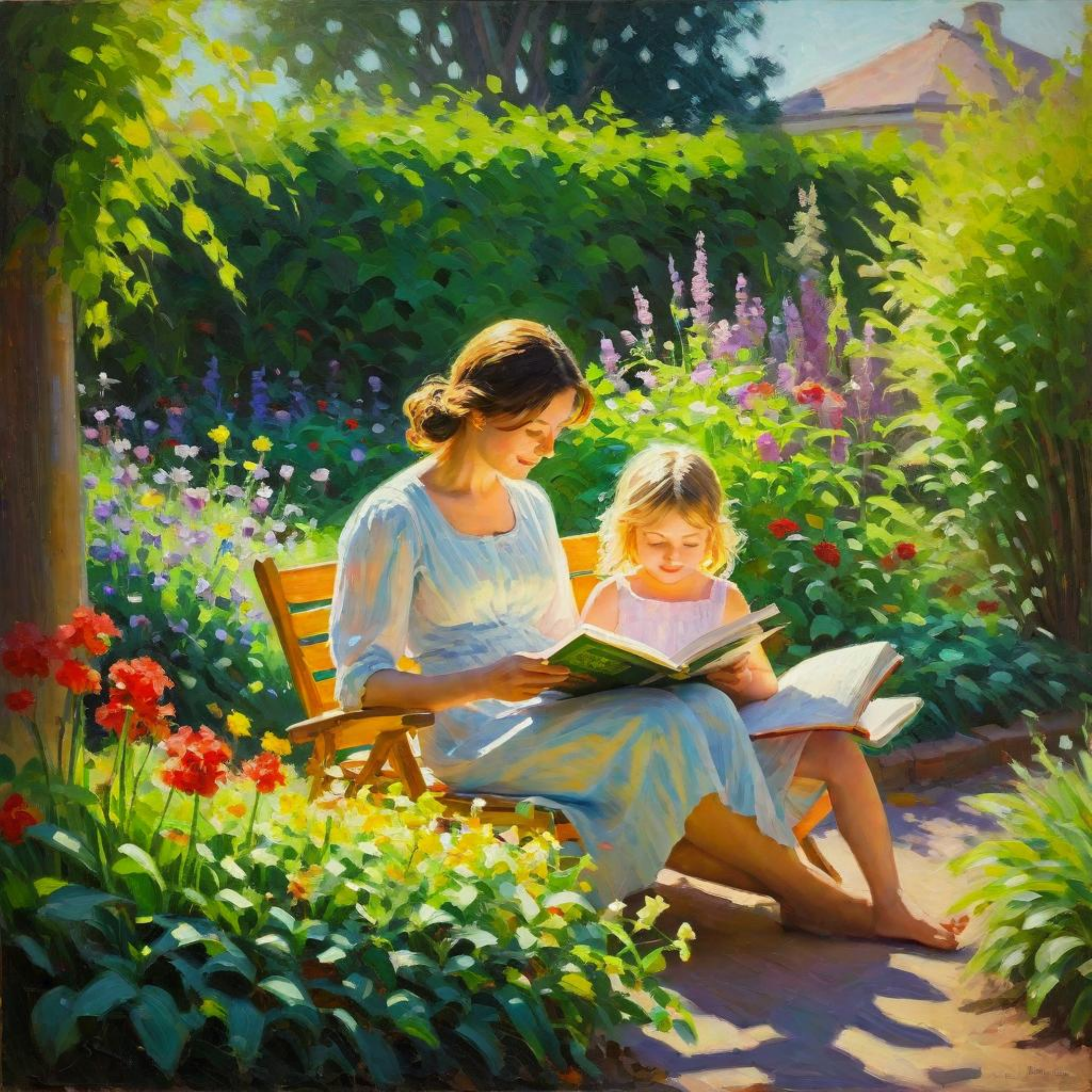} &
    \includegraphics[width=0.15\linewidth]{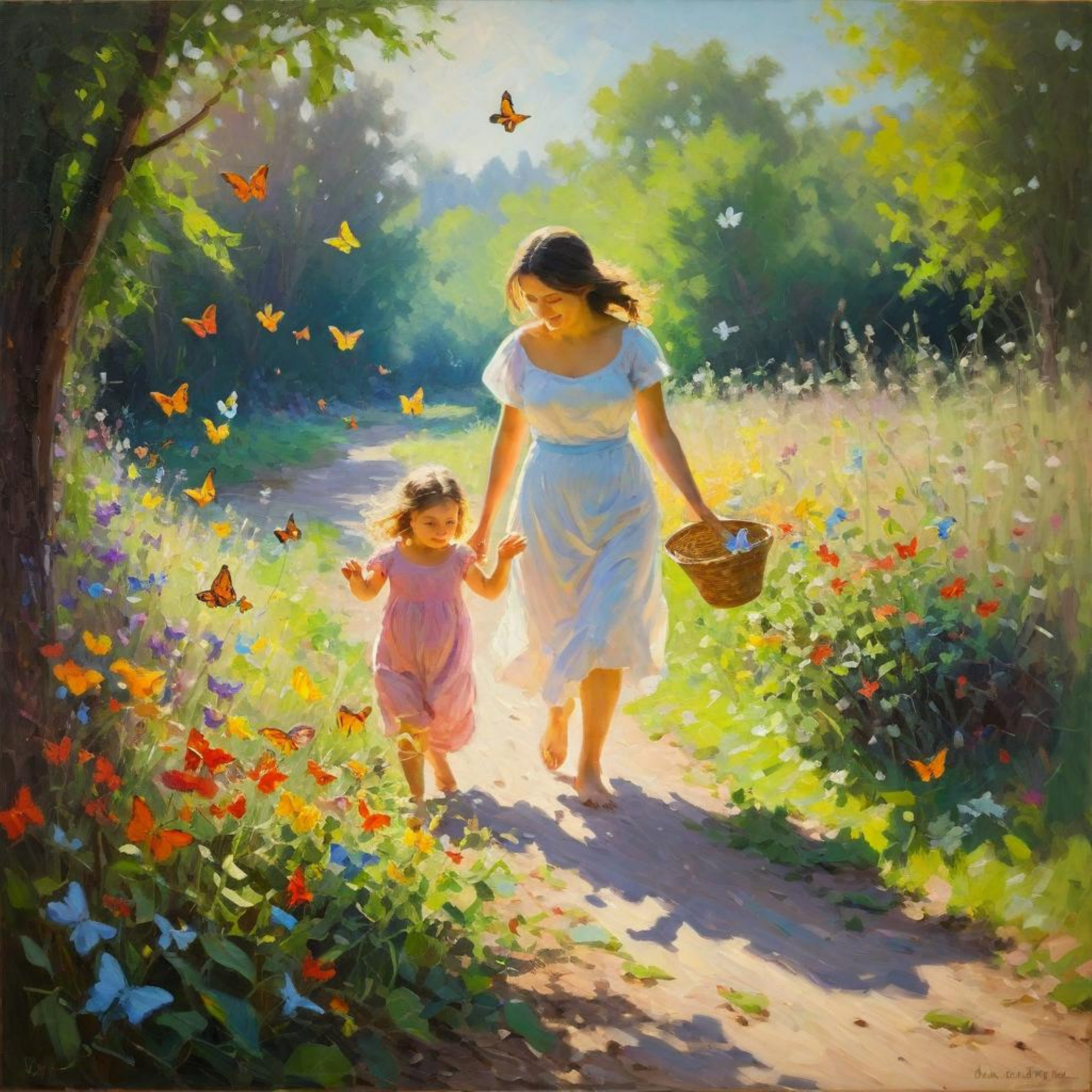} &
\end{tabular}
&
\begin{tabular}{cccc}
    \includegraphics[width=0.15\linewidth]{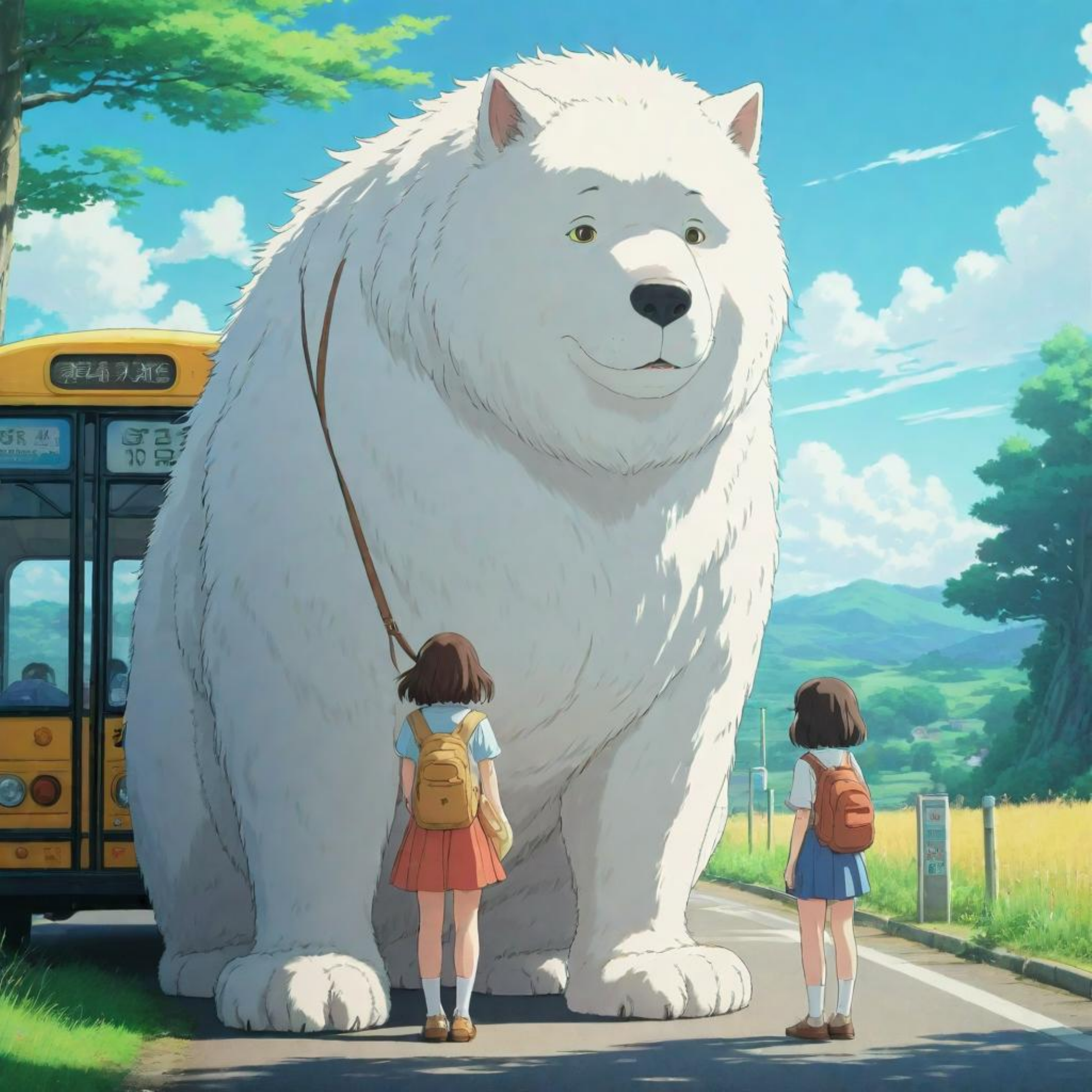} &
    \includegraphics[width=0.15\linewidth]{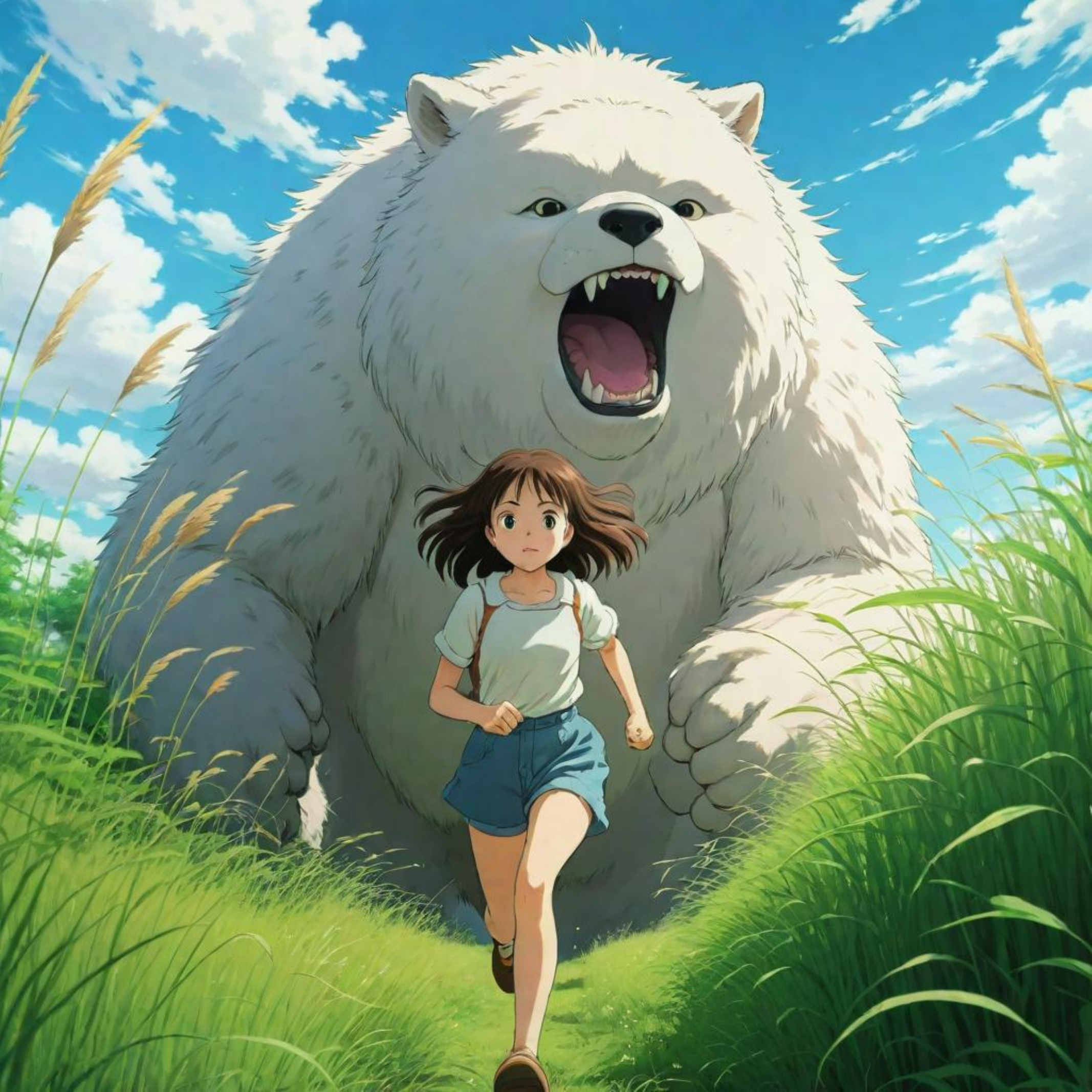} &
    \includegraphics[width=0.15\linewidth]{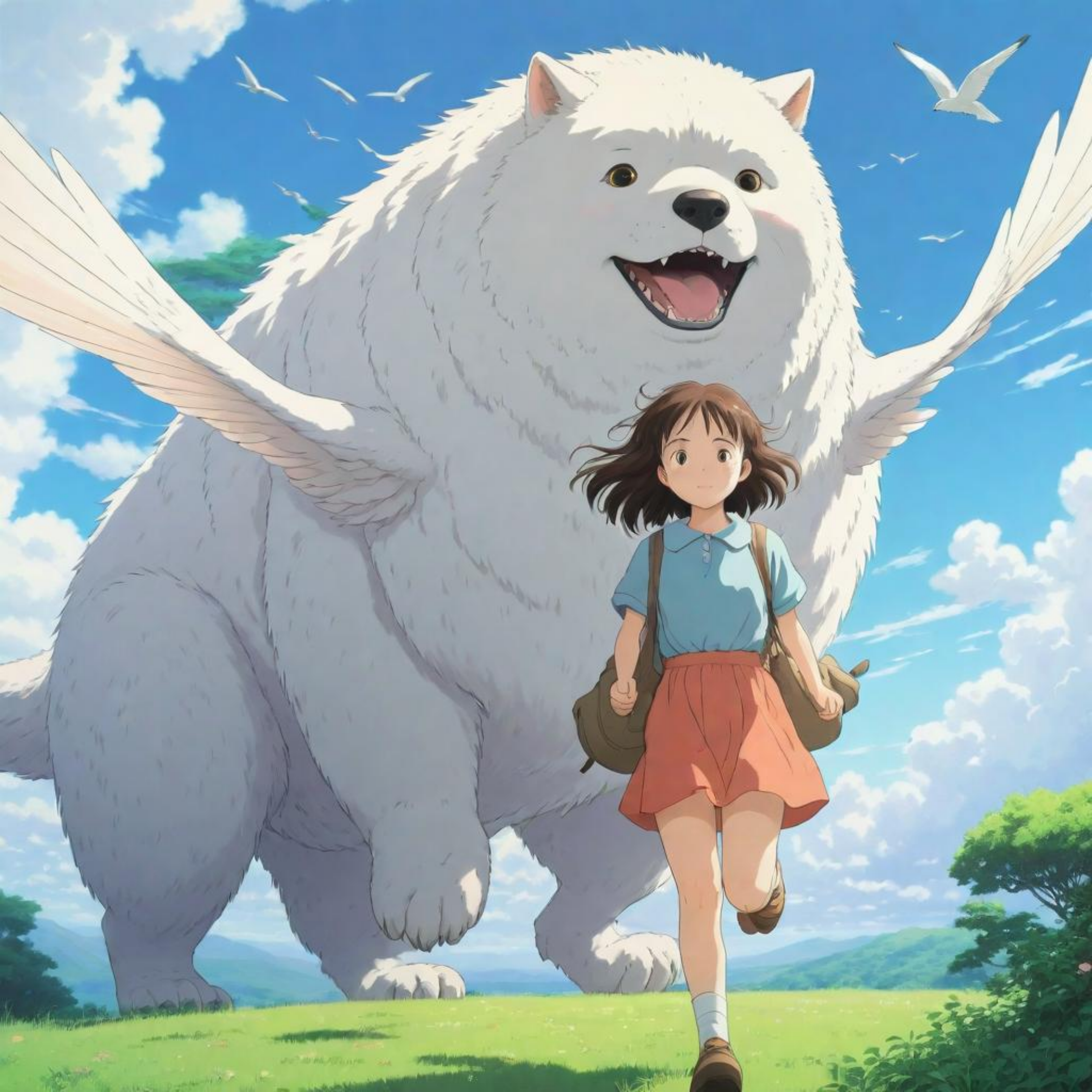}
\end{tabular}
\\
\vspace{-13.5mm}\rotatebox[origin=c]{90}{\small 1P1S} &
\begin{tabular}{cccc}
    \includegraphics[width=0.15\linewidth]{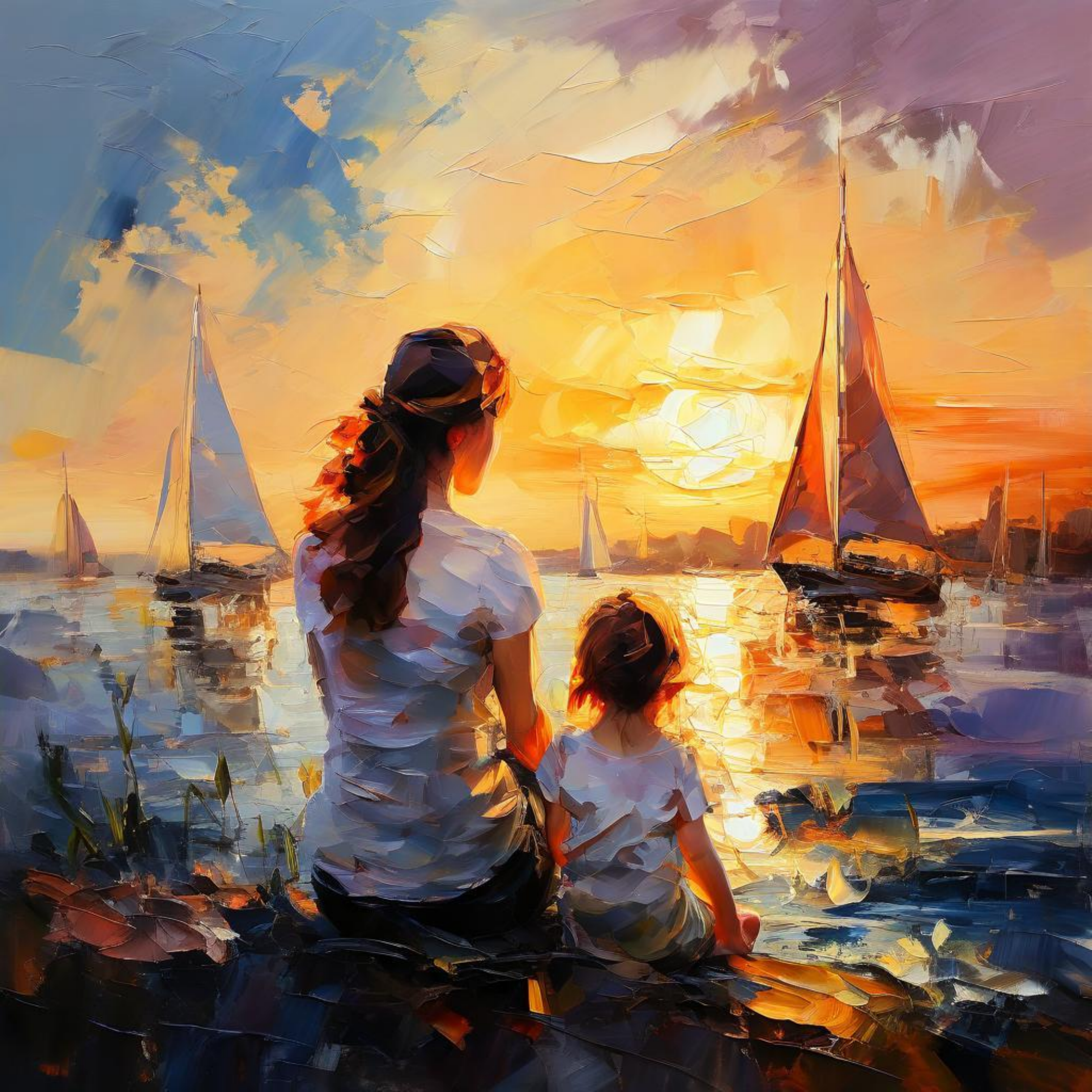} &
    \includegraphics[width=0.15\linewidth]{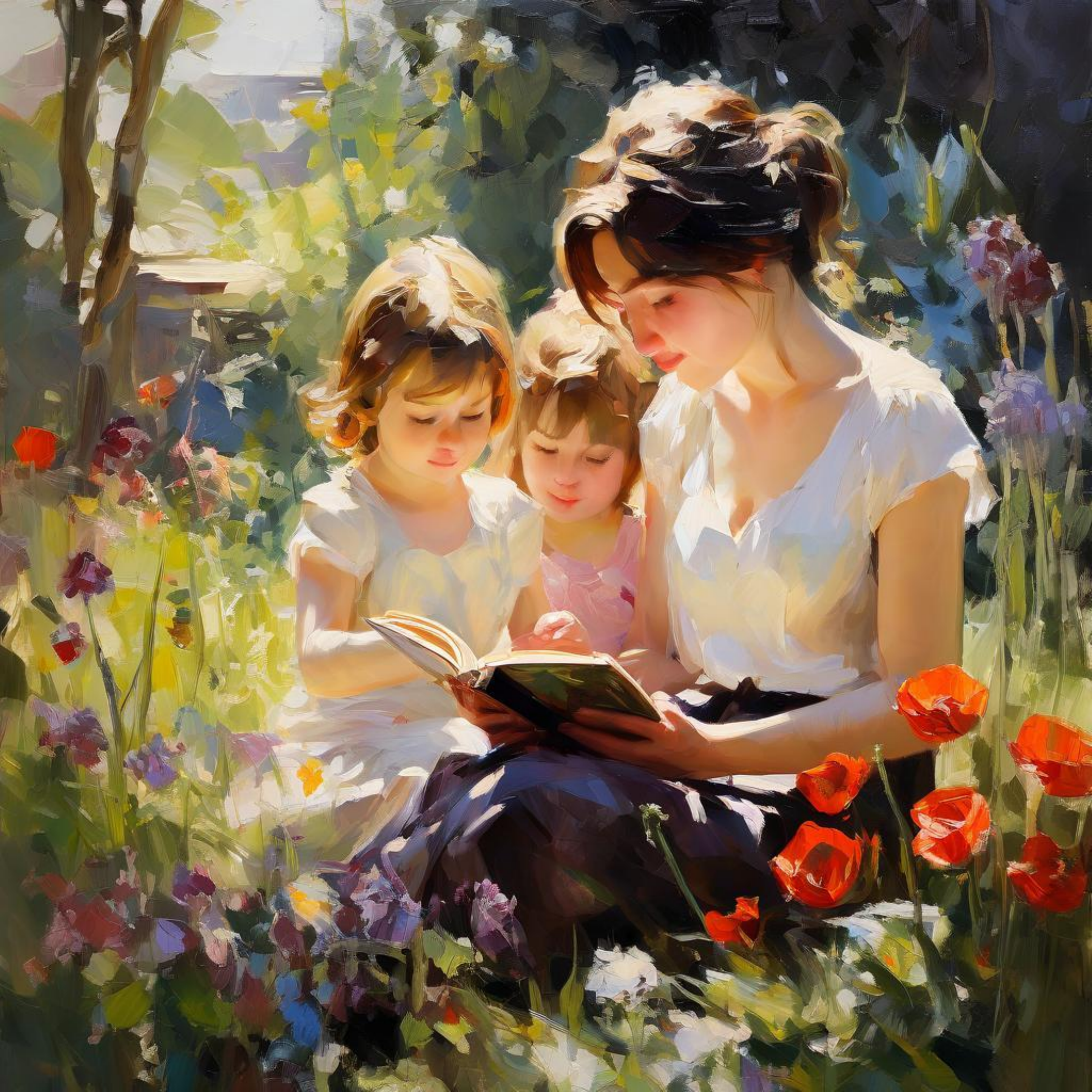} &
    \includegraphics[width=0.15\linewidth]{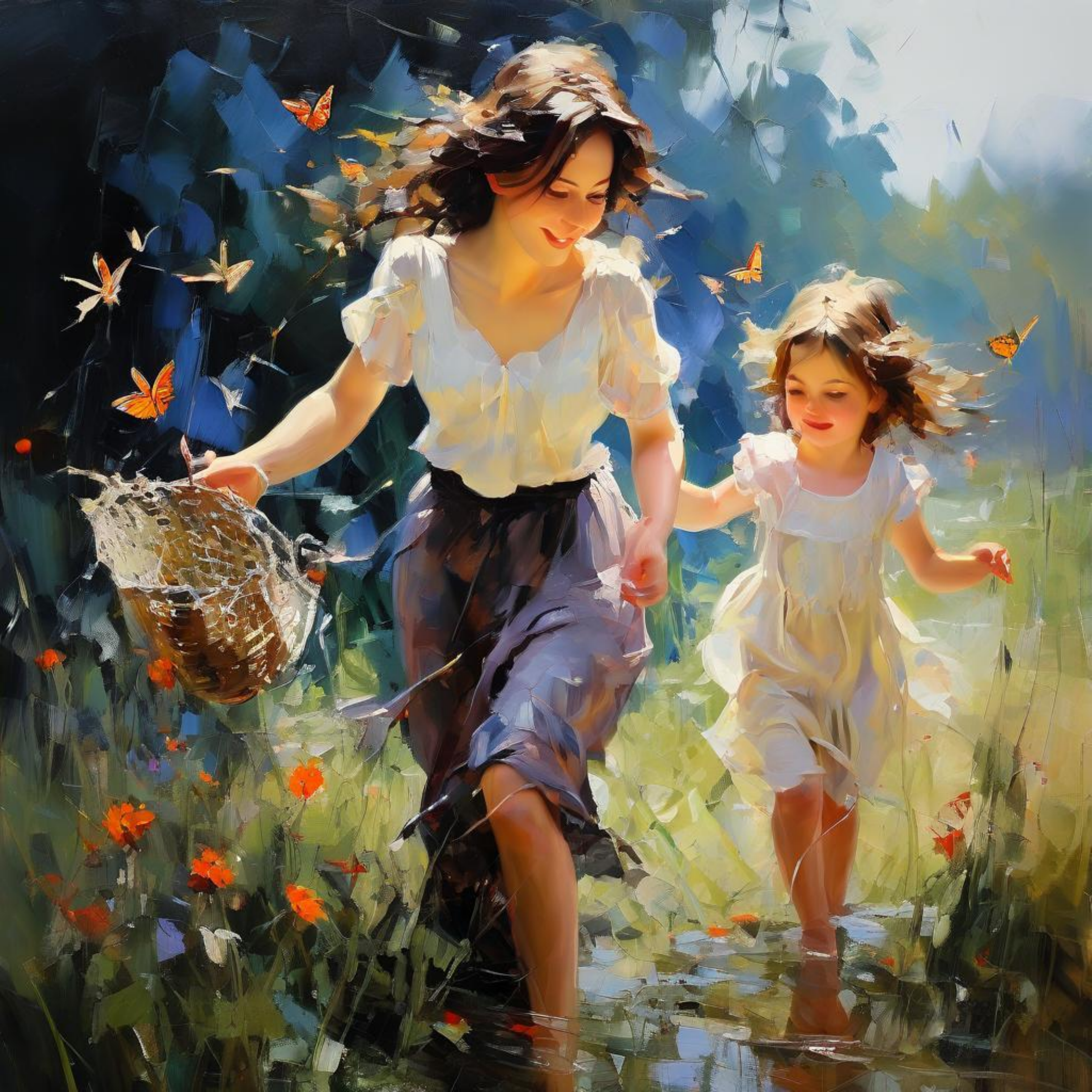} &
\end{tabular}
&
\begin{tabular}{cccc}
    \includegraphics[width=0.15\linewidth]{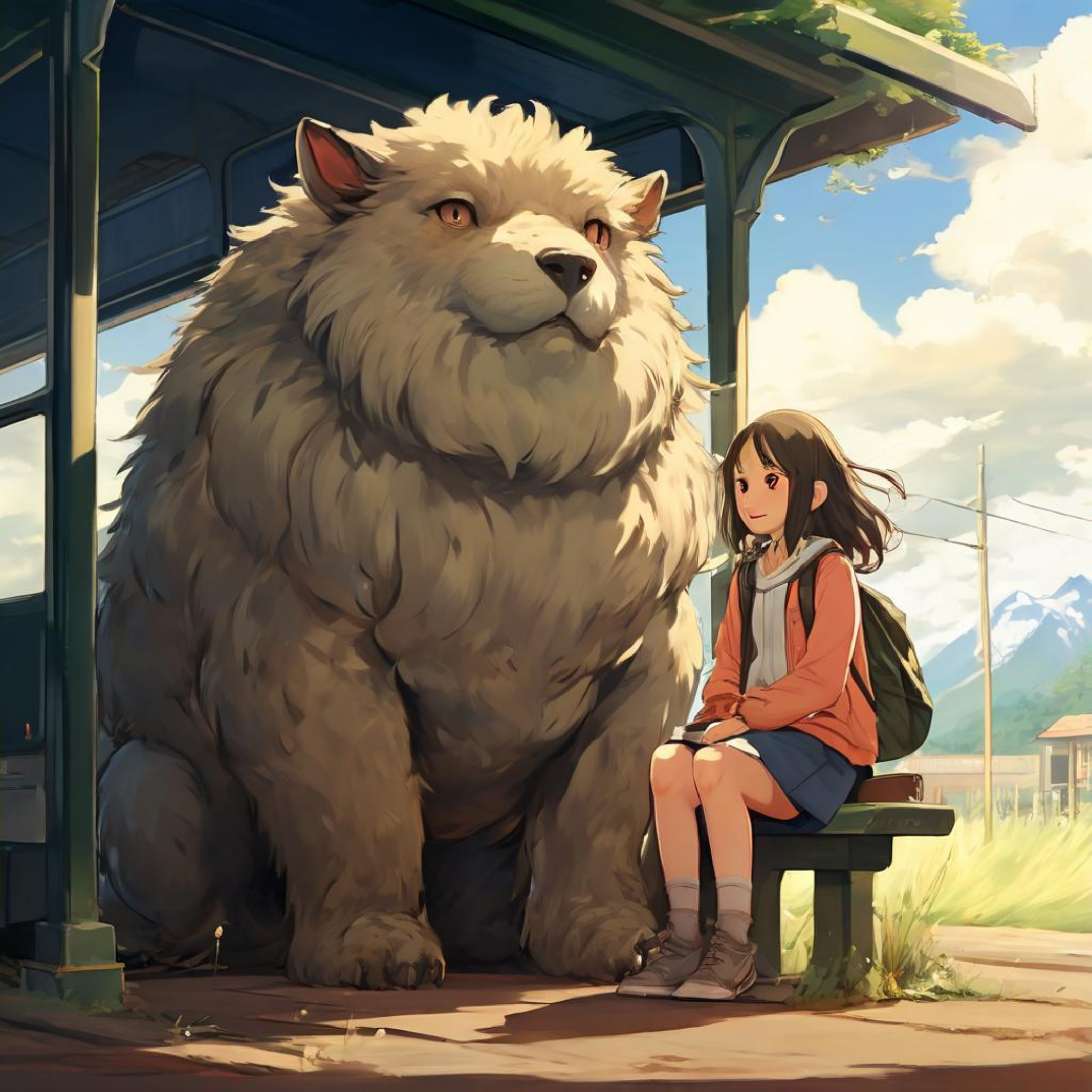} &
    \includegraphics[width=0.15\linewidth]{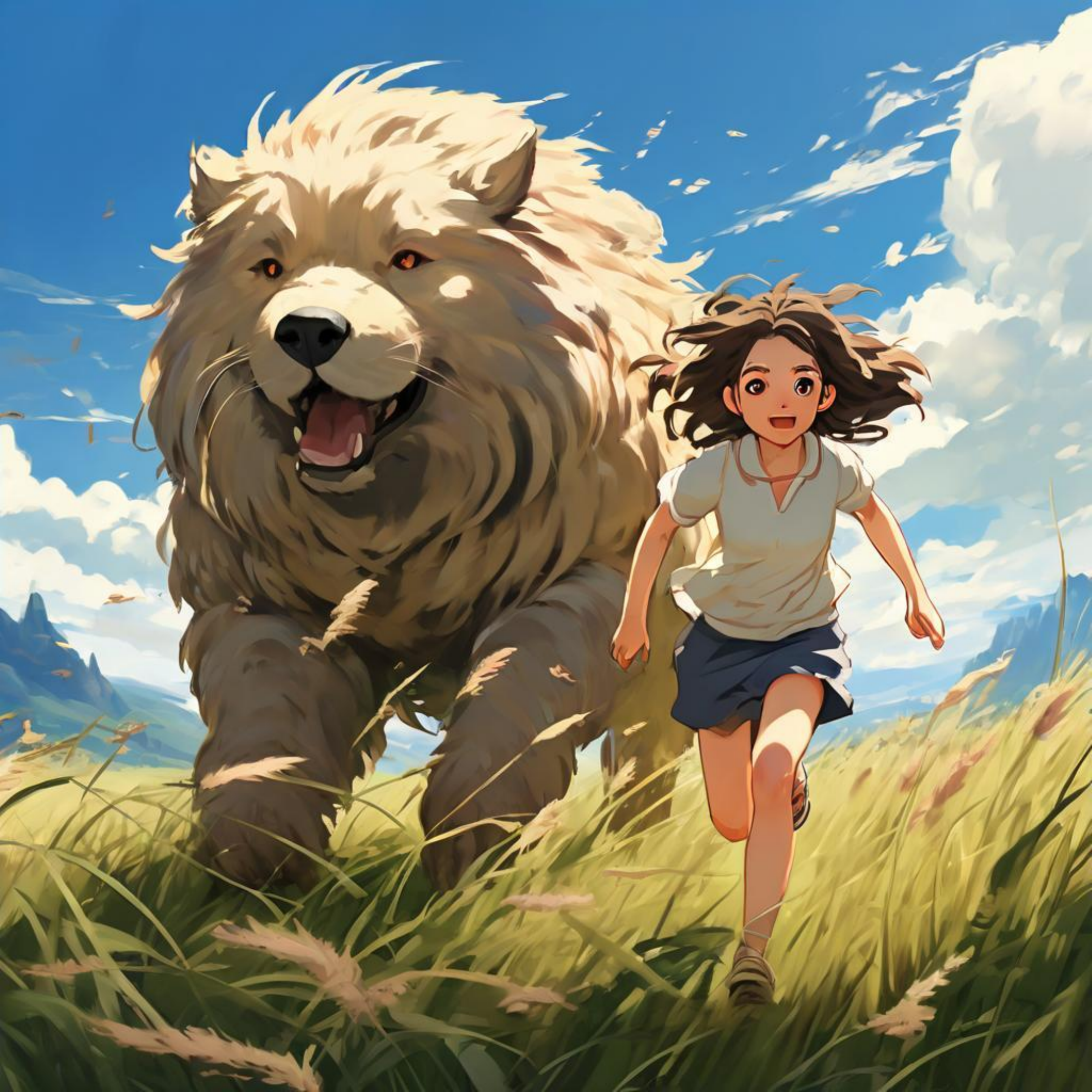} &
    \includegraphics[width=0.15\linewidth]{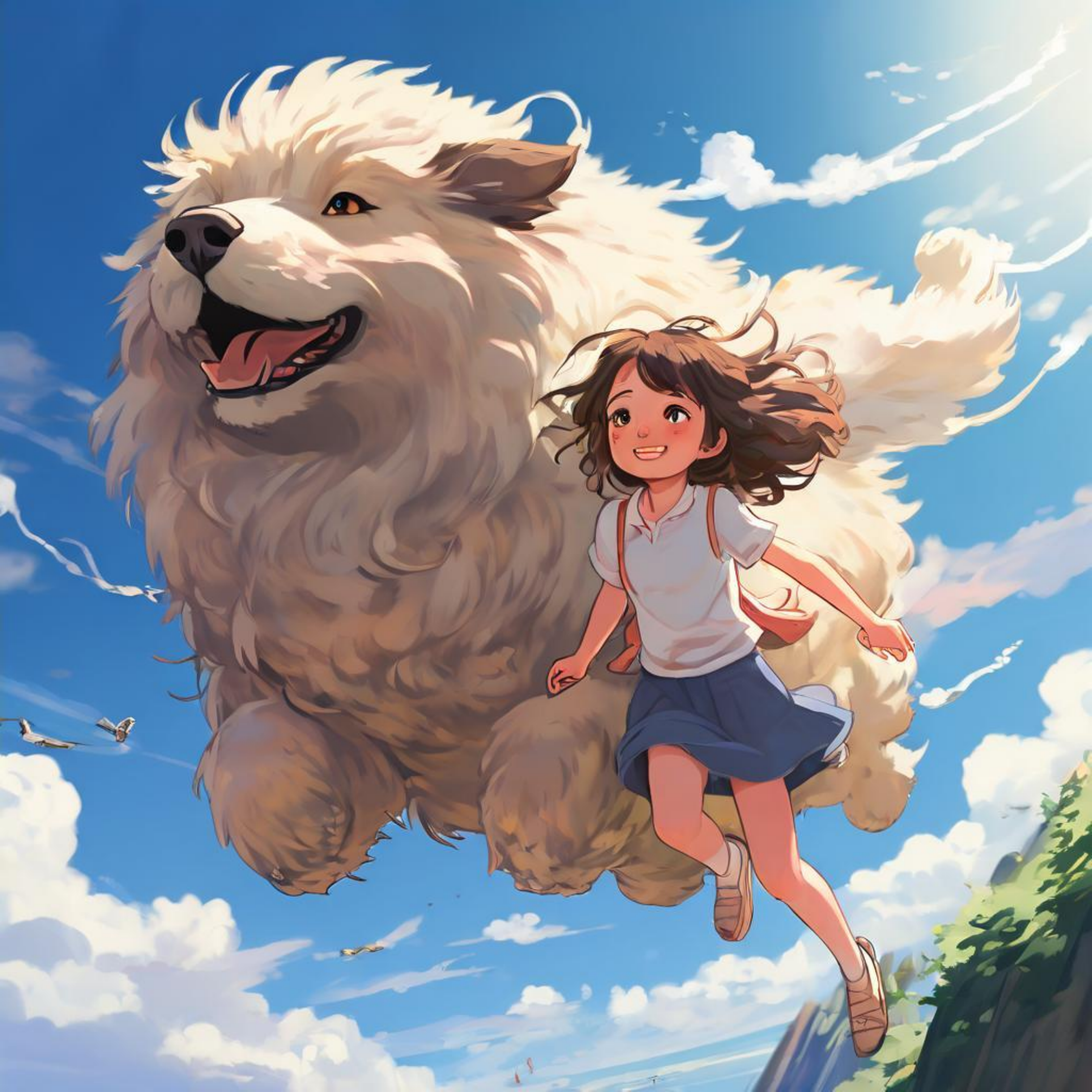}
\end{tabular}
\\ 
\specialrule{\heavyrulewidth}{2pt}{2pt}

%---------------------------------------------------
% OUR MODEL: RealDiffusion
%---------------------------------------------------
\vspace{-13.5mm}\rotatebox[origin=c]{90}{\small\textbf{Ours}} &
\begin{tabular}{cccc}
    \includegraphics[width=0.15\linewidth]{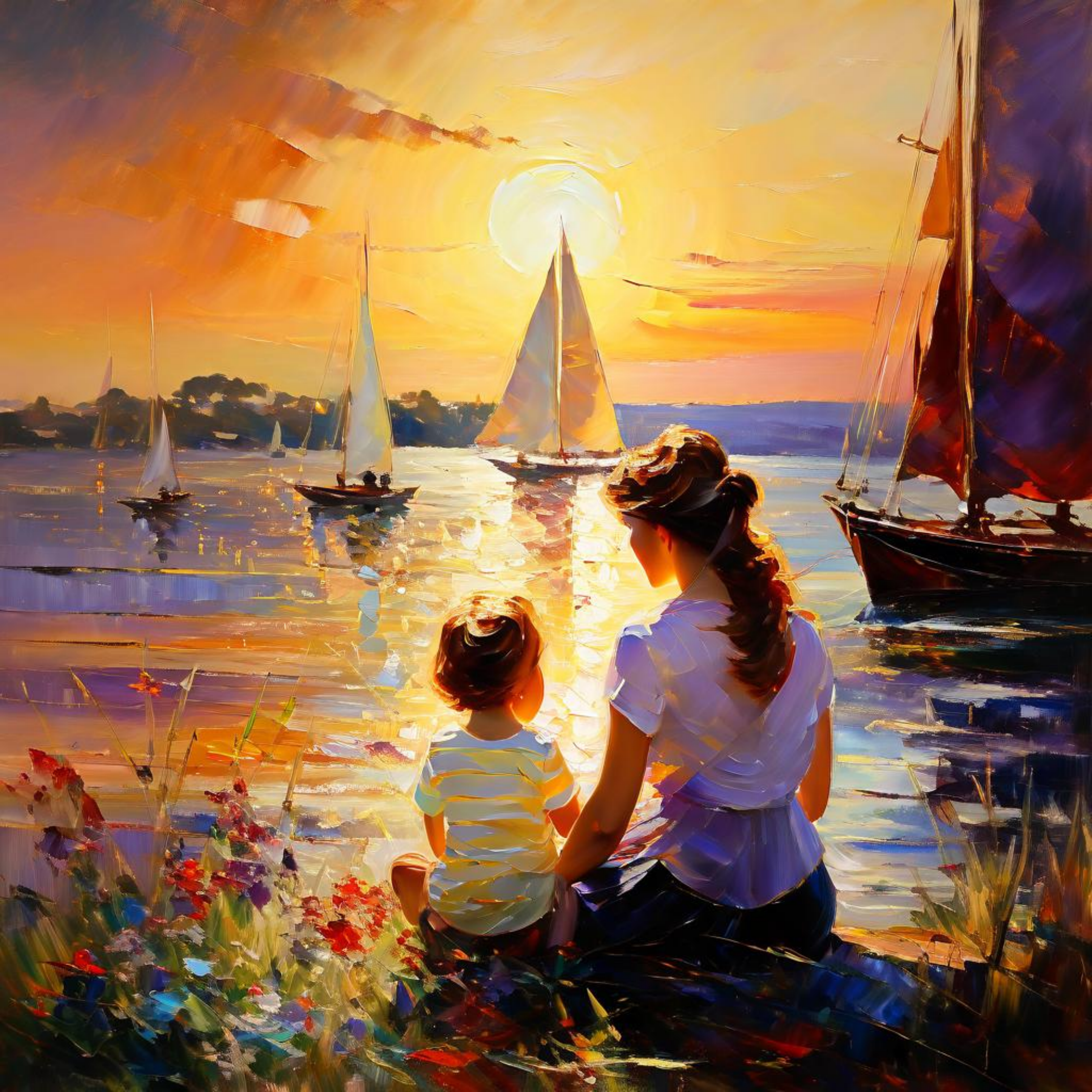} &
    \includegraphics[width=0.15\linewidth]{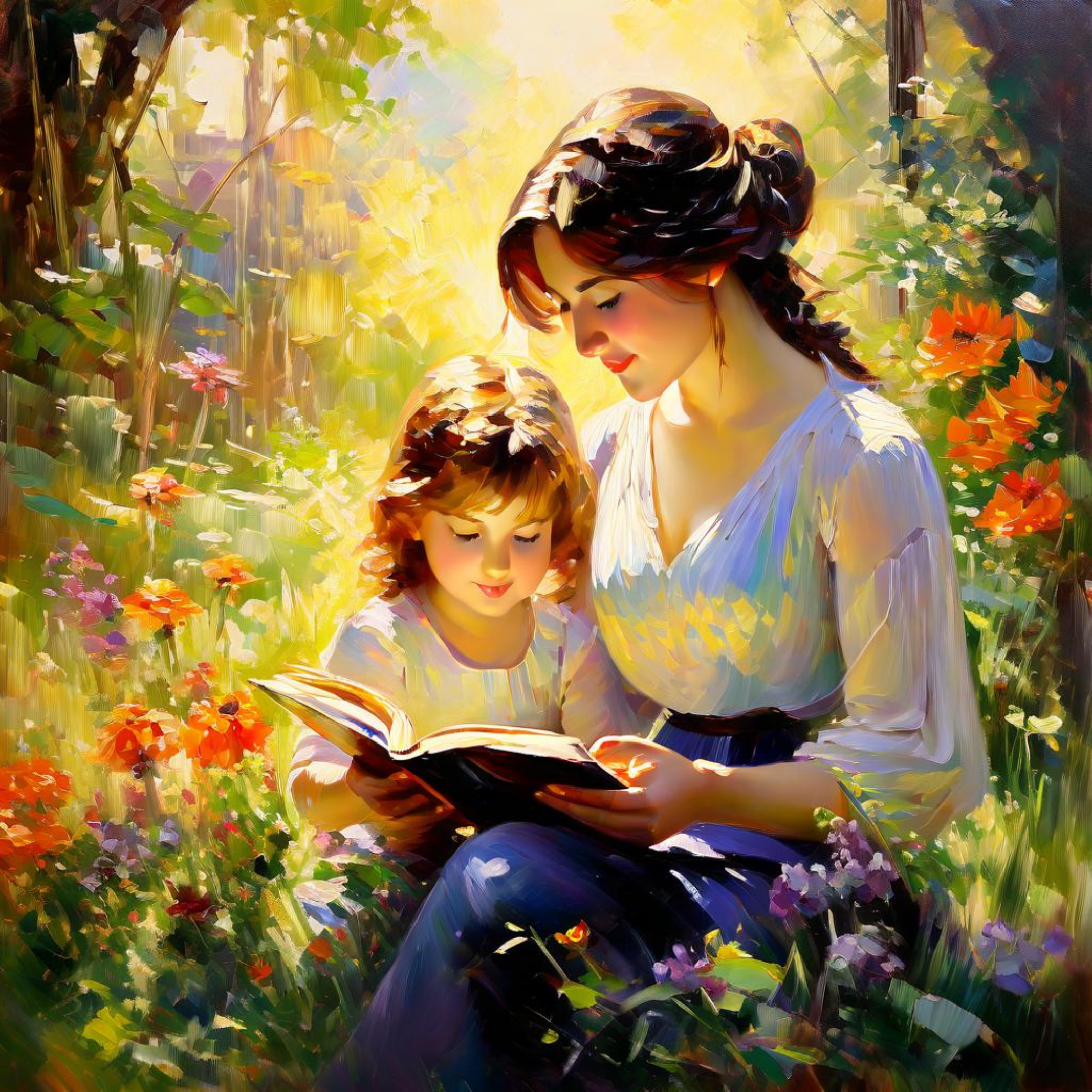} &
    \includegraphics[width=0.15\linewidth]{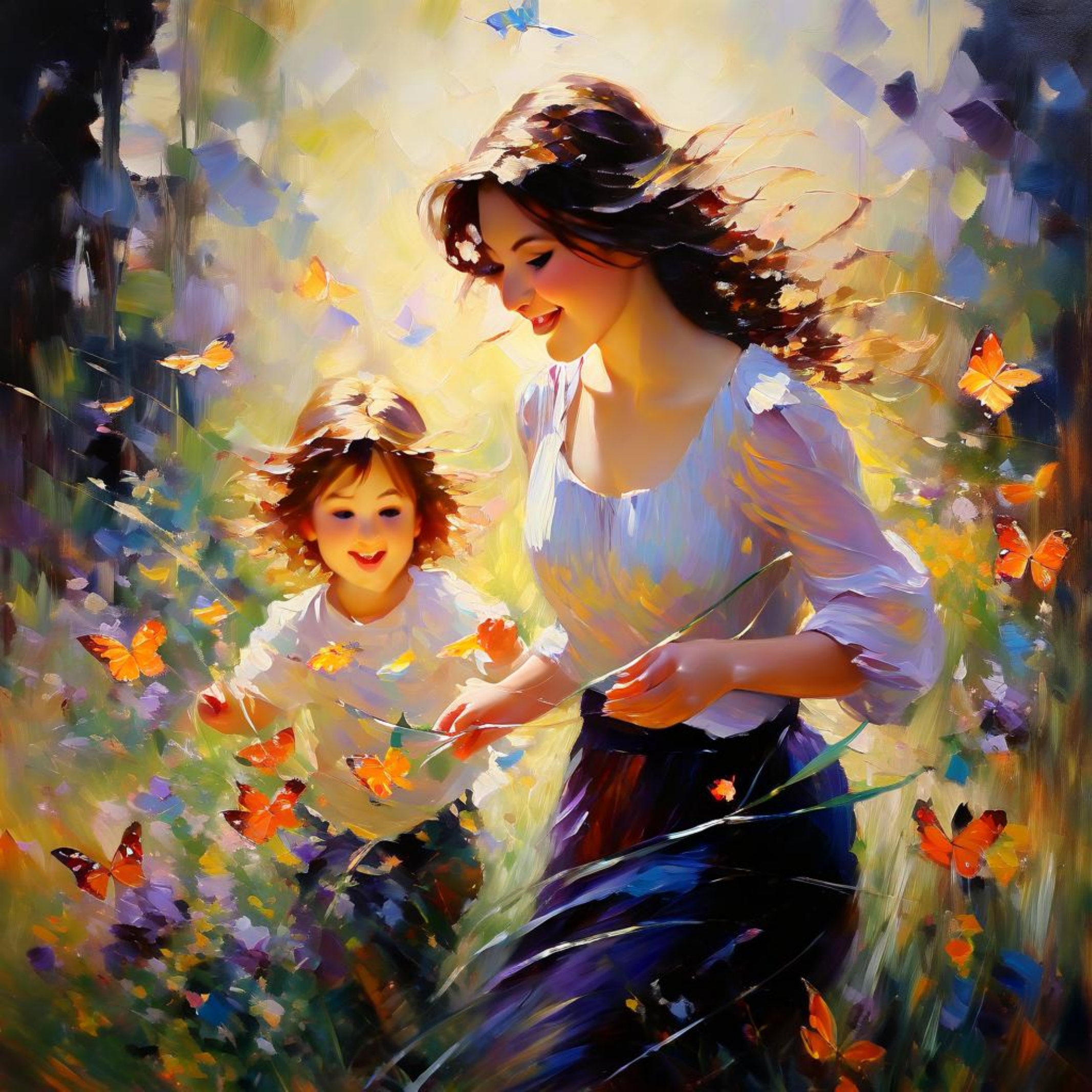} &
\end{tabular}
&
\begin{tabular}{cccc}
    \includegraphics[width=0.15\linewidth]{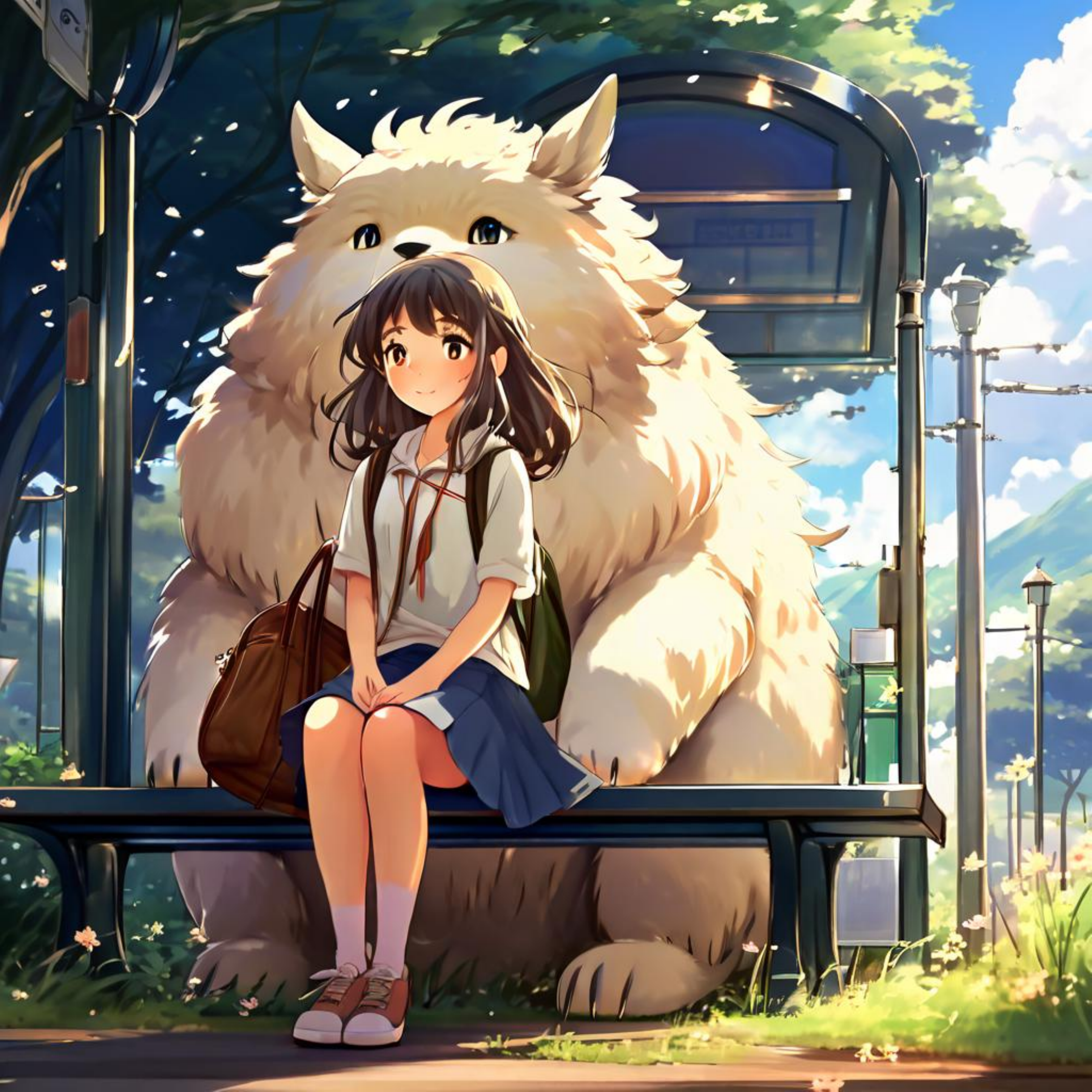} &
    \includegraphics[width=0.15\linewidth]{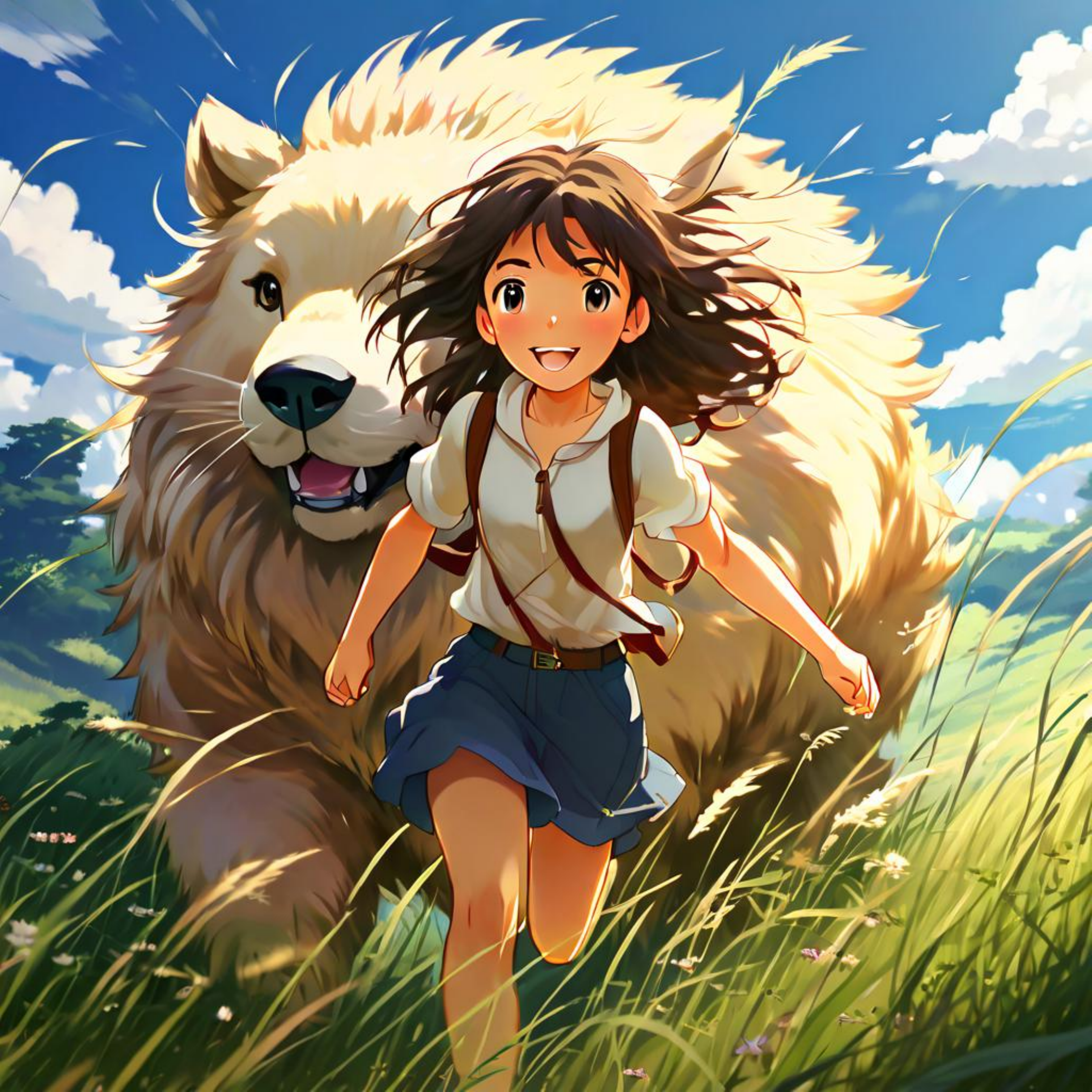} &
    \includegraphics[width=0.15\linewidth]{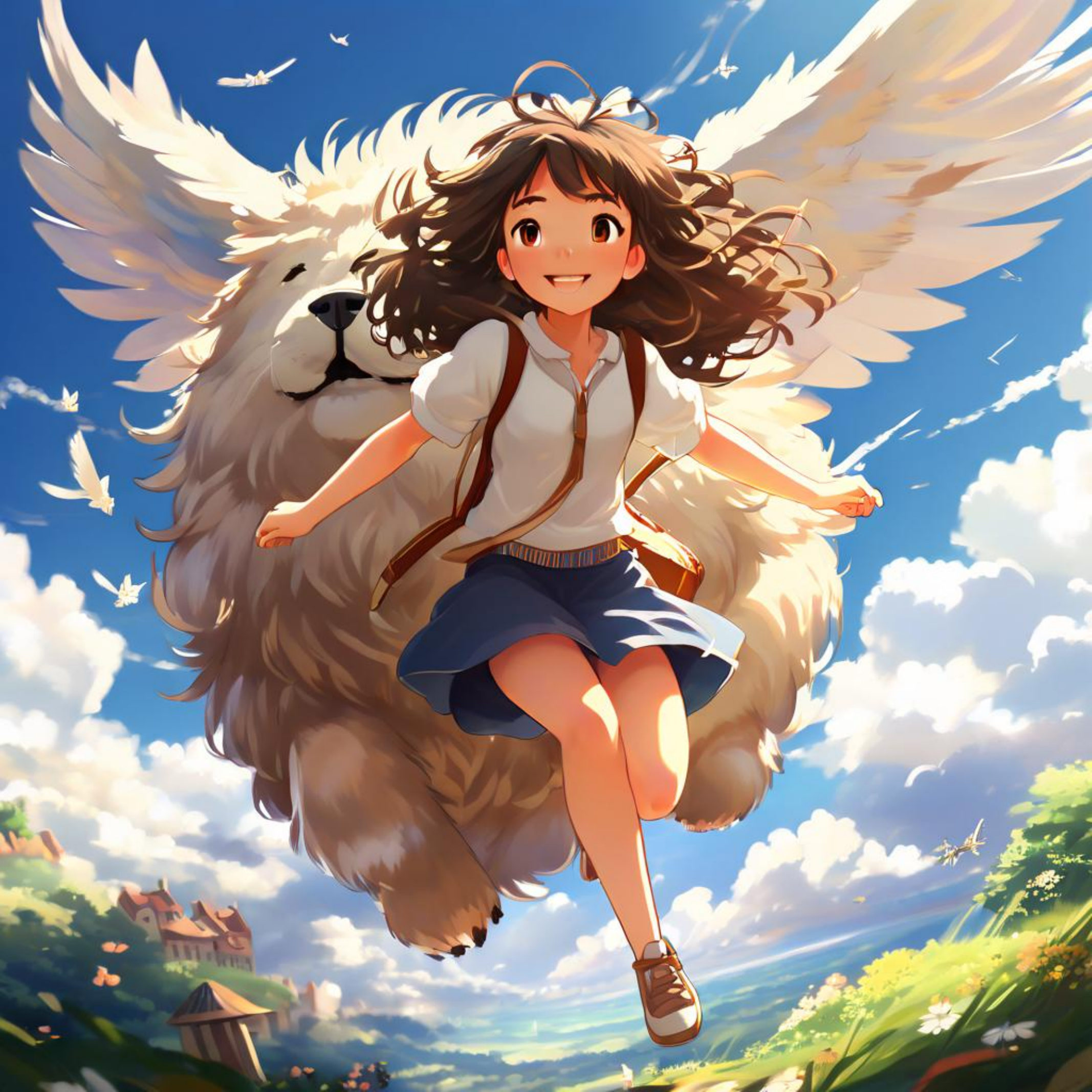}
\end{tabular}
\\

\end{tabular}
\caption{
    A qualitative comparison of our RealDiffusion against five state-of-the-art baseline models.
}
\label{fig:main_comparison}
\end{figure*}

\begin{figure*}[!ht] 
    \centering
    \setlength{\tabcolsep}{1pt}
    \renewcommand{\arraystretch}{0.5}

    \begin{tabular}{cccccc}
        \small Ori & 
        \small Burgers & 
        \small Wave &
        \small Conservation & 
        \small Elasticity & 
        \small \textbf{Ours (Heat Diff.)} \\

        \includegraphics[width=0.15\linewidth]{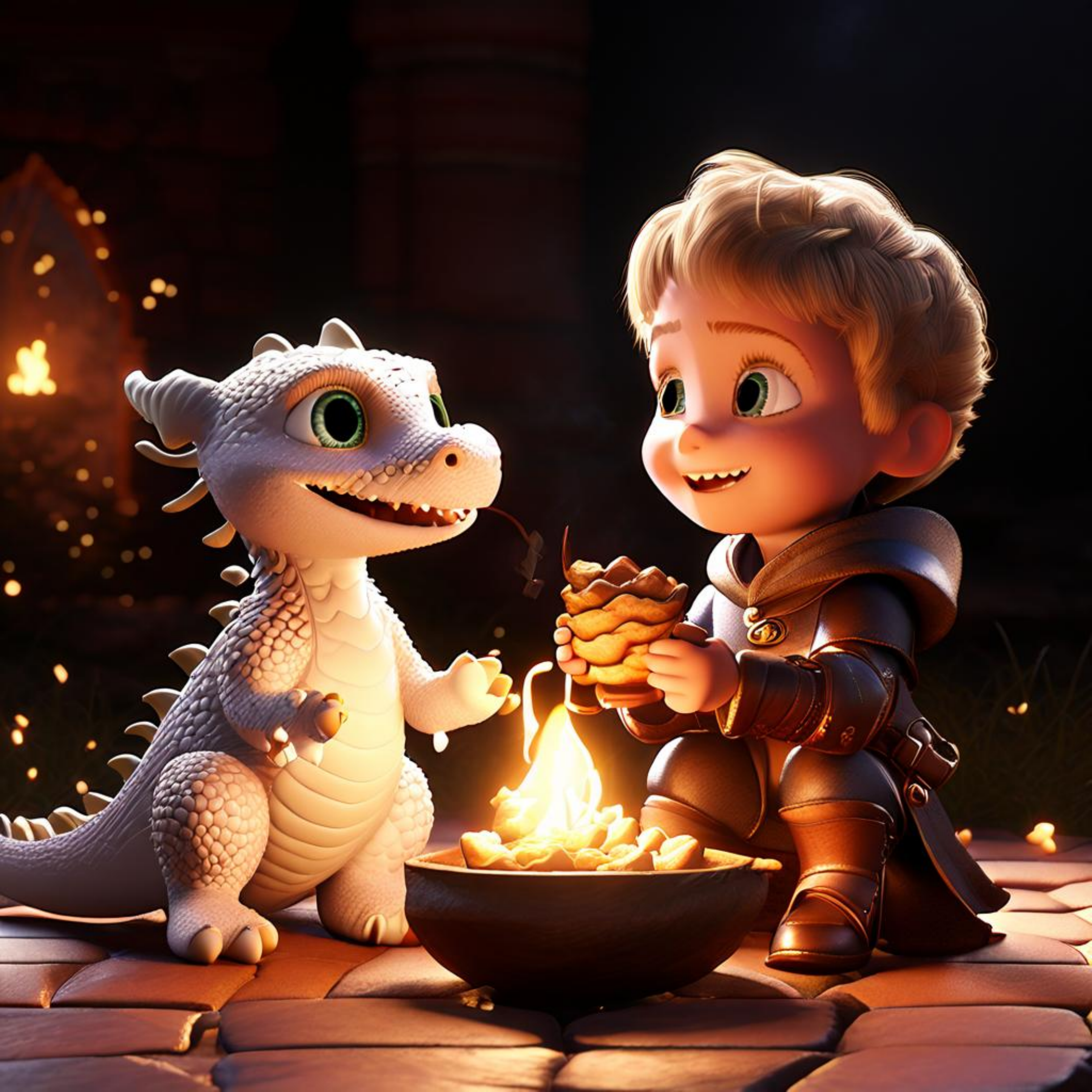} &
        \includegraphics[width=0.15\linewidth]{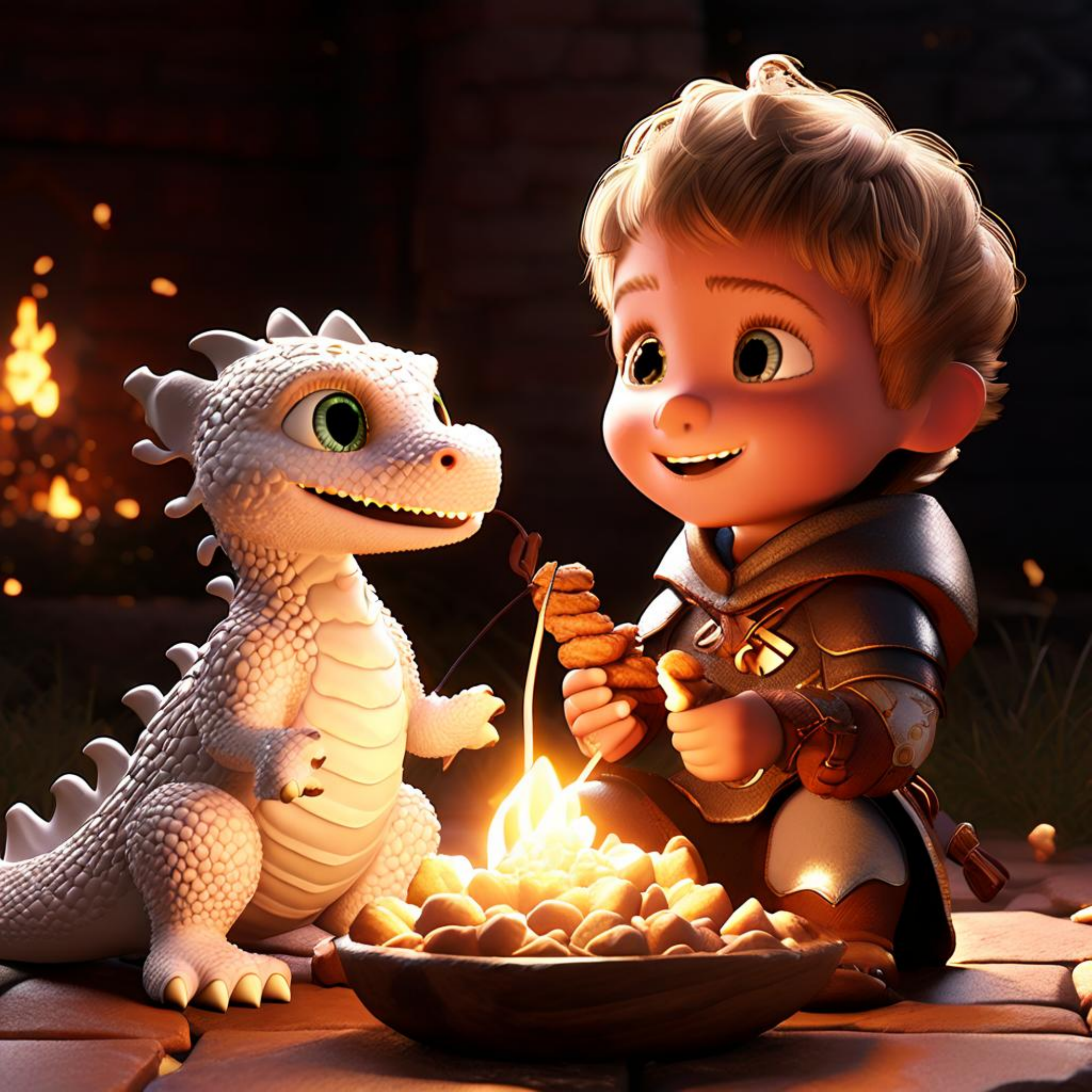} &
        \includegraphics[width=0.15\linewidth]{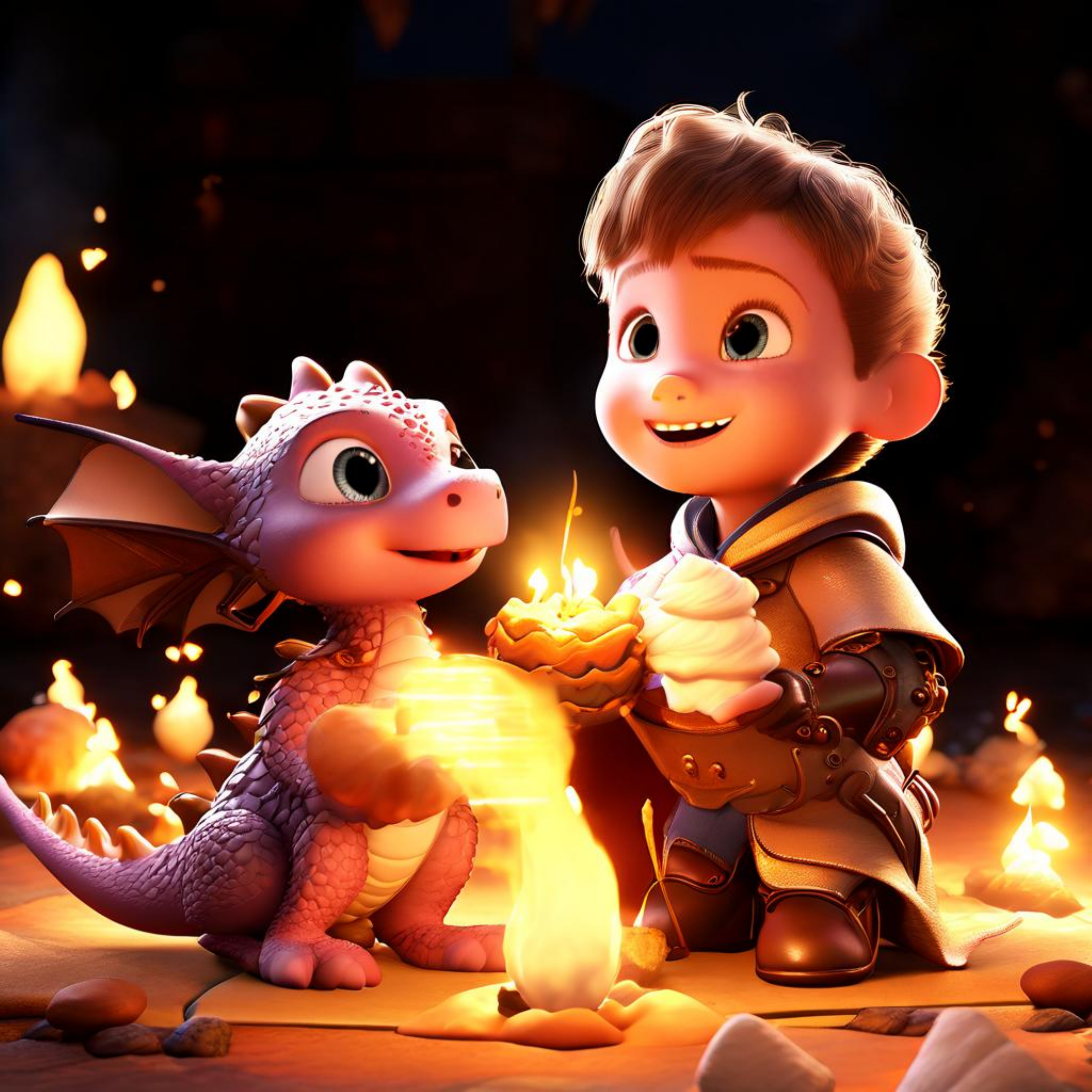} &

        \includegraphics[width=0.15\linewidth]{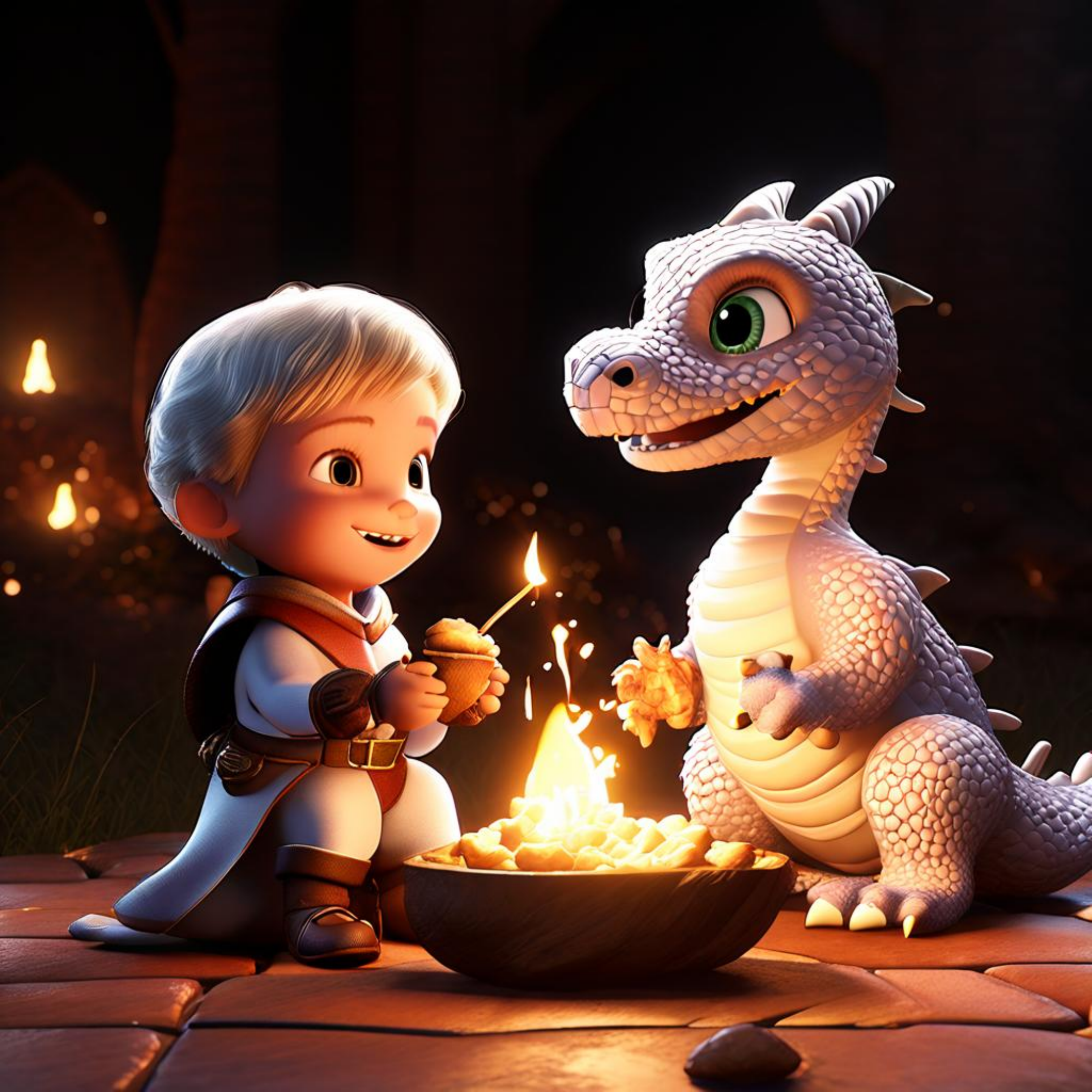} &
        \includegraphics[width=0.15\linewidth]{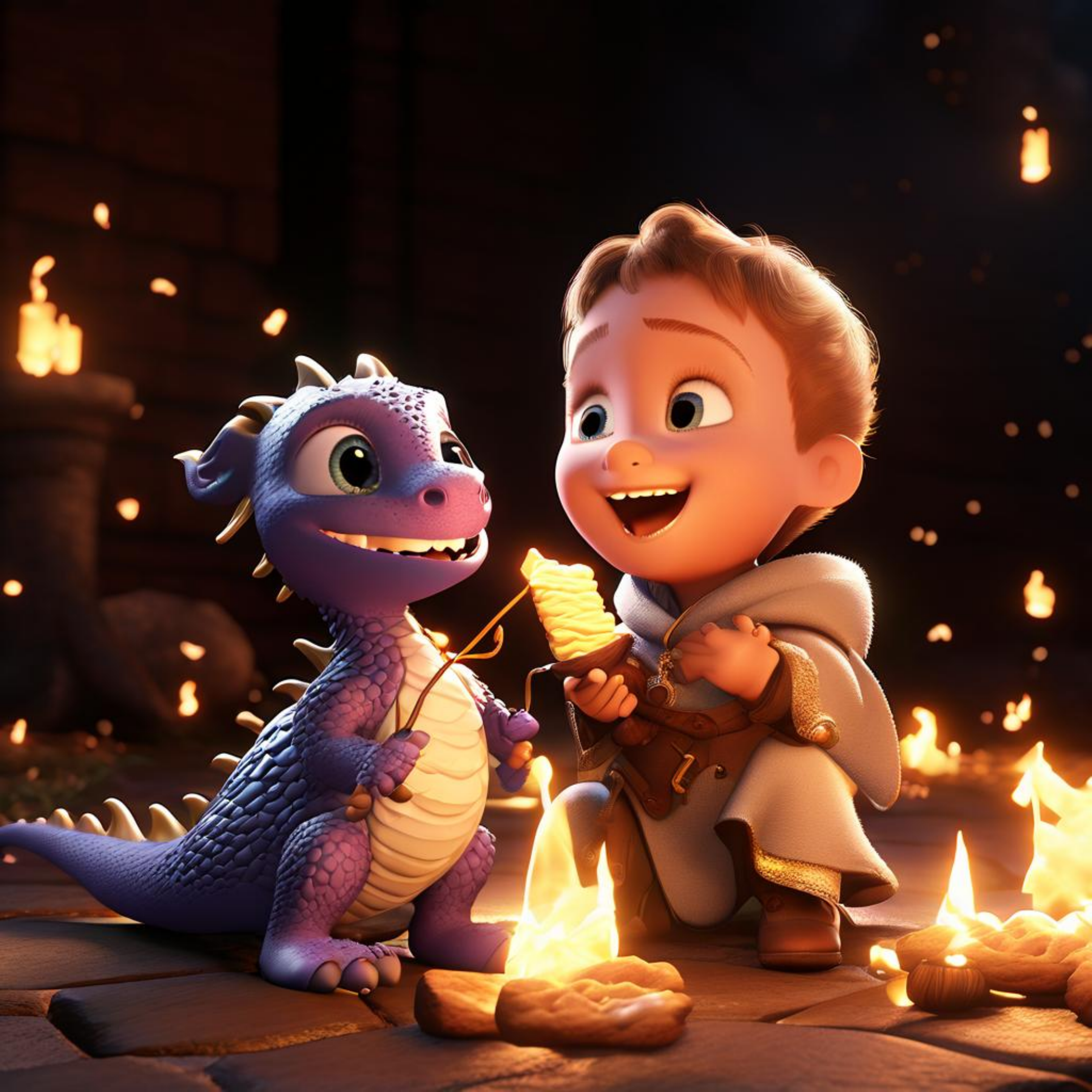} &
        \includegraphics[width=0.15\linewidth]{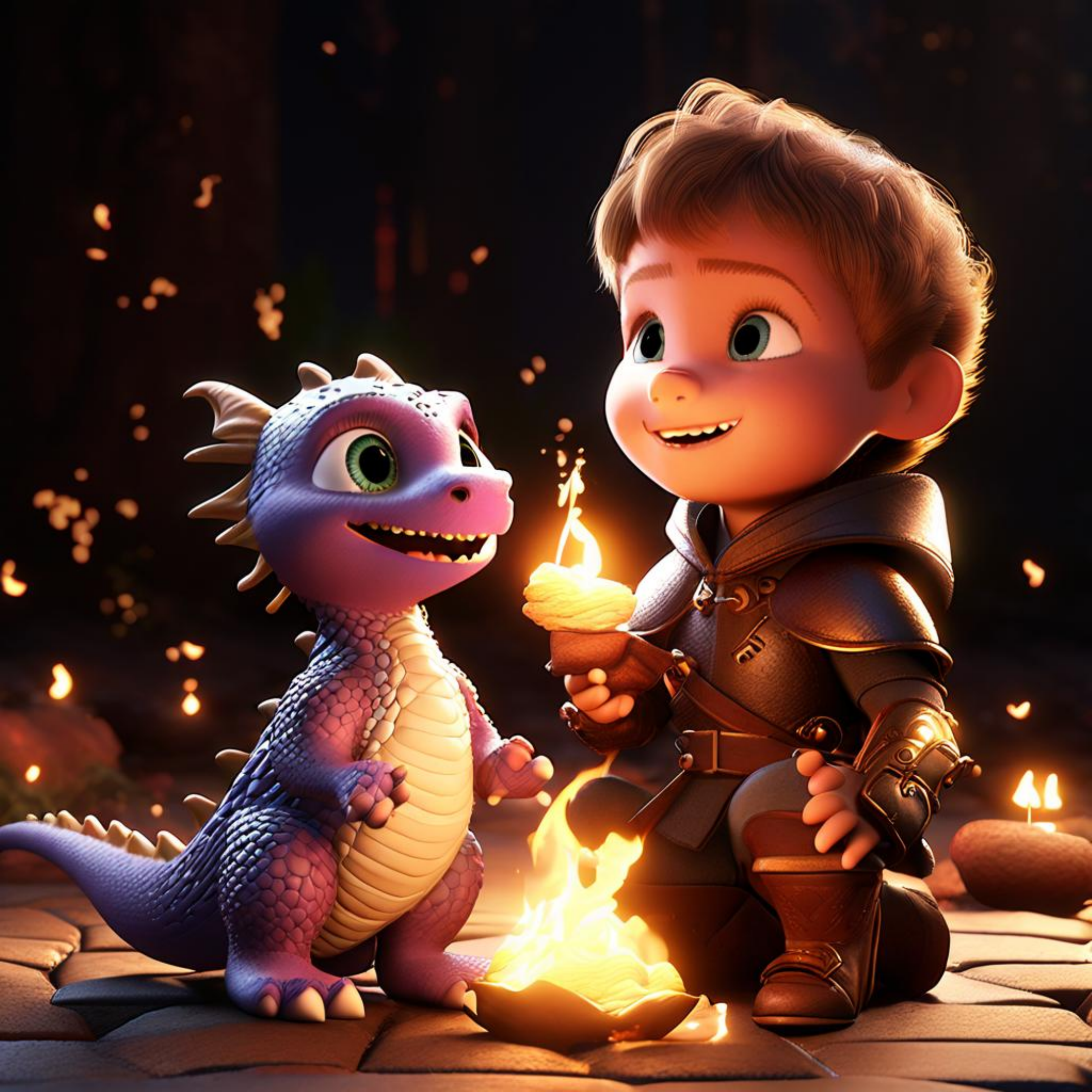} \\

        \includegraphics[width=0.15\linewidth]{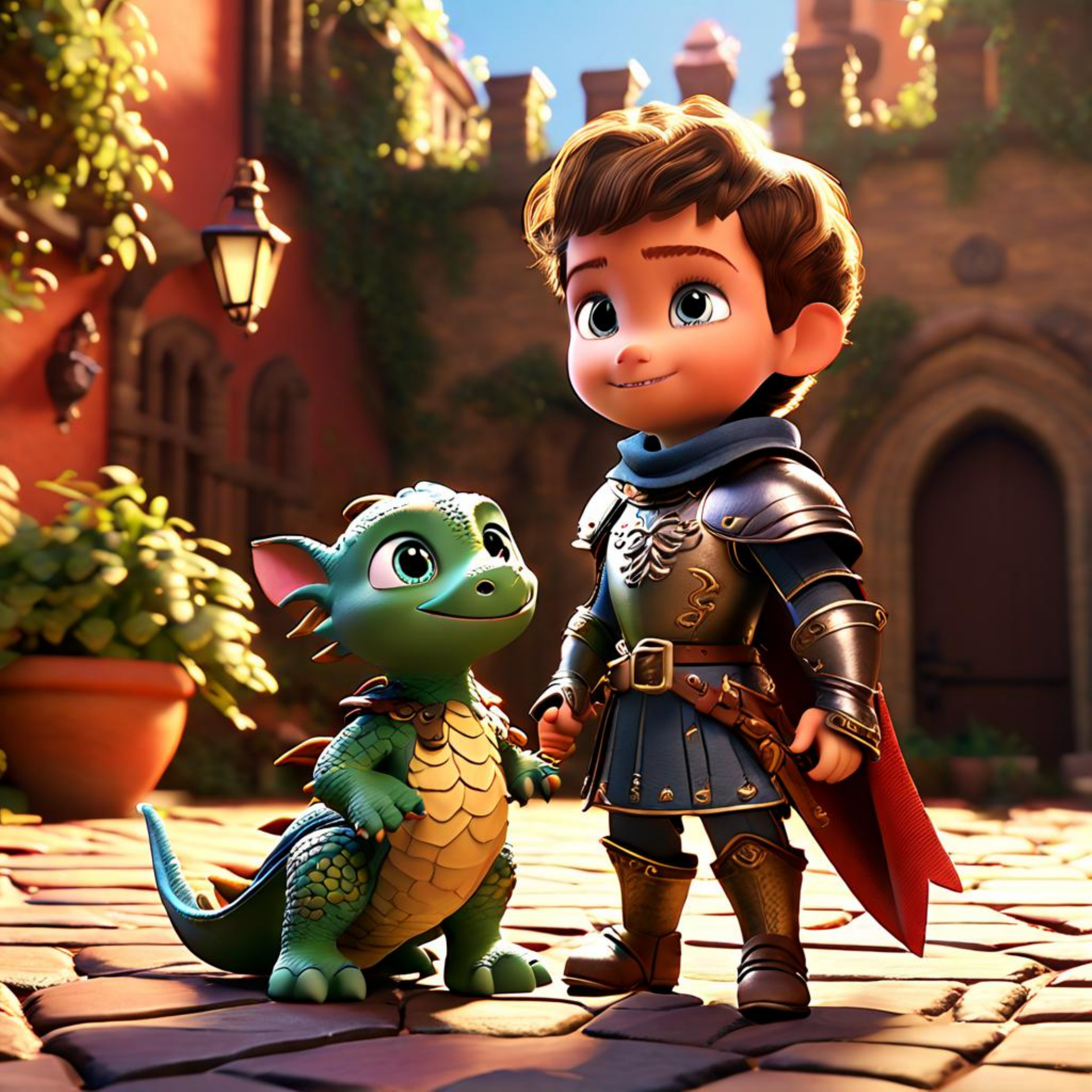} &
        \includegraphics[width=0.15\linewidth]{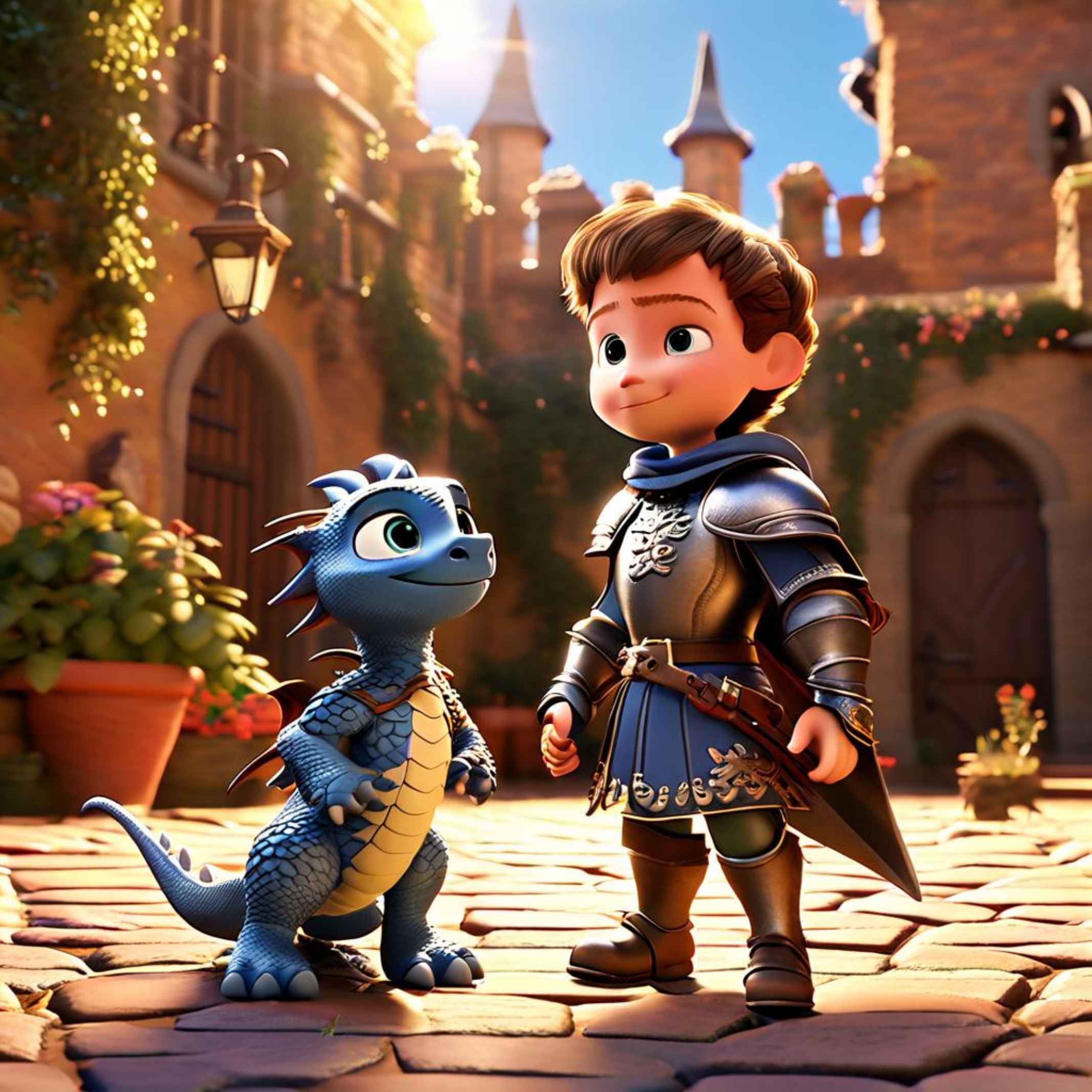} &
        \includegraphics[width=0.15\linewidth]{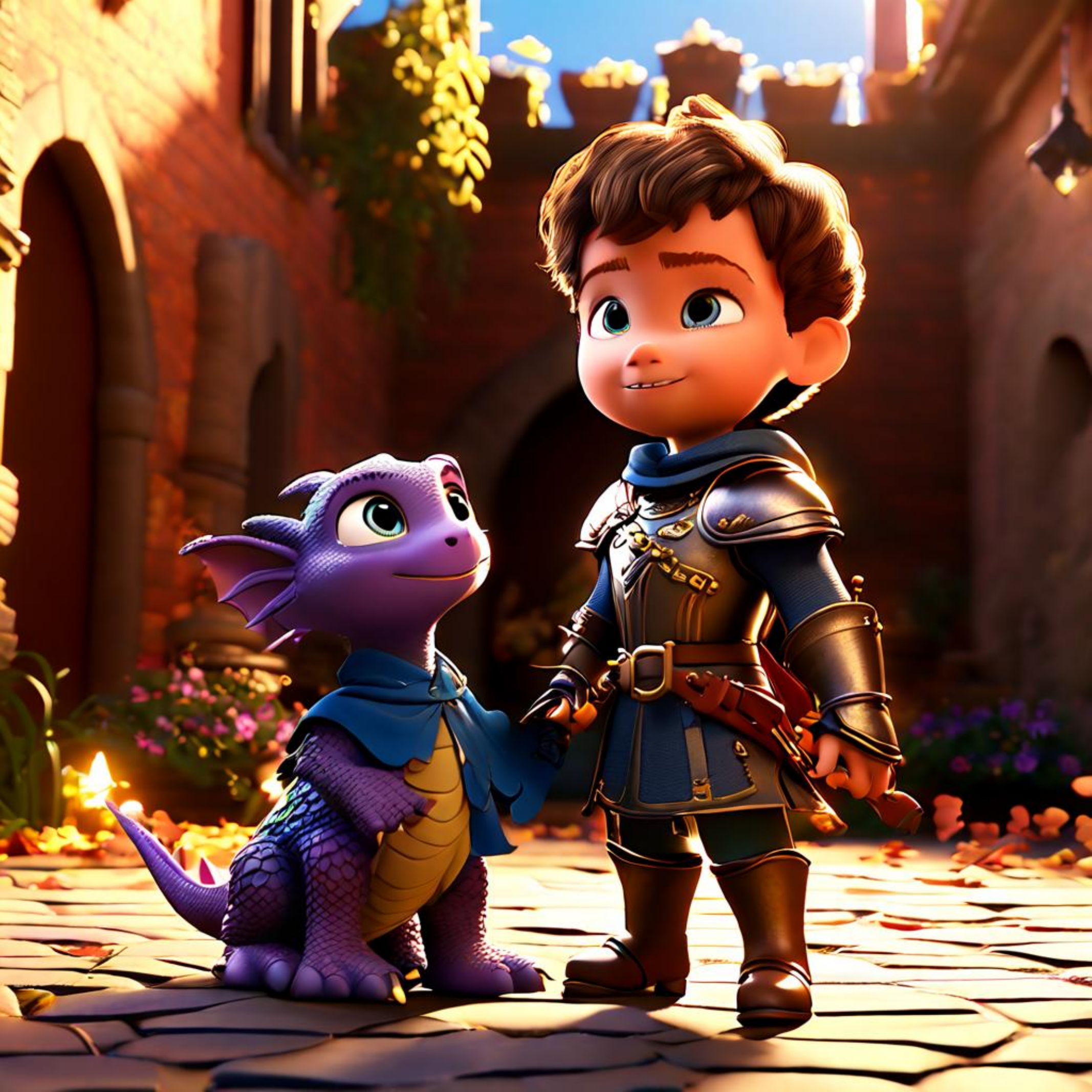} &

        \includegraphics[width=0.15\linewidth]{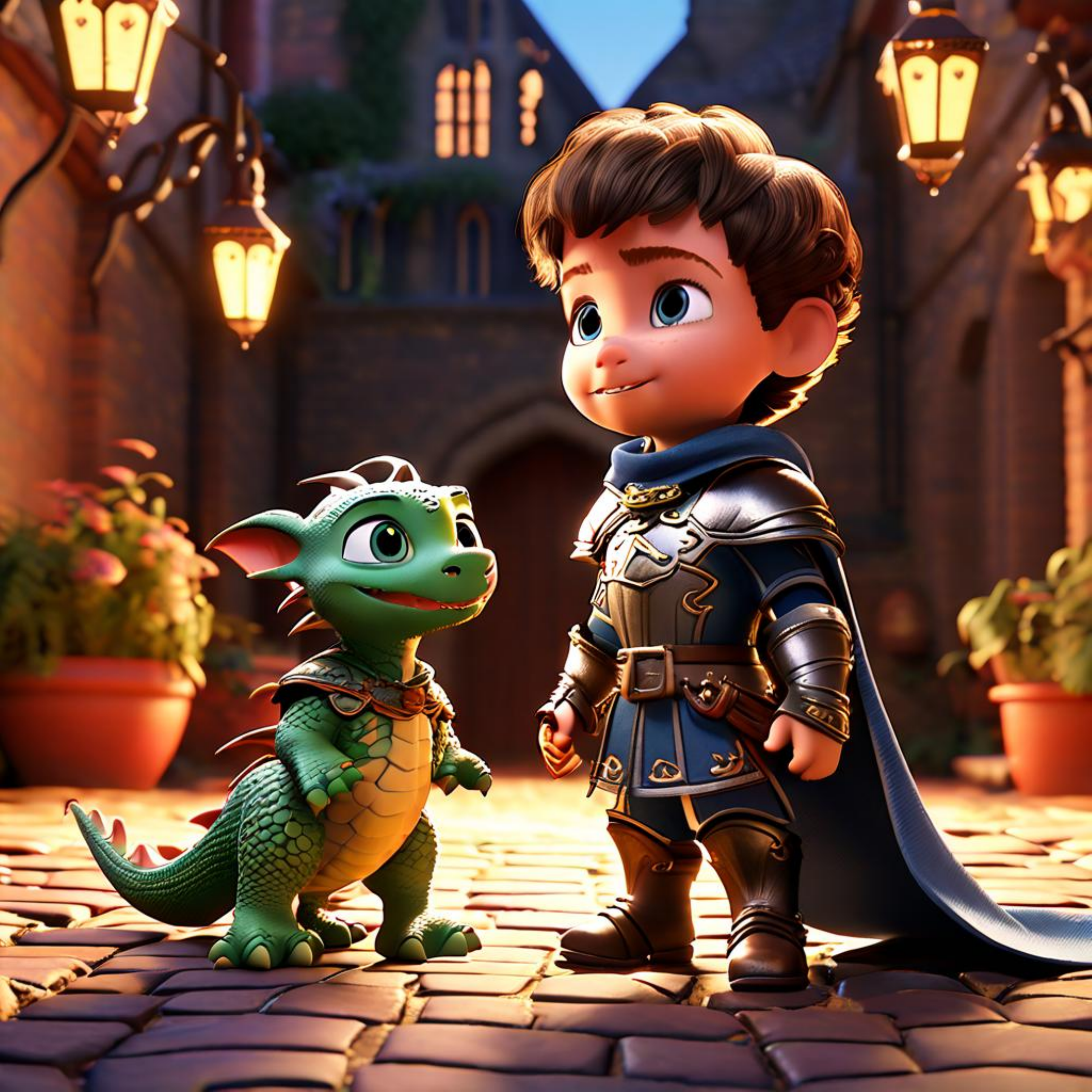} &
        \includegraphics[width=0.15\linewidth]{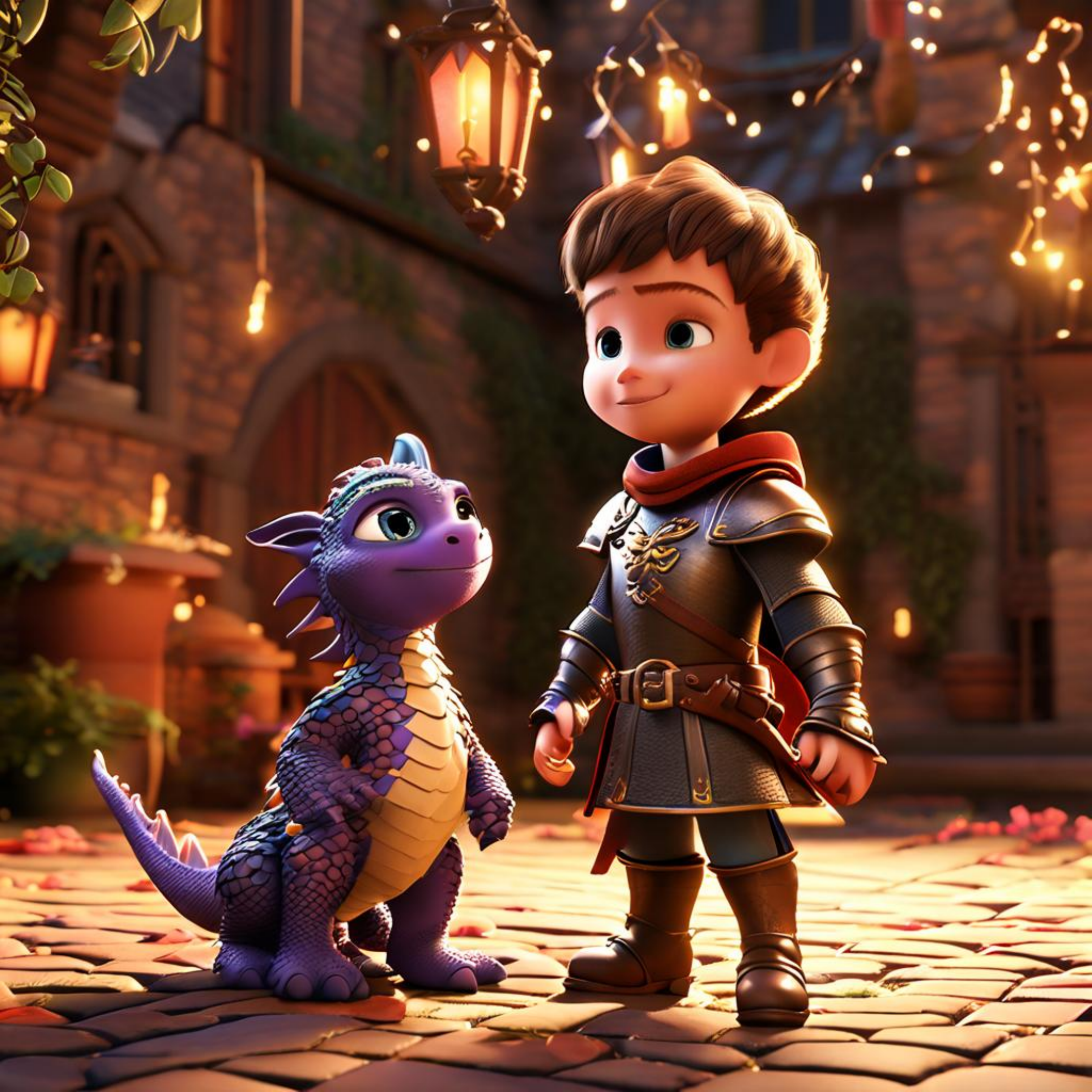} &
        \includegraphics[width=0.15\linewidth]{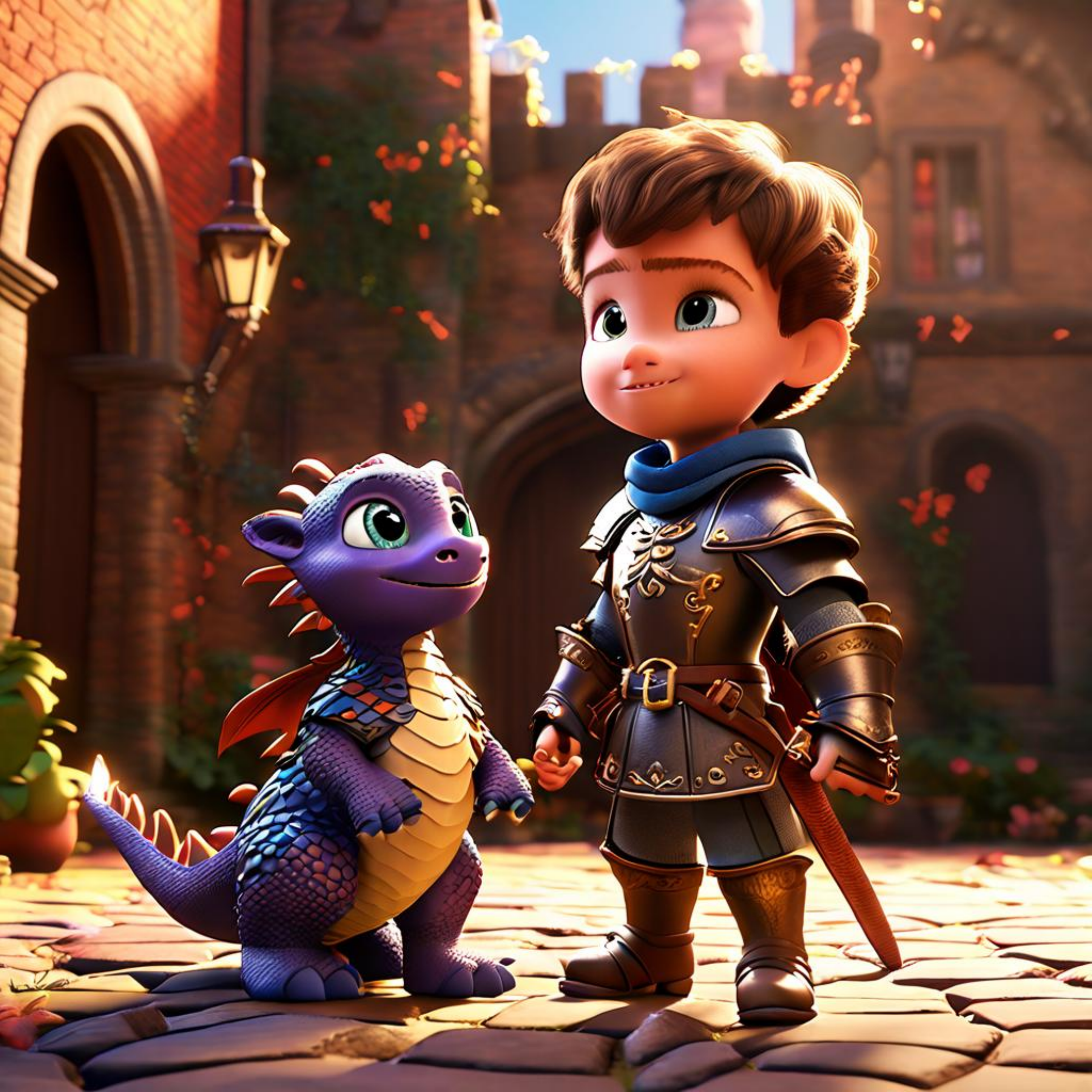} \\
        
        \includegraphics[width=0.15\linewidth]{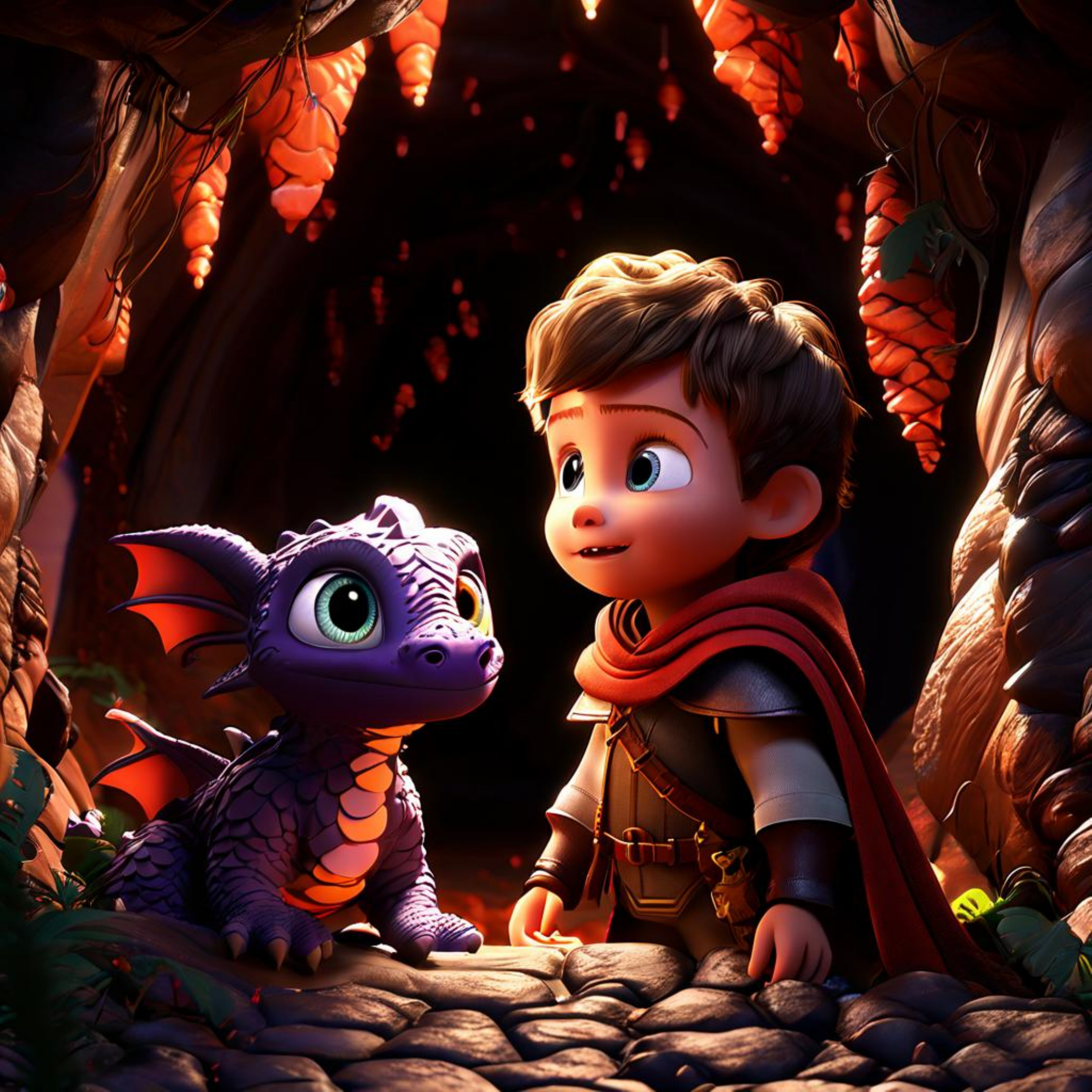} &
        \includegraphics[width=0.15\linewidth]{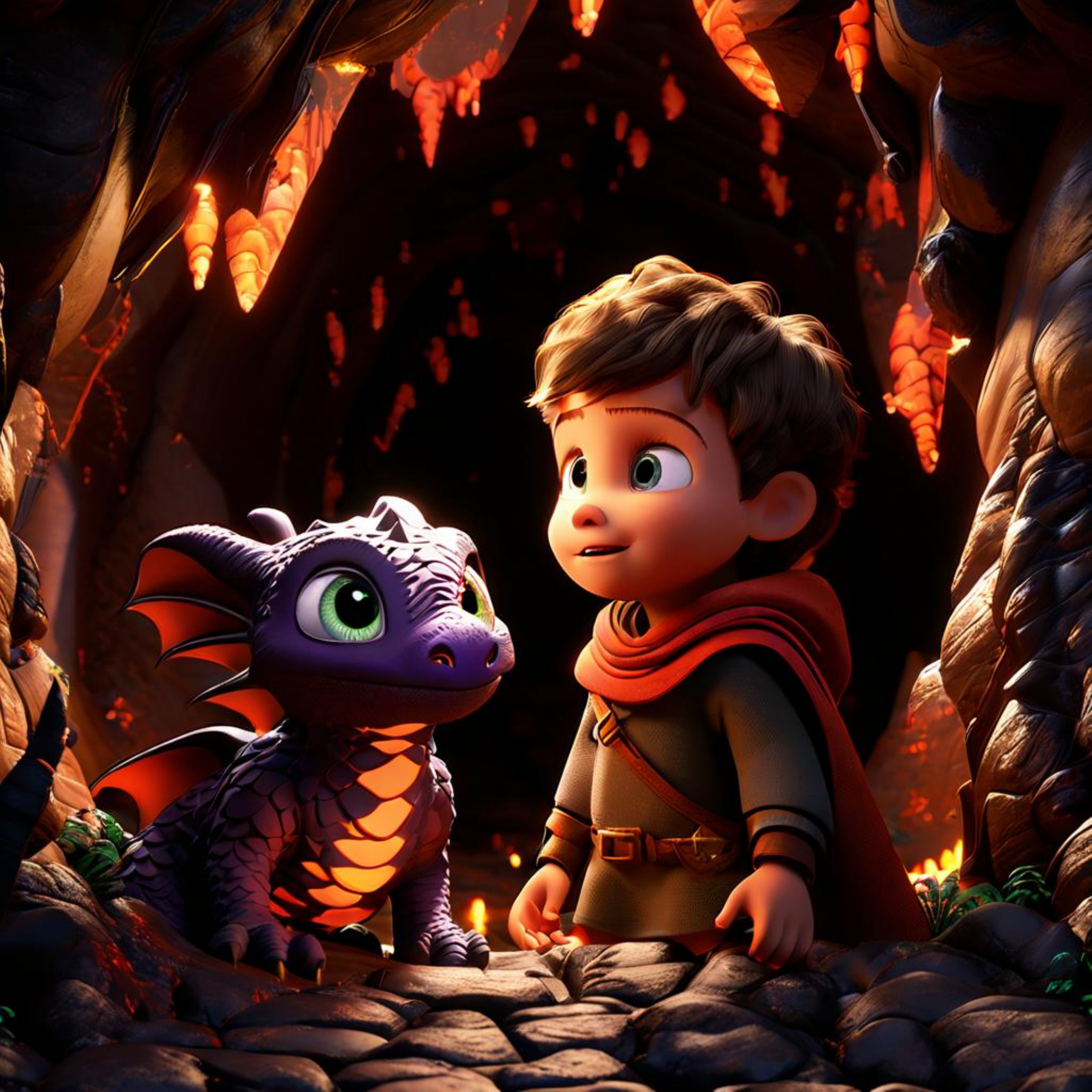} &
        \includegraphics[width=0.15\linewidth]{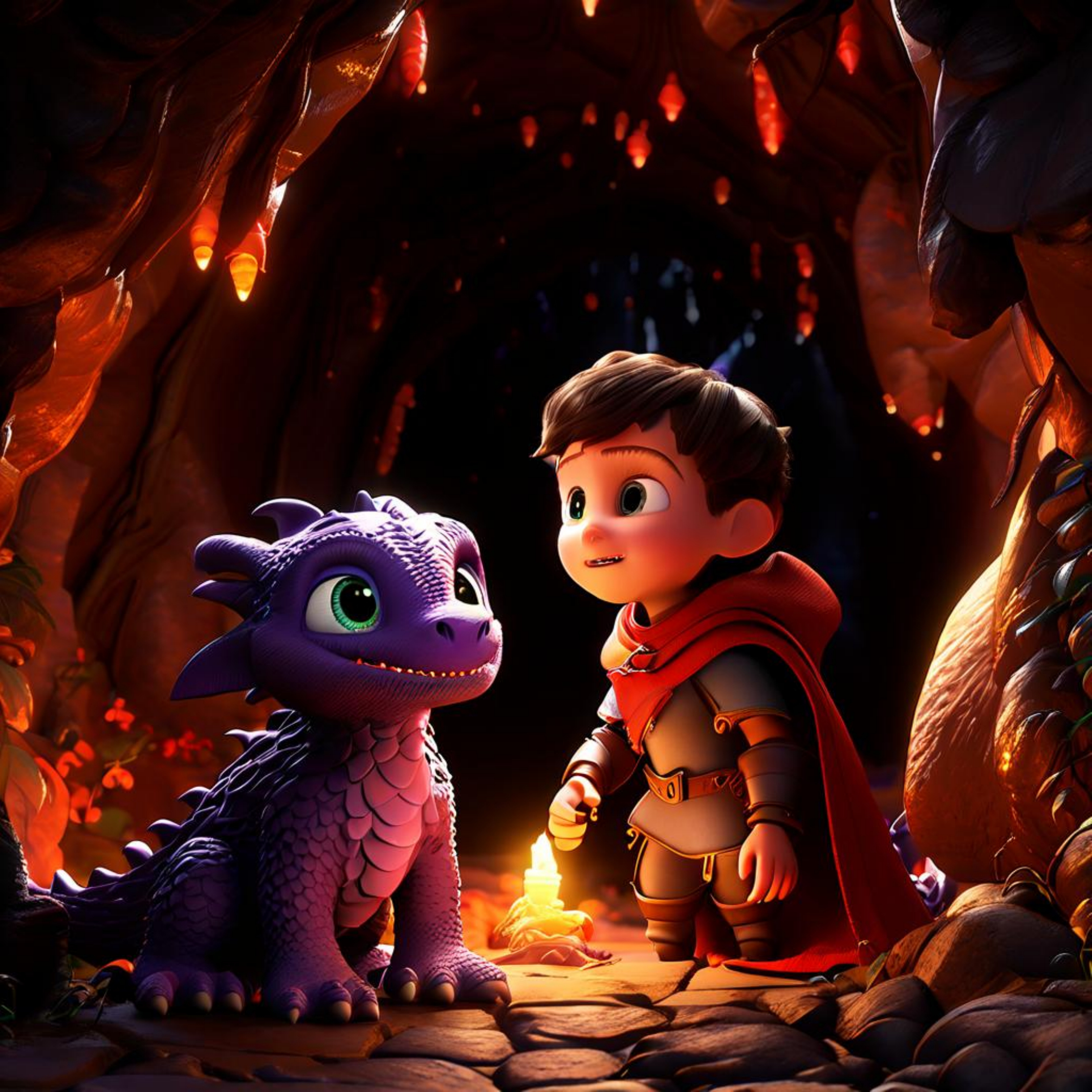} &

        \includegraphics[width=0.15\linewidth]{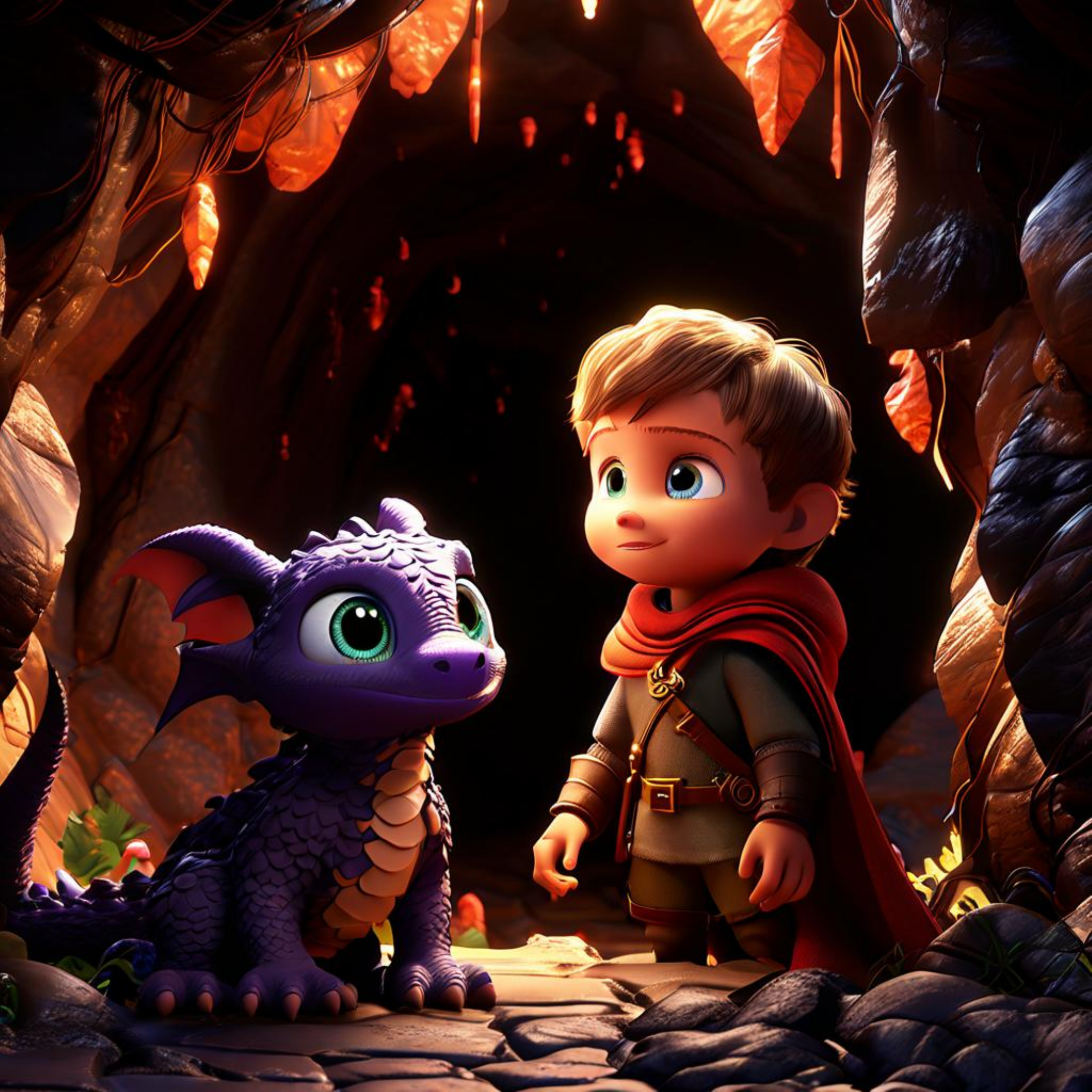} &
        \includegraphics[width=0.15\linewidth]{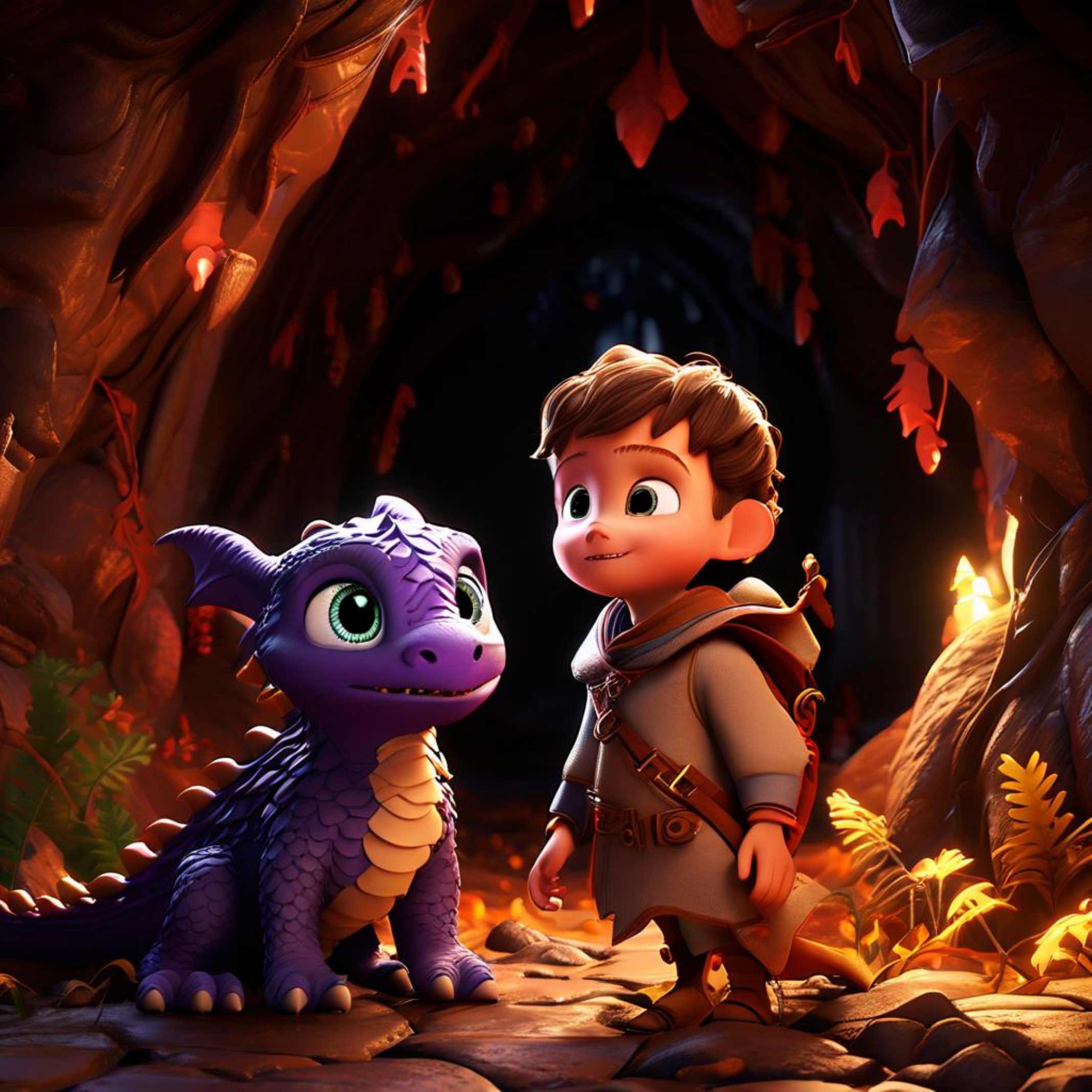} &
        \includegraphics[width=0.15\linewidth]{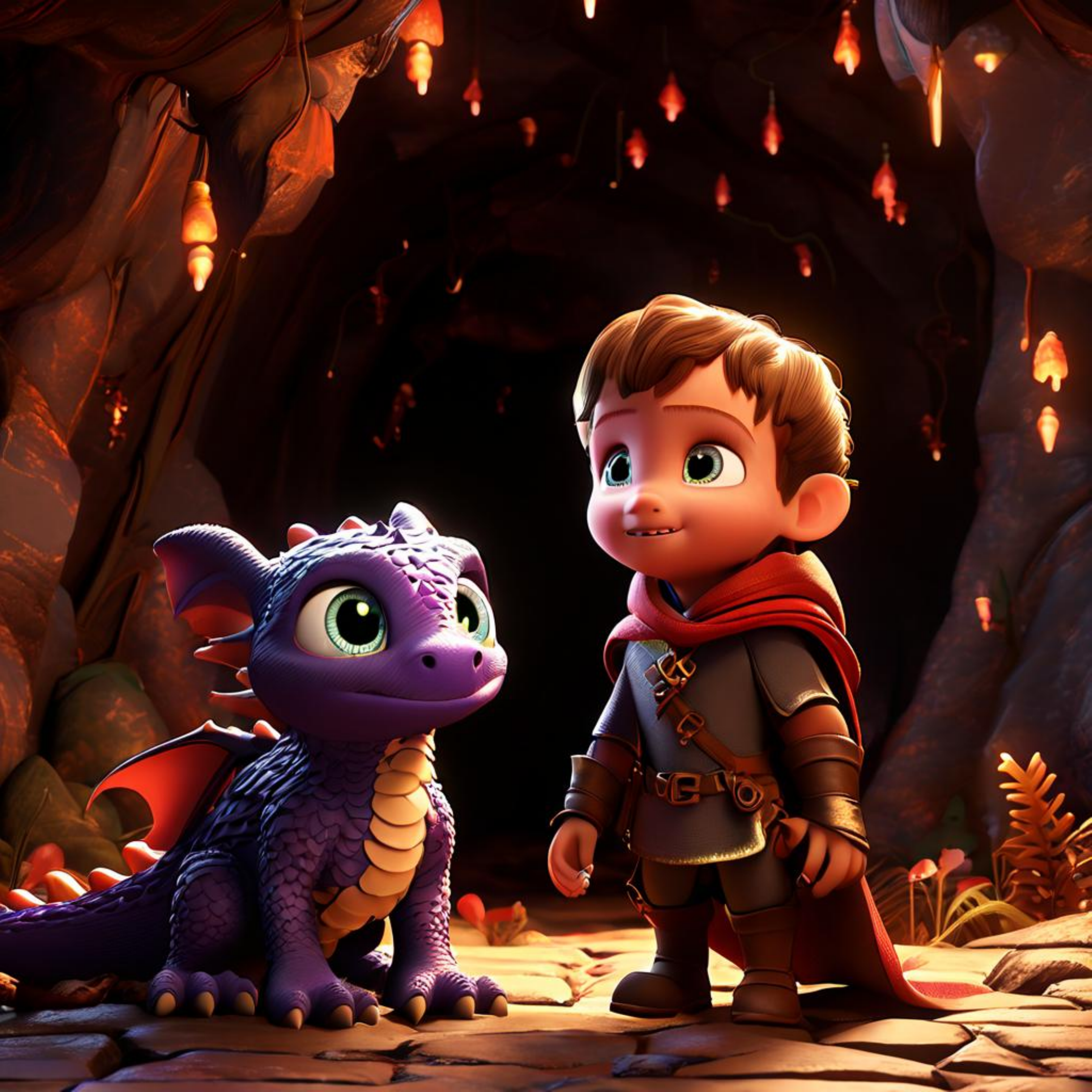} \\

        \includegraphics[width=0.15\linewidth]{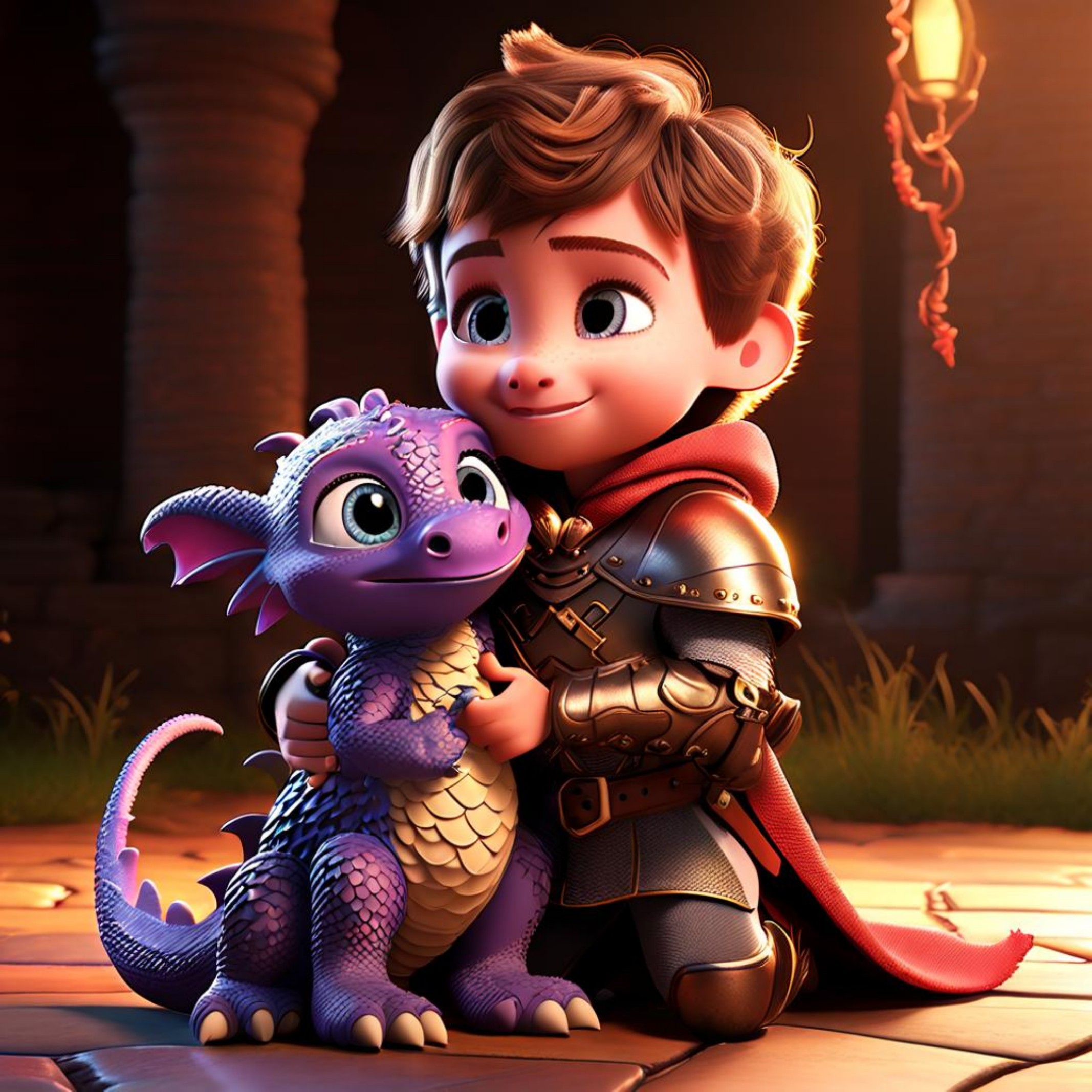} &
        \includegraphics[width=0.15\linewidth]{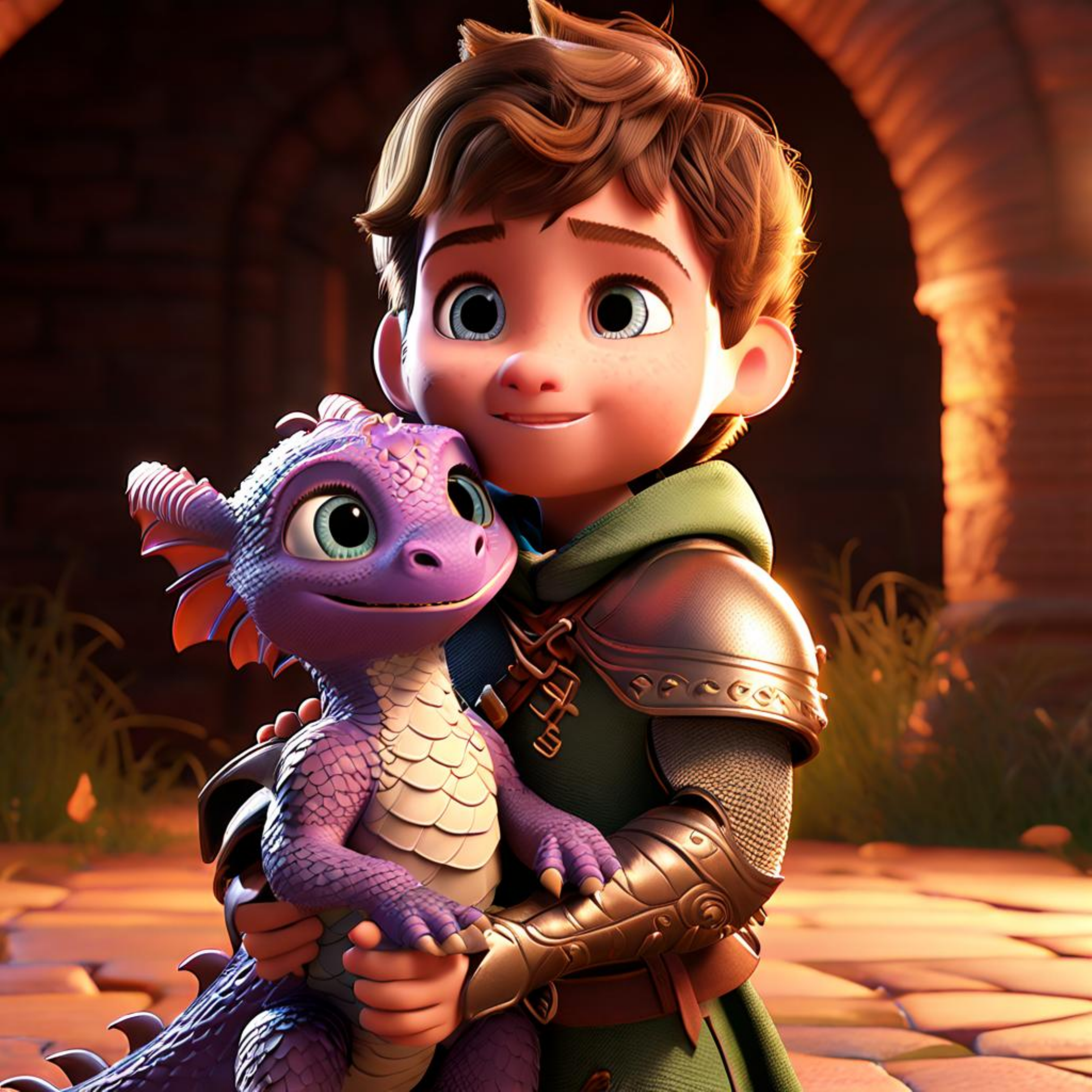} &
        \includegraphics[width=0.15\linewidth]{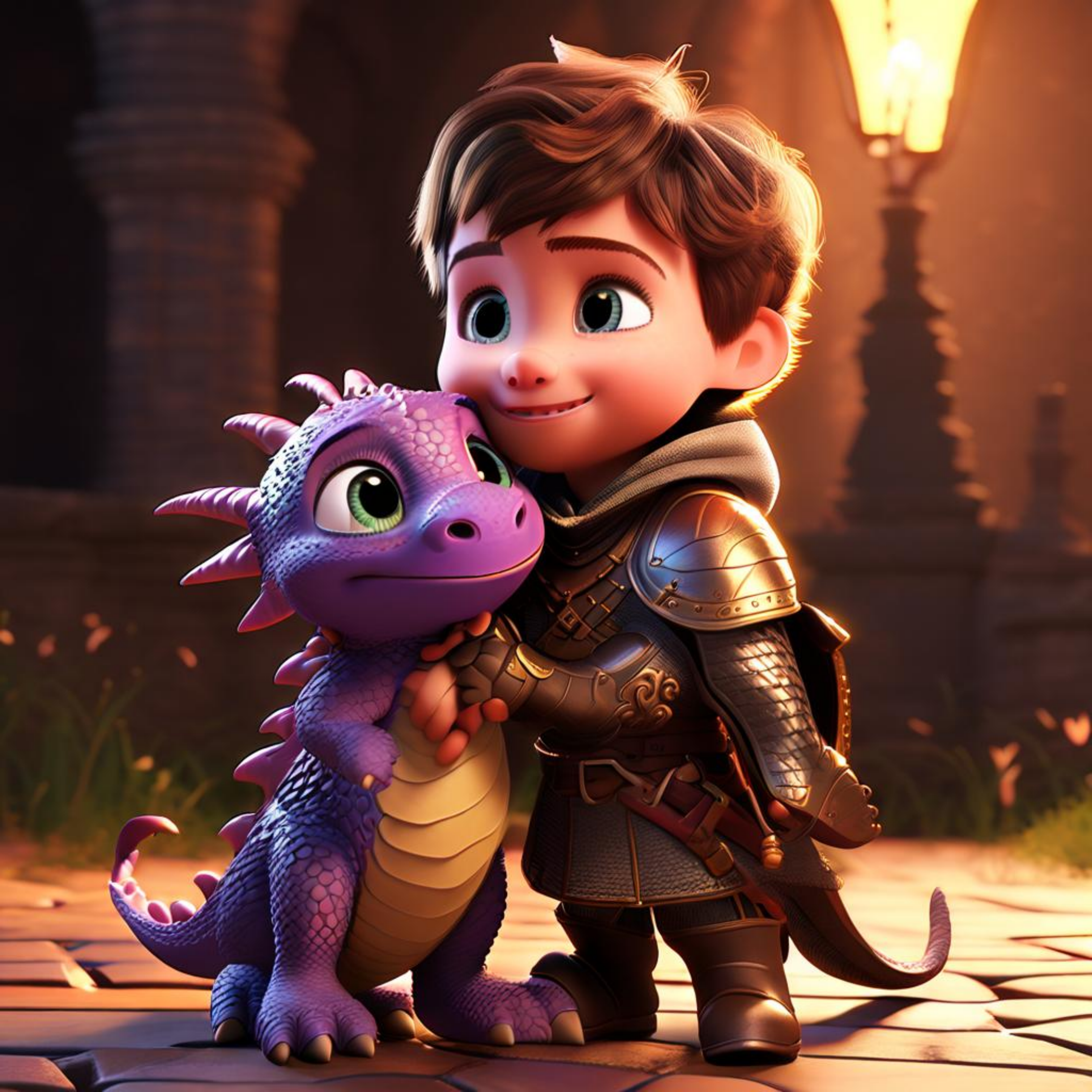} &

        \includegraphics[width=0.15\linewidth]{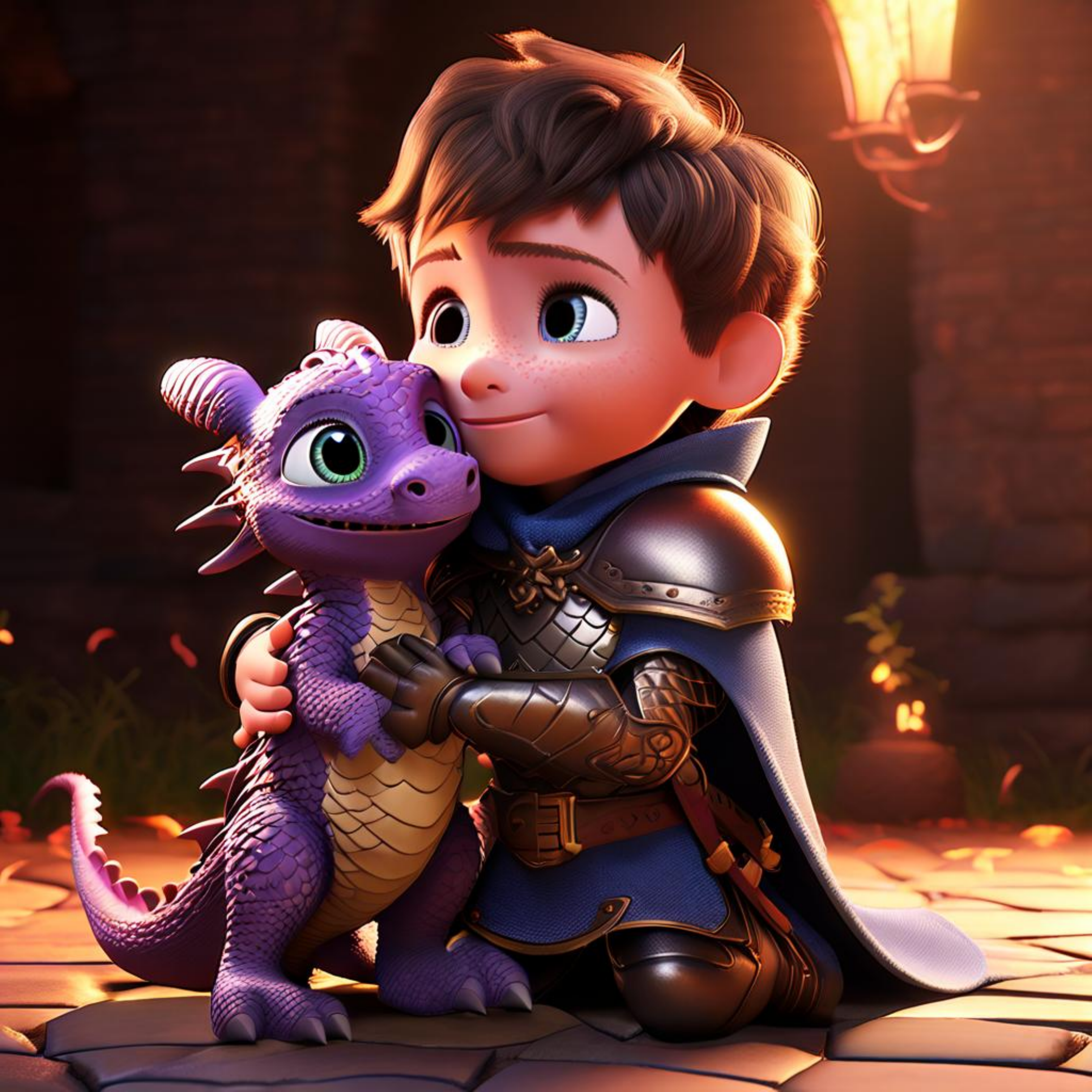} &
        \includegraphics[width=0.15\linewidth]{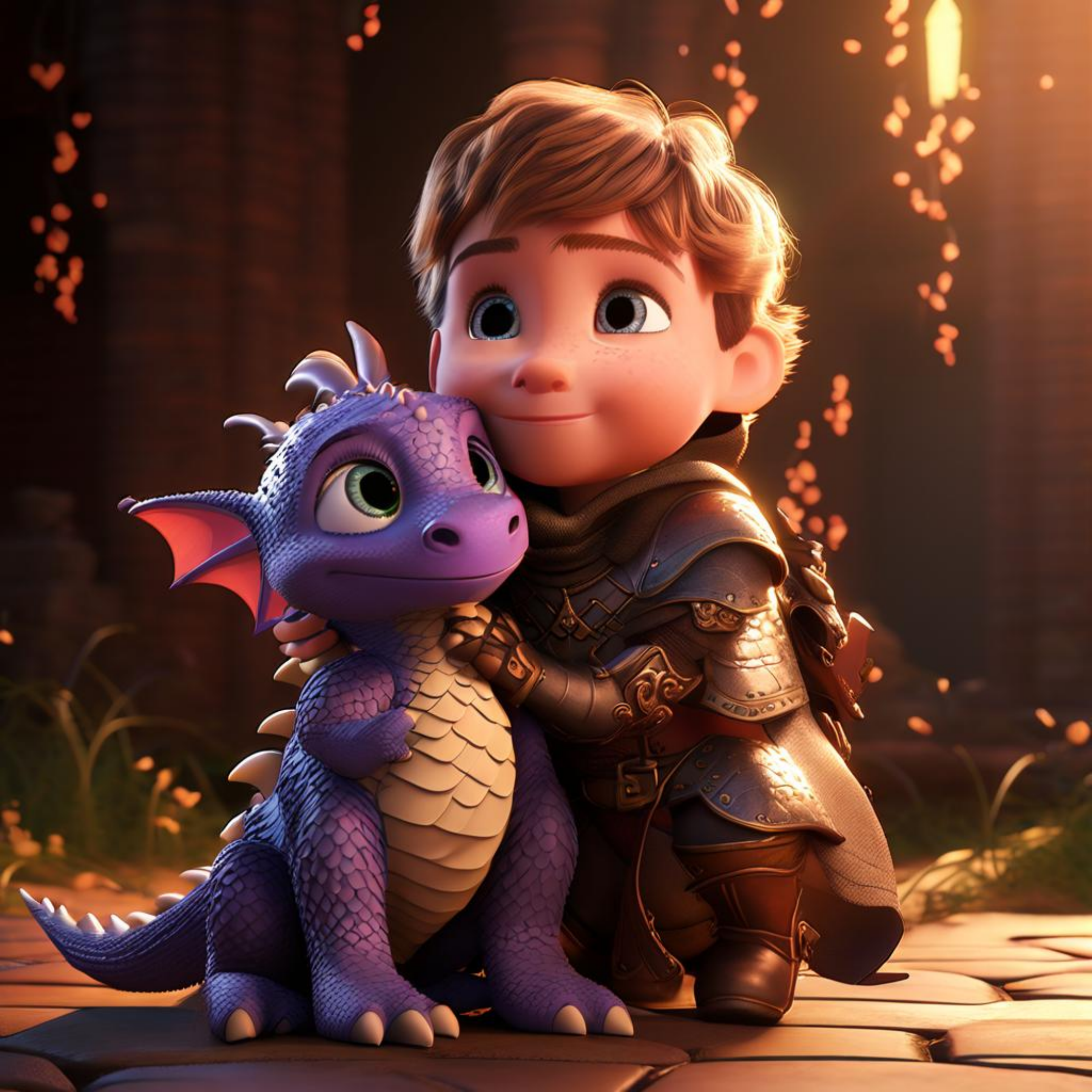} &
        \includegraphics[width=0.15\linewidth]{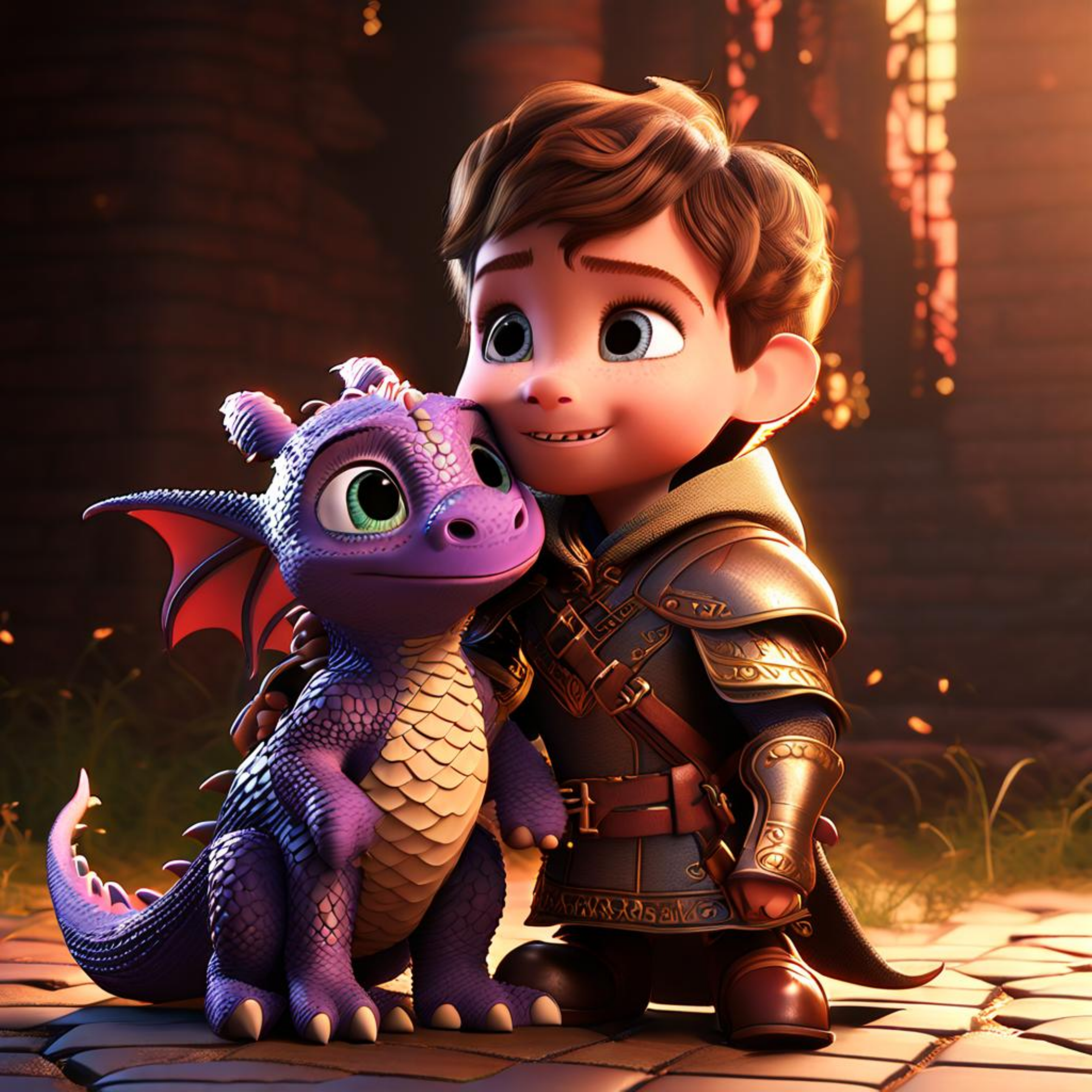} \\

    \end{tabular}
\caption{Ablation on physical priors. A high level of character coherence can be readily achieved by using heat diffusion as the coherence kernel. Prompt: \promptpart{idcolor}{A 3d art style of Pixar, knight and baby dragon;} \promptpart{action1color}{roasting marshmallow,} \promptpart{action2color}{practicing,} \promptpart{action3color}{peeking into cave,} \promptpart{action4color}{hugging each other.}}

    \label{fig:pde_comparison_grid}
\end{figure*}

\begin{table*}[ht]
\centering
\small
\caption{Quantitative comparison with baselines. `↑` higher is better, `↓` lower is better. \textbf{Bold}: best.}
\label{tab:main_comparison}
\resizebox{0.9\textwidth}{!}{% 
\begin{tabular}{l|ccc|cc|cc}
\toprule
\multirow{2}{*}{\textbf{Method}} & \multicolumn{3}{c|}{\textbf{Standard Metrics}} & \multicolumn{2}{c|}{\textbf{Proposed Metrics}} & \multicolumn{2}{c}{\textbf{Cost Metrics}} \\
\cmidrule(lr){2-4} \cmidrule(lr){5-6} \cmidrule(lr){7-8}
& \textbf{CLIP-T} $\uparrow$ & \textbf{CLIP-I}$\uparrow$ & \textbf{DreamSim} $\downarrow$ & \textbf{$\mathcal{R}_t$} $\downarrow$   & \textbf{$\mathcal{S}_t$} $\uparrow$ & \textbf{Time (s)}  & \textbf{VRAM (GB)}  \\
\midrule
IP-Adapter~\cite{ye2023ip} & 0.2950 & 0.7664 & 0.3768 & 21.2323 & 0.1527 & 14.0 & 18.4 \\
PhotoMaker~\cite{li2024photomaker} & 0.2770 & 0.7826 & 0.3536 & 20.2335 & 0.1691 & 14.1 & 17.2 \\
StoryDiffusion~\cite{zhou2024storydiffusion} & 0.3121 & 0.8011 & 0.3286 & 19.2786 & 0.1837 & 16.7 & 20.4 \\
ConsiStory~\cite{tewel2024training} & \textbf{0.3355} & 0.7905 & 0.3560 & 19.7387 & 0.1742 & 29.6& 17.9 \\
OnePromptOneStory~\cite{liu2025one} & 0.3103 & 0.7860 & 0.3823 & 20.2241 & 0.1920 & 20.8 & 18.4 \\
Zigzag Sampling~\cite{liconsistent} & 0.3172 & 0.7706 & 0.3898 & 20.7667 & 0.1621 & 15.2 & 16.4 \\
CharaConsist~\cite{wang2025characonsist} & 0.3167 & 0.7715 & 0.3847 & 20.6927 & 0.1617 & 15.5 & 16.5 \\
\midrule
\textbf{RealDiffusion (Ours)} & 0.3161 & \textbf{0.8242} & \textbf{0.2780} & \textbf{17.9104} & \textbf{0.2132} & 20.5 & 17.8 \\
\bottomrule
\end{tabular}%
}
\end{table*}

\subsection{Experimental Setup}
\label{sec:setup}

\paragraph{Implementation Details.}
All experiments are built upon Stable Diffusion XL and run on a single NVIDIA RTX 3090. For fair comparison, all methods use a $1024 \times 1024$ resolution, 50 DDIM~\cite{song2020denoising} sampling steps, and a 7.5 classifier-free guidance scale. For our PhysicsOperator, we use $n=10$ iterations with a virtual time step of $\Delta\tau = 0.1$. The main experiments use $\alpha=0.5$, and the base coefficients are set to $c_{\text{id}}=1$, $c_{\text{heat}}=2$, $c_\mathrm{s}=0.1$, and $c_\mathrm{b}=0.1$.

\paragraph{Metrics.}
 \textbf{CLIP-T} measures text alignment using the average CLIPScore~\cite{hessel2021clipscore}. For identity preservation, \textbf{IDSim} computes the average CLIP image similarity between each frame and the reference ID image. For temporal smoothness, \textbf{CLIP-I} measures adjacent-frame cosine similarity, while \textbf{DreamSim}~\cite{fu2023dreamsim} reports the perceptual distance.

To capture the trade-off central to our work, we introduce two sequence-aware measures on the sequence of per-frame CLIP features $\{f_t\}_{t=1}^T$: \textbf{Temporal Regularity} ($\mathcal{R}_t$) and \textbf{Storytelling Quality} ($\mathcal{S}_t$). $\mathcal{R}_t$ (lower is better) is designed to measure \textbf{coherence} by quantifying the smoothness of the feature trajectory. It penalizes temporal fluctuations by calculating the average L2 norm of the second-order differences, which acts as a discrete Laplacian:
\begin{equation}
    \mathcal{R}_t \;=\; \frac{1}{T-2}\sum_{t=2}^{T-1}\,\big\|\,f_{t+1}-2f_t+f_{t-1}\big\|_2 .
    \label{eq:metric_R}
\end{equation}
Our second metric, $\mathcal{S}_t$, is designed to holistically evaluate the \textbf{balance between coherence and dynamism}. It is calculated by first defining an auxiliary first-order variation, $\mathcal{D}_t = \frac{1}{T-1}\sum_{t=1}^{T-1}\|f_{t+1}-f_t\|_2$, to quantify dynamism. We then convert $(\mathcal{R}_t,\mathcal{D}_t)$ into bounded scores $\widehat{\mathcal{R}}_t=\exp(-\gamma_r\,\mathcal{R}_t)$ and $\widehat{\mathcal{D}}_t=1-\exp(-\gamma_d\,\mathcal{D}_t)$, using small positive scaling constants $\gamma_r$ and $\gamma_d$. $\mathcal{S}_t$ (higher is better) then uses a soft-min aggregator. This function approaches the minimum of the two scores at high $p$ values and acts as a balanced mean for moderate values. We set $p=4, \gamma_r=\gamma_d=0.1$ in all experiments:
\begin{equation}
\label{eq:metric_S_final}
    \mathcal{S}_t^{(p)} \;=\;
    \left(\frac{\widehat{\mathcal{R}}_t^{-p}+\widehat{\mathcal{D}}_t^{-p}}{2}\right)^{-1/p}.
\end{equation}

\subsection{Qualitative Results}

\noindent\textbf{Comparison with State-of-the-Art.}
Our evaluation, conducted on a comprehensive test set of 300 multi-character prompts generated by GPT-5~\cite{openai2025gpt5systemcard}, highlights RealDiffusion's strengths in complex scenarios. The qualitative results in Figure~\ref{fig:main_comparison} show that our method generates sequences with significantly higher character coherence. While baseline methods often suffer from attribute swapping or identity drift in multi-character interactions, our approach maintains robust visual fidelity, preserving individual character features across diverse scenes. This focus on coherence is validated by our quantitative analysis in Table~\ref{tab:main_comparison}. RealDiffusion achieves the best performance on our proposed Storytelling Quality ($\mathcal{S}_t$) metric, which specifically rewards a balance between coherence and dynamism. This indicates that our method excels at managing this crucial trade-off. These gains in narrative coherence are achieved with minimal added computational cost in time and VRAM usage, confirming the lightweight nature of our framework.%极少有待商榷

\subsection{Ablation Studies}

\noindent\textbf{Choice of Physical Prior.}
An ablation study, run with $\alpha=0.5$, validates our choice of the heat diffusion equation. As shown in Figure~\ref{fig:pde_comparison_grid} and Table~\ref{tab:ablation_physics}, heat diffusion is the most effective prior. Conservative systems like the Wave equation cause unstable feature oscillations, while non-linear systems like Burgers' risk unpredictable artifacts. In contrast, the dissipative nature of heat diffusion provides a smooth and stable regularization, enhancing coherence without visual degradation.

\begin{figure*}[t]
\centering
\setlength{\tabcolsep}{1pt} 
\small 

\begin{tabular}{cc} % Outer table for vertical label + image row

%---------------------------------------------------
% ROW for alpha = 0.0
%---------------------------------------------------
\begin{tabular}{c}
    \vspace{1mm}\rotatebox[origin=c]{90}{\textbf{$\alpha = 0.0$}}
\end{tabular}
&
\begin{tabular}{cccccc}
    \includegraphics[width=0.15\linewidth]{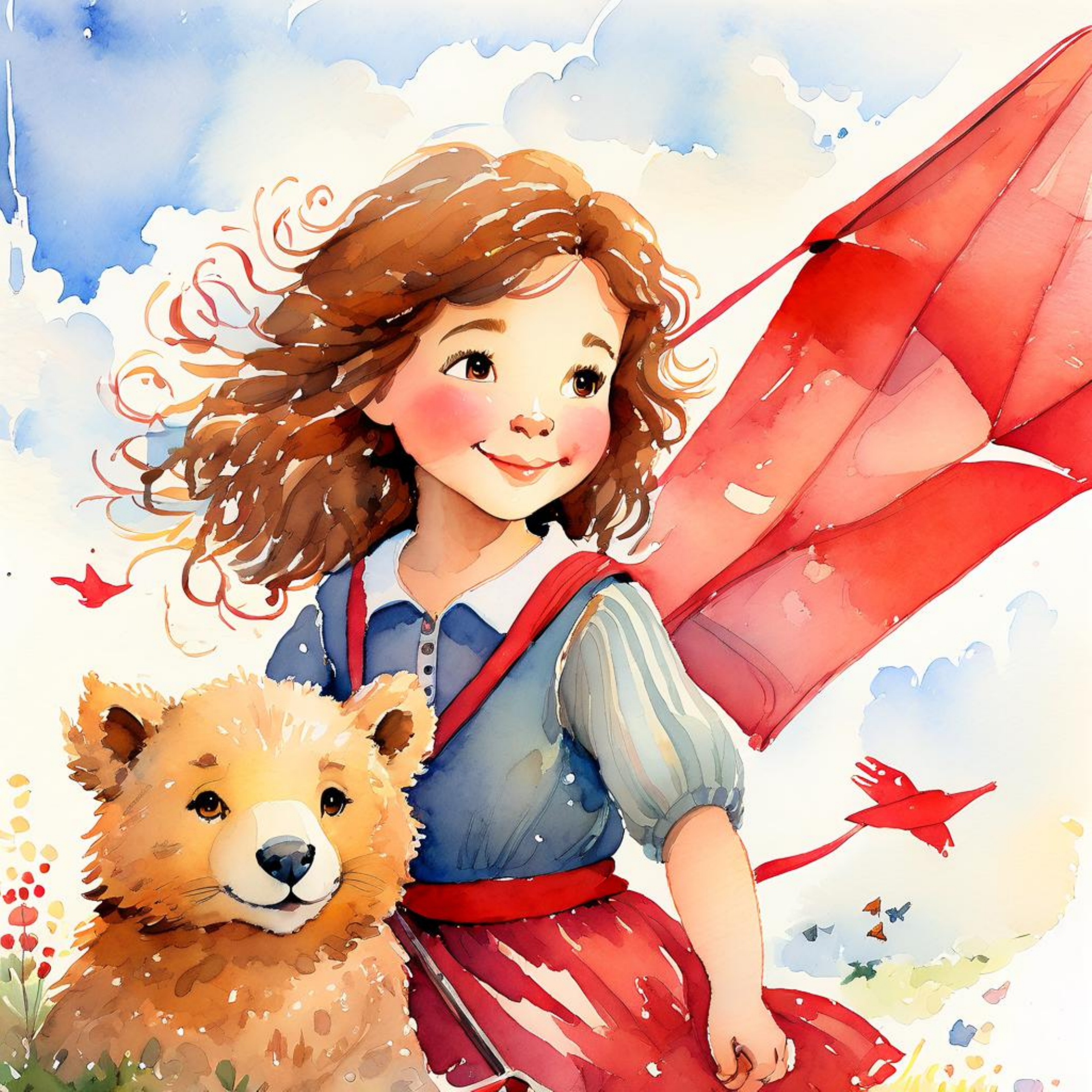} &
    \includegraphics[width=0.15\linewidth]{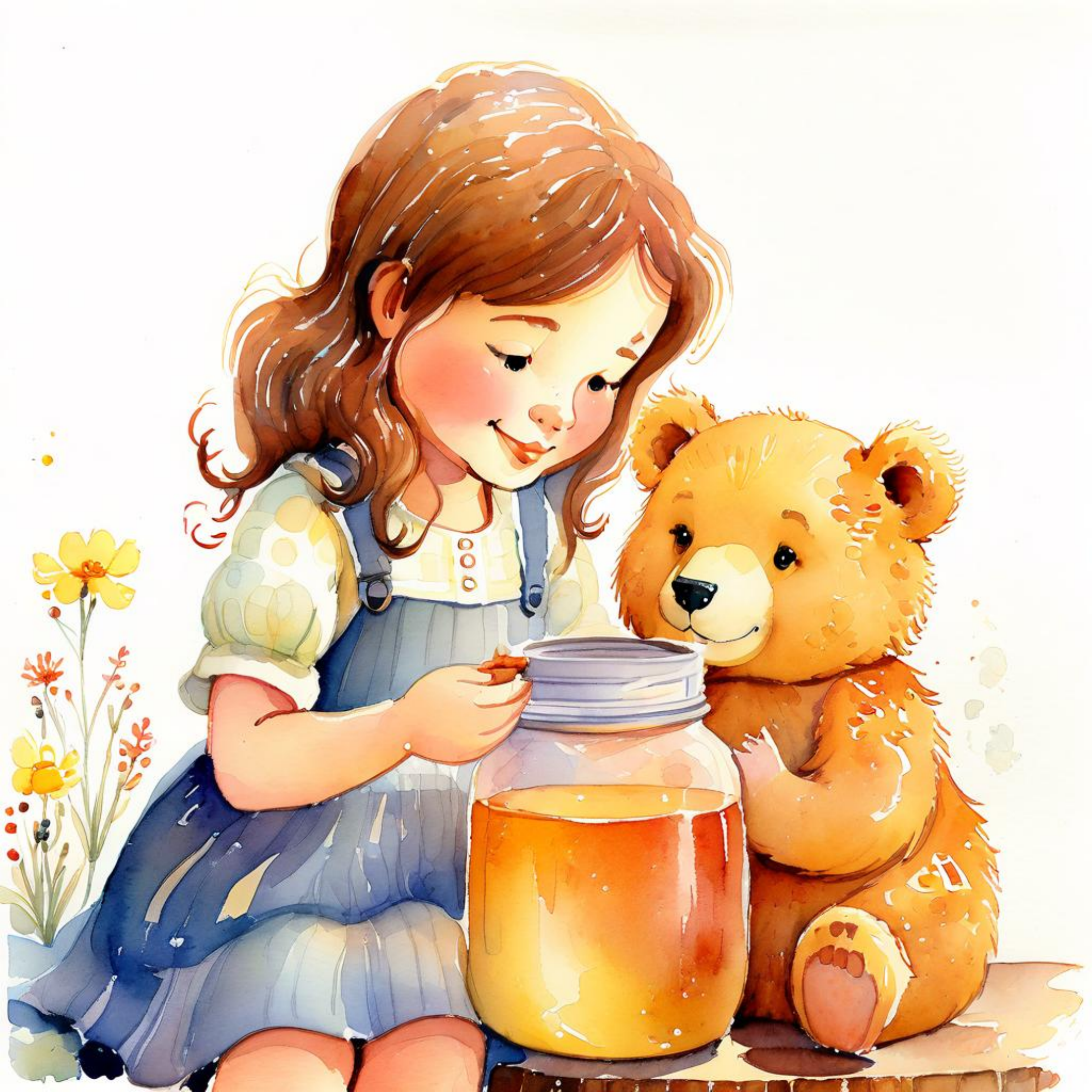} &
    \includegraphics[width=0.15\linewidth]{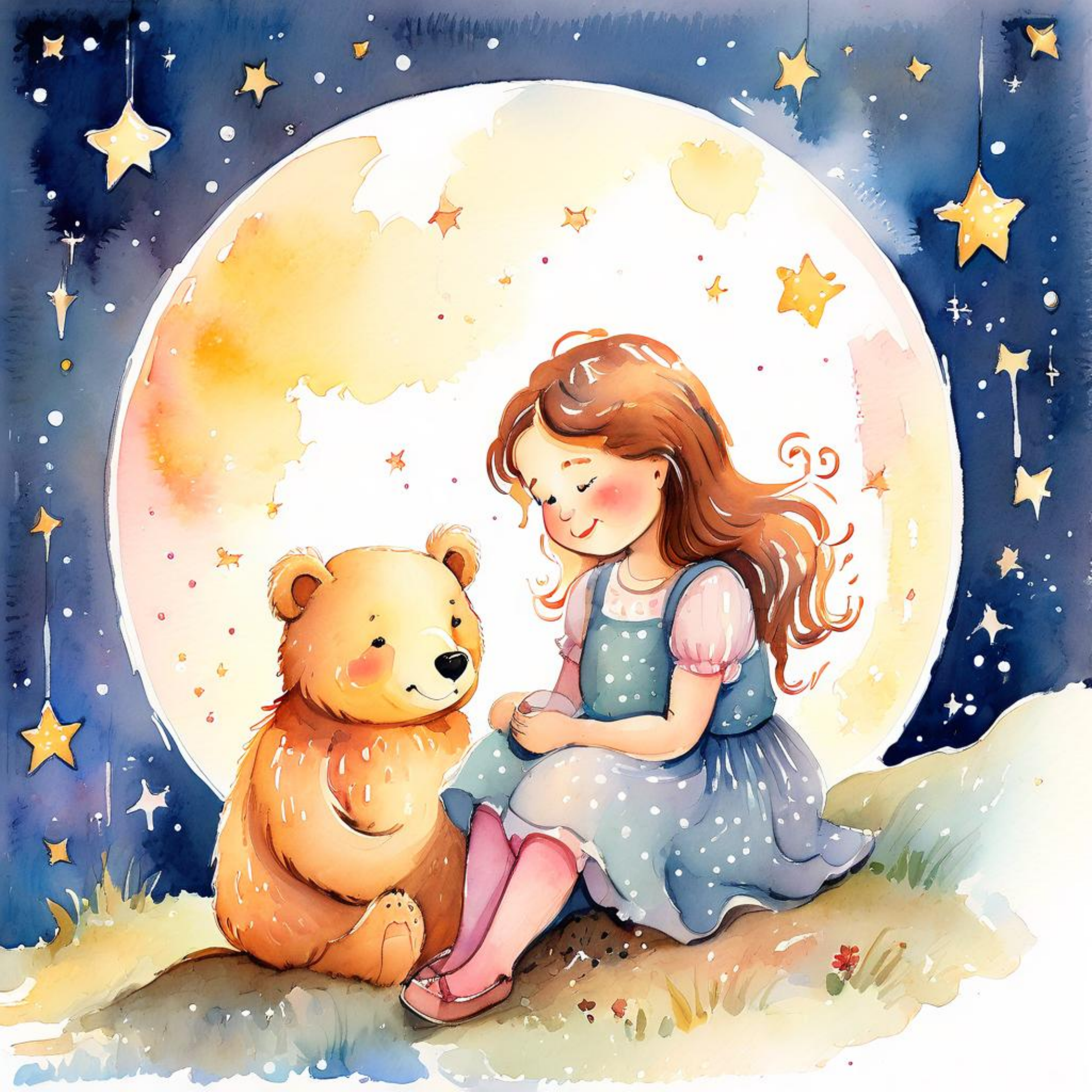} &
    \includegraphics[width=0.15\linewidth]{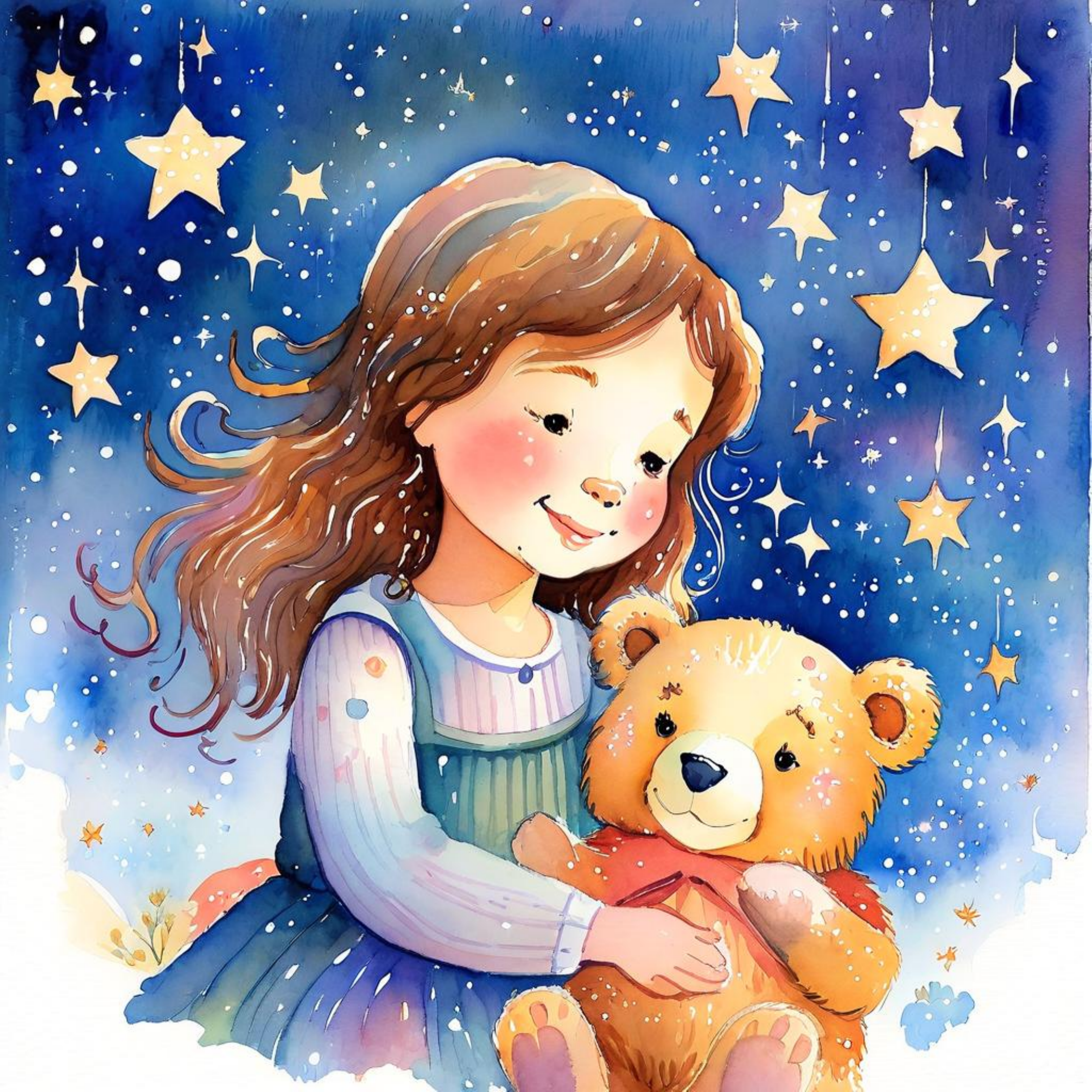} &
    \includegraphics[width=0.15\linewidth]{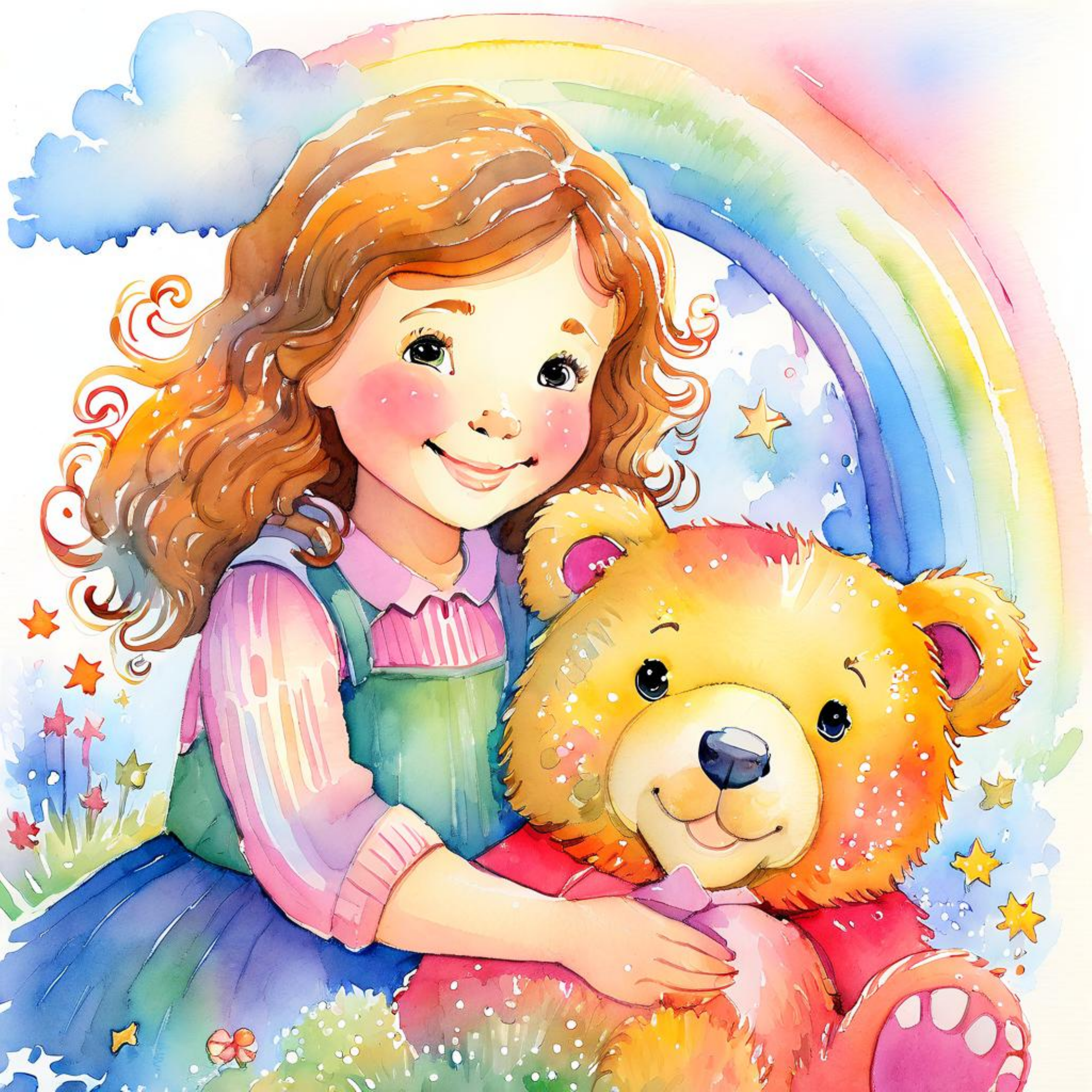} &
    \includegraphics[width=0.15\linewidth]{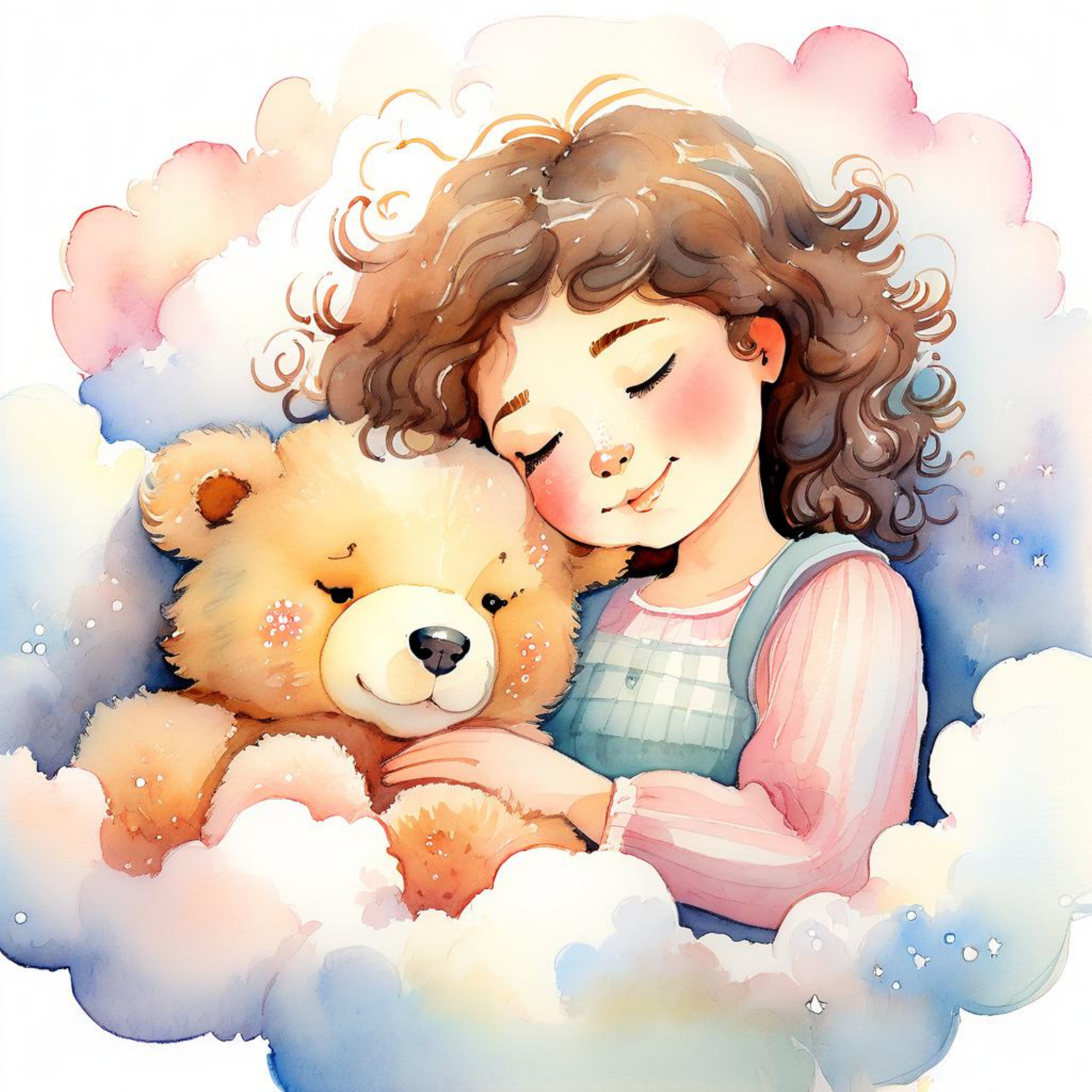}
\end{tabular}
\\ 

%---------------------------------------------------
% ROW for alpha = 0.25
%---------------------------------------------------
% \begin{tabular}{c}
%     \vspace{1mm}\rotatebox[origin=c]{90}{\textbf{$\alpha = 0.25$}}
% \end{tabular}
% &
% \begin{tabular}{cccccc}
%     \includegraphics[width=0.15\linewidth]{figure_cvpr/alpha_1/row_2_col_1.pdf} &
%     \includegraphics[width=0.15\linewidth]{figure_cvpr/alpha_1/row_2_col_2.pdf} &
%     \includegraphics[width=0.15\linewidth]{figure_cvpr/alpha_1/row_2_col_4.pdf} &
%     \includegraphics[width=0.15\linewidth]{figure_cvpr/alpha_1/row_2_col_5.pdf} &
%     \includegraphics[width=0.15\linewidth]{figure_cvpr/alpha_1/row_2_col_6.pdf} &
%     \includegraphics[width=0.15\linewidth]{figure_cvpr/alpha_1/row_2_col_7.pdf}
% \end{tabular}
% \\

%---------------------------------------------------
% ROW for alpha = 0.5
%---------------------------------------------------
\begin{tabular}{c}
    \vspace{1mm}\rotatebox[origin=c]{90}{\textbf{$\alpha = 0.5$}}
\end{tabular}
&
\begin{tabular}{cccccc}
    \includegraphics[width=0.15\linewidth]{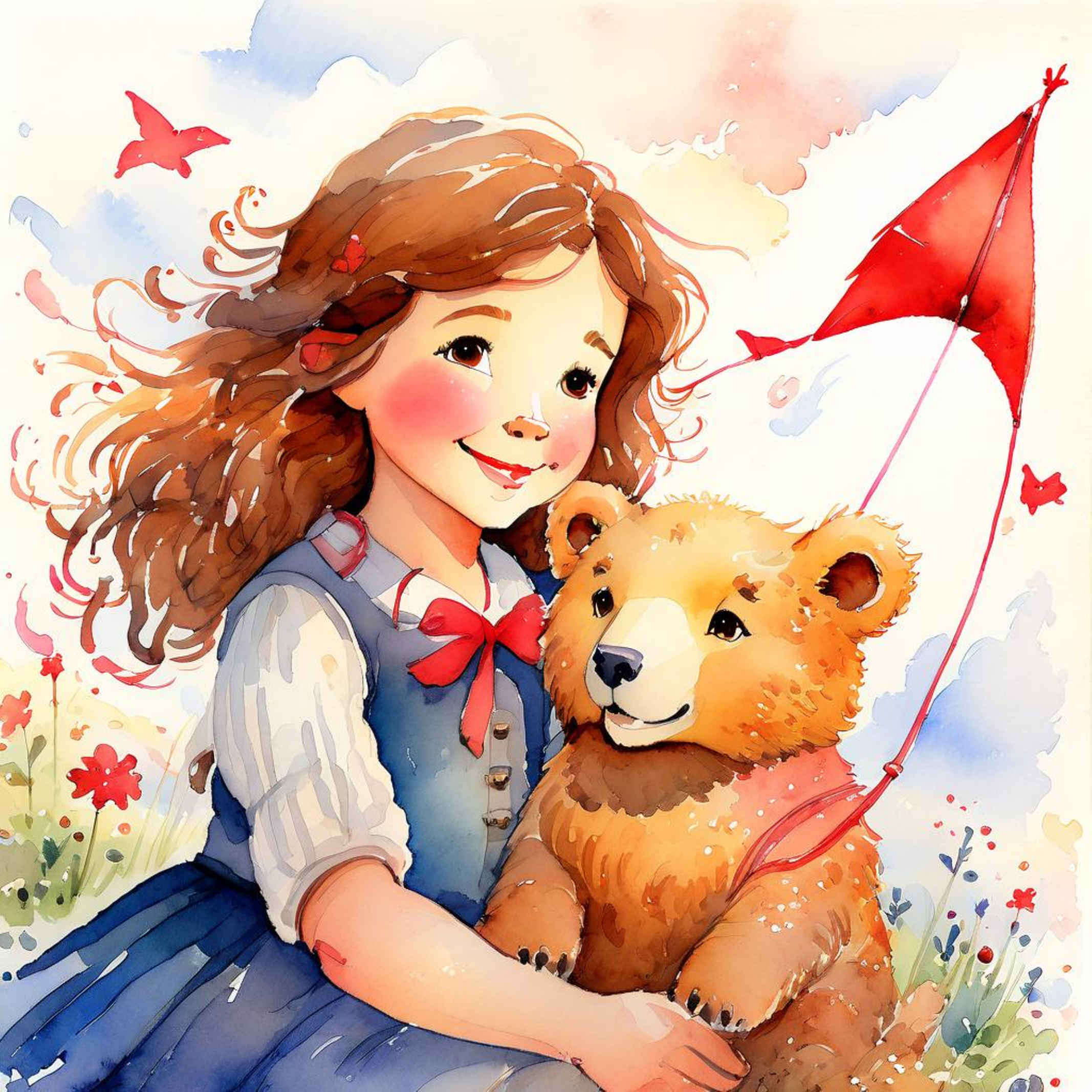} &
    \includegraphics[width=0.15\linewidth]{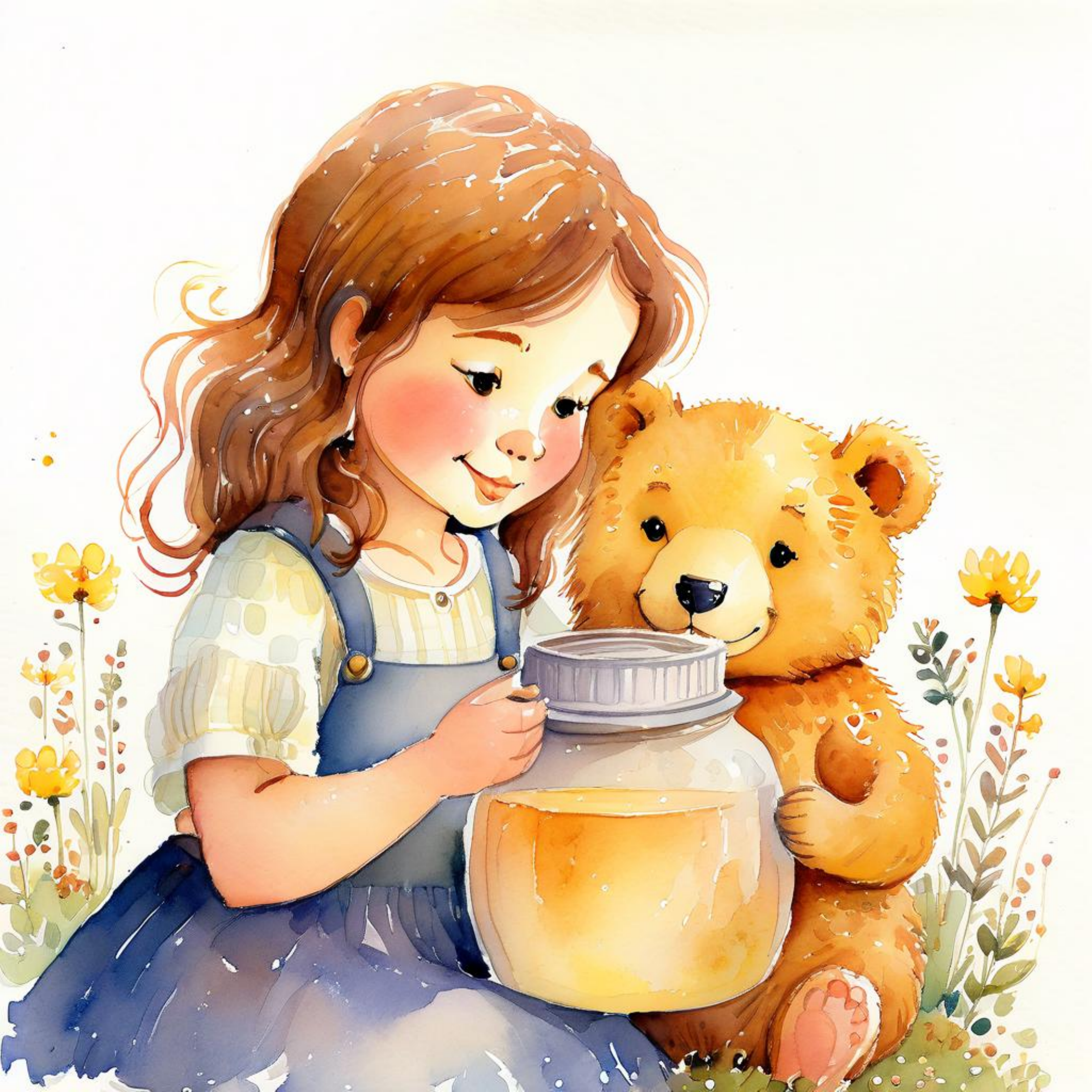} &
    \includegraphics[width=0.15\linewidth]{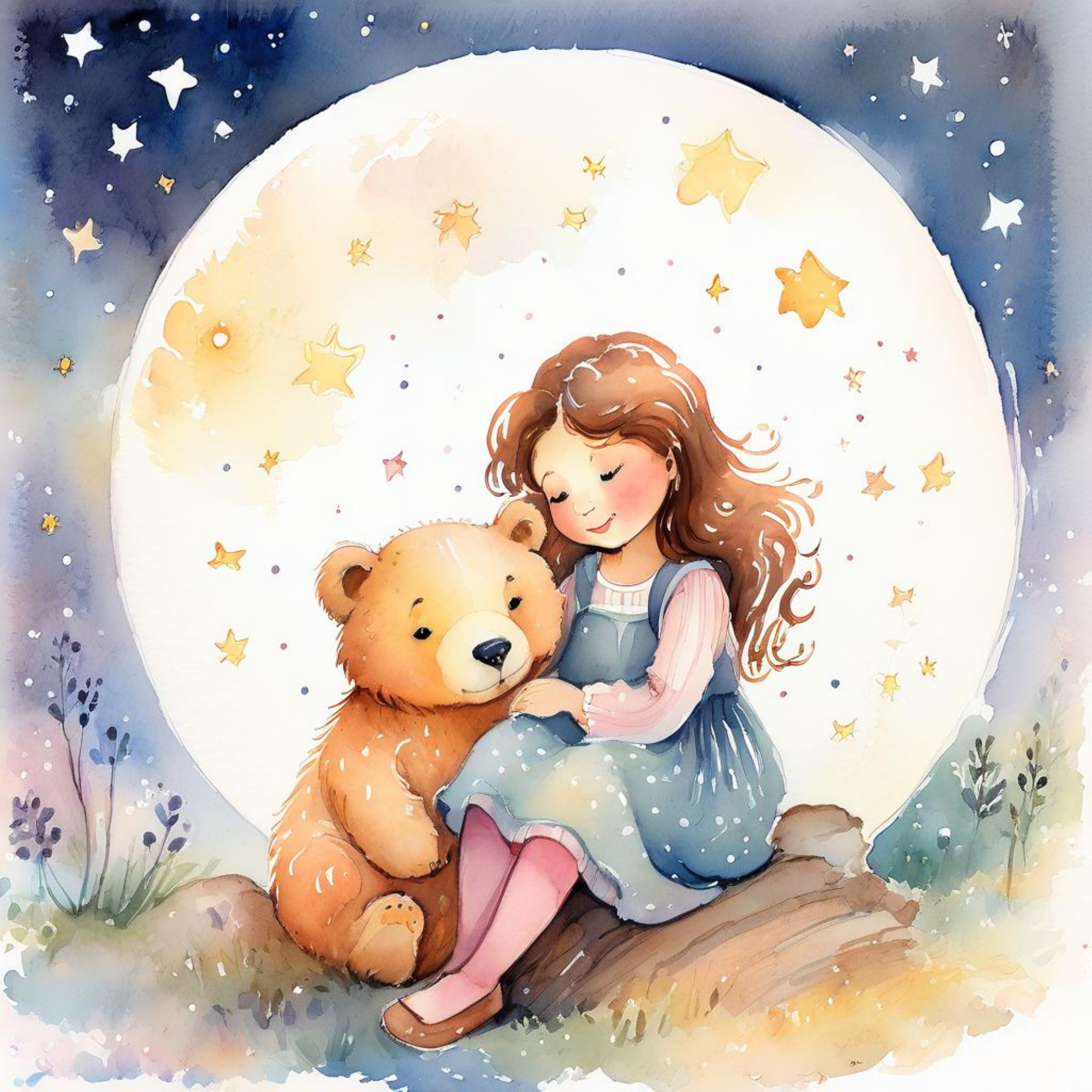} &
    \includegraphics[width=0.15\linewidth]{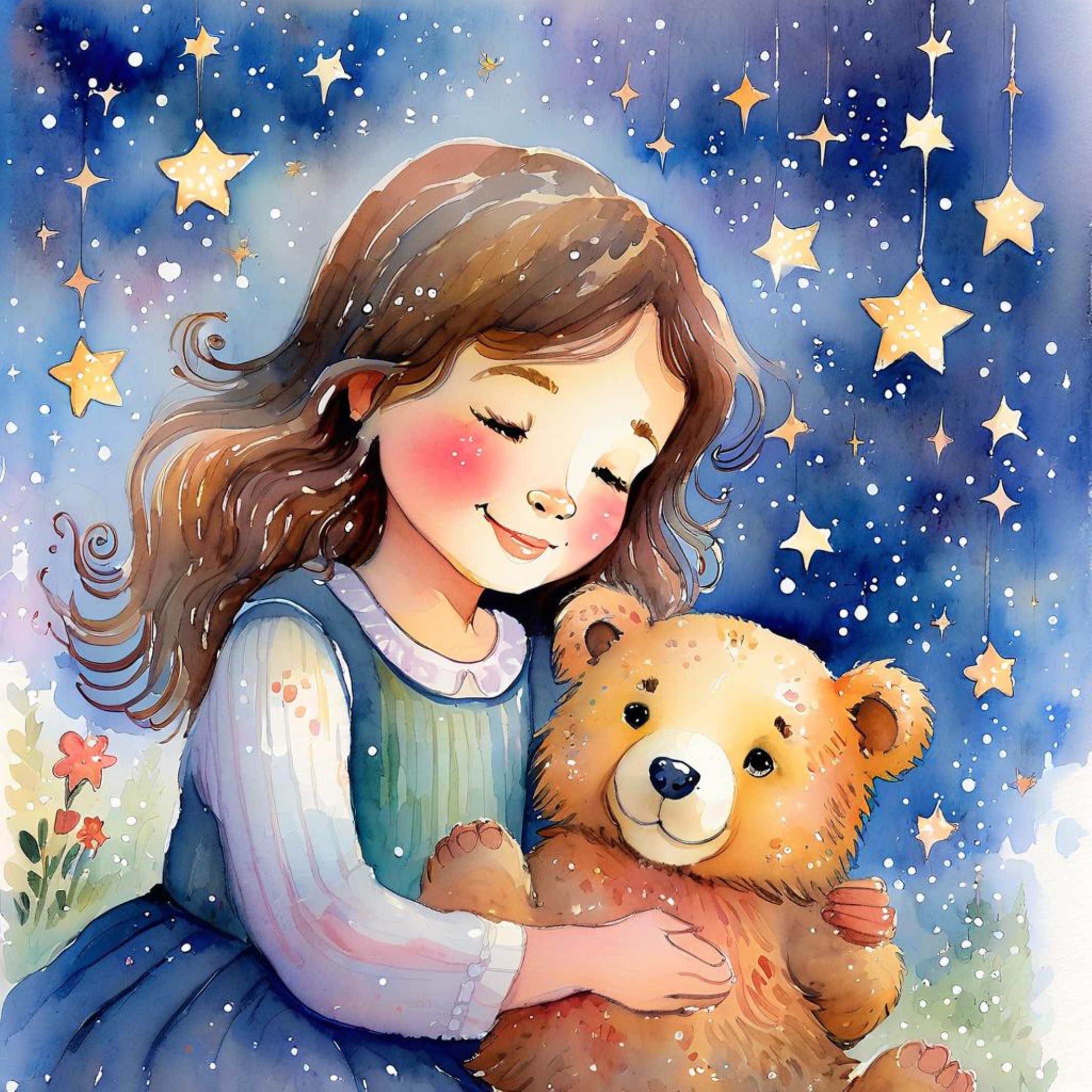} &
    \includegraphics[width=0.15\linewidth]{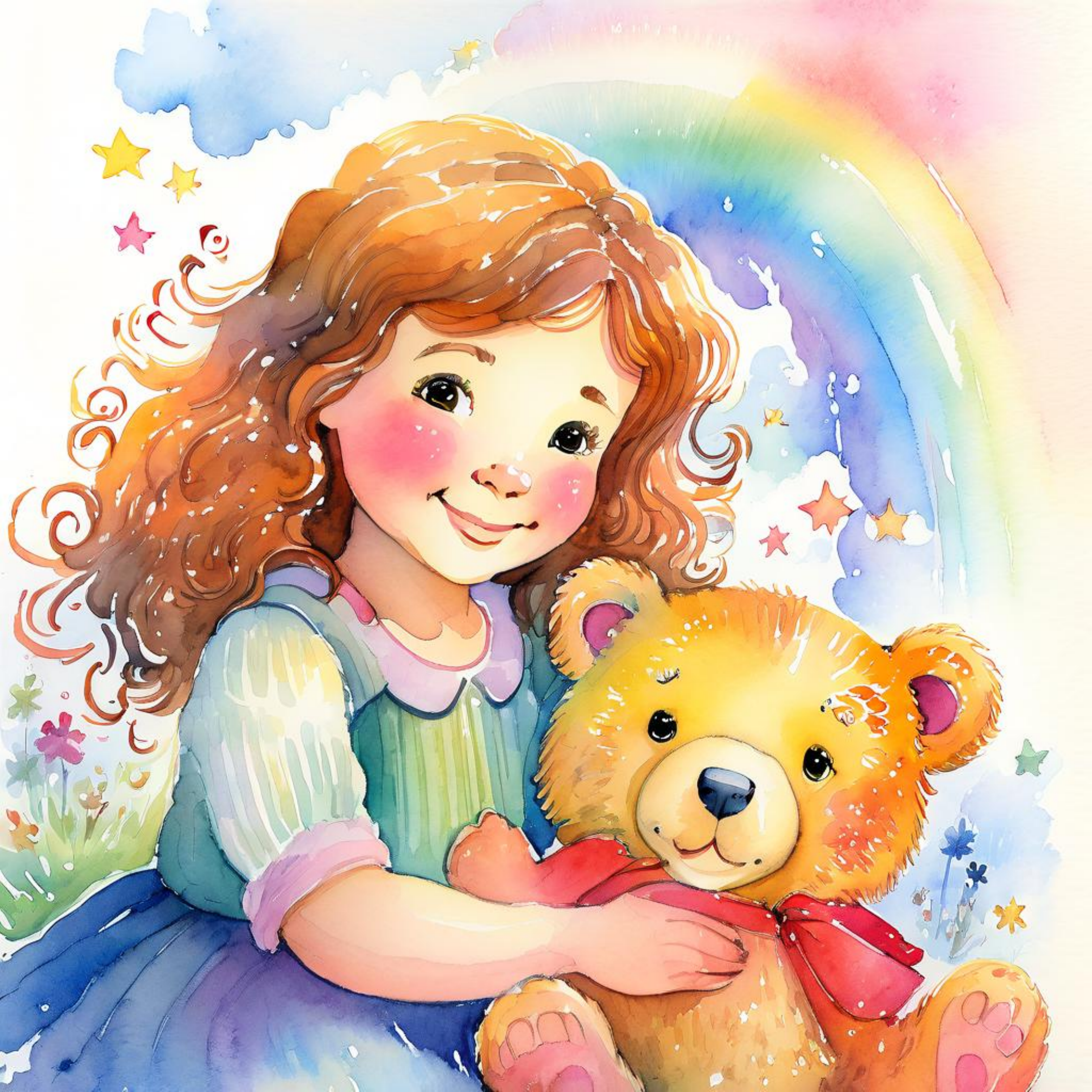} &
    \includegraphics[width=0.15\linewidth]{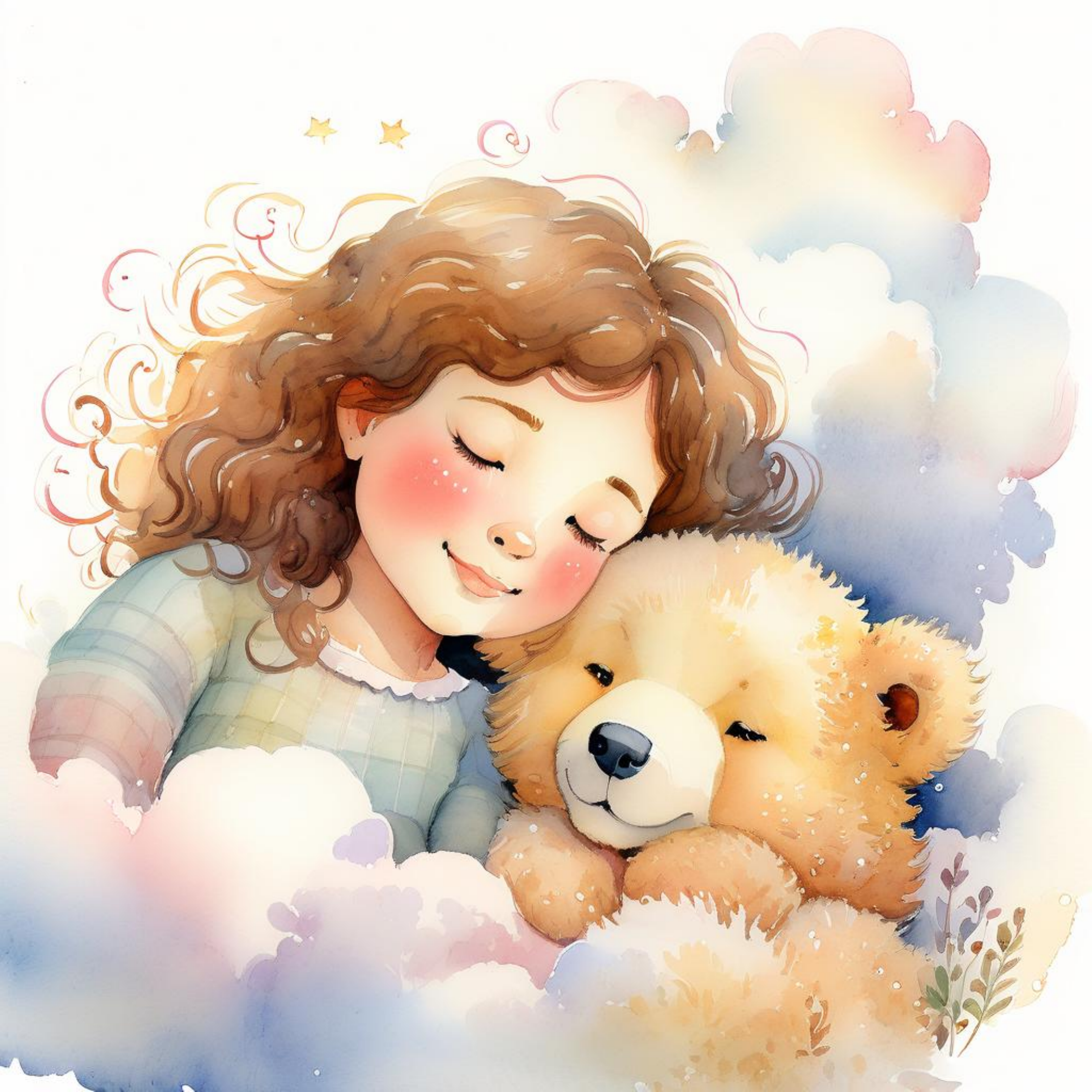}
\end{tabular}
\\

%---------------------------------------------------
% ROW for alpha = 0.75
%---------------------------------------------------
% \begin{tabular}{c}
%     \vspace{1mm}\rotatebox[origin=c]{90}{\textbf{$\alpha = 0.75$}}
% \end{tabular}
% &
% \begin{tabular}{cccccc}
%     \includegraphics[width=0.15\linewidth]{figure_cvpr/alpha_3/row_2_col_1.pdf} &
%     \includegraphics[width=0.15\linewidth]{figure_cvpr/alpha_3/row_2_col_2.pdf} &
%     \includegraphics[width=0.15\linewidth]{figure_cvpr/alpha_3/row_2_col_4.pdf} &
%     \includegraphics[width=0.15\linewidth]{figure_cvpr/alpha_3/row_2_col_5.pdf} &
%     \includegraphics[width=0.15\linewidth]{figure_cvpr/alpha_3/row_2_col_6.pdf} &
%     \includegraphics[width=0.15\linewidth]{figure_cvpr/alpha_3/row_2_col_7.pdf}
% \end{tabular}
% \\

%---------------------------------------------------
% ROW for alpha = 1.0
%---------------------------------------------------
\begin{tabular}{c}
    \vspace{1mm}\rotatebox[origin=c]{90}{\textbf{$\alpha = 1.0$}}
\end{tabular}
&
\begin{tabular}{cccccc}
    \includegraphics[width=0.15\linewidth]{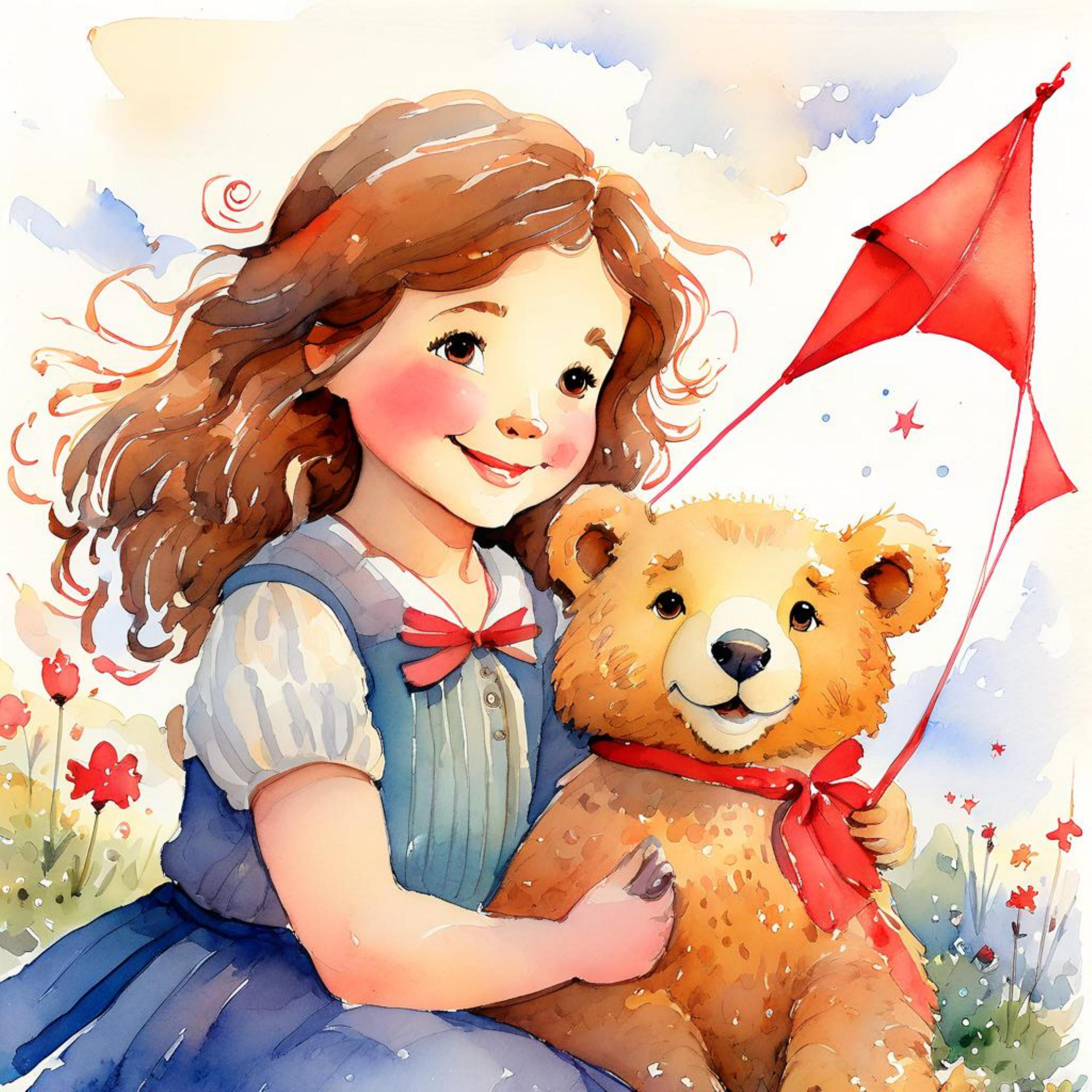} &
    \includegraphics[width=0.15\linewidth]{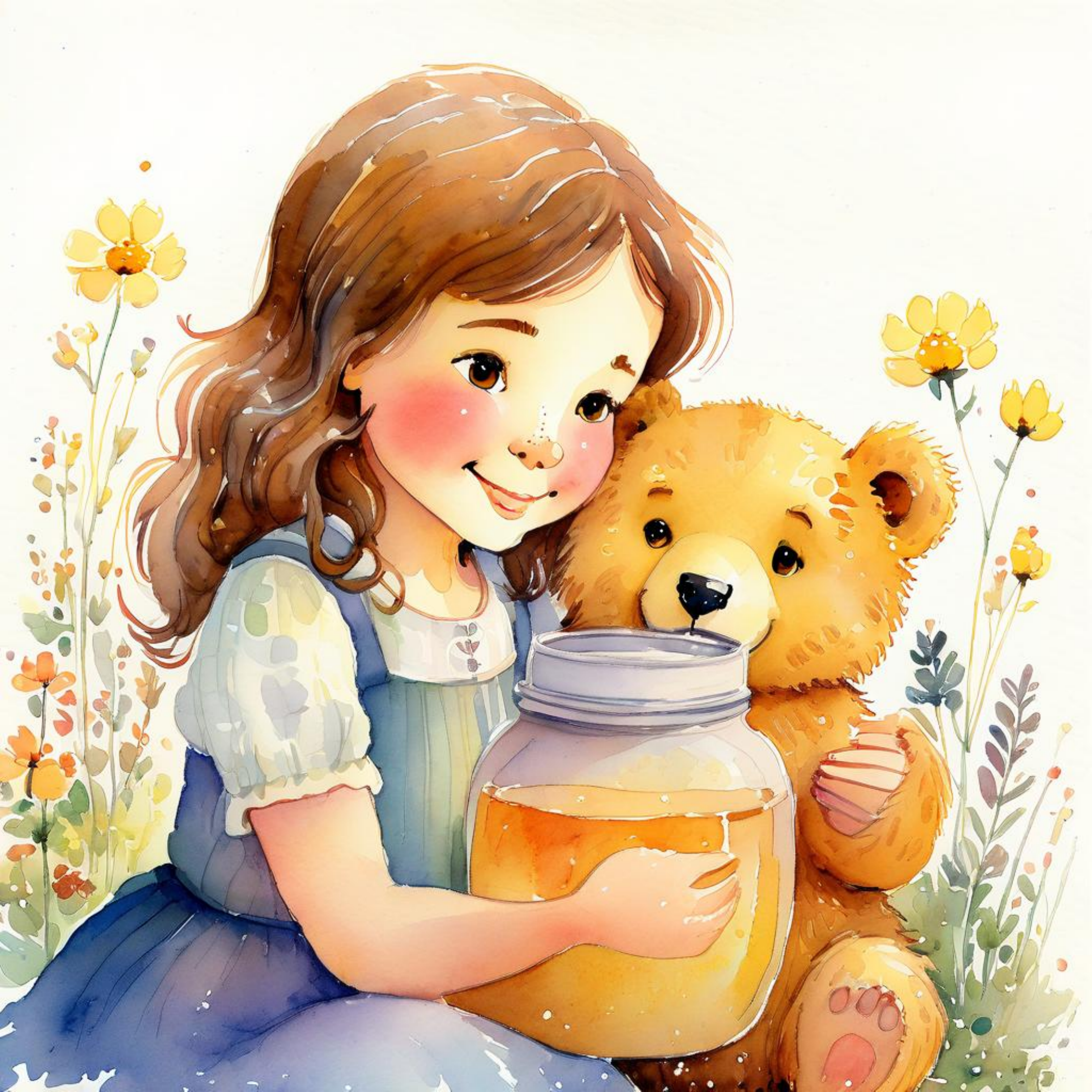} &
    \includegraphics[width=0.15\linewidth]{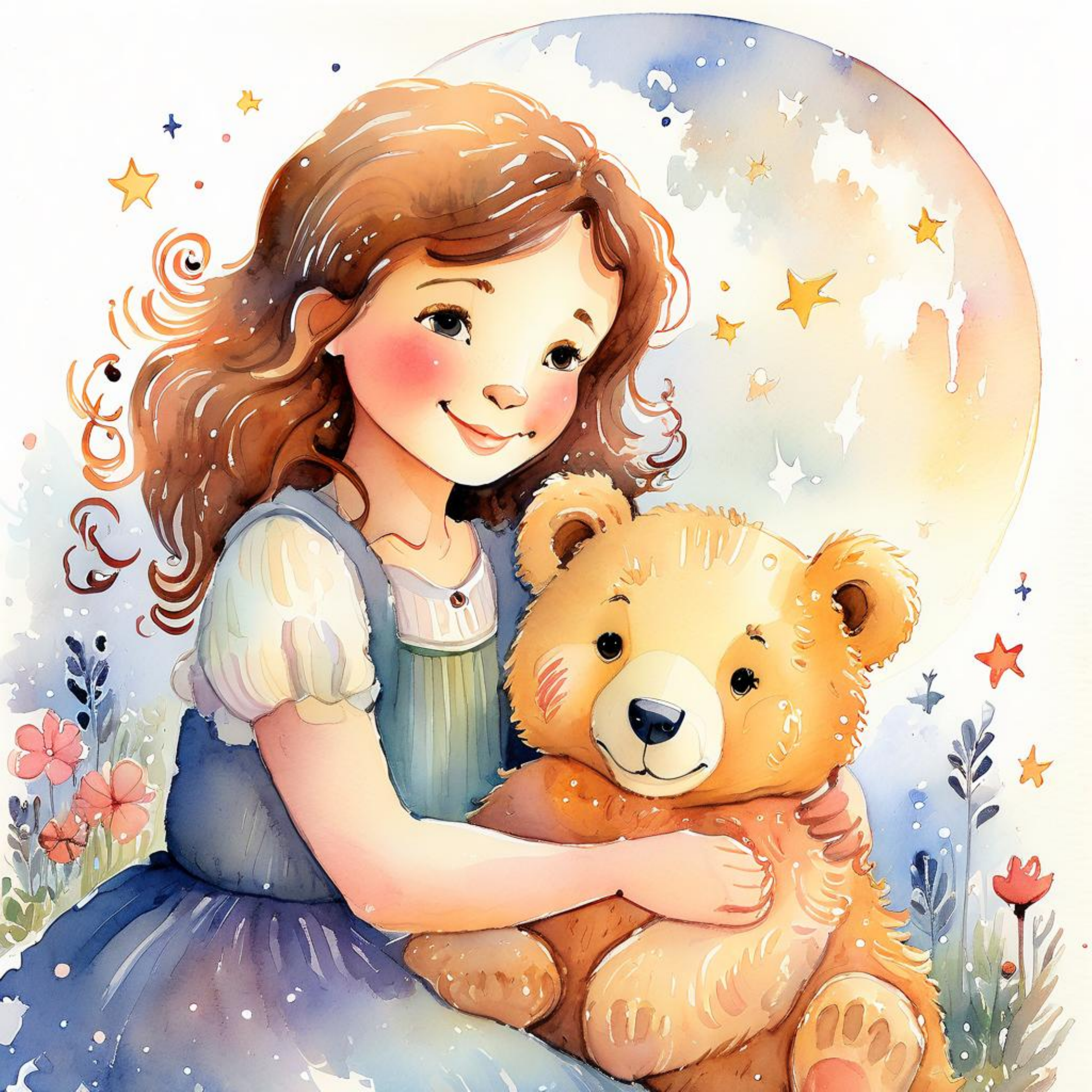} &
    \includegraphics[width=0.15\linewidth]{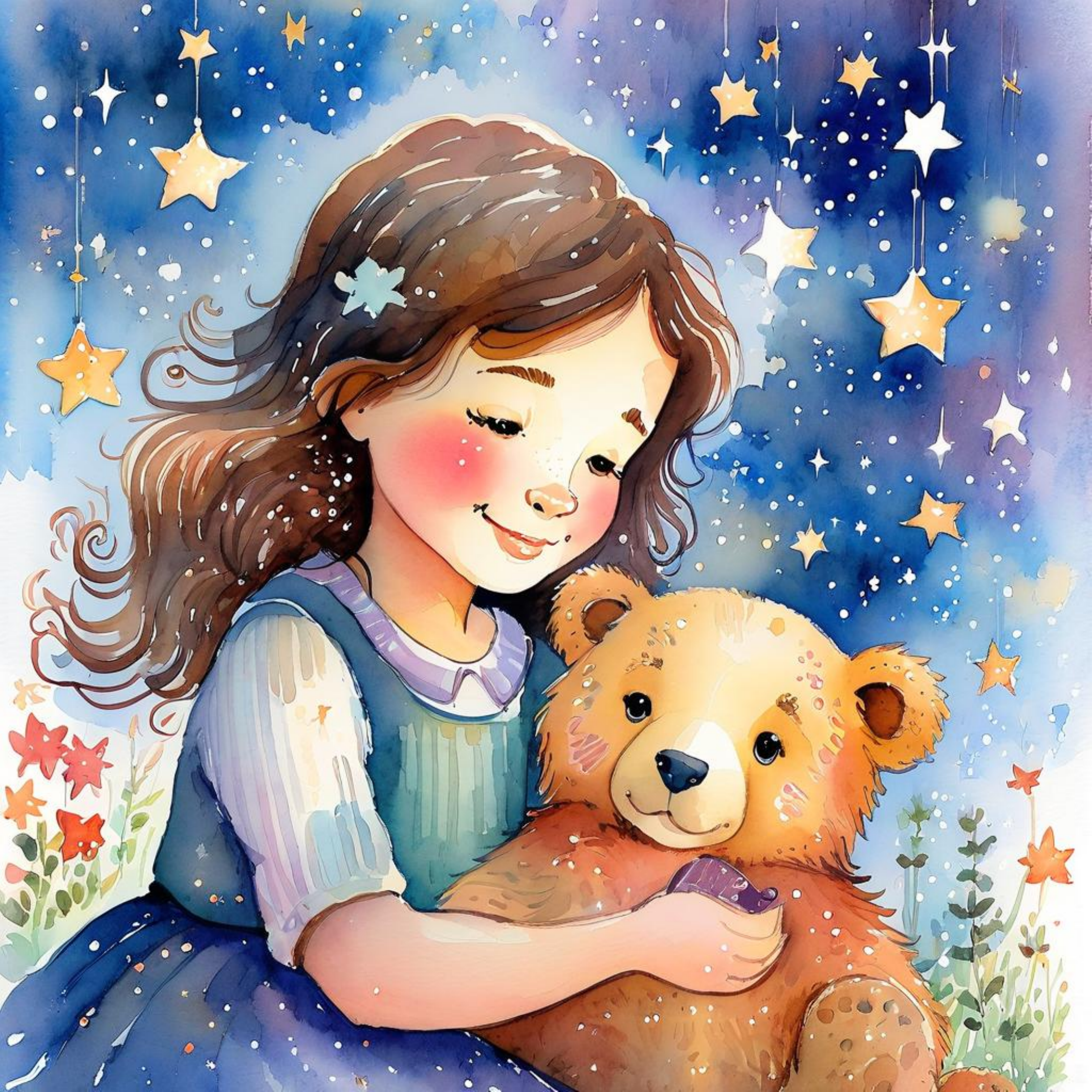} &
    \includegraphics[width=0.15\linewidth]{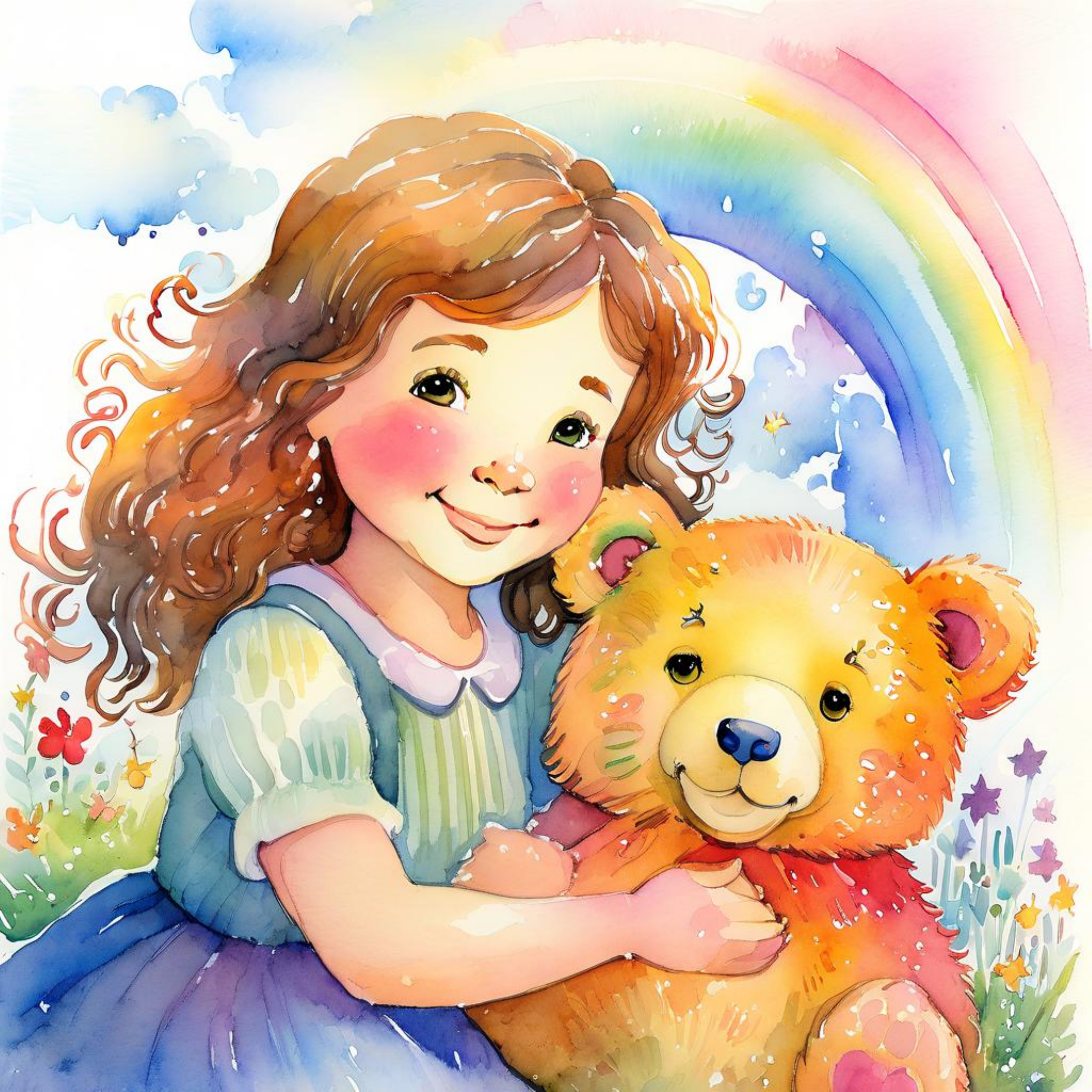} &
    \includegraphics[width=0.15\linewidth]{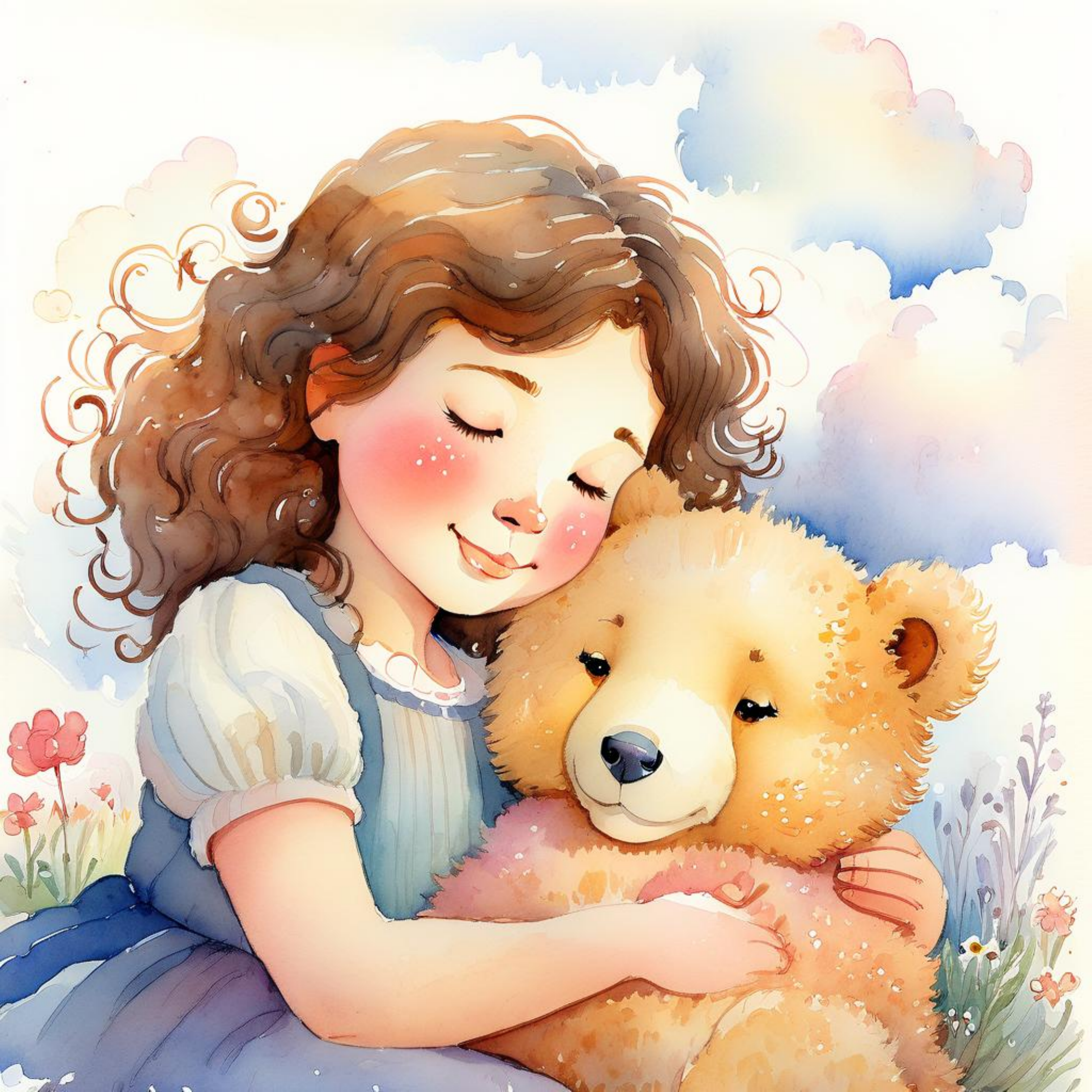}
\end{tabular}
\\

\end{tabular}
\caption{Effect of $\alpha$ on the trade-off between coherence and dynamism. Prompt: \promptpart{idcolor}{Storybook watercolor, girl and bear,} \promptpart{action1color}{flying a kite,} \promptpart{action2color}{sharing honey,} \promptpart{action3color}{sitting on the moon,} \promptpart{action4color}{catching stars,} \promptpart{action5color}{sliding down a rainbow,} \promptpart{action6color}{and napping on a cloud.}} 
\label{fig:alpha_ablation}
\end{figure*}

\begin{figure}[!ht] % [h!] tries to place the figure "here"
    \centering
    \includegraphics[width=\linewidth]{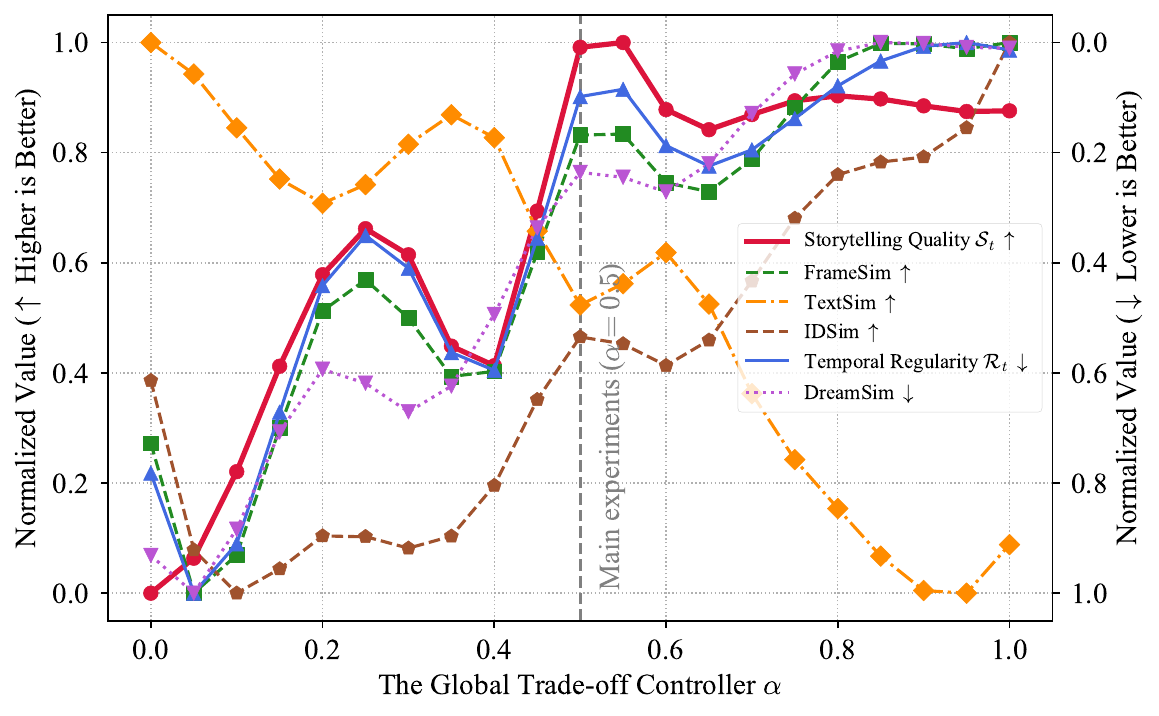} 
    \caption{Quantitative impact of the controller $\alpha$ on metrics.}
    \label{fig:alpha_metrics_plot} % A unique label for cross-referencing
\end{figure}

\noindent\textbf{Impact of the Global Trade-off Controller $\alpha$.}
A core feature of our framework is the intuitive and powerful controllability offered by $\alpha$. As illustrated in Figure~\ref{fig:alpha_ablation}, users can intuitively tune the balance between narrative dynamism and character coherence. A low $\alpha$ prioritizes dynamic poses and actions, while a high $\alpha$ strengthens the physical prior to enforce robust visual fidelity. This crucial flexibility is notably absent in previous approaches. This trade-off is quantitatively validated in Figure~\ref{fig:alpha_metrics_plot}, which presents our normalized metrics. The plot demonstrates that $\alpha$ allows for adjusting the balance between dynamism and coherence; selecting a value near 0.5, for instance, allows $\mathcal{S}_t$ to reach a favorable level without significantly compromising CLIP-T.

\begin{table}[ht]
\centering

\caption{Ablation study on the choice of physical prior. Heat diffusion achieves the best performance.}
\label{tab:ablation_physics}
\resizebox{\linewidth}{!}{%
\begin{tabular}{l|cccc|cc}
\toprule
\multirow{2}{*}{\textbf{Physical Law}} & \multicolumn{4}{c|}{\textbf{Standard Metrics}} & \multicolumn{2}{c}{\textbf{Proposed Metrics}} \\
\cmidrule(lr){2-5} \cmidrule(lr){6-7}
& \textbf{CLIP-T} & \textbf{CLIP-I}  & \textbf{IDSim}  & \textbf{DreamSim}  & \textbf{$\mathcal{R}_t$} & \textbf{$\mathcal{S}_t$} \\
\midrule
Ori (Baseline) & \textbf{0.3195} & 0.7812 & 0.8133 & 0.3603 & 20.3531 & 0.1670 \\
Burgers' Equation & 0.3077 & 0.7753 & 0.8146 & 0.3550 & 20.8394 & 0.1581 \\
Wave Equation & 0.3092 & 0.7835 & 0.7913 & 0.3725 & 20.2250 & 0.1671 \\
Conservation Law & 0.2969 & 0.7574 & 0.7842 & 0.3554 & 20.2755 & 0.1666 \\
Elasticity & 0.3106 & 0.7788 & 0.7918 & 0.3826 & 21.6029 & 0.1527 \\
\midrule
\textbf{Ours (Heat Diff.)} & 0.3161 & \textbf{0.8242} & \textbf{0.8564} & \textbf{0.2780} & \textbf{17.9104} & \textbf{0.2132} \\
\bottomrule
\end{tabular}%
}
\end{table}

\subsection{Discussion}
Our results demonstrate that RealDiffusion makes a deliberate trade-off. While accepting a minimal computational overhead and a slight decrease in CLIP-T, our method excels on metrics measuring temporal coherence in CLIP-I, DreamSim and our proposed $\mathcal{R}_t$. This confirms our framework effectively prioritizes the core challenge of multi-character coherence. The ablation studies underscore the necessity of our chosen physical prior. The dissipative nature of the heat diffusion equation is essential for guiding the generation towards a stable, coherent state. In contrast, conservative systems induce feature oscillations and non-linear priors introduced instability, confirming that a stable, dissipative process is crucial for temporal regularization. Finally, the controllability of our framework is centered on the parameter $\alpha$. Our experiments reveal its practical implications: a very high $\alpha$ can lead to static characters lacking narrative dynamism, while a very low $\alpha$ diminishes the prior's influence, causing a loss of coherence. This shows $\alpha$ is an intuitive tool for navigating the fundamental trade-off between dynamism and coherence.

\section{Conclusion}
We introduce RealDiffusion, a novel framework specifically designed to tackle the critical challenge of coherent multi-character storybook generation. Our core is a Physics-informed Attention mechanism that injects a dual physical prior directly during inference. Heat diffusion acts as a dissipative prior suppressing attribute drift and identity swapping common in multi-character scenes. A complementary stochastic process preserves narrative dynamism allowing for natural character evolution and interaction. This dual system provides fine-grained control over the trade-off between multi-character coherence and narrative dynamism. The entire intervention operates at inference time avoiding costly retraining and ensuring seamless integration with standard, pre-trained SDXL backbones. Extensive experiments on complex multi-character prompts demonstrate substantial gains in identity coherence and sustained narrative dynamism across frames. Our work thus paves the way for controllable and coherent long-form visual storytelling.

\section*{Acknowledgment}
This work was supported by the Key R\&D Program of Zhejiang (No.\ 2026C02A1234). This work was completed during the internship of Qi Zhao at SGIT AI Lab, State Grid Corporation of China. The authors would like to thank the anonymous reviewers for their constructive comments.

{
    \small
    \bibliographystyle{ieeenat_fullname}
    \bibliography{main}
}

% WARNING: do not forget to delete the supplementary pages from your submission 
% \input{sec/X_suppl1}

\end{document}